\documentclass[11pt,oneside,openany]{book}
\usepackage[a4paper, top=1.25in, bottom=1.25in, left=1.25in, right=1.25in]{geometry}
\usepackage{ragged2e}
\usepackage{amsmath}
\usepackage{amssymb}
\usepackage{todonotes}
\usepackage{graphicx}
\usepackage{subcaption}
\usepackage{caption}
\usepackage{amsfonts}
\usepackage{algorithmic}
\usepackage{xcolor}
\usepackage{url}
\usepackage{fancyhdr}
\usepackage{amsfonts}
\usepackage[most]{tcolorbox}
\usepackage[linesnumbered,ruled,vlined]{algorithm2e}
\usepackage{subcaption}
\usepackage{lastpage}
\usepackage{lipsum} 
\usepackage{lastpage}
\usepackage[pdfencoding=auto]{hyperref}

\newif\ifincludeexercises
\includeexercisesfalse
\includeexercisestrue

\hypersetup{
    pdftitle={An Introduction to Deep Reinforcement and Imitation Learning},
    pdfauthor={Pedro Santana},
    pdfsubject={Embodied agents, such as robots and virtual characters, must continuously select actions to execute tasks effectively, solving complex sequential decision-making problems. Given the difficulty of designing such controllers manually, learning-based approaches have emerged as promising alternatives, most notably Deep Reinforcement Learning (DRL) and Deep Imitation Learning (DIL). DRL leverages reward signals to optimize behavior, while DIL uses expert demonstrations to guide learning. This document introduces DRL and DIL in the context of embodied agents, adopting a concise, depth-first approach to the literature. It is self-contained, presenting all necessary mathematical and machine learning concepts as they are needed. It is not intended as a survey of the field; rather, it focuses on a small set of foundational algorithms and techniques, prioritizing in-depth understanding over broad coverage. The material ranges from Markov Decision Processes to REINFORCE and Proximal Policy Optimization (PPO) for DRL, and from Behavioral Cloning to Dataset Aggregation (DAgger) and Generative Adversarial Imitation Learning (GAIL) for DIL.},
    pdfkeywords={Deep Reinforcement Learning, Deep Imitation Learning, Markov Decision Processes, REINFORCE, Proximal Policy Optimization (PPO),  Behavioral Cloning (BC), Dataset Aggregation (DAgger), Generative Adversarial Imitation Learning (GAIL).}
}

\makeatother
\begin{document}

\begin{titlepage}
	\centering

	{\LARGE \textbf{An Introduction to\\[0.5em]
			Deep Reinforcement and Imitation Learning}}\par
	\vspace{0.3cm}

	\vspace{1cm}

	{\Large Pedro Santana\par}
	\vspace{0.5cm}

	{\large ISCTE -- University Institute of Lisbon\par}
	\vspace{0.3cm}

	{\small
		\url{pedro.santana@iscte-iul.pt}\\
		\url{https://ciencia.iscte-iul.pt/authors/pedro-santana/}
		\par}
	\vspace{2cm}

	\textbf{Abstract}

	\begin{justify}
		Embodied agents, such as robots and virtual characters, must continuously select actions to execute tasks effectively, solving complex sequential decision-making problems. Given the difficulty of designing such controllers manually, learning-based approaches have emerged as promising alternatives, most notably Deep Reinforcement Learning (DRL) and Deep Imitation Learning (DIL). DRL leverages reward signals to optimize behavior, while DIL uses expert demonstrations to guide learning. This document introduces DRL and DIL in the context of embodied agents, adopting a concise, depth-first approach to the literature. It is self-contained, presenting all necessary mathematical and machine learning concepts as they are needed. It is not intended as a survey of the field; rather, it focuses on a small set of foundational algorithms and techniques, prioritizing in-depth understanding over broad coverage. The material ranges from Markov Decision Processes to REINFORCE and Proximal Policy Optimization (PPO) for DRL, and from Behavioral Cloning to Dataset Aggregation (DAgger) and Generative Adversarial Imitation Learning (GAIL) for DIL.
	\end{justify}

	\vspace{0.5cm}

	\begin{justify}
	\end{justify}

	\vfill

\end{titlepage}

\newpage
\tableofcontents
\newpage

\mainmatter

\pagenumbering{arabic}
\chapter{Introduction}

Embodied agents, such as robots and virtual characters, must continuously decide which action to take to execute a task effectively. This requires solving a sequential decision-making problem: selecting actions over time that control the agent's actuators so it can move, perceive, and manipulate its environment with the ultimate goal of completing the assigned task.

Manually designing such sequential decision-making mechanisms is notoriously difficult. Challenges include constructing feature extractors that can interpret the agent's high-dimensional, multi-modal sensory data, and devising optimal non-linear mappings from these features to actuator commands. In many situations, the controller must also incorporate memory and actively manage the sensing process itself, further complicating manual engineering.

A powerful alternative is to rely on machine learning. Learning-based methods construct control policies end-to-end, jointly learning the sensory feature extractors and the mappings from those features to actuator commands. To do so, these methods require a feedback signal that indicates how well the agent is performing the task, whether it involves locomotion, object manipulation, navigation, or other skills. This feedback is typically provided by a human expert.

In some settings, the human expert can directly evaluate the agent's behavior, supplying positive rewards for desirable actions and negative rewards for undesirable ones. Deep Reinforcement Learning algorithms use this reward signal to train end-to-end control policies. In other settings, the expert demonstrates how to perform the task. The discrepancy between the agent's behavior and the expert demonstrations provides a strong learning signal, which Deep Imitation Learning algorithms exploit to train end-to-end controllers.

\section{Overview}

The goal of this document is to introduce the reader to Deep Reinforcement Learning and Deep Imitation Learning in the context of embodied agents. The document starts by presenting the formalism of Markov Decision Processes (MDPs) and describing both exact and approximate solution methods. Among approximate methods, this document covers the classical algorithm REINFORCE \cite{williams1992reinforce} and then focus on Proximal Policy Optimization (PPO) \cite{schulman2017ppo}, one of the most widely used and effective reinforcement learning algorithms for controlling embodied agents. The document then transitions to Deep Imitation Learning, introducing three foundational approaches: (a) Behavioral Cloning~\cite{Pomerleau1991,Bojarski2016}; (b) its interactive extension, Dataset Aggregation (DAgger)~\cite{dagger}; and (c) Generative Adversarial Imitation Learning (GAIL)~\cite{gail}.

\section{Readership}

This document is intended for students and researchers with a college-level background in mathematics and computer science who wish to learn about Deep Reinforcement Learning and Deep Imitation Learning applied to embodied agents. Although a basic mathematical background is assumed, this document includes a chapter reviewing the core principles necessary to understand the subsequent material. Prior knowledge of machine learning is helpful but not mandatory, as all necessary concepts are introduced as they arise. The document is designed to be self-contained: from mathematics to machine learning, all relevant concepts are presented when they are needed, following a pedagogical and progressive approach.

\section{Scope}

This document grew from lecture notes prepared by the author, adopting a concise, depth-first approach to the literature. It is not intended to be a survey of the field. The focus is deliberately placed on a small set of foundational algorithms and techniques, prioritizing in-depth analysis over broad coverage.
This choice reflects the belief that a deep understanding of core methods provides a stronger foundation for independently learning the many existing and future variants than a superficial, breadth-first overview.

\section{How to Read}

Readers who need to revisit basic principles of probability, information theory, and calculus should begin with Chapter~\ref{cha:Mathematical_Primer} (Mathematical Background). After that, depending on their interests, the document can be followed along two main reading paths, described below.

Readers interested in Deep Reinforcement Learning should proceed to Chapter~\ref{cha:MDP} (Markov Decision Processes) and then to Chapter~\ref{cha:drl} (Deep Reinforcement Learning). Those who also wish to learn about Deep Imitation Learning should continue to Chapter~\ref{cha:dil} (Deep Imitation Learning). Such readers may skip Section~\ref{sec:back_concepts}, which provides a concise recap of the key ideas from Chapters~\ref{cha:MDP} and \ref{cha:drl} that are essential for understanding imitation learning algorithms.

Readers primarily interested in Deep Imitation Learning, but not necessarily in Deep Reinforcement Learning, may jump directly to Chapter~\ref{cha:dil} (Deep Imitation Learning), which, as noted, includes a recap of the essential concepts from the earlier chapters.

\section{Further Readings}

The material in this document can be complemented with several recommended resources for readers seeking a deeper theoretical or practical understanding of the topics covered.

For Deep Reinforcement Learning, Stable Baselines3 (SB3) \cite{sb32024} offers reliable PyTorch implementations of widely used reinforcement learning algorithms, making it an excellent platform for experimentation. In addition, OpenAI's educational resources \cite{openai2018spinningup} provide a valuable combination of theory and hands-on guidance. The textbook by Sutton and Barto \cite{sutton2018reinforcement} remains a foundational and highly recommended reference, offering a comprehensive introduction to reinforcement learning principles and methods.

For Deep Imitation Learning, several surveys are recommended for broader context and historical perspective~\cite{Hussein2017,Zheng2022,zare2024}, along with domain-specific reviews focusing on virtual characters~\cite{Mourot2022,Kwiatkowski2022}. Practical resources include the \texttt{imitation} library~\cite{imitation} and \texttt{ML-Agents} \cite{Juliani2018}, both of which provide PyTorch-based implementations of modern imitation learning algorithms.

\subsection{Credits}

The scientific content presented in this document is entirely drawn from the cited literature; no original scientific claims should be attributed to the author of this document. Any original contributions are purely pedagogical, limited to new examples, exercises, and to the manner in which well-established scientific concepts are explained and communicated to the reader. 

\chapter{Mathematical Background}\label{cha:Mathematical_Primer}

Mathematics and computational modeling lie at the heart of Machine Learning. The field welcomes both mathematically inclined readers and those with a strong background in programming, offering a natural meeting point between theory and practice. It also provides an excellent opportunity to see mathematical ideas at work, demonstrating how concepts taught in the classroom translate into real-world problem-solving, especially in modern systems powered by Machine Learning.

Fortunately, the mathematical tools required to gain a solid understanding of most deep reinforcement learning and imitation learning algorithms are typically covered in undergraduate engineering and computer science curricula. This chapter introduces the essential concepts while intentionally avoiding the level of formal rigor expected in a mathematics textbook. Instead, the focus is on building intuition and comfort with the material so that the reader is well prepared for the chapters that follow. To support this goal, the exposition includes illustrative numerical examples and concludes with a set of exercises designed to reinforce the key ideas.

\section{Probability}

\subsection{Sample Space}

When some kind of experiment is performed in the real-world, it is always necessary to take into account the possibility of randomness. Although it is not possible to be fully sure of the outcome of the experiment, the set of all possible outcomes is often known. The set of all possible outcomes of an experiment is known as \textit{sample space}, $\Omega$. For instance, in an experiment consisting of tossing fair coins, then $\Omega=\{H, T\}$, where $H$ and $T$ mean that the outcome of the toss is head and tail, respectively. If the experiment is tossing two fair coins, then its sample space is $\Omega=\{(HH), (HT),(TH),(TT)\}$.

Any subset $E$ of the sample space $\Omega$, $E\subset \Omega$, is known as an \textit{event}. For instance, $E = \{(H,H), (H,T)\}$ refers to the event that a head appears on the first coin. The probability of every event $E$ of the sample space $\Omega$ to occur is $P(E)$, which is subject to the following conditions:
\[ 0 \leq P(E) \leq 1 \quad \text{and} \quad P(\Omega)=\sum_{o\in \Omega} P(o) = 1. \]

For example, if two fair coins are tossed, the probability of getting two heads is $P(\{HH\})=\frac{1}{4}$, since the sample space has four equally probable elements. In the same experiment, the probability of getting a single head is $P(\{HT,TH\})=P(\{HT\})+P(\{TH\})=\frac{1}{4}+\frac{1}{4}=\frac{1}{2}$.

\subsection{Discrete Random Variables}

We may be interested on certain functions of experiment outcomes rather than the outcomes themselves. For instance, when rolling two dice, we might be more interested in the sum of the dice than the specific numbers rolled. We might want to know that the sum is seven, without caring whether the roll was (1, 6), (2, 5), (3, 4), (4, 3), (5, 2), or (6, 1). These functions of outcomes are known as \textit{random variables}. They are called random because their specific values are determined by chance.

Formally, a \textit{discrete random variable} \( X \) is a deterministic function that assigns a real number to each outcome in a sample space of a random experiment, \( X : \Omega \to \mathbb{R} \), where the random variable takes on a countable number of distinct values. The complete set of values that the random variable $X$ can take is denoted by $R(X)$.

The probability that a discrete random variable \( X \) takes a specific value \( x \in R(X) \) is defined by its Probability Mass Function (PMF), denoted as \( P(X = x) \). For simplicity, \( P(X = x) \) will be referred to hereafter simply as the probability of \( X \) taking the value \( x \).

For the sake of simplicity, instead of $\sum_{x\in R(X)}$, we will use $\sum_x$ to denote a summation over all possible values that the random variable $X$ can take. Following the definition of probability, the following conditions must hold:
\[ 0 \leq P(X = x) \leq 1 \quad \text{and} \quad \sum_{x} P(X = x) = 1. \]

Let us revisit the simple example in which a fair coin is tossed twice. In this experiment, the sample space is \( \Omega = \{HH, HT, TH, TT\} \). Let the discrete random variable \( X \) be the number of heads observed in these two coin tosses. Therefore, \( X \) can take the values 0, 1, or 2:
\[
	X(TT) = 0, X(HT) = 1, X(TH) = 1, X(HH) = 2.
\]

Note that this example clearly shows that the random variable $X$ is a simple deterministic function that maps an occurred outcome to a real-valued scalar. The randomness is fully contained in the experiment, that is, in determining which outcome actually occurs. This approach allows to decouple the randomness of the experiment from its deterministic mathematical treatment. Let us get back to our example. Since the coin is fair, each of the four outcomes is equally likely and, thus,
\begin{align*}
	P(X = 0) & = P(\{TT\}) = \frac{1}{4},      \\
	P(X = 1) & = P(\{HT, TH\}) =  \frac{1}{2}, \\
	P(X = 2) & = P(\{HH\}) = \frac{1}{4}.
\end{align*}

Note that the probabilities sum up to 1, satisfying the condition for probability distributions,
\[
	\sum_{x} P(X = x) = P(X = 0) + P(X = 1) + P(X = 2) = \frac{1}{4} + \frac{1}{2} + \frac{1}{4} = 1.
\]

\subsection{Continuous Random Variables}\label{sec:cont_rand_var}

In some experiments, we may be interested in outcomes that vary over a continuous range rather than distinct discrete values. In these cases, the random variable is called of \textit{continuous random variables} as it takes one of an uncountable number of distinct values.

Formally, a \textit{continuous random variable} \( X \) is a deterministic function that assigns a real number to each outcome in a sample space of a random experiment, \( X : \Omega \to \mathbb{R} \). Unlike discrete random variables, a continuous random variable can take any value within a certain range or interval. The probability that \( X \) takes a specific value is always zero, as there are infinitely many possible values in any continuous range. Instead, probabilities are defined over intervals and are determined using a \textit{probability density function (PDF)}, denoted by \( f_X(x) \). The function \( f_X(x) \) satisfies the following conditions:
\[
	f_X(x) \geq 0 \quad \text{for all } x \in \mathbb{R},
\]
\[
	\int_{-\infty}^{\infty} f_X(x) \, dx = 1.
\]

The probability that \( X \) falls within an interval \([a, b]\) is given by the integral of the PDF over that interval:
\[
	P(a \leq X \leq b) = \int_a^b f_X(x) \, dx.
\]
\subsubsection{Uniform Distribution}

Suppose \( X \) represents the random outcome of selecting a point uniformly at random on the interval \([0, 1]\). In this case, the PDF of \( X \) is defined as follows:
\[
	f_X(x) =
	\begin{cases}
		1 & \text{if } 0 \leq x \leq 1 \\
		0 & \text{otherwise}
	\end{cases},
\]

\noindent which is a valid PDF because \( f_X(x) \geq 0 \) for all \( x \) and it integrates to 1 over the entire domain (recall that the primitive of 1 is \( x \)):
\[
	\int_{-\infty}^\infty f_X(x) \, dx = \int_0^1 1 \, dx = [x]_0^1 = 1 - 0 = 1.
\]

The probability that \( X \) lies within a subinterval \([a, b]\), where \( 0 \leq a < b \leq 1 \), is:
\[
	P(a \leq X \leq b) = \int_a^b f_X(x) \, dx = \int_a^b 1 \, dx = [x]_a^b = b - a.
\]

For example, the probability that \( X \) lies in \([0.25, 0.75]\) is:
\[
	P(0.25 \leq X \leq 0.75) = \int_{0.25}^{0.75} 1 \, dx = [x]_{0.25}^{0.75} = 0.75 - 0.25 = 0.5.
\]

Since the distribution is uniform over \([0, 1]\), each subinterval of equal length has the same probability. The length of the interval \([0.25, 0.75]\) is \( 0.75 - 0.25 = 0.5 \), which is half the length of the entire interval \([0, 1]\). Therefore, the probability of \( X \) falling within this subinterval is exactly \( 0.5 \), as obtained.

\subsubsection{Gaussian Distribution}\label{sec:gaussian-distribution}

\begin{figure}[h]
	\begin{center}
		\includegraphics[width=11cm]{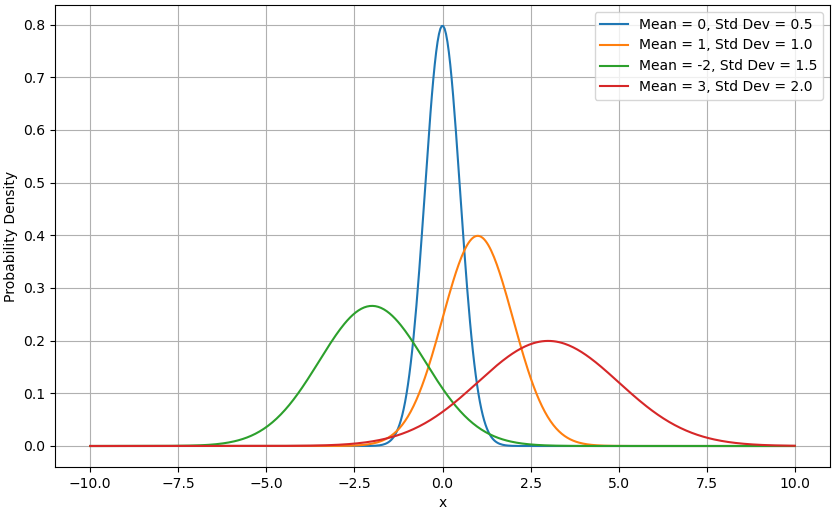}
		\caption{PDFs of Gaussian distributions with different means and standard deviations.}
		\label{fig:gaussians}
	\end{center}
\end{figure}

Let us now analyze the case of a random variable \( X \) that follows a Gaussian (normal) distribution with mean $\mu$ and standard deviation $\sigma$. Such a distribution is denoted by $\mathcal{N}(\mu, \sigma^2)$, where $\sigma^2$ is the variance. The mean defines the center of the distribution, which is the most probable outcome, while the standard deviation defines the spread or dispersion of the values around the mean. Figure~\ref{fig:gaussians} illustrates the PDFs of Gaussian distributions with different means and standard deviations.

Formally, the PDF of a continuous random variable \( X \) that follows a Gaussian distribution with mean $\mu$ and standard deviation $\sigma$, is defined as:
\begin{equation}\label{equ:gaussian}
	f_X(x) = \frac{1}{\sigma \sqrt{2 \pi}} \exp \left( -\frac{(x - \mu)^2}{2 \sigma^2} \right).
\end{equation}

This equation expresses that the probability density decreases as $x$ moves further away from the mean $\mu$, and that the density is scaled according to the dispersion $\sigma$. In other words, a larger standard deviation spreads the distribution, lowering the peak of the density, while a smaller standard deviation concentrates the probability mass around the mean.

\subsection{Sampling Distributions}

Sampling is the process of obtaining a specific value of a random variable according to a given probability distribution. Let us denote the sampling process with \( \sim \). Sampling draws values stochastically rather than deterministically. This means that values with higher probability under the distribution are more likely to be selected than those with lower probability, although no single outcome is guaranteed. In other words, the likelihood of drawing a particular value is proportional to its probability mass or density under the specified distribution.

For example, obtaining a sample \( x \) from a Gaussian probability density function (PDF) is denoted as:
\[
	x \sim \mathcal{N}(\mu, \sigma^2).
\]

The distribution, whether Gaussian or not, can arise naturally in real-world phenomena. In such cases, sampling is performed by measuring some quantity, such as the current temperature at the reader's location. Alternatively, distributions can be represented computationally, where samples are generated programmatically.

In computational sampling, a procedure is used to randomly select a value according to the PMF for discrete random variables or the PDF for continuous random variables. For example, the Python package NumPy can be used to generate eight samples from a Gaussian distribution centered at 10 with a standard deviation of 1, $\mathcal{N}(10, 1)$ with the following function call:

\[\text{\texttt{print(numpy.random.normal(loc=10, scale=1, size=8))}}\]

A possible output of this function call is:

\[\text{\texttt{[10.08  9.79  9.22 10.60 10.12 10.37 9.47 9.84]}},\]

\noindent where most sampled values lie close to 10, although they are not exactly equal to it, reflecting the inherent randomness of the sampling process. As the standard deviation increases, the variability of the observed values also increases.

\subsection{Joint Probability of Discrete Random Variables}

For discrete random variables \( X \) and \( Y \), the joint probability \( P(X = x, Y = y) \) represents the probability that \( X \) takes the value \( x \) and \( Y \) takes the value \( y \) simultaneously. This can be viewed as the probability of the event where both conditions \( X = x \) and \( Y = y \) are satisfied at the same time. Formally, it is expressed as
\[
	P(X = x, Y = y).
\]

The sum of the joint probabilities over all possible pairs of values \( (x, y) \) must be 1, as it encompasses all possible outcomes of the random variables \( X \) and \( Y \). This is a requirement for the joint probability distribution to be valid. Mathematically, this is represented as
\[
	\sum_{x} \sum_{y} P(X = x, Y = y) = 1.
\]

If the random variables \( X \) and \( Y \) are independent, the joint probability is the product of the individual probabilities,
\[
	P(X = x, Y = y) = P(X = x) P(Y = y).
\]

\subsubsection*{Example: ChatBot}

Let us imagine a study in which 100 potential users were surveyed to determine their preferences for interacting with a chatbot that has a cartoonish appearance. The objective of the study was to analyze whether these users would like or dislike such a ChatBot and how their preferences are distributed between children and adults. Assuming the sample size is representative of the overall population, the statistics derived from this survey can be generalized.

Table~\ref{tab:chatbot-example-users} shows the distribution of users based on their age group and preferences. For example, 30 individuals are children who like the cartoonish character. By dividing each number in the table by the total sample size (100), we can calculate the joint probability distribution over the random variables $X$ and $Y$, denoted as $P(X=x, Y=y)$, where $x \in \{\text{like}, \text{dislike}\}$ and $y \in \{\text{child}, \text{adult}\}$. For instance, the 30 children who like the cartoonish character represent a joint probability of $P(X=\text{like}, Y=\text{child}) = 0.3$, which can be interpreted as 30\%.

Table~\ref{tab:chatbot-example-joint} provides the complete joint probability distribution. Please remember that this is a hypothetical survey (no actual user study was conducted) and most likely it does not accurately reflect real-world preferences.

\begin{table}[h]
	\centering
	\caption{Distribution of user preferences for the ChatBot example.}
	\label{tab:chatbot-example-users}
	\begin{tabular}{|c|c|c|c|}
		\hline
		               & \textbf{Like} & \textbf{Dislike} & Total \\
		\hline
		\textbf{Child} & 30            & 10               & 40    \\
		\hline
		\textbf{Adult} & 20            & 40               & 60    \\
		\hline
		Total          & 50            & 50               & 100   \\
		\hline
	\end{tabular}
\end{table}

\begin{table}[h]
	\centering
	\caption{Joint probability distribution for the ChatBot example.}
	\label{tab:chatbot-example-joint}
	\begin{tabular}{|c|c|c|}
		\hline
		               & \textbf{Like}                            & \textbf{Dislike}                            \\
		\hline
		\textbf{Child} & $P(X=\text{like}, Y=\text{child}) = 0.3$ & $P(X=\text{dislike}, Y=\text{child}) = 0.1$ \\
		\hline
		\textbf{Adult} & $P(X=\text{like}, Y=\text{adult}) = 0.2$ & $P(X=\text{dislike}, Y=\text{adult}) = 0.4$ \\
		\hline
	\end{tabular}
\end{table}

\subsection{The Log Trick}

When dealing with a sequence of events for time steps $t=1,2,\ldots,n$ for a set of independent random variables \( X_1, X_2, \ldots, X_n \), the joint probability of these events occurring is given by
\[
	P(X_1 = x_1, X_2 = x_2, \ldots, X_n = x_n) = \prod_{i=1}^{n} P(X_i = x_i),
\]

\noindent which represents the probability of observing the specific values $x_1,x_2,\ldots,x_n$ for the corresponding random variables \( X_1, X_2, \ldots, X_n \).

When many very small probabilities are multiplied together, numerical underflow may occur. To prevent this, we use the log trick. Taking the logarithm of the joint probability transforms the product of probabilities into a sum of logarithms, as a consequence of the logarithmic identity \( \ln(xy) = \ln(x) + \ln(y) \), where $\ln(z)$ denotes the natural logarithm of $z$, i.e., the power to which $e \approx 2.71828$ must be raised to yield $z$. This transformation simplifies calculations and enhances numerical stability, as it is easier to handle sums of logarithms than products of small numbers:
\begin{align*}
	\ln P(X_1 = x_1, X_2 = x_2, \ldots, X_n = x_n) & = \ln \left( \prod_{i=1}^{n} P(X_i = x_i) \right) \\
	                                               & =\sum_{i=1}^{n} \ln P(X_i = x_i).
\end{align*}

\begin{figure}[h]
	\begin{center}
		\includegraphics[width=10cm]{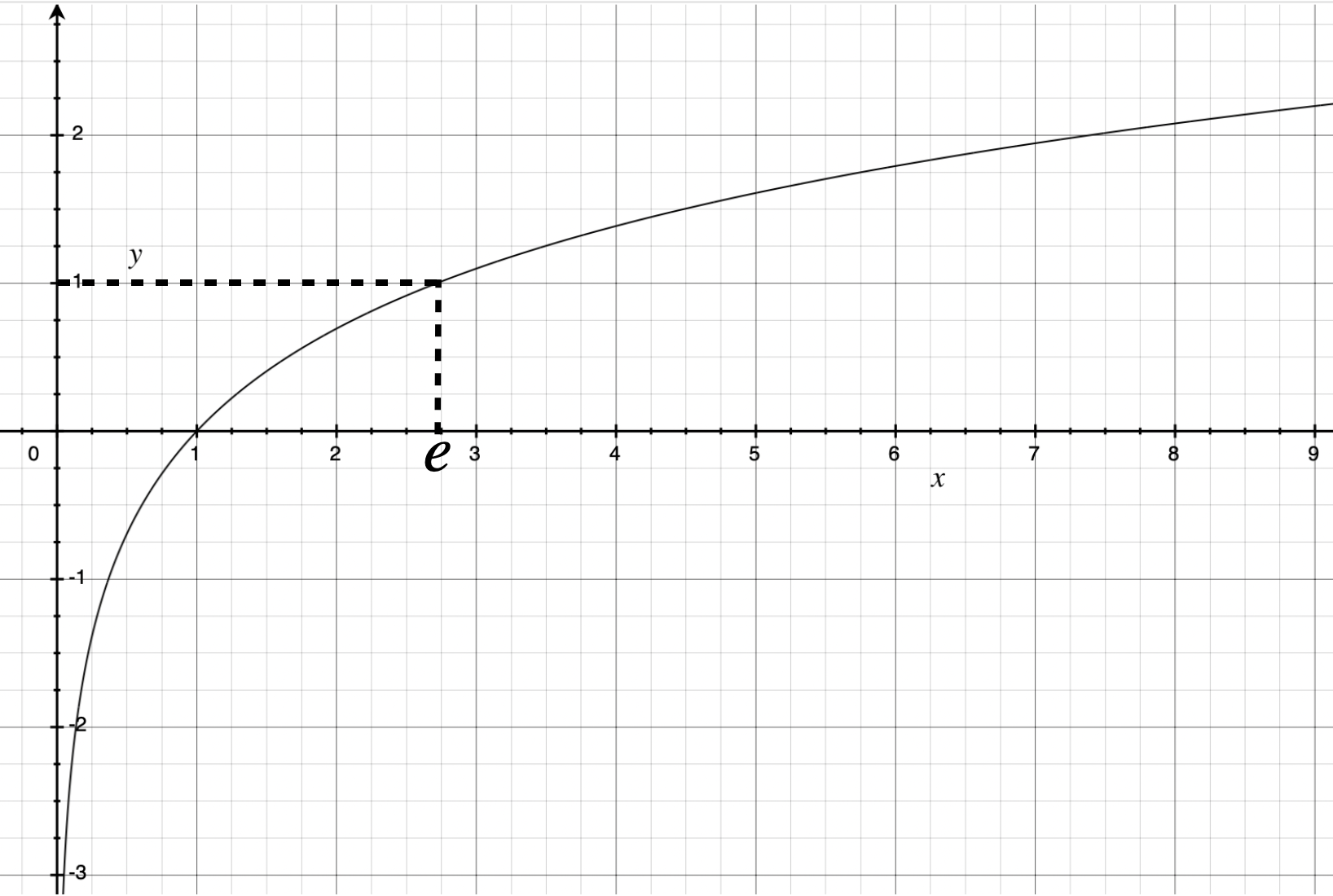}
		\caption{Plot of the natural logarithm function, $\ln(x)$.}
		\label{fig:nt-log}
	\end{center}
\end{figure}

In optimization problems, the log trick is particularly useful. Optimizing the log probability yields the same result as optimizing the probability itself because the logarithm is a monotonically increasing function (see Figure~\ref{fig:nt-log}). This means that the values of the probability and the log probability increase and decrease together. Therefore, finding the maximum of the log probability will also find the maximum of the probability. This simplifies the optimization process and often makes it more computationally efficient.

\subsection{Probability Marginalization}

Marginalization is the process of summing the joint probability distribution over one of the variables to obtain the marginal probability distribution of the other variable. This process gives the probabilities of each variable individually, without reference to the other variable. For instance, to find the marginal probability of \( X \), we sum over all possible values of \( Y \):
\[ P(X = x) = \sum_{y} P(X = x, Y = y). \]

Similarly, to find the marginal probability of \( Y \), we sum over all possible values of \( X \):
\[ P(Y = y) = \sum_{x} P(X = x, Y = y). \]

In these equations, \( P(X = x) \) is the probability that \( X \) takes the value \( x \) regardless of the value of \( Y \) while \( P(Y = y) \) is the probability that \( Y \) takes the value \( y \) regardless of the value of \( X \).

\subsubsection*{Example: ChatBot (continued)}

Table~\ref{tab:chatbot-example-marginal} illustrates both the joint and marginal probability distributions for the ChatBot example. Recall that this hypothetical survey examines user preferences for a ChatBot with a cartoonish appearance, segmented by age groups (children and adults) and their likes or dislikes.

The marginal probabilities are derived by summing the joint probabilities across the different categories of the other variable. For instance, the marginal probability $P(X=\text{child})=0.4$ is obtained by adding $P(X=\text{child}, Y=\text{like})$ and $P(X=\text{child}, Y=\text{dislike})$, which represent the total probability of being a child regardless of preference. The probability of liking the ChatBot regardlessness of age, $P(Y=\text{like})=0.5$, is found by summing the probabilities of both children and adults who like it.

\begin{table}[h]
	\centering
	\caption{Joint and marginal probability distributions for the ChatBot example.}
	\label{tab:chatbot-example-marginal}
	\resizebox{\textwidth}{!}{
		\begin{tabular}{|c|c|c|c|}
			\hline
			               & \textbf{Like}                          & \textbf{Dislike}                          & Total                   \\\hline
			\textbf{Child} & $P(X=\text{child}, Y=\text{like})=0.3$ & $P(X=\text{child}, Y=\text{dislike})=0.1$ & $P(X=\text{child})=0.4$ \\\hline
			\textbf{Adult} & $P(X=\text{adult}, Y=\text{like})=0.2$ & $P(X=\text{adult}, Y=\text{dislike})=0.4$ & $P(X=\text{adult})=0.6$ \\\hline
			Total          & $P(Y=\text{like})=0.5$                 & $P(Y=\text{dislike})=0.5$                 &                         \\\hline
		\end{tabular}
	}
\end{table}

\subsection{Conditional Probability}

For discrete random variables \( X \) and \( Y \), the conditional probability \( P(X = x \mid Y = y) \) represents the probability that \( X \) takes the value \( x \) given that \( Y \) has taken the value \( y \). This can be interpreted as the proportion of times \( X \) takes the value \( x \) whenever \( Y \) takes the value \( y \). Formally, this is defined as
\[
	P(X = x \mid Y = y) = \frac{P(X = x, Y = y)}{P(Y = y)}, \quad \text{with } P(Y = y) > 0.
\]

Moreover, the conditional probability \( P(X = x \mid Y = y) \) itself forms a probability distribution over \( X \) for any given \( y \). Therefore, the sum of \( P(X = x \mid Y = y) \) over all possible values of \( x \) must be 1:
\[
	\sum_{x} P(X = x \mid Y = y) = 1,\quad \text{with } P(Y = y) > 0.
\]

\subsubsection*{Example: ChatBot (continued)}

Let us proceed with the analyses of our ChatBot example. Table~\ref{tab:chatbot-example-marginal} shows that the marginal probabilities for liking and disliking the ChatBot are $P(Y=\text{like}) = 0.5$ and $P(Y=\text{dislike}) = 0.5$, indicating that, overall, the probability of a user liking the ChatBot is equal to the probability of disliking it. However, examining the conditional probabilities provides additional insights. The conditional probability of a user liking the ChatBot given they are a child is:
\[
	P(Y=\text{like} \mid X=\text{child}) = \frac{P(X=\text{child}, Y=\text{like})}{P(X=\text{child})} = \frac{0.3}{0.4} = 0.75.
\]

This can be interpreted as 75\% of children liking the ChatBot. Similarly, the conditional probability of a user liking the ChatBot given they are an adult is:
\[
	P(Y=\text{like} \mid X=\text{adult}) = \frac{P(X=\text{adult}, Y=\text{like})}{P(X=\text{adult})} = \frac{0.2}{0.6} = 0.33(3).
\]

These conditional probabilities reveal that children are more likely to like the ChatBot compared to adults. Specifically, 75\% of children like the ChatBot, while only 33.3\% of adults do. Note that, although the marginal probabilities indicate a higher likelihood of encountering an adult ($P(X=\text{adult}) = 0.6$) than a child ($P(X=\text{child}) = 0.4$), the conditional probabilities highlight that children have a stronger preference for the ChatBot. Thus, this example shows how different aspects of the data can be revealed through different types of probability distributions.

\subsection{Chain Rule of Probability}

The chain rule of probability, which follows directly from the definition of conditional probability, states how joint distributions over random variables can be decomposed into conditional distributions over only one variable:
\begin{align*}
	P(X = x, Y = y) & = P(X = x \mid Y = y)P(Y = y) \\
	                & =P(Y = y \mid X = x)P(X = x)
\end{align*}

Via simple recursive substitutions, the rule generalizes to $n$ random variables as follows:
\begin{equation*}
	P(X_1 = x_1, \ldots X_n = x_n) = P(X_1 = x_1)\prod_{i=2}^nP(X_i = x_i \mid X_1=x_1,\ldots X_{i-1}=x_{i-1}).
\end{equation*}

When stating probability equations, it is common to treat the joint probability of groups of variables as a single entity. For example, for $n=3$, the following decompositions can be derived:
\begin{align*}
	P(X = x, Y = y, Z = z) & =
	P(X=x)P(Y=y \mid X=x)P(Z=z \mid X=x,Y=y)                          \\
	                       & =P(X=x \mid Y = y, Z = z)P(Y = y, Z = z) \\
	                       & =P(X=x,Y = y \mid  Z = z)P(Z = z).
\end{align*}

\subsection{Independence and Conditional Independence}

Two random variables \(X\) and \(Y\) are considered \textit{independent} if their joint probability distribution can be expressed as the product of their individual marginal probability distributions:
\begin{equation}
	P(X = x, Y = y) = P(X = x) P(Y = y), \quad \forall x\in R(x), \forall y\in R(y).
	\label{equ:joint_prob_def}
\end{equation}

Note Equation~\ref{equ:joint_prob_def} defines joint probability without referring to any conditional probability. That is, when two random variables are independent, knowing the value of one provides no information about the value of the other. This can also be expressed as follows:
\[
	P(X = x \mid Y = y) = P(X = x), \quad \forall x \in R(X), \ \forall y \in R(Y) \text{ with } P(Y = y) > 0,
\]
\[
	P(Y = y \mid X = x) = P(Y = y), \quad \forall y \in R(Y), \ \forall x \in R(X) \text{ with } P(X = x) > 0.
\]

Two random variables $X$ and $Y$ are \textit{conditionally independent} given a random variable $Z$ if, for every value of $Z$, the conditional probability distribution of $X$ and $Y$ can be expressed as a product of the individual conditional distributions of $X$ given $Z$ and $Y$ given $Z$:
\begin{align*}
	P(X=x, & Y=y \mid Z=z)=                  \\
	       & P(X=x \mid Z=z)P(Y=y \mid Z=z).
\end{align*}

\subsubsection*{Example: ChatBot (continued)}

We can use the probabilities expressed in Table~\ref{tab:chatbot-example-marginal} to determine whether the random variables \(X\) (age group) and \(Y\) (preference) in our ChatBot example are independent. For the random variables to be independent, the joint probability must equal the product of the marginal probabilities. However, this is not the case in our example. For instance, from the table, if follows that:
\[
	P(X=\text{child}, Y=\text{like}) \neq P(X=\text{child}) P(Y=\text{like}) \quad \text{(because } 0.3 \neq 0.4 \cdot 0.5).
\]

This discrepancy indicates that knowing the user's age group provides information about their likelihood of liking or disliking the ChatBot. This conclusion is further supported by examining the conditional probability:
\[
	P(Y=\text{like} \mid X=\text{child}) \neq P(Y=\text{like}) \quad \text{(because } 0.75 \neq 0.5).
\]

Hence, the preferences are not independent of the age group, as the likelihood of liking the ChatBot differs when conditioned on the age group.

\subsection{Expected Value}\label{sec:expected_value}

The expected value, or \textit{mean}, of a random variable provides a measure of the central tendency of the distribution of the random variable. It represents the long-term average outcome of the random variable when the process is repeated infinitely many times.

\subsubsection*{Discrete Random Variables}

For a discrete random variable \(X\), the expected value is defined as the weighted average of all possible values that \(X\) can take, with the weights being the probabilities of each value:
\begin{equation*}
	\mathbb{E}[X] = \sum_{x} x \cdot P(X = x).
\end{equation*}

For example, let us consider a random variable \( X \) that can take the values \( X = 1 \) and \( X = 2 \) with probabilities of 70\% and 30\%, respectively. In this case, the expected value of \( X \) is calculated as \( 0.7 \cdot 1 + 0.3 \cdot 2 = 1.3 \). This means that, on average, the values that \( X \) takes in repeated random experiments will converge to 1.3 over time. While \( X \) may never actually take the value 1.3, this is the average outcome expected in the long run.

\subsubsection*{Continuous Random Variables}

For a continuous random variable \(X\), the expected value is calculated using the probability density function, \(f_X(x)\), which describes the distribution of \(X\) (see Section~\ref{sec:cont_rand_var}). In this case the summation used for discrete random variables is replaced by an integral because the possible values of \(X\) are spread over a continuous range ($x$ represents specific realizations (values) of the random variable
$X$):
\begin{equation*}
	\mathbb{E}[X] = \int_{-\infty}^{\infty} x \cdot f_X(x) \, dx.
\end{equation*}

\subsubsection*{Sampling-Based Estimates}
\label{sec:sampling-based}

When the probabilities of each outcome are unknown, taking $n$ independent samples from the underlying distribution allows the sample mean to serve as an approximation to the distribution's true expected value. Assuming that the samples are obtained under the same distribution (i.e., its statistical properties are invariant over time), the Law of Large Numbers provides us with the guarantee that the sample mean will converge to the mean of that distribution with probability $1$ as $n$ approaches infinity.

Formally, let $X_1, X_2, \ldots$ be a sequence of independent random variables  having a common distribution with mean $\mu$, $\mathbb{E}[X_i]=\mu$. Each of these random variables takes a specific value when sampled from the underlying distribution. The law of large numbers tells us that, with probability 1,
\[
	\frac{X_1+X_2+\ldots+X_n}{n} \rightarrow \mu \quad \text{as} \quad n\rightarrow \infty
\]

Let us analyze a simple example. Consider rolling a fair six-sided die with outcomes \(\{1, 2, 3, 4, 5, 6\}\), each having an equal probability of \(\frac{1}{6}\). Given that the outcomes probabilities are known, the true expected value can be calculated: $\mathbb{E}(X) = \sum_{i=1}^{6} x_i P(X = x_i) = \frac{1}{6} (1 + 2 + 3 + 4 + 5 + 6) = 3.5$. Now let us suppose we do not have access to the actual probabilities but we are able to draw the die $n=10$ times, that is, to take 10 samples. Suppose we get the outcomes \{3, 5, 2, 6, 4, 1, 5, 6, 3, 4\}. The sample mean is $(3 + 5 + 2 + 6 + 4 + 1 + 5 + 6 + 3 + 4)/10 = 3.9$, which is roughly the true expected value of 3.5. As $n$ increases, the sample mean approaches the true expected value of 3.5.

\subsubsection*{Properties}

Expectation is a linear operator and, thus, satisfies the properties of linearity. Concretely, if $X$ and $Y$ are random variables with finite expected values and $a$ is a constant, then
\[
	\mathbb{E}[X + Y] = \mathbb{E}[X] + \mathbb{E}[Y],
\]
\[
	\mathbb{E}[aX] = a\mathbb{E}[X].
\]

The linearity property extends to multiple random variables $X_1, X_2, \ldots, X_n$:
\[
	\mathbb{E}\left[\sum_{i=1}^{n} X_i\right] = \sum_{i=1}^{n} \mathbb{E}[X_i].
\]

The variance of a random variable measures the spread or dispersion of its possible values around the mean (expected value). Specifically, it measures the squared expected deviation from the mean:
\[
	\text{Var}(X)=\mathbb{E}[(X-\mathbb{E}[X])^2]=\mathbb{E}[X^2] - \mathbb{E}[X]^2.
\]

\section{Information Theory}\label{sec:info-entropy}

When analyzing random variables, we are often interested in quantifying uncertainty, that is, how unpredictable the outcomes of the variable are. This uncertainty depends on how the probabilities are distributed across all possible outcomes. If the probabilities are evenly spread, uncertainty is high because every outcome is equally likely. Conversely, when the probabilities are concentrated around a few outcomes, uncertainty is lower because the results are more predictable.

\subsection{Surprise / Information Content}

Consider a scenario where the daily temperature in a specific city is measured over time. If the temperature consistently stays around $25^\circ\mathrm{C}$, the outcomes are predictable, and uncertainty is low. However, if temperatures range widely, from $15^\circ\mathrm{C}$ to $40^\circ\mathrm{C}$, with varying probabilities, uncertainty is higher because the potential outcomes are less predictable. To quantify this uncertainty, it is necessary to understand \emph{surprise} and \emph{information content}.

Surprise reflects how unexpected an observed outcome is. For example, if temperatures in the city typically range between $20^\circ\mathrm{C}$ and $30^\circ\mathrm{C}$, observing a value like $25^\circ\mathrm{C}$ is unsurprising because it aligns with our expectations. On the other hand, observing a rare spike of $40^\circ\mathrm{C}$ is highly surprising. The more surprising an outcome, the more \emph{information content} that outcome conveys regarding the underlying probability distribution. For instance, repeatedly observing $25^\circ\mathrm{C}$ adds little new information because it merely confirms our expectations. Conversely, observing $40^\circ\mathrm{C}$ significantly updates our understanding of temperature variability, suggesting that previous observations in the $20^\circ\mathrm{C}$ and $30^\circ\mathrm{C}$ range may have been an unrepresentative or biased sample.

The idea of less probable events being more surprising and bearing more information content is captured by the following expression (recall that $\ln(a/b)=\ln(a)-\ln(b))$:
\[
	I(x) \doteq \ln\left(\frac{1}{P(X = x)}\right) = \ln(1)-\ln(P(X = x))=-\ln(P(X = x)),
\]

\noindent where $x$ is the observed value taken by random variable $X$. Intuitively, this expression states that the higher the probability of observing $x$ the lower its information content. The logarithm non-linearly squashes information content as the probability decreases, reflecting the fact that extremely rare events do not carry unreasonable amount of information about the probability distribution. Note that $I(x)$ is not defined for non-occurring events (null probability).\\

\noindent \textbf{Example.} Suppose the probabilities of two temperature events are \( P(T = 25^\circ\mathrm{C}) = 0.8 \) (highly probable) and \( P(T = 40^\circ\mathrm{C}) = 0.01 \) (unlikely). The information content \( I(T) \) for an observed temperature \( T = t \) is given by \( I(t) = -\ln(P(T = t)) \). For the highly probable event \( T = 25^\circ\mathrm{C} \), the information content is \( I(25) = -\ln(0.8) \approx 0.22 \), reflecting low surprise. For the unlikely event \( T = 40^\circ\mathrm{C} \), the information content is \( I(40) = -\ln(0.01) \approx 4.61 \), indicating high surprise. This demonstrates that observing rare events provides significantly more information compared to common ones.

\subsection{Entropy of a Discrete Random Variable}

The concept of \emph{entropy} captures the overall uncertainty of a random variable, that is, the average surprise or information content over all possible outcomes. Formally, the entropy of a \textit{discrete} random variable $X$ is defined as:
\begin{equation}\label{equ:entropy_discrete}
	\mathcal{H}(X) = \mathbb{E}[I(x)] = \mathbb{E}[-\ln(P(X = x))]=-\sum_{x} P(X = x) \ln(P(X = x)).
\end{equation}

\noindent \textbf{Example.} Suppose the discrete random variable \( T_1 \) represents the temperature with three possible outcomes: \(20^\circ\mathrm{C} \), \(25^\circ\mathrm{C} \), and \(30^\circ\mathrm{C} \), having probabilities \( P(T_1 = 20) = 0.2 \), \( P(T_1 = 25) = 0.5 \), and \( P(T_1 = 30) = 0.3 \). The entropy \( \mathcal{H}(T_1) \) is given by \( \mathcal{H}(T_1) = -\sum_t P(T_1 = t) \ln(P(T_1 = t)) \). Substituting the probabilities we obtain: $\mathcal{H}(T_1) = -(0.2 \ln(0.2) + 0.5 \ln(0.5) + 0.3 \ln(0.3))  \approx 1.03$. Now let us apply the same reasoning to a discrete random variable $T_2$ with much less spread:  \( P(T_2 = 20) = 0.8 \), \( P(T_2 = 25) = 0.1 \), and \( P(T_2 = 30) = 0.1 \). In this case, the entropy is $\mathcal{H}(T_2) = -(0.8 \ln(0.8) + 0.1 \ln(0.1) + 0.1 \ln(0.1))  \approx 0.64$. These examples show that the entropy is higher for spreader probability distributions (1.03 vs. 0.64).

\subsection{Entropy of a Continuous Random Variable}

The summation in Equation~\ref{equ:entropy_discrete} can be substituted by an integral to compute the entropy of a \textit{continuous} random variable $X$ with PDF denoted by $f_X(x)$:
\begin{equation}\label{equ:entropy}
	\mathcal{H}(X) = -\int_{-\infty}^\infty f_X(x) \ln(f_X(x)) \, dx.
\end{equation}

The entropy of a Gaussian (normal) distribution $X$ with mean $\mu$ and standard deviation $\sigma$ can be computed by solving the integral from Equation~\ref{equ:entropy} with the Gaussian PDF (Equation~\ref{equ:gaussian}), resulting in:
\begin{equation}\label{equ:entropy_gaussian}
	\mathcal{H}(X) = \frac{1}{2}+\frac{1}{2} \ln(2\pi) + \ln(\sigma).
\end{equation}

Note that the entropy of a Gaussian distribution depends on its standard deviation $\sigma$. A higher standard deviation (i.e., greater spread) leads to higher entropy, reflecting greater uncertainty in the distribution.\\

\noindent \textbf{Example.} Consider two temperature distributions: one with a mean of \( 25^\circ\mathrm{C} \) and standard deviation \( \sigma_1 = 3 \), and another with a mean of \( 30^\circ\mathrm{C} \) and standard deviation \( \sigma_2 = 5 \). For the first distribution, substituting \( \sigma_1 = 3 \) in Equation~\ref{equ:entropy_gaussian}, we get \( \mathcal{H}(T_1) = \frac{1}{2} + \frac{1}{2} \ln(2\pi) + \ln(3) \approx 2.52 \). For the second distribution, substituting \( \sigma_2 = 5 \), we get \( \mathcal{H}(T_2) = \frac{1}{2} + \frac{1}{2} \ln(2\pi) + \ln(5) \approx 3.03 \). The entropy of the second distribution is higher, reflecting greater uncertainty due to its larger standard deviation.

\subsection{Kullback-Leibler (KL) Divergence}\label{sec:kl-divergence}

Kullback-Leibler (KL) divergence measures the difference between the probability distribution \( P_X \) of a random variable \( X \) and an approximation \( Q_X \), which is also a probability distribution over \( X \). It quantifies how much information is lost when \( Q_X \) is used to approximate \( P_X \). A smaller KL divergence indicates that \( Q_X \) is more similar to \( P_X \), and thus a better approximation.

\subsubsection{Discrete Random Variables}

For a discrete random variable \( X \) with probability distributions \( P_X \) and \( Q_X \), the KL divergence is defined as:
\[
	D_{KL}(P_X, Q_X) = \sum_{x} P_X(X = x) \left( \ln(P_X(X = x)) - \ln(Q_X(X = x)) \right),
\]

\noindent where \( P_X(X = x) \) and \( Q_X(X = x) \) represent the probabilities of \( X \) taking the value \( x \) under probability distributions \( P_X \) and \( Q_X \), respectively. The term \( \ln(P_X(X = x)) - \ln(Q_X(X = x)) \) represents the difference in information content for the outcome \( x \) under the two distributions. Hence, the KL divergence is the \textit{expected value of this difference}, weighted by \( P_X(X = x) \), which represents the actual proportion of realizations of \( x \) under the true distribution.

Recalling that \( \ln(a/b) = \ln(a) - \ln(b) \), the KL divergence can be rewritten in its most commonly used formulation:
\begin{equation}\label{equ:KL-div-discrete}
	\begin{split}
		D_{KL}(P_X, Q_X)
		&= \mathbb{E}_{x\sim P_X}\!\left[
			\ln\!\left(\frac{P_X(X=x)}{Q_X(X=x)}\right)
			\right] \\
		&= \sum_{x} P_X(X=x)\,
		\ln\!\left(\frac{P_X(X=x)}{Q_X(X=x)}\right).
	\end{split}
\end{equation}

\noindent \textbf{Example.} Consider a discrete random variable \( X \) representing weather conditions with the outcomes \{Sunny, Cloudy, Rainy\}. The true distribution \( P_X \) is given as \( P_X(X = \text{Sunny}) = 0.7 \), \( P_X(X = \text{Cloudy}) = 0.2 \), \( P_X(X = \text{Rainy}) = 0.1 \). Two approximate distributions are considered: \( Q_{X,1} \) with \( Q_{X,1}(X = \text{Sunny}) = 0.5 \), \( Q_{X,1}(X = \text{Cloudy}) = 0.3 \), \( Q_{X,1}(X = \text{Rainy}) = 0.2 \), and \( Q_{X,2} \) with \( Q_{X,2}(X = \text{Sunny}) = 0.4 \), \( Q_{X,2}(X = \text{Cloudy}) = 0.4 \), \( Q_{X,2}(X = \text{Rainy}) = 0.2 \). Using the KL divergence formula, we compute for \( Q_{X,1} \): $0.7 \ln\left(\frac{0.7}{0.5}\right) \approx 0.2355, \quad 0.2 \ln\left(\frac{0.2}{0.3}\right) \approx -0.0810, \quad 0.1 \ln\left(\frac{0.1}{0.2}\right) \approx -0.0693$. Summing these, we get: $D_{KL}(P_X \| Q_{X,1}) \approx 0.2355 - 0.0810 - 0.0693 = 0.0852$. For \( Q_{X,2} \): $0.7 \ln\left(\frac{0.7}{0.4}\right) \approx 0.3917, \quad 0.2 \ln\left(\frac{0.2}{0.4}\right) \approx -0.1386, \quad 0.1 \ln\left(\frac{0.1}{0.2}\right) \approx -0.0693$. Summing these, we get: $D_{KL}(P_X \| Q_{X,2}) \approx 0.3917 - 0.1386 - 0.0693 = 0.1838$. The comparison shows that \( Q_{X,2} \), with higher divergence, is a poorer approximation of \( P_X \) than \( Q_{X,1} \).

\subsubsection{Continuous Random Variables}

To handle a continuous random variable \( X \), the KL divergence is computed by replacing the summation with an integral and using \( p_X(X = x) \) and \( q_X(X = x) \) as PDFs of \( P_X \) and \( Q_X \), respectively:
\begin{equation}\label{equ:kl__}
	D_{KL}(P_X, Q_X) = \int_{-\infty}^\infty p_X(X = x) \ln\left(\frac{p_X(X = x)}{q_X(X = x)}\right) dx.
\end{equation}

If the probability distributions \( P_X \) and \( Q_X \) are Gaussian with means (\( \mu_P, \mu_Q \)) and standard deviations (\( \sigma_P, \sigma_Q \)), the KL divergence is derived by solving the integral from Equation~\ref{equ:kl__} with the Gaussian PDF (Equation~\ref{equ:gaussian}), resulting in the following expression:
\begin{equation}\label{equ:kl-continuous}
	D_{KL}(P_X, Q_X) = \ln\left(\frac{\sigma_Q}{\sigma_P}\right) + \frac{\sigma_P^2 + (\mu_P - \mu_Q)^2}{2\sigma_Q^2} - \frac{1}{2}.
\end{equation}

This formula captures contributions from both the difference in means (\( \mu_P, \mu_Q \)) and standard deviations (\( \sigma_P, \sigma_Q \)). If the distributions have identical means and variances, the KL divergence is zero, reflecting no difference between the two distributions.\\

\noindent \textbf{Example.} Consider a continuous random variable \( X \) following the true Gaussian distribution \( P_X \) with \( \mu_P = 0 \) and \( \sigma_P = 1 \). Two approximate Gaussian distributions are considered: \( Q_{X,1} \) with \( \mu_{Q_1} = 1 \) and \( \sigma_{Q_1} = 1.8 \), and \( Q_{X,2} \) with \( \mu_{Q_2} = 0.5 \) and \( \sigma_{Q_2} = 1.3 \). For \( Q_{X,1} \): $D_{KL}(P_X, Q_{X,1}) = \ln\left(\frac{1.8}{1}\right) + \frac{1 + (0 - 1)^2}{2 \cdot 1.8^2} - \frac{1}{2} \approx 0.3964$. For \( Q_{X,2} \): $D_{KL}(P_X, Q_{X,2}) = \ln\left(\frac{1.3}{1}\right) + \frac{1 + (0 - 0.5)^2}{2 \cdot 1.3^2} - \frac{1}{2} \approx 0.1322$. Comparing the divergences, \( Q_{X,2} \), with a smaller KL divergence (0.1322 vs 0.3964), is a closer approximation to \( P_X \) than \( Q_{X,1} \).

\section{Chain Rule of Calculus}

The derivative of a function measures how its output changes in response to a small change in its input, that is, it quantifies the rate at which the function grows or decreases as the variable changes. The derivative of the composition of two functions can be found using the chain rule of calculus, which systematically accounts for how changes in the inner function propagate through the outer function. This reasoning extends directly to partial derivatives and gradients, since a gradient is a vector composed of partial derivatives, capturing the rates of change with respect to each input dimension.

Given a function \( y = f(u) \) and another function \( u = g(x) \), that is, $y=f(g(x))$, the derivative of \( y \) with respect to \( x \) according to the chain rule is:
\[
	\frac{d y}{d x} = \frac{d y}{d u} \frac{d u}{d x}.
\]

The chain rule can be applied recursively. For instance, in the case of three composite functions \( y = f(u), u=g(v), v=h(x) \), that is, $y=f(g(h(x)))$, the chain rule applies as follows:
\[
	\frac{d y}{d x} = \frac{d y}{d u} \frac{d u}{d v} \frac{d v}{d x}.
\]

Let us analyze the example of a quadratic function of a single variable $x$:
\[
	y=(3+4x)^2.
\]

This function can be decomposed as the composite of the following two functions:
\begin{align*}
	y & =f(u)=u^2   \\
	u & =g(x)=3+4x.
\end{align*}

The derivatives of these functions are:
\begin{align*}
	\frac{dy}{du} & =f'(u)=2u=2(3+4x)=6+8x \\
	\frac{du}{dx} & =g'(x)=4.
\end{align*}

The chain rule can be  applied  to obtain the derivative of $y$ with respect to $x$:
\[
	\frac{d y}{d x} = \frac{d y}{d u}\cdot\frac{d u}{d x}=(6+8x)\cdot 4=24+32x.
\]

Let us analyze another example, in this case a quadratic function of two variables $x$ and $z$:
\[
	y=(3+4x+2z)^2.
\]

As before, this function can be decomposed as the composite of the following two functions:
\begin{align*}
	y & =f(u)=u^2       \\
	u & =g(x,z)=3+4x+2z
\end{align*}

The next step is to compute the derivatives of these functions. However, since $g$ has two variables, its \textit{partial derivatives} with respect to each variable are computed instead:
\begin{align*}
	\frac{d y}{d u} & =f'(u)=2u=2(3+4x+2z)=6+8x+4z                      \\
	\frac{d u}{d x} & =g_x'(x, z)=\frac{\partial g}{\partial x}(x,z)=4  \\
	\frac{d u}{d z} & =g_z'(x, z)=\frac{\partial g}{\partial z}(x,z)=2.
\end{align*}

Finally, by applying the chain rule, it is possible to obtain the partial derivatives of $y$ with respect to $x$ and $z$:
\begin{align*}
	\frac{d y}{d x} & = \frac{d y}{d u}\cdot\frac{d u}{d x}=(6+8x+4z)\cdot 4=32x+16z+24. \\
	\frac{d y}{d z} & = \frac{d y}{d u}\cdot\frac{d u}{d z}=(6+8x+4z)\cdot 2=16x+8z+12.
\end{align*}

\ifincludeexercises
\section{Test Your Knowledge}

To consolidate your understanding, work through the exercises in Section~\ref{sec:exercises} on your own. After completing each exercise, review the step-by-step solution in Section~\ref{sec:solutions} to confirm your reasoning. Avoid consulting the detailed solutions before obtaining an answer yourself to ensure stronger problem-solving skills and deeper learning.

\subsection{Exercises}\label{sec:exercises}

\noindent \textbf{Exercise 1.} A robot explores a 2D grid. It moves twice, each move being either \textit{Right} (R) or \textit{Up} (U). The probability of moving \textit{Right} is \( \frac{2}{3} \).
\begin{itemize}
	\item[a.] Specify the sample space of possible paths.
	\item[b.] Determine the probability of moving \textit{Up} twice.
	\item[c.] Determine the probability of making exactly one \textit{Right} move.
\end{itemize}

\noindent \textbf{Exercise 2.}  A robot navigates a corridor and scans its surroundings twice. Each scan can detect either an \textit{obstacle} or \textit{no obstacle}. The probability of detecting an obstacle is \(0.6\).
\begin{itemize}
	\item[a.] Specify the sample space of possible scan outcomes.
	\item[b.] Specify random variable \( X \) representing the number of obstacles detected.
	\item[c.] Determine the probability of detecting 0, 1, or 2 obstacles.
\end{itemize}

\noindent \textbf{Exercise 3.} A robot is performing tasks with duration \( T \) according to an exponential distribution with PDF  $f_T(t) = e^{-t}, \quad \text{for } t \geq 0$.
\begin{itemize}
	\item[a.] Calculate the probability that \( T \) is between 0.5 s and 1.5 s.
\end{itemize}

\noindent \textbf{Exercise 4.} A robot's sensor measures distances with noise modeled by a Gaussian distribution \( X \sim \mathcal{N}(10, 1^2) \). Determine which of the following sample sets is most plausible given this distribution:
\begin{itemize}
	\item[a.] \{12.5, 9.5, 10.7, 15.4, 9.3\} ~ c. \{9.0, 5.2, 8.8, 2.1, 8.9\}
	\item[b.] \{15.5, 28.5, 12.0, 1.8, 9.0\} ~~ d. \{10.0, 10.1, 9.9, 10.2, 9.8\}
\end{itemize}

\noindent \textbf{Exercise 5.} A robot has 3 independent sensors, each with a probability of failing to detect obstacles: {\small\(P(X_1 = \text{fail}) = 10^{-5}, P(X_2 = \text{fail}) = 10^{-6}, P(X_3 = \text{fail}) = 10^{-4}\)}.
\begin{itemize}
	\item[a.] Determine the joint probability that all 3 sensors fail.
\end{itemize}

\noindent \textbf{Exercise 6.} The terrain type (T) and robot available energy (E) were collected over multiple missions, resulting in the joint probabilities:  $P(\text{T=flat, E=high}) = 0.3$, $P(\text{T=flat, E=low}) = 0.1$, $P(\text{T=sloped, E=high}) = 0.2$, $P(\text{T=sloped, E=low}) = 0.4$.
\begin{itemize}
	\item[a.] How probable was to find the robot on flat terrain?
	\item[b.] How probable was to find the robot with high energy?
\end{itemize}

\noindent \textbf{Exercise 7.} A robot moves successfully (S) depending on its battery (B) level with joint probabilities \(P(\text{S=yes, B=high}) = 0.6, P(\text{S=yes, B=low}) = 0.1\) and marginal probabilities \(P(\text{B=high}) = 0.8, P(\text{B=low}) = 0.2\).
\begin{itemize}
	\item[a.] What is the probability of success given it has a high battery?
	\item[b.] What is the probability of success given it has a low battery?
\end{itemize}

\noindent \textbf{Exercise 8.}  A robot has 3 sensors \( S_1, S_2, S_3 \) that detect obstacles according to the following probabilities: \( P(S_1 = \text{detect}) = 0.8 \), \( P(S_2 = \text{detect} \mid S_1 = \text{detect}) = 0.7 \), \( P(S_3 = \text{detect} \mid S_1 = \text{detect}, S_2 = \text{detect}) = 0.6 \).
\begin{itemize}
	\item[a.] Compute the joint probability \( P(S_1 = \text{detect}, S_2 = \text{detect}, S_3 = \text{detect}) \).
\end{itemize}

\noindent \textbf{Exercise 9.} A robot has two sensors \( S_1 \) and \( S_2 \) that detect obstacles according to the following probabilities: \( P(S_1 = \text{detect}) = 0.6 \), \( P(S_2 = \text{detect}) = 0.7 \), \( P(S_1 = \text{detect}, S_2 = \text{detect}) = 0.5 \).
\begin{itemize}
	\item[a.] Check if \( S_1 \) and \( S_2 \) are independent.
\end{itemize}

\noindent \textbf{Exercise 10.} A robot completes tasks within time \( T \) (in hours) with the following probabilities: $P(T = 1) = 0.5$, $P(T = 2) = 0.3$, $P(T = 3) = 0.2$.
\begin{itemize}
	\item[a.] Compute the mean task completion time.
\end{itemize}

\noindent \textbf{Exercise 11.} A robot uses a sensor to detect obstacles (O) with probabilities: $P(O=no)=0.8, P(O=yes)=0.2$.
\begin{itemize}
	\item[a.] Calculate the information content of detecting an obstacle.
	\item[b.] Compare with the information content of not detecting an obstacle.
\end{itemize}

\noindent \textbf{Exercise 12.} A robot measures  distances polluted with Gaussian noise in Environment 1 with \( \sigma_1 = 2 \)  and in Environment 2  with \( \sigma_2 = 4 \).
\begin{itemize}
	\item[a.] Compute the entropy of the sensor noise in Environment 1.
	\item[b.] Compare with the entropy of the sensor noise in Environment 2.
\end{itemize}

\noindent \textbf{Exercise 13.} A robot includes two models to predict the presence of an obstacle (O).  True environment probabilities are: \(P(O=yes) = 0.7, P(O=no) = 0.3\). Model 1 probabilities: \(Q_1(O=yes) = 0.6, Q_1(O=no) = 0.4\). Model 2 probabilities: \(Q_2(O=yes) = 0.5, Q_2(O=no) = 0.5\).
\begin{itemize}
	\item[a.] Which model better approximates the true probability distribution?
\end{itemize}

\noindent \textbf{Exercise 14.} A  sensor provides raw uncalibrated values \( x \), which are calibrated according to the function:  \(y = (5 + 3x)^3\).
\begin{itemize}
	\item[a.] Determine how much a small variation in $x$ impacts $y$.
\end{itemize}

\noindent \textbf{Exercise 15.} A robot estimates environmental variable $y$ based on data from sensors $x$ and $z$ according to the function: \(y = (2 + 3x + 5z)^2\).
\begin{itemize}
	\item[a.] Variable $y$ is more sensible to $x$ or $z$?
\end{itemize}

\subsection{Step-by-Step Solutions}\label{sec:solutions}

\noindent \textbf{Solution to Exercise 1.}
\begin{itemize}
	\item[a.] The robot makes two moves, and each move can be either \textit{Right} (R) or \textit{Up} (U). The possible paths are:
		\[
			\Omega = \{\text{RR}, \ \text{RU}, \ \text{UR}, \ \text{UU}\}
		\]

	\item[b.] The probability of moving \textit{Up} in any step is \( 1 - \frac{2}{3} = \frac{1}{3} \).
		Since the moves are independent, the probability of taking the path \textit{UU} is:
		\[
			P(\text{UU}) = \left( \frac{1}{3} \right) \times \left( \frac{1}{3} \right) = \frac{1}{9}
		\]

	\item[c.] Making exactly one \textit{Right} move corresponds to the paths \textit{RU} and \textit{UR}. We calculate their probabilities:
		\[
			P(\text{RU}) = \left( \frac{2}{3} \right) \times \left( \frac{1}{3} \right) = \frac{2}{9}
		\]
		\[
			P(\text{UR}) = \left( \frac{1}{3} \right) \times \left( \frac{2}{3} \right) = \frac{2}{9}
		\]
		Therefore, the total probability of making exactly one \textit{Right} move is:
		\[
			P(\{RU, UR\}) = P(\text{RU}) + P(\text{UR}) = \frac{2}{9} + \frac{2}{9} = \frac{4}{9}
		\]
\end{itemize}

\noindent \textbf{Solution to Exercise 2.}
\begin{itemize}
	\item[a.] Each scan can either detect an \textit{obstacle} (O) or \textit{no obstacle} (N). The possible outcomes for the two scans are:
		\[
			\Omega = \{\text{OO}, \ \text{ON}, \ \text{NO}, \ \text{NN}\}.
		\]

	\item[b.] \( X \) as the number of obstacles detected in the two scans:
		\[
			X(\text{OO}) = 2, \quad X(\text{ON}) = 1, \quad X(\text{NO}) = 1, \quad X(\text{NN}) = 0.
		\]

	\item[c.] Let \( p = 0.6 \) be the probability of detecting an obstacle and \( 1 - p = 0.4 \) the probability of detecting no obstacle.

		Probability of detecting 0 obstacles, which corresponds to outcome \textit{NN}:
		\[
			P(X = 0) = (0.4) \times (0.4) = 0.16.
		\]

		Probability of detecting 1 obstacle, which corresponds to outcomes \textit{ON} and \textit{NO}:
		\[
			P(X = 1) = P(\text{ON}) + P(\text{NO}),
		\]
		\[
			P(\text{ON}) = (0.6) \times (0.4) = 0.24,
		\]
		\[
			P(\text{NO}) = (0.4) \times (0.6) = 0.24,
		\]
		\[
			P(X = 1) = 0.24 + 0.24 = 0.48.
		\]

		Probability of detecting 2 obstacles, which corresponds to outcome \textit{OO}:
		\[
			P(X = 2) = (0.6) \times (0.6) = 0.36.
		\]

		Therefore, the probabilities of detecting 0, 1, or 2 obstacles are:
		\[
			P(X = 0) = 0.16, \quad P(X = 1) = 0.48, \quad P(X = 2) = 0.36.
		\]

\end{itemize}

\noindent \textbf{Solution to Exercise 3.}

\begin{itemize}
	\item[a.] The probability that \( T \) lies between \( 0.5 \) s and \( 1.5 \) s is given by:
		\[
			P(0.5 \leq T \leq 1.5) = \int_{0.5}^{1.5} f_T(t) \, dt = \int_{0.5}^{1.5} e^{-t} \, dt = \left[ -e^{-t} \right]_{0.5}^{1.5}
		\]

		Substituting the limits:
		\[
			P(0.5 \leq T \leq 1.5) = -e^{-1.5} + e^{-0.5} \approx 0.3834
		\]

\end{itemize}

\noindent \textbf{Solution to Exercise 4.} \\

\noindent The most plausible sample set is option (d), as it contains values that are closely centered around the mean ($\mu=10$) and fall well within the expected range for a normal distribution with \( \sigma = 1 \).\\

\noindent \textbf{Solution to Exercise 5.}

\begin{itemize}

	\item[a.] As sensors failures are independent, their joint probability of failing is:
		\begin{align*}
			P(X_1 = \text{fail}, X_2 = \text{fail}, X_3 = \text{fail}) & = P(X_1 = \text{fail}) \times P(X_2 = \text{fail}) \times P(X_3 = \text{fail}) \\
			                                                           & = (10^{-5}) \times (10^{-6}) \times (10^{-4})                                  \\
			                                                           & = 10^{-15}
		\end{align*}

\end{itemize}

\noindent \textbf{Solution to Exercise 6.}

\begin{itemize}

	\item[a.] The marginal probability is obtained by summing the joint probabilities involving flat terrain:
		\[
			P(T=\text{flat}) = P(\text{T=flat, E=high}) + P(\text{T=flat, E=low}) = 0.3 + 0.1 = 0.4.
		\]

	\item[b.] The marginal probability is obtained by summing the joint probabilities involving high-energy:
		\[
			P(\text{E=high}) = P(\text{T=flat, E=high}) + P(\text{T=sloped, E=high}) = 0.3 + 0.2 = 0.5.
		\]

\end{itemize}
\newpage
\noindent \textbf{Solution to Exercise 7.}
\begin{itemize}

	\item[a.] The conditional probability is:
		\[
			P(\text{S=yes} \mid \text{B=high}) = \frac{P(\text{S=yes, B=high})}{P(\text{B=high})} = \frac{0.6}{0.8} = 0.75.
		\]

	\item[b.] The conditional probability is:
		\[
			P(\text{S=yes} \mid \text{B=low}) = \frac{P(\text{S=yes, B=low})}{P(\text{B=low})} = \frac{0.1}{0.2} = 0.5.
		\]

\end{itemize}

\noindent \textbf{Solution to Exercise 8.}

\begin{itemize}

	\item[a.] The joint probability \( P(S_1 = \text{detect}, S_2 = \text{detect}, S_3 = \text{detect}) \) is calculated using the chain rule of probability:
		\begin{align*}
			P(S_1 = \text{detect}, & S_2 = \text{detect}, S_3 = \text{detect}) =                                                              \\
			                       & P(S_1 = \text{detect}) \times                                                                            \\
			                       & P(S_2 = \text{detect} \mid S_1 = \text{detect}) \times                                                   \\
			                       & P(S_3 = \text{detect} \mid S_1 = \text{detect}, S_2 = \text{detect}) = 0.8 \times 0.7 \times 0.6 = 0.336
		\end{align*}

\end{itemize}

\noindent \textbf{Solution to Exercise 9.}

\begin{itemize}

	\item[a.] Two events \( S_1 \) and \( S_2 \) are independent if:
		\[
			P(S_1 = \text{detect}, S_2 = \text{detect}) = P(S_1 = \text{detect}) \times P(S_2 = \text{detect}).
		\]

		In our case:
		\[
			P(S_1 = \text{detect}) \times P(S_2 = \text{detect}) = 0.6 \times 0.7 = 0.42,
		\]
		\[
			P(S_1 = \text{detect}, S_2 = \text{detect}) = 0.5.
		\]

		Since $0.42 \neq 0.5$, the events \( S_1 \) and \( S_2 \) are not independent. In other words, observing the outcome of one sensor provides information about the likelihood of detection by the other, such as when their detection regions overlap.

\end{itemize}

\noindent \textbf{Solution to Exercise 10.}

\begin{itemize}

	\item[a.] The mean task execution time is the expected value:
		\[
			\mathbb{E}[T] = 1 \times P(T = 1) + 2 \times P(T = 2) + 3 \times P(T = 3) = 1 \times 0.5 + 2 \times 0.3 + 3 \times 0.2 = 1.7 \text{ hours}.
		\]

\end{itemize}

\noindent \textbf{Solution to Exercise 11.}

\begin{itemize}

	\item[a.] The information content for detecting an obstacle is:
		\[
			I(O=yes) = -\ln P(O=yes) = -\ln (0.2) \approx 1.61.
		\]

	\item[b.] The information content for not detecting an obstacle is:
		\[
			I(O=no) = -\ln P(O=no) = -\ln (0.8) \approx 0.22
		\]

		The information content for detecting an obstacle is significantly higher than for not detecting. This means that detecting an obstacle is far more surprising, as it is less likely, providing more information when it occurs.

\end{itemize}

\noindent \textbf{Solution to Exercise 12.}

\begin{itemize}

	\item[a.] The entropy for Environment 1 with \( \sigma_1 = 2 \):
		\[
			\mathcal{H}_1 = \frac{1}{2}+\frac{1}{2} \ln(2\pi)+\ln(2) \approx 2.11.
		\]

	\item[b.] The entropy for Environment 2 with \( \sigma_2 = 4 \):
		\[
			\mathcal{H}_2 = \frac{1}{2}+\frac{1}{2} \ln(2\pi)+\ln(4) \approx 2.81.
		\]

		The entropy in Environment 2 is higher than in Environment 1 \( (2.81 > 2.11) \), meaning that in Environment 2, the robot's sensor readings are less reliable (higher variability).

\end{itemize}

\noindent \textbf{Solution to Exercise 13.}

\begin{itemize}

	\item[a.] The KL divergence between the true distribution $P$ and model $Q_1$ is:
		\begin{align*}
			D_{\text{KL}}(P, Q_1) & =  P(O=yes) \ln \frac{P(O=yes)}{Q_1(O=yes)} + P(O=no) \ln \frac{P(O=no)}{Q_1(O=no)} \\ &=0.7 \ln \frac{0.7}{0.6} + 0.3 \ln \frac{0.3}{0.4} \approx 0.022.
		\end{align*}

		The KL divergence between the true distribution $P$ and model $Q_1$ is:
		\begin{align*}
			D_{\text{KL}}(P, Q_2) & =  P(O=yes) \ln \frac{P(O=yes)}{Q_2(O=yes)} + P(O=no) \ln \frac{P(O=no)}{Q_2(O=no)} \\ &=0.7 \ln \frac{0.7}{0.5} + 0.3 \ln \frac{0.3}{0.5} \approx 0.082.
		\end{align*}

		Since \( D_{\text{KL}}(P, Q_1) \) is smaller than \( D_{\text{KL}}(P, Q_2)\), $0.022 < 0.082$, model $Q_1$ better approximates the true distribution $P$.

\end{itemize}

\noindent \textbf{Solution to Exercise 14.}

\begin{itemize}

	\item[a.] The impact is calculated by applying the chain rule:
		\[
			\frac{dy}{dx} = 3(5 + 3x)^2 \cdot \frac{d}{dx}(5 + 3x)= 3(5 + 3x)^2 \cdot 3 = 9(5 + 3x)^2
		\]

		Therefore, a small variation in \( x \) results in a change in \( y \) at a rate proportional to \( 9(5 + 3x)^2 \).

\end{itemize}

\noindent \textbf{Solution to Exercise 15.}

\begin{itemize}

	\item[a.]  The partial derivative of \( y \) with respect to \( x \) is obtained with the chain rule:
		\[
			\frac{\partial y}{\partial x} = 2(2 + 3x + 5z) \cdot \frac{\partial}{\partial x}(2 + 3x + 5z)=2(2 + 3x + 5z)\cdot 3= 6(2 + 3x + 5z)
		\]

		The partial derivative of \( y \) with respect to \( z \) is obtained with the chain rule:
		\[
			\frac{\partial y}{\partial z} = 2(2 + 3x + 5z) \cdot \frac{\partial}{\partial z}(2 + 3x + 5z) = 2(2 + 3x + 5z) \cdot 5 = 10(2 + 3x + 5z)
		\]

		Given that $10>6$,  \( y \) is more sensitive to changes in \( z \) than in \( x \), that is, a small change in $z$ induces a larger change in $y$ than a small change in $x$.

\end{itemize}
\fi

\chapter{Markov Decision Processes}\label{cha:MDP}

Reinforcement learning builds upon the idea that it is possible to formulate goals in terms of reward signals, given that the agent then seeks to maximize the amount of reward it receives in the long term. This means that instead of explicitly specifying goals in the agent, the designer defines reward functions that, when maximized, induce the agent to accomplish the designer goals. This idea that goals can be defined in terms of reward signals is known as the \textit{reward hypothesis}.

As will be shown, rewards are scalar values that can be positive or negative. For example, in a maze, an agent could receive a reward of -1 per time step to encourage finding the shortest path. It is also possible to define sparse rewards, such as +100 for finding the goal and -10 for hitting an obstacle.

To earn rewards, an agent's "brain" must choose the best possible action at each step based on the current context. For example, an agent might reason: "I need to stop moving because I've encountered an obstacle; otherwise, I'll receive a -10 penalty." Markov Decision Processes (MDPs) provide a foundational framework for modeling such sequential decision-making tasks. They serve as an idealized mathematical model for reinforcement learning problems. In an MDP, the decision-maker is referred to as the agent, while everything external that the agent interacts with is collectively called the environment

The rest of this chapter introduces the MDP formalism and explains how it can be used to derive optimal agent behavior using exact solution methods. When applicable, this chapter adopts the notation and equations of Sutton and Barto \cite{sutton2018reinforcement}, ensuring coherence with the reference textbook in reinforcement learning.

\section{Definition}

MDPs assume that the environment can be fully described by its \textit{state}. A state is a snapshot of all the information the agent needs in order to decide which action to take. What counts as a state depends on the problem.
For example, in a navigation task, the state may include the agent's position and the positions of nearby obstacles. In robotic manipulation, the state might also contain the current joint angles and velocities of the robot's arm. In a 2D video-game setting, the state could be as raw as the RGB pixel values of the rendered image shown to the player.
All states the agent might ever encounter form the state space, denoted by $\mathcal{S}$.

MDPs also assume that the agent can choose \textit{actions} from a predefined set, denoted by $\mathcal{A}$. An action represents a decision the agent takes at a given state, and its form depends on the nature and level of control required by the task. Actions may be low-level, such as specifying torques or velocities for a robot's actuators, or high-level, such as issuing symbolic commands that trigger complex behaviors. In some settings, actions can even correspond to internal cognitive operations, like shifting the agent's attention within an input. The structure of the action space $\mathcal{A}$ can vary: (a) in discrete action spaces the agent is able to select from a finite set of distinct choices, such as $\{left, right, jump\}$; (b) in continuous action spaces actions are real-valued vectors, such as continuous torques applied to robot joints or steering angles in autonomous driving. Continuous actions allow for fine-grained control but typically require more sophisticated learning algorithms.

\begin{figure}[h]
	\begin{center}
		\includegraphics[width=8cm]{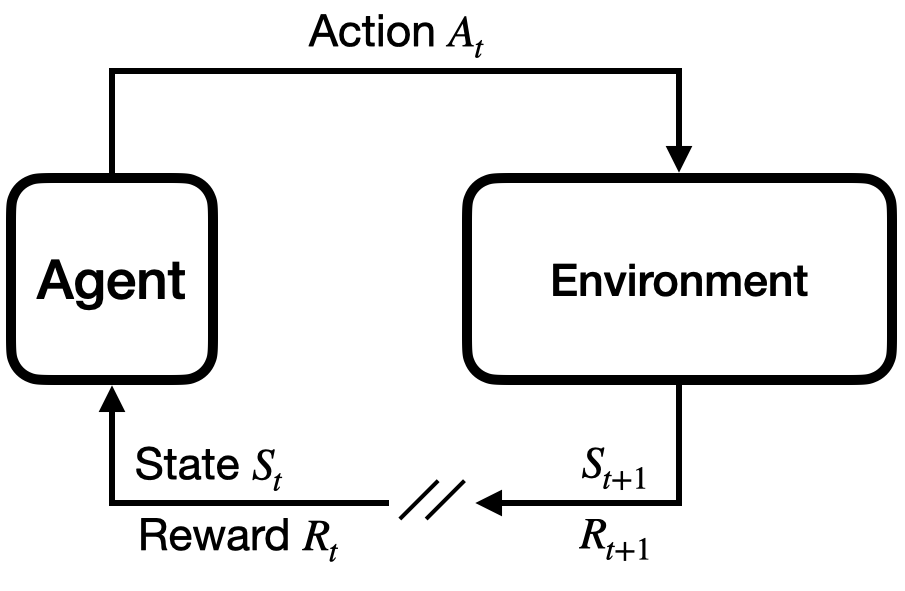}
		\caption{ The agent-environment interaction cycle in a MDP (adapted from \cite{sutton2018reinforcement}).}
		\label{fig:agent-env}
	\end{center}
\end{figure}

In an MDP, the agent and the environment interact in a continuous cycle. Concretely, at each discrete time step $t = 0, 1, 2, 3, \ldots$, the agent observes the environment's state, $S_t\in\mathcal{S}$, and based on this observation, selects an available action in the current state, $A_t\in\mathcal{A}$. After taking action $A_t$, at the next time step, the agent receives a numerical reward, $R_{t+1} \in \mathbb{R}$, and the environment transitions to a new state, $S_{t+1}\in\mathcal{S}$. This state transition can occur as a consequence of the agent's action or due to environment's intrinsic dynamics. Figure~\ref{fig:agent-env} illustrates the agent-environment interaction cycle in a MDP. State $S_t$, action $A_t$, and reward $R_{t}$ are random variables, reflecting the stochastic nature of the environment and agent-environment interactions.

The agent--environment interaction unfolds over time and gives rise to a \textit{history} or \textit{trajectory}, denoted by $\tau$. A trajectory is the sequence of states, actions, and rewards experienced by the agent as it interacts with the environment:
\begin{equation*}
	\tau \doteq (S_0, A_0, R_1, S_1, A_1, R_2, S_2, A_2, R_3, \ldots),
\end{equation*}

\noindent where $S_0$ is sampled from an initial state distribution, typically denoted by $p(S_0)$ or $\rho_0$. At each time step $t$, the agent observes the current state $S_t$, selects an action $A_t$, and the environment responds by generating a reward $R_{t+1}$ and transitioning to a new state $S_{t+1}$. This process continues either indefinitely (in continuing tasks) or until a terminal state is reached (in episodic tasks).

\subsection*{Example}

\begin{figure}[h]
	\centering
	\includegraphics[width=12cm]{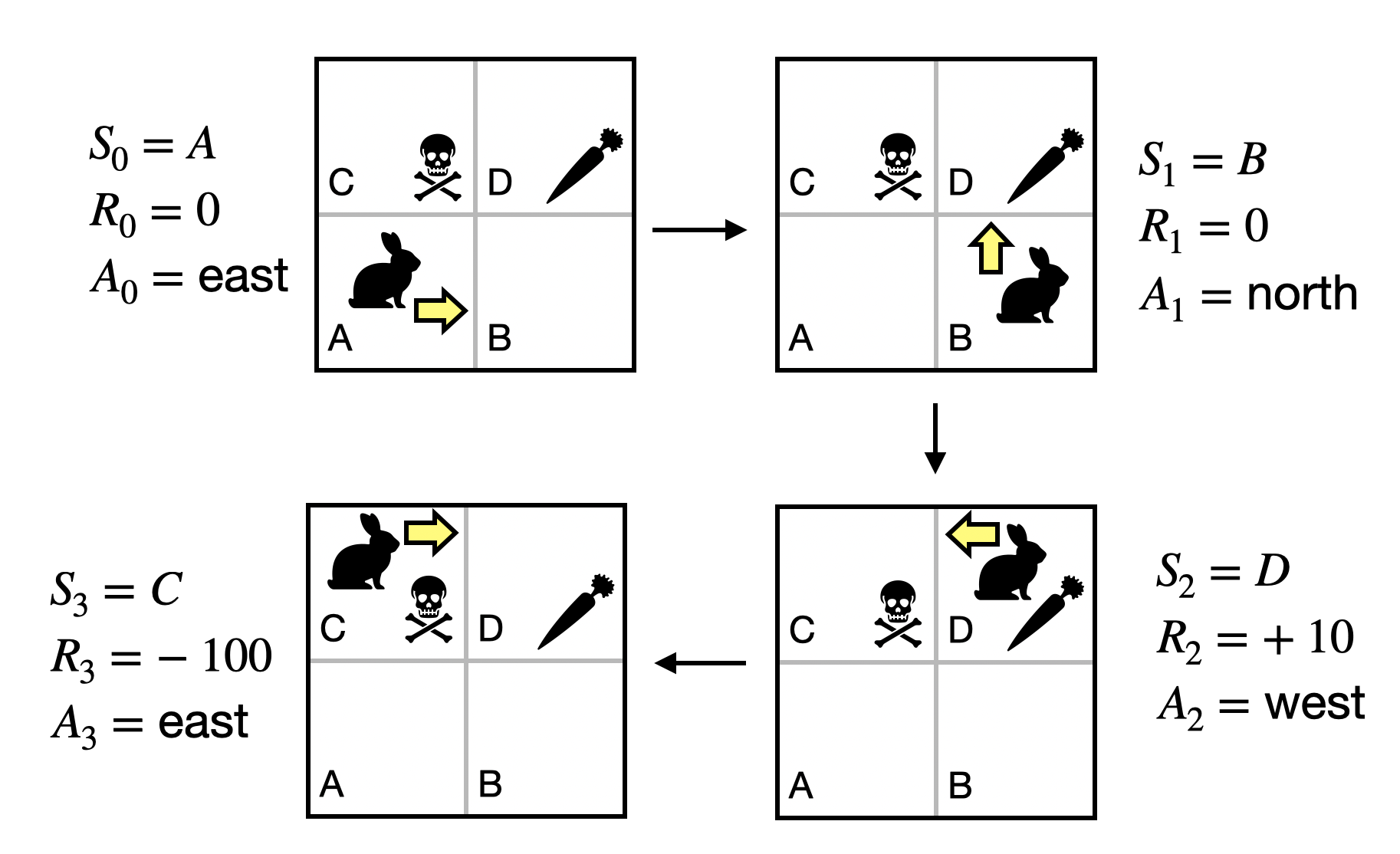}
	\caption{Example of the MDP interaction cycle: a rabbit-agent randomly moving in a $2\times2$ grid environment. Each panel shows the state the agent occupies, the action taken represented by an arrow, and the reward obtained, illustrating how sequential decisions and their consequences unfold in an MDP.}
	\label{fig:mdp_int_cycle_example}
\end{figure}

Figure~\ref{fig:mdp_int_cycle_example} illustrates an example trajectory of a rabbit-agent in a simple grid-world MDP. The agent, represented by the rabbit, begins in state A and selects the action 'move east', transitioning to state B with reward 0. From B, it chooses 'move north', reaching state D, where it receives a positive reward of +10 for finding the carrot. The agent then takes the action 'move west', moving into state C, which contains a hazardous cell and results in a large penalty of -100. Finally, from C, the agent again chooses 'move east', completing the sequence. The sequence of actions taken by the agent appears arbitrary, as it clearly does not correspond to the strategy that would maximize reward (e.g., moving from D to C is clearly a poor choice, at is results in a -100 penalty). In fact, identifying which action should be taken in each state to earn the highest possible cumulative reward is precisely the goal of the learning process.

\subsection{Markov Property}

In an MDP, all states are assumed to satisfy the \textit{Markov property}. This property asserts that the conditional probability distribution of the next state depends only on the current state (and action), and not on any earlier states or actions. In other words, the current state encapsulates all information from the past that is relevant for predicting the future. Consequently, past and future states are conditionally independent given the present state.

Formally, for a sequence of random variables $\{X_t\}_{t \in \mathbb{N}_0}$, the Markov property is expressed as \cite{sutton2018reinforcement}
\begin{equation*}
	P(X_{t+1} = x_{t+1} \mid X_t = x_t, X_{t-1} = x_{t-1}, \ldots, X_0 = x_0)
	= P(X_{t+1} = x_{t+1} \mid X_t = x_t).
\end{equation*}

In the context of MDPs, this translates to the transition dynamics depending only on the current state $S_t$ and the action $A_t$ selected by the agent \cite{sutton2018reinforcement}:
\begin{equation*}
	P(S_{t+1} = s' \mid S_t = s, A_t = a, S_{t-1}, A_{t-1}, \ldots)=P(S_{t+1} = s' \mid S_t = s, A_t = a).
\end{equation*}

The Markov property is fundamental for most reinforcement learning algorithms, as it enables the design of recursive methods that do not require explicit storage of the entire interaction history. By ensuring that the current state contains all information needed to predict future dynamics, algorithms can operate using only the most recent state-action pair, greatly simplifying computation.

\subsection*{Example}

\begin{figure}[h]
	\centering
	\includegraphics[width=10cm]{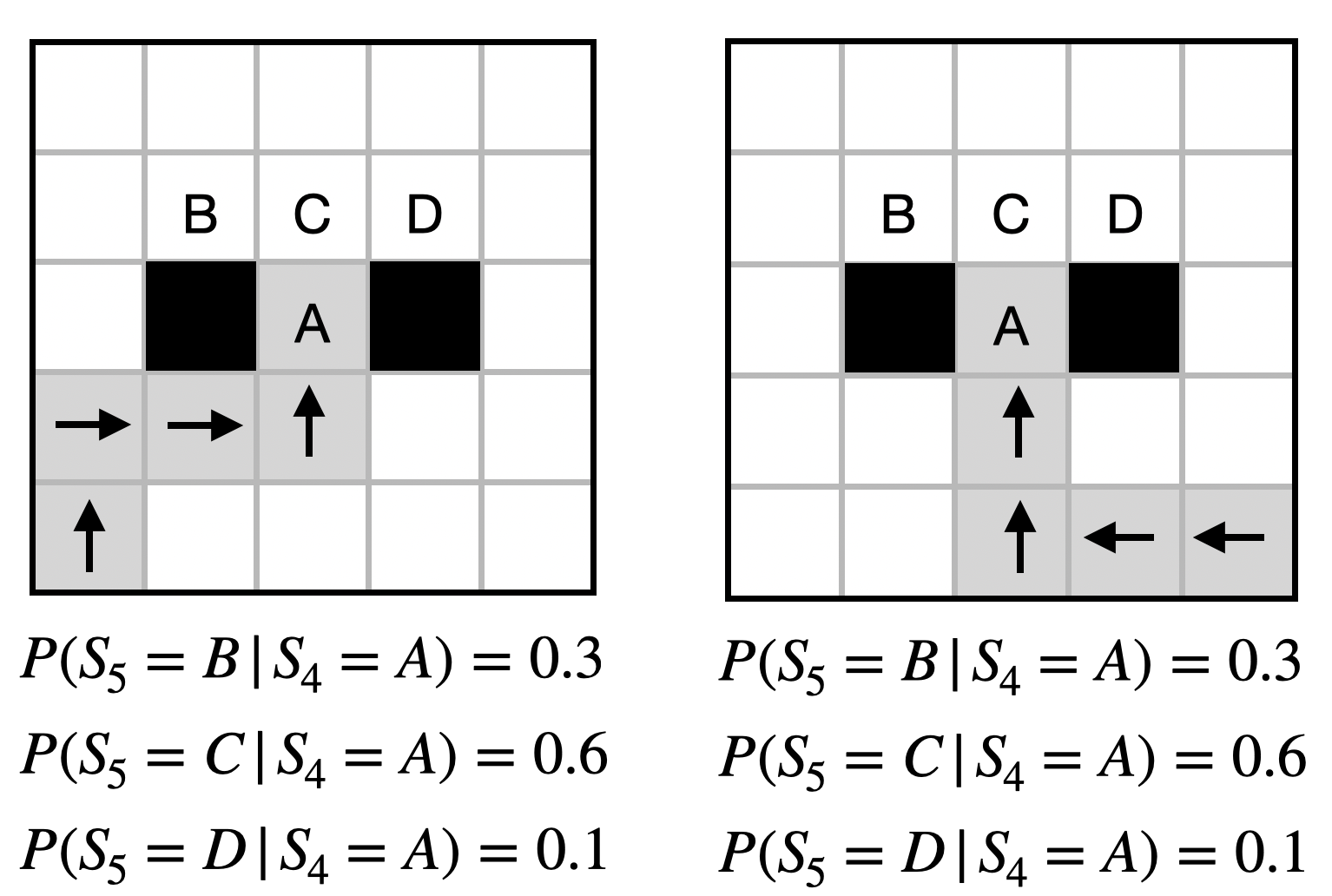}
	\caption{Illustrative grid-world showing a local transition neighbourhood around state $A$ (each cell is a possible state). The black squares are obstacles (unavailable states). Arrows on the grid indicate actions took by the agent and the shaded cells the path taken by the agent.}
	\label{fig:markov_property}
\end{figure}

Figure \ref{fig:markov_property} illustrates the Markov property in a grid-world environment, highlighting a scenario where the distribution of the next state depends only on the current state and not on the sequence of states that preceded it. The two panels show the same local configuration around a central state $A$. From
$A$, the agent can transition to three neighboring states, $B$ (above-left), $C$ (above), and $D$ (above-right), while the remaining adjacent cells contain walls and are therefore inaccessible. As depicted, if the agent occupies state $A$ at time $t$, the probabilities of moving to $B$, $C$, or $D$ at time $t+1$ are 0.3, 0.6, and 0.1, respectively. These probabilities depend solely on the current state $A$ and remain the same regardless of the path or sequence of states visited before reaching $A$ (represented by shaded cells).

\subsection{Partial Observability}

In many practical scenarios, the Markov property may not hold. In such cases, instead of observing the full environment state $s_t$, the agent receives only partial, egocentric \emph{observations} $o_t \in \mathcal{O}$. For example, in a 3D navigation task, the full state may include the global positions and velocities of all entities in the scene, while the agent's observation is limited to sensory information within its field of view. Under these conditions, the agent's input is insufficient to satisfy the Markov property, and the problem is more appropriately modeled as a Partially Observable Markov Decision Process (POMDP).

To maintain focus and conceptual clarity, this document assumes \emph{full observability}, treating the agent's input as the full state $s_t$. Nevertheless, all algorithms can also be applied in partially observable settings by substituting $s_t$ with the observation $o_t$. While this simplification facilitates analysis and implementation, it may introduce a performance gap due to missing information. These limitations can be partially mitigated by augmenting the observation with recent past observations (\emph{observation stacking}) or with the \emph{hidden state} of a recurrent neural network (e.g., LSTM or GRU) that summarizes relevant history.

\subsection{Dynamics Function}

The MDP relies on a \textit{dynamics function} to describe how the environment evolves over time as a result of its internal physical processes and the agent's actions. For example, an object may move because of gravity (internal dynamics) or because the agent pushes it (action-dependent dynamics). This function can be used to simulate how the environment changes from one state to the next, and it can also serve as an internal model that allows an agent to reason about the future consequences of its actions. In some settings, the dynamics function is fully known (e.g., in a simulator or a well-specified physics model). In others, it is unknown and must be estimated from data. In yet more challenging scenarios, the dynamics may be too complex or too costly to learn explicitly, forcing the agent to make decisions without relying on a learned model at all. Different reinforcement learning algorithms make different assumptions about whether the dynamics are known, learnable, or should be treated as a black box, and these assumptions fundamentally shape how the algorithms operate.

Formally, and in compliance with the Markov property, it is possible to define the MDP's \textit{dynamics function} $p: \mathcal{S}\times \mathcal{R}\times\mathcal{S}\times\mathcal{A}\rightarrow[0,1]$ as the probability of the current state $S_t$ and current reward $R_t$ taking the specific values $s'$ and $r$ passed as arguments of the function, given the specific values of the preceding state $s$ and action $a$ also passed as arguments of the function \cite{sutton2018reinforcement}:
\begin{equation*}
	p(s', r \mid s, a) \doteq P(S_t = s', R_t = r \mid S_{t-1} = s, A_{t-1} = a).
\end{equation*}

Given that $p$ specifies a probability distribution for each choice of $s$ and $a$, it follows that \cite{sutton2018reinforcement}:
\begin{equation*}
	\sum_{s' \in \mathcal{S}} \sum_{r \in \mathcal{R}} p(s', r \mid s, a) = 1, \quad \text{for all } s \in \mathcal{S}, a \in \mathcal{A}.
\end{equation*}

By marginalising over all reward values in the dynamics function, one obtains the \textit{state-transition function} $p: \mathcal{S} \times \mathcal{S} \times \mathcal{A} \to [0, 1]$, which specifies the probability of reaching state $s'$ when action $a$ is executed in state $s$. This function captures the consequences of the agent's actions by indicating which next states are more likely to occur given the current state-action pair. Formally, $p$ is defined as  \cite{sutton2018reinforcement}:
\begin{align*}
	p(s' \mid s, a) & \doteq P(S_t = s' \mid S_{t-1} = s, A_{t-1} = a) \\
	                & = \sum_{r \in \mathcal{R}} p(s', r \mid s, a).
\end{align*}

\subsection{Reward Function}

As mentioned, in an MDP the agent may receive rewards from the environment either for reaching a desirable state or for selecting a particular action in a given state. The reward indicates the desirability, or "goodness", of states and actions. Formally, the reward received by the agent can be described by a function $r: \mathcal{S} \times \mathcal{A} \to \mathbb{R}$, which returns a scalar value corresponding to the reward obtained by taking a given action in a given state.

Given the dynamics function, it is possible to define the reward function exactly. First, we need to determine the probability of each possible reward, which is obtained by marginalizing over all possible next states in the dynamics function \cite{sutton2018reinforcement}:
\begin{equation*}
	p(r \mid s, a) = \sum_{s' \in \mathcal{S}} p(s', r \mid s, a).
\end{equation*}

Using this, the reward function can be defined as the expected reward obtained by taking action $a$ in state $s$. The need to consider the expected value arises from the stochastic nature of the rewards: the outcome of taking the same action in the same state may vary according to the environment's dynamics. In other words, if the agent interacts with the environment multiple times from the same state-action pair, the observed rewards may differ, with some outcomes being more probable than others. The reward function captures this uncertainty by returning the mean reward under the probability distribution induced by the dynamics function.

By directly applying the definition of expected value (see Section~\ref{sec:expected_value}), the reward function can be expressed as the weighted sum of all possible rewards, where the weights correspond to the probabilities of each reward value \cite{sutton2018reinforcement}:

\begin{align*}
	r(s, a) & \doteq \mathbb{E}[R_t \mid S_{t-1} = s, A_{t-1} = a]                       \\
	        & = \sum_{r \in \mathcal{R}} r \, p(r \mid s, a)                             \\
	        & = \sum_{r \in \mathcal{R}} r \sum_{s' \in \mathcal{S}} p(s', r \mid s, a).
\end{align*}

\subsection*{Example}

\begin{figure}[h]
	\begin{center}
		\includegraphics[width=12cm]{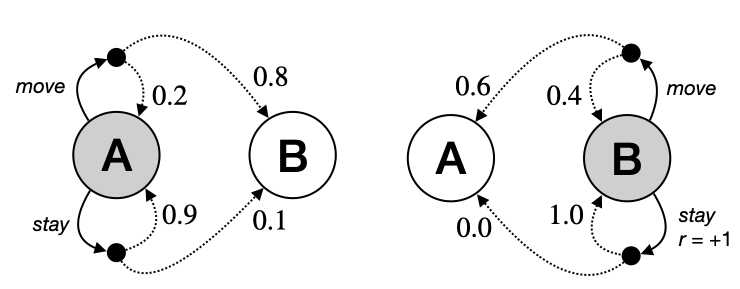}
		\caption{Illustrative MDP example. For simplicity, the visualization of the MDP is split into two parts: the left side depicts the transitions and actions when the agent is in state $A$, $S_t=A$, and the right side depicts the transitions and actions when the agent is in state $B$, $S_t=B$. Solid arrows represent actions, while dashed arrows indicate state transitions along with their respective conditional probabilities. The small dots highlight the stochastic outcomes of the actions.}
		\label{fig:mdp-example}
	\end{center}
\end{figure}

Consider an MDP where an agent can be in one of two states \( S = \{A, B\} \) and can take two actions \( \mathcal{A} = \{\text{move}, \text{stay}\} \). State \( A \) corresponds to a location that is easy to move away from (e.g., hard terrain), whereas state \( B \) corresponds to a location that is very hard to move away from (e.g., sandy terrain). A visual representation of this MDP is provided in Figure~\ref{fig:mdp-example}, while Table~\ref{tab:mdp-transition} presents the corresponding dynamics and reward functions.

If the agent is in state \( A \) and selects the \text{move} action, then it is highly likely to reach \( B \) with probability \( P(S_{t} = B \mid S_{t-1} = A, A_{t-1} = \text{move}) = 0.8 \). There is a small chance of slipping and staying in \( A \) with probability \( P(S_{t} = A \mid S_{t-1} = A, A_{t-1} = \text{move}) = 0.2 \). If the agent opts to stay in state \( A \), it is very likely to remain there with probability \( P(S_{t} = A \mid S_{t-1} = A, A_{t-1} = \text{stay}) = 0.9 \). There is a small chance that it will slip to state \( B \) with probability \( P(S_{t} = B \mid S_{t-1} = A, A_{t-1} = \text{stay}) = 0.1 \).

If the agent is at state \( B \) and selects the \text{move} action towards state \( A \), it is likely to be able to reach \( A \) with probability \( P(S_{t} = A \mid S_{t-1} = B, A_{t-1} = \text{move}) = 0.6 \), but there is a significant probability of remaining in \( B \) with probability \( P(S_{t} = B \mid S_{t-1} = B, A_{t-1} = \text{move}) = 0.4 \). The difficulty of leaving \( B \) is reflected in the deterministic consequence of staying in \( B \) with probability \( P(S_{t} = B \mid S_{t-1} = B, A_{t-1} = \text{stay}) = 1.0 \).

The reward is defined to be \( 0 \) everywhere except when the agent decides to stay in state \( B \), where it can consume food and earn a \( +1 \) reward. To accumulate reward, the agent needs to learn that when in \( A \), it should move to \( B \) and, once reached \( B \), it should stay there forever.

\begin{table}[h]
	\centering
	\caption{Dynamics and reward functions for the MDP example. Here, $S_t = s'$, $S_{t-1} = s$, and $A_{t-1} = a$.}
	\label{tab:mdp-transition}
	\begin{tabular}{|c|c|c|c|c|}
		\hline
		\textbf{$s$} & \textbf{$a$} & \textbf{$s'$} & \textbf{$p(s'\mid s,a)$} & \textbf{$r(s, a)$} \\ \hline
		$A$          & move         & $B$           & $0.8$                    & $0$                \\ \hline
		$A$          & move         & $A$           & $0.2$                    & $0$                \\ \hline
		$A$          & stay         & $A$           & $0.9$                    & $0$                \\ \hline
		$A$          & stay         & $B$           & $0.1$                    & $0$                \\ \hline
		$B$          & move         & $A$           & $0.6$                    & $0$                \\ \hline
		$B$          & move         & $B$           & $0.4$                    & $0$                \\ \hline
		$B$          & stay         & $B$           & $1.0$                    & $+1$               \\ \hline
	\end{tabular}
\end{table}

\section{Expected Return}

Since rewards indicate the desirability of states and actions, it is natural that the agent's objective is to accumulate reward over time. However, in a stochastic environment, it is often more important for the agent to secure immediate rewards than to rely on delayed rewards, because the longer the delay, the greater the uncertainty about the outcomes.

To distinguish between immediate and accumulated rewards, the cumulative reward over time is commonly referred to as the \textit{return}. The return represents the total accumulated reward that an agent can obtain starting from a given time step. Under these considerations, it is standard in reinforcement learning to define the agent's goal as the maximization of the expected \textit{discounted return} \cite{sutton2018reinforcement}:

\begin{equation}\label{equ:exp_return}
	G_t \doteq R_{t+1} + \gamma R_{t+2} + \gamma^2 R_{t+3} + \ldots = \sum_{k=t+1}^{T} \gamma^{k-t-1} R_k,
\end{equation}

\noindent where $T$ is the final time step and $0\le\gamma\le 1$ is the \textit{discount factor}. Tasks are called \textit{continuing} when $T=\infty$ and \textit{episodic} when otherwise.

If $\gamma = 0$ then the agent only takes into account the immediate reward, meaning that it is not able to take into account that one's action influence also future rewards (e.g., if the agent opts for staying still it will never be able to reach the object that will provide reward). As $\gamma$ approaches $1$ future rewards are progressively being more accounted for, although the more recent the reward is, the more important it is. On the other extreme, if $\gamma = 1$, then all rewards into the future are taken into account, meaning that if $T=\infty$ the return would be infinite. Hence, if $T=\infty$, it is necessary to ensure that $\gamma < 1$.

As mentioned, when $\gamma < 1$, the return is finite, despite the fact it is the sum of infinite terms. For instance, if the agent collects at each time step a constant reward $R$ (note the absence of time-based subscript), the return is a geometric series with the following closed form solution \cite{sutton2018reinforcement}:
\begin{equation}
	G_t = \sum_{k=0}^\infty R\gamma^k = \frac{R}{1-\gamma} \quad (\text{e.g., if } \gamma=0.9 \text{ and } R=1 \text{ then } G_t = 10 ).
\end{equation}

\subsection*{Example}

\begin{figure}[h]
	\centering
	\includegraphics[width=12cm]{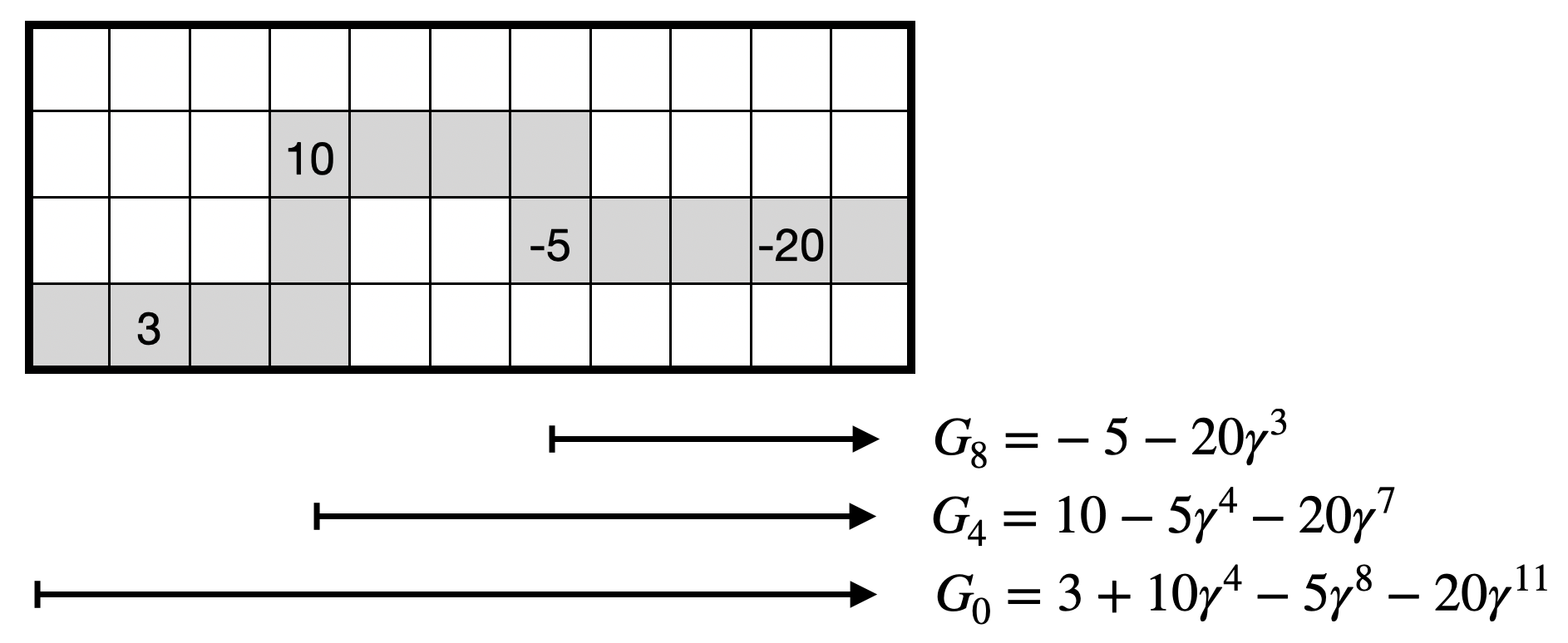}
	\caption{Discounted return at $t=0$, $t=4$, and $t=8$ along an agent trajectory on a grid-world, with a given $\gamma$.}
	\label{fig:return_example}
\end{figure}

Figure~\ref{fig:return_example} shows the path an agent takes in a grid-world, along with the non-zero rewards received in each state (cell). It also illustrates the discounted return at three points along the trajectory: $t=0$, $t=4$, and $t=8$, for a given discount factor $\gamma$. The discounted return at a time step $t$ represents the total reward the agent expects to accumulate from that moment onward, with each future reward multiplied by successive powers of $\gamma$ to account for temporal discounting. For example, the return at $t=0$ sums all future rewards starting from the beginning of the trajectory, with rewards farther in the future being weighted less due to discounting, while the return at $t=8$ only sums rewards from that later point onward. As a result, the immediate reward of $-20$ at the end of the trajectory has a much larger effect on the return at $t=8$ (closer in time) than on the return at $t=0$ (farther in time).

\section{Exact Methods for MDPs}

This section explores the problem of solving MDPs using exact methods, which are guaranteed to find the optimal mapping from states to action probabilities. These methods assume complete knowledge of the MDP, which can limit their practical application in real-world problems. However, they form the foundational concepts of reinforcement learning and are crucial for understanding the building blocks of optimal sequential decision making.

\subsection{Policy Functions}

The optimal mapping from states to actions is the one that allows the agent to achieve the highest cumulated reward over time, assuming the agent always follows it exactly. Any such mapping is called \textit{policy}, even if it is not optimal. In other words, a policy simply describes how the agent chooses actions in each state, while the optimal policy is the one that leads to the best long-term cumulated reward.

Formally, if the agent is in state $s\in \mathcal{S}$ at time $t$ (i.e., $S_t = s$) and is following a \textit{stochastic} policy $\pi$, then $\pi(a \mid s)$ is the probability of the agent selecting action $a\in \mathcal{A}$ at time $t$ (i.e., $A_t = a$). Concretely, $\pi: \mathcal{A}\times \mathcal{S}\rightarrow[0,1]$ is an ordinary function that defines a probability distribution over $a\in \mathcal{A}$ for each $s\in \mathcal{S}$ \cite{sutton2018reinforcement}:
\begin{equation*}
	\pi(a \mid s) \doteq P(A_t = a \mid S_t = s), \quad \text{with} \quad \sum_{a\in\mathcal{A}} \pi(a \mid s) = 1 \quad \text{for all } s\in \mathcal{S}.
\end{equation*}

\subsection*{Example}

\begin{figure}[h]
	\centering
	\includegraphics[width=13cm]{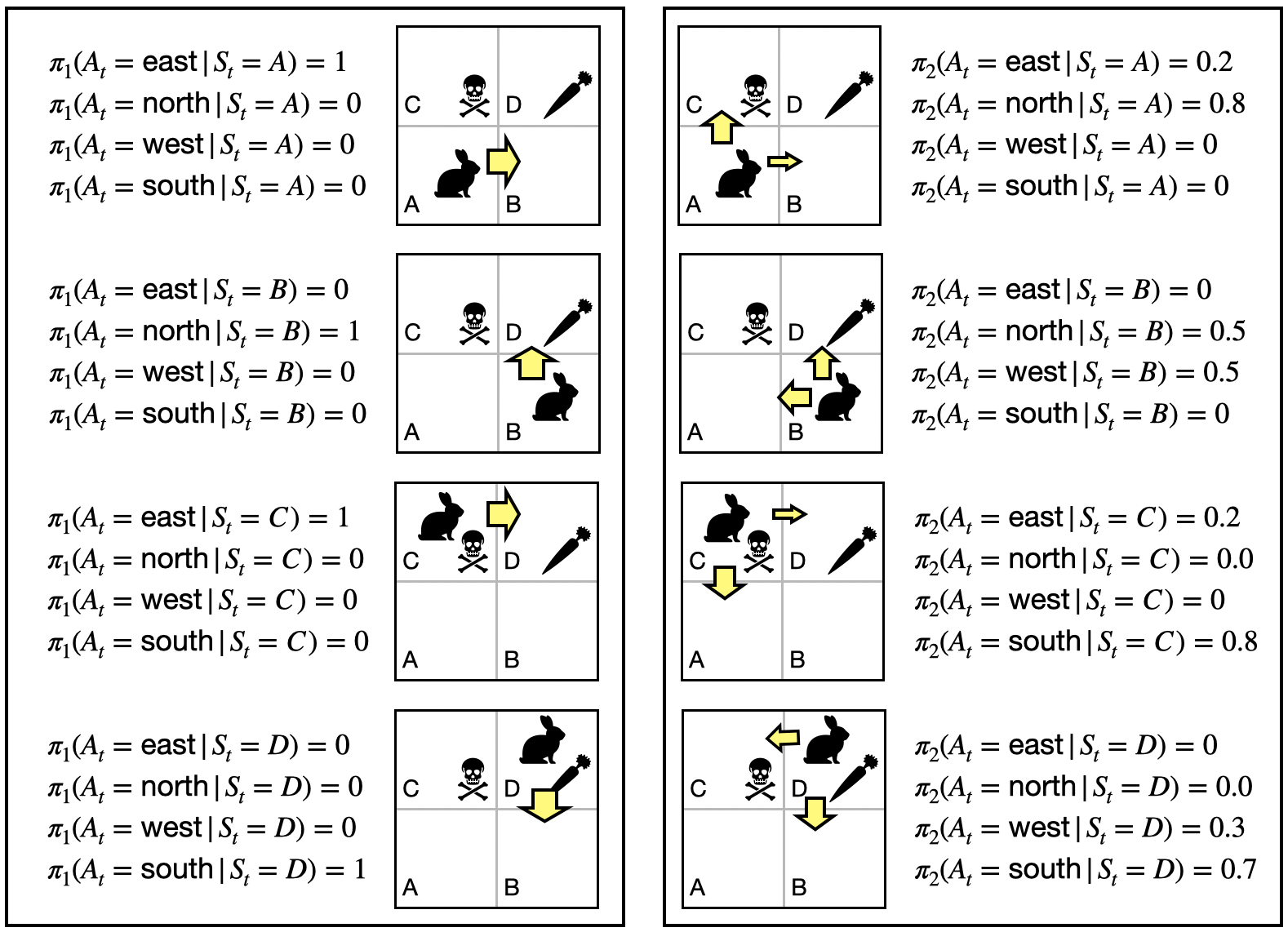}
	\caption{Example of the optimal policy $\pi_1$ (left) and a sub-optimal policy $\pi_2$ (right) for the problem of a rabbit-agent moving the a grid-world with rewards as follows: landing on a carrot = +5, landing on poison = -5, moving out-of-bounds = -10. See the main text for a complete explanation.}
	\label{fig:policy_example_two}
\end{figure}

Recall that Figure~\ref{fig:mdp_int_cycle_example} illustrates the problem of a rabbit-agent moving on a grid world populated with a carrot and poison. In that figure, the agent follows a sub-optimal policy. Figure~\ref{fig:policy_example_two} revisits this problem by comparing two policies: the optimal policy $\pi_1$ and a sub-optimal policy $\pi_2$.

Each panel represents a possible situation defined by the agent's state (its position) and the distribution over actions induced by one of the two policies. Arrows indicate the actions, with their thickness proportional to the corresponding action probabilities. The optimal policy consistently guides the agent toward the carrot (positive reward) and away from the poison (negative reward). In contrast, the sub-optimal policy distributes actions more evenly, making it more likely for the agent to move away from the carrot and toward the poison.

\subsection{Value Functions}

The goal of learning is to discover the optimal policy through experience. A stepping stone toward this goal is to learn how valuable a given state is under a policy $\pi$. That is, we want to understand how much reward the agent is expected to collect if it visits a particular state and then continues to follow $\pi$.
This tells the agent how desirable it is to visit that state, and also how good a policy is in terms of the states it tends to reach. Some states lead to higher long-term reward than others, and some policies drive the agent toward more valuable states than alternative policies.

The goodness, or value, of a state $s$ under a given policy $\pi$ is  defined by the \textit{state-value function}. This function represents the discounted return the agent is expected to receive starting from state $s \in \mathcal{S}$, assuming it follows $\pi$ from that point onward \cite{sutton2018reinforcement}:
\[v_{\pi}(s) \doteq \mathbb{E}_{\pi} [G_t \mid S_t = s],\]

\noindent where $\mathbb{E}_{\pi}[\cdot]$ denotes the expected value of a random variable under the assumption that the agent acts according to policy $\pi$. In practice, this expectation averages over both the actions chosen according to $\pi$ and the subsequent states visited as a result of those actions.

It is worth pausing to clarify the distinction between the return $G_t$ and the value function $v_{\pi}(s)$. The return $G_t$ represents the actual cumulative reward an agent obtains when interacting with the environment starting from time step $t$. In contrast, the value function $v_{\pi}(s)$ predicts the expected return if the agent starts in state $s$ and follows policy $\pi$. In other words, $G_t$ is a realized quantity that depends on the specific trajectory taken, whereas $v_{\pi}(s)$ provides an expectation of $G_t$, averaging over all possible future trajectories consistent with the policy and the environment dynamics.

It is also useful to consider the \textit{action-value function} of a policy $\pi$, $q_{\pi}:\mathcal{S}\times\mathcal{A}\rightarrow\mathbb{R}$, which is defined as the expected discounted return obtained by the agent when it starts in state $s\in \mathcal{S}$, takes action $a\in \mathcal{A}$, and then follows policy $\pi$ thereafter \cite{sutton2018reinforcement}:
\begin{equation}\label{equ:return}
	q_{\pi}(s, a) \doteq \mathbb{E}_{\pi} [G_t \mid S_t = s, A_t = a].
\end{equation}

The \textit{state-value function} can then be represented as the expectation of the \textit{action-value} function under the action probabilities specified by the policy $\pi$ \cite{sutton2018reinforcement}:
\begin{equation}\label{equ:state_action}
	v_\pi(s) = \sum_{a\in\mathcal{A}} \pi(a \mid s) q_\pi(s, a).
\end{equation}

This equation has exactly the form of an expectation: it computes a weighted average of the action-value function \(q_\pi(s, a)\) over all possible actions in \(\mathcal{A}\), where each weight corresponds to the probability of selecting that action under the policy, \(\pi(a \mid s)\).

This formalization shows that the \textit{state-value} function allows the agent to reason about the desirability of a given state, independently of which action will be selected next. On the other hand, the \textit{action-value function} enables more fine-grained reasoning, as it allows the agent to evaluate and compare the desirability of specific state-action pairs. Both functions play an important role in the design and analysis of reinforcement learning algorithms.

\subsection*{Examples}

As mentioned, the value function indicates how desirable it is for the agent, under policy $\pi$, to find itself in a particular state. For example, if $v_{\pi}(A) > v_{\pi}(B)$, then an agent following $\pi$ is expected to obtain more long-term reward by visiting state $A$ rather than state $B$. This difference may arise because state $B$ leads to a dead-end, making it difficult for the policy to accumulate further reward, whereas state $A$ might be an open region from which the agent can easily navigate toward rewarding areas. In this way, the value function reflects not only the immediate reward of a state but also its strategic potential for future returns.

\begin{figure}[h]
	\centering
	\includegraphics[width=14cm]{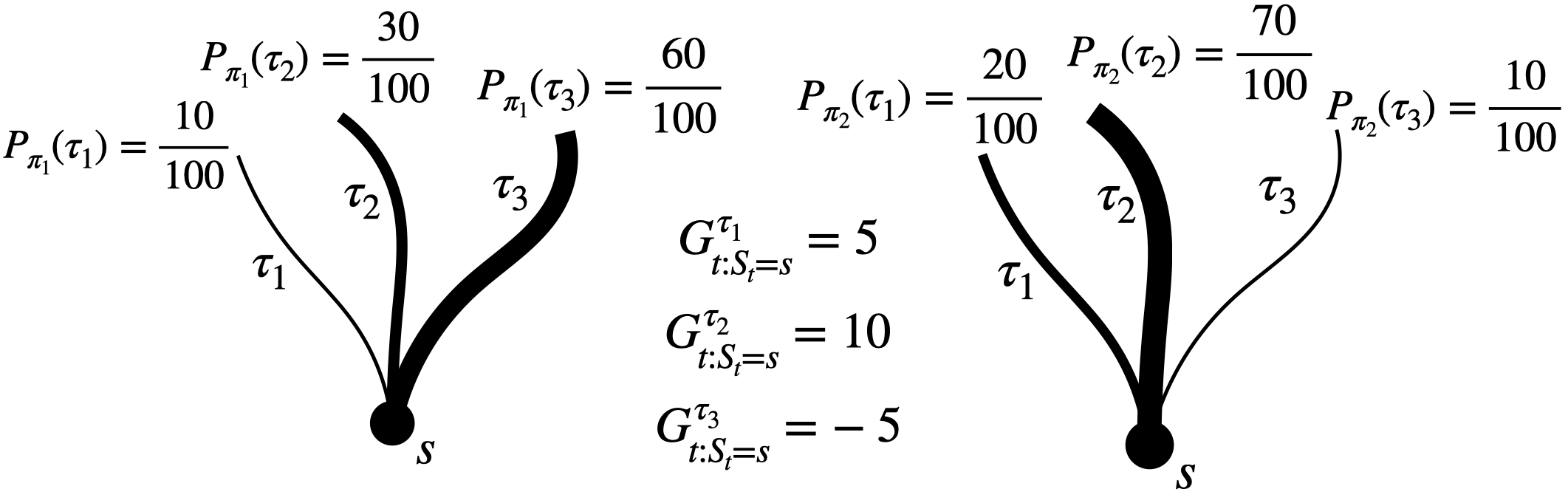}
	\caption{Example of trajectories induced by two different policies, $\pi_1$ and $\pi_2$. The thicker a curve, the more frequently the corresponding trajectory was followed by the agent.}
	\label{fig:state_value_example}
\end{figure}

Let us now consider a different example. Figure~\ref{fig:state_value_example} illustrates the state-value function for a particular state $s$ under two different policies, $\pi_1$ and $\pi_2$. In this scenario, it is assumed that only three  trajectories can follow from state $s$, $\tau_1, \tau_2, \tau_3$.

Depending on the policy  controlling the agent, the probabilities of the possible trajectories change. For example, suppose the agent runs 100 episodes following policy $\pi_1$. We can then estimate the probability of trajectory $\tau_1$ under $\pi_1$ as the proportion of runs the agent actually followed the trajectory $\tau_1$. If this occurs 10 times, then the empirical probability is $P_{\pi_1}(\tau_1)=10/100=0.1$.

Let us assume that each trajectory yields a different discounted return: $G^{\tau_1}_{t:S_t=s}=5$, $G^{\tau_2}_{t:S_t=s}=10$, and $G^{\tau_3}_{t:S_t=s}=-5$. Thus, depending on which trajectory is followed, the agent experiences different sequences of immediate rewards and therefore different total returns. The state-value, defined as the expected discounted return when starting in state $s$, is influenced by how frequently each trajectory is visited under the policy.

Using data from the 100 episodes, we can approximate these expectations by the sample mean. In this way, we can estimate the expected return under each policy and thereby determine which policy yields higher reward from state $s$ onward:

\begin{align*}
	v_{\pi_1}(s) & \approx0.1\cdot5+0.3\cdot10+0.6\cdot(-5)=0.5, \\
	v_{\pi_2}(s) & \approx0.2\cdot5+0.7\cdot10+0.1\cdot(-5)=7.5.
\end{align*}

In this example, the estimated state value under $\pi_2$ is higher ($7.5 > 0.5$), suggesting that $\pi_2$ yields a greater long-term return from state $s$. This indicates that $\pi_2$ is, under the available data, a more promising candidate for controlling the agent than $\pi_1$.

\subsection{Bellman Equation}

Let us now derive the state-value function $v_{\pi}(s)$ to the point where it can be computed. The key in this derivation is a recursive formulation for the discounted return, because decisions, and their consequences, unfold step by step: the return from a state equals the immediate reward plus the discounted return from the next state onward.

Intuitively, if the agent receives a reward now and expects further rewards in the future, the total discounted return can be accumulated along the trajectory one step at a time. With this intuition, a recursive formulation of the \textit{discounted return} can be derived, assuming $G_T = 0$ (recall that $T$ is the final time step) \cite{sutton2018reinforcement}:
\begin{align}
	G_t & \doteq R_{t+1} + \gamma R_{t+2} + \gamma^2 R_{t+3} + \gamma^3 R_{t+4} + \cdots \nonumber         \\
	    & = R_{t+1} + \gamma \left( R_{t+2} + \gamma R_{t+3} + \gamma^2 R_{t+4} + \cdots \right) \nonumber \\
	    & = R_{t+1} + \gamma G_{t+1}.
	\label{equ:return_recursive}
\end{align}

Given the recursive formulation of the discounted return $G_t$, a recursive formulation for the state-value function $v_{\pi}$ can be derived as follows \cite{sutton2018reinforcement}:
\begin{align}\label{equ_bellman_equation}
	v_{\pi}(s) & \doteq \mathbb{E}_{\pi} [G_t \mid S_t = s] \nonumber                                                                                                                                      \\
	           & = \mathbb{E}_{\pi} [R_{t+1} + \gamma G_{t+1} \mid S_t = s] \quad (\text{using Equation}~\ref{equ:return_recursive}) \nonumber                                                             \\
	           & = \sum_{a\in\mathcal{A}} \pi(a \mid s) \sum_{s'\in\mathcal{S}} \sum_{r\in\mathcal{R}} p(s', r \mid s, a) \left[ r + \gamma \mathbb{E}_{\pi} [G_{t+1} \mid S_{t+1} = s'] \right] \nonumber \\
	           & = \sum_{a\in\mathcal{A}} \pi(a \mid s) \sum_{s'\in\mathcal{S}} \sum_{r\in\mathcal{R}}  p(s', r \mid s, a) \left[ r + \gamma v_{\pi}(s') \right], \quad \text{for all } s \in S.
\end{align}

Equation~\ref{equ_bellman_equation} is known as the \textit{Bellman equation} for $v_\pi$. To better understand its role, let us take a moment to develop some intuition about what this equation expresses. According to the chain rule of probabilities, $\pi(a|s)p(s',r | s,a)$ is the joint probability of the triplet $a$, $s'$, and $r$, given $s$. In the equation, this probability is multiplied by $\left[ r + \gamma v_{\pi}(s') \right]$ for every $a$, $s'$, and $r$ to obtain the expected value (mean) of $\left[ r + \gamma v_{\pi}(s') \right]$. Hence, the Bellman equation states that the expected value of the start state must equal the reward obtained at that state plus the (discounted) value of the expected next state (recursive call of $v_{\pi}$).

Comparing the Bellman equation (Equation~\ref{equ_bellman_equation}) with our earlier definition of the state-value function, $v_{\pi}(s) = \sum_{a} \pi(a|s) q_{\pi}(s,a)$ (Equation~\ref{equ:state_action}), we observe that both equations share the same structure: a summation over actions weighted by $\pi(a \mid s)$. By aligning the terms inside the summation, we identify the expression within the brackets as $q_{\pi}(s, a)$. Thus, the state-action-value function, representing the expected discounted return obtained by the agent when starting from state $s$, taking action $a$, and following  policy $\pi$ thereafter, is defined as:
\begin{equation}
    q_\pi(s,a) = \sum_{s'\in\mathcal{S}} \sum_{r\in\mathcal{R}} p(s', r \mid s, a) \left[ r + \gamma v_{\pi}(s') \right].
\end{equation}

The Bellman equation forms a system of equations, with one equation corresponding to each state. For a problem with $n$ states, the system consists of $n$ equations with $n$ unknowns. If the dynamics function $p$ is known, this system can, in principle, be solved directly to obtain $v_{\pi}$. Alternatively, the solution can be found iteratively using \textit{dynamic programming} techniques.

\subsection{Iterative Policy Evaluation Algorithm}

A well-known dynamic programming algorithm for computing $v_\pi$ is known as \textit{iterative policy evaluation}. The algorithm begins with an arbitrary initial estimate of $v_\pi$, denoted $v_0$. It then generates successive approximations by updating each estimate using the Bellman equation (\ref{equ_bellman_equation}). Formally, for $k = 0,1,2,\dots$, the next approximation $v_{k+1}$ is computed from the previous one $v_k$ as follows \cite{sutton2018reinforcement}:
\begin{align}
	v_{k+1}(s) & \doteq \mathbb{E}_{\pi} [R_{t+1} + \gamma v_{k}(S_{t+1}) \mid S_t = s] \nonumber                                                                                              \\
	           & = \sum_{a\in\mathcal{A}} \pi(a \mid s) \sum_{s'\in\mathcal{S}} \sum_{r\in\mathcal{R}}  p(s', r \mid s, a) \left[ r + \gamma v_{k}(s') \right], \quad \text{for all } s \in S.
\end{align}

The appeal of \textit{iterative policy evaluation} lies on the guarantee that, if there is a $v_\pi$, the sequence $\{v_k\}$  converges to $v_\pi$ as $k\rightarrow \infty$. In practice the sequence is stopped as soon as the difference between successive approximations is below a small enough threshold $\psi$. Algorithm~\ref{alg:iterative-policy-evaluation} presents the pseudo-code of \textit{iterative policy evaluation}.

\begin{algorithm}[h]
	\caption{Iterative Policy Evaluation (adapted from \cite{sutton2018reinforcement})}
	\begin{algorithmic}[1]
		\STATE \textbf{Input:} the policy to be evaluated, $\pi$, and the dynamics function, $p$
		\STATE \textbf{Parameters:} threshold $\psi > 0$
		\STATE \textbf{Initialize} arbitrarily $V(s)$, for all $s \in \mathcal{S}^+$, except for $V(terminal) = 0$
		\STATE \textbf{Loop:}
		\begin{ALC@g}
			\STATE $\Delta \leftarrow 0$
			\STATE \textbf{Loop for each $s \in \mathcal{S}$:}
			\begin{ALC@g}
				\STATE $v \leftarrow V(s)$
				\STATE $V(s) \leftarrow \sum_{a\in\mathcal{A}} \pi(a \mid s) \sum_{s'\in\mathcal{S}} \sum_{r\in\mathcal{R}}  p(s', r \mid s, a) \left[ r + \gamma V(s') \right]$
				\STATE $\Delta \leftarrow \max(\Delta, |v - V(s)|)$
			\end{ALC@g}
		\end{ALC@g}
		\STATE \textbf{Until $\Delta < \psi$}  \textcolor{blue}{// condition occurs when $V(s) \approx v_{\pi}(s)$ for all $s \in \mathcal{S}$}
		\STATE \textbf{Return} $V$
	\end{algorithmic}
	\label{alg:iterative-policy-evaluation}
\end{algorithm}

\subsubsection*{Example: Grid world (adapted from \cite{sutton2018reinforcement})}

Let us assume a regular $5\times 5$ grid world in which an agent inhabits. The moves deterministically (i.e., given a state and an action, only one successive state has non-null probability) from its current cell to another by executing an action: north, south, east, or west. Actions that push the agent off the grid result in no position change and award a reward of $-1$. The reward is elsewhere 0. The state is the agent's position in the grid world (i.e., cell index). Hence, in this problem, $\mathcal{A}=\{\text{north}, \text{south}, \text{east}, \text{west}\}$ and $\mathcal{S}=\{1, \ldots, 25\}$. Let us assume a discount factor of $\gamma=0.9$.

\begin{figure}[t]
	\centering
	\begin{subfigure}[b]{0.3\textwidth}
		\centering
		\includegraphics[width=\textwidth]{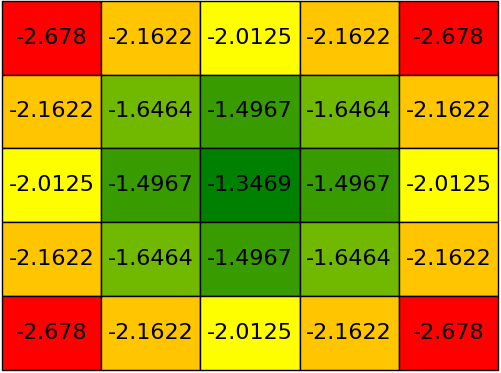}
		\caption{$k=74$ (at convergence)}
		\label{fig:sub1}
	\end{subfigure}
	\hfill
	\begin{subfigure}[b]{0.3\textwidth}
		\centering
		\includegraphics[width=\textwidth]{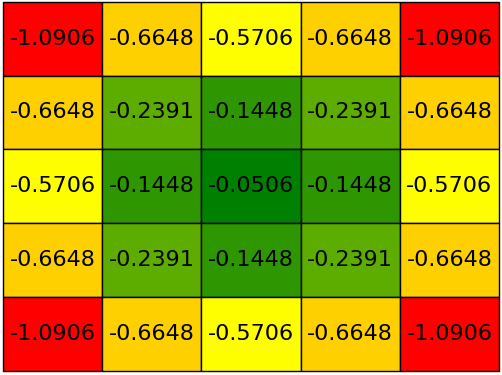}
		\caption{$k=3$}
		\label{fig:sub1}
	\end{subfigure}
	\hfill
	\begin{subfigure}[b]{0.3\textwidth}
		\centering
		\includegraphics[width=\textwidth]{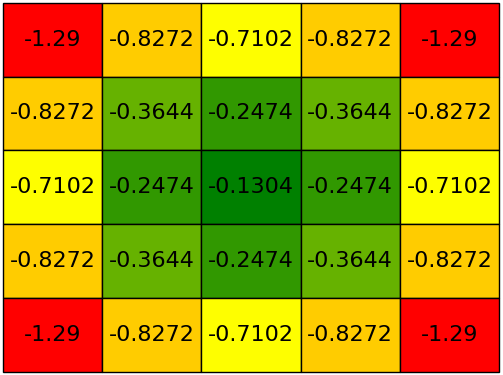}
		\caption{$k=4$ }
		\label{fig:sub2}
	\end{subfigure}
	\caption{Iterative policy evaluation for grid world with $P(A_t=a)=0.25, \forall t, a\in\mathcal{A}$.}
	\label{fig:policy-evaluation-example}
\end{figure}

Figure~\ref{fig:policy-evaluation-example} shows the grid world with the values of $v_{74}(s), \forall s \in \mathcal{S}$ (left), $v_3(s), \forall s \in \mathcal{S}$ (middle), $v_4(s), \forall s \in \mathcal{S}$ (right), that is, the values of the iterative policy evaluation in its final iteration (convergence with $\psi=10^{-4}$), in its third iteration and in its fourth iteration. These values were obtained with a uniformly random policy, meaning that all actions are equally probable, $P(A_t=a)=0.25, \forall t, a\in\mathcal{A}$. The final iteration shows that the further the agent is from the borders, the higher the value of those states, that is, the less likely it is to find the negative rewarding border. Note that $k=4$ is determined by $k=3$. For example, the central cell in $k=4$ is calculated with a weighted average of the north, south, east, and west neighbors in $k=3$, the weights being the probability of reaching those states times the discount factor: $-0.1448\times0.25\times0.9 + -0.1448\times0.25\times0.9 + -0.1448\times0.25\times0.9 + -0.1448\times0.25\times0.9 \approx -0.13$.

\begin{figure}[h]
	\centering
	\begin{subfigure}[b]{0.3\textwidth}
		\centering
		\includegraphics[width=\textwidth]{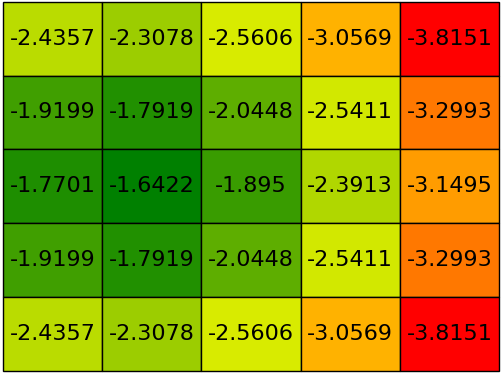}
		\caption{$k=78$ (at convergence)}
		\label{fig:sub2}
	\end{subfigure}
	\hfill
	\begin{subfigure}[b]{0.3\textwidth}
		\centering
		\includegraphics[width=\textwidth]{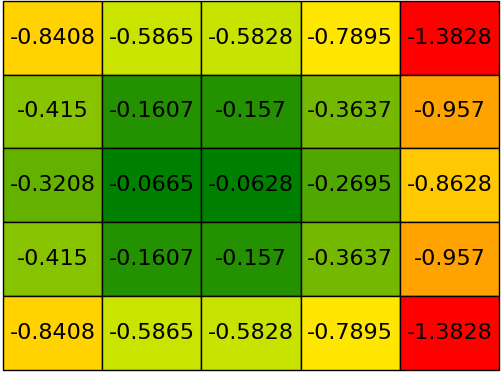}
		\caption{$k=3$}
		\label{fig:sub1}
	\end{subfigure}
	\hfill
	\begin{subfigure}[b]{0.3\textwidth}
		\centering
		\includegraphics[width=\textwidth]{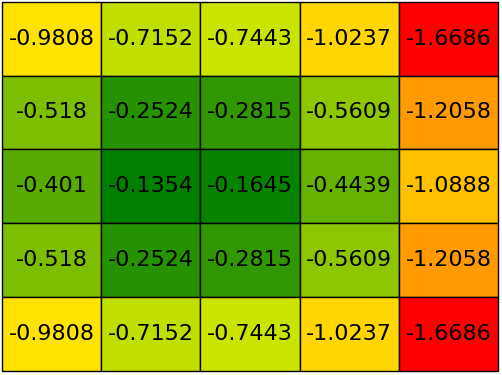}
		\caption{$k=4$}
		\label{fig:sub2}
	\end{subfigure}
	\caption{Iterative policy evaluation with $P(A_t=\text{north})=0.25, P(A_t=\text{south})=0.25, P(A_t=\text{left})=0.15, P(A_t=\text{right})=0.35, \forall t$.}
	\label{fig:policy-evaluation-example-wind}
\end{figure}

Figure~\ref{fig:policy-evaluation-example-wind} shows the same procedure but for a different action distribution. In this case the random walk is biased towards moving to right as a consequence of wind presence, $P(A_t=\text{north})=0.25, P(A_t=\text{south})=0.25, P(A_t=\text{left})=0.15, P(A_t=\text{right})=0.35, \forall t$. Note how the states with higher value shifted to the left. That is the because starting away from the right boundary renders less likely and delays the encounter with the negative rewards than starting away from the left boundary. Delaying the encounter with rewards impacts the final return due to the discount factor. For the sake of completeness, let us calculate the value of the central cell in $k=4$, which is done by taking into account the values at iteration $k=3$:  $-0.0665\times0.15\times0.9 + -0.2695\times0.35\times0.9 + -0.157\times0.25\times0.9 + -0.157\times0.25\times0.9 \approx -0.16$.

\begin{figure}[h]
	\centering
	\begin{subfigure}{0.225\textwidth}
		\centering
		\includegraphics[width=\textwidth]{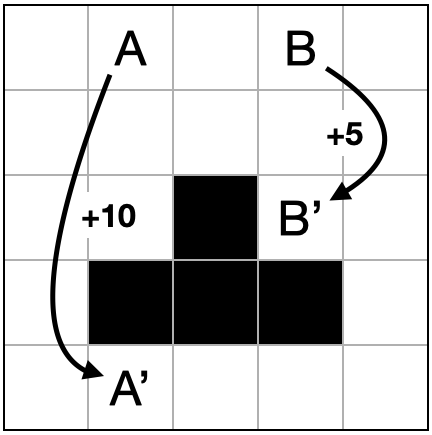}
		\caption{World rules.}
		\label{fig:sub1}
	\end{subfigure}
	\hfill
	\begin{subfigure}{0.3\textwidth}
		\centering
		\includegraphics[width=\textwidth]{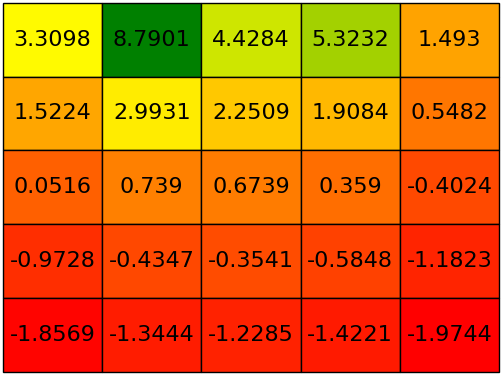}
		\caption{$k=47$ (at convergence)}
		\label{fig:sub1}
	\end{subfigure}
	\hfill
	\begin{subfigure}{0.3\textwidth}
		\centering
		\includegraphics[width=\textwidth]{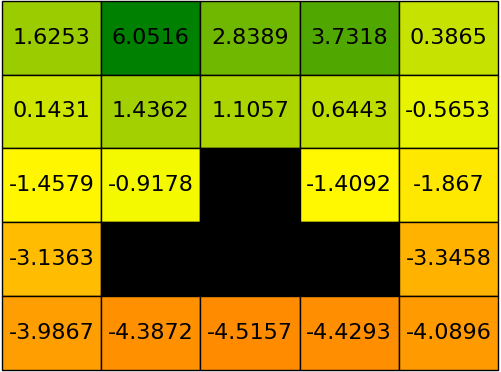}
		\caption{$k=76$ (at convergence)}
		\label{fig:sub2}
	\end{subfigure}
	\caption{Iterative policy evaluation for grid world with $P(A_t=a)=0.25, \forall t, a\in\mathcal{A}$. Left: the rules of the world: jumps positions $A\leftarrow A'$ and $B\leftarrow B'$ and walls positions (black cells). Middle: converged state values considering jumps $A$ and $B$ but without walls. Right: converged state values considering jumps $A$ and $B$ and walls. (Adapted and extended from Sutton and Barto \cite{sutton2018reinforcement})}
	\label{fig:policy-evaluation-example-A-B}
\end{figure}

Figure~\ref{fig:policy-evaluation-example-A-B} shows where iterative policy evaluation converges for a somewhat more complex environment under the uniform random policy, $P(A_t=a)=0.25, \forall t, a\in\mathcal{A}$. The environment now contains two special cells $A$ and $B$. Whatever the action taken by the agent in $A$ takes the agent immediately to cell $A'$, awarding a $+10$ reward. Whatever the action taken by the agent in $B$ takes the agent immediately to cell $B'$, awarding a $+5$ reward.

Let us now assume that some cells are designated as walls. Moving to a wall results exactly in the same consequences of moving towards a world boundary: stays in the same state and gets a $-1$ reward. In the environment without walls, state $A$, although considered optimal, has an expected return lower than 10, its immediate reward, because the agent transitions from $A$ to $A'$, where it is likely to encounter the boundary of the grid, which awards a negative reward. Conversely, State $B$ is valued higher than 5, its immediate reward, as the agent moves from $B$ to $B'$, which has a positive value. The potential penalty for hitting an edge from $B'$ is outweighed by the potential benefits of reaching $A$ or $B$ from there.

This example extends the classical grid world introduced by Sutton and Barto \cite{sutton2018reinforcement}, which considered only uniform random policies, negative rewards at the boundaries, and positive rewards at the "jump" states.

\subsection{Value Iteration Algorithm}

As mentioned, reinforcement learning problems are generally solved by seeking a policy that maximizes the cumulative reward over time. As previously noted, this can be achieved by comparing policies in terms of their value functions: better policies assign higher values to states. Thus, in finite MDPs, value functions induce a partial ordering over different policies, providing a principled way to identify which policies are preferable.

A policy \(\pi\) is considered superior to, or on par with, another policy \(\pi'\) if its expected return is at least as high as that of \(\pi'\) for every state. There always exists at least one policy that outperforms or matches all other policies, known as the \emph{optimal policy}, denote by \(\pi_*\). The Bellman equation for its state-value function \(v_* : \mathcal{S} \rightarrow \mathbb{R}\), also called the \textit{Bellman optimality equation}, is given by \cite{sutton2018reinforcement}:
\begin{align}\label{equ:optimal-bellman}
	v_*(s) & = \max_{a \in \mathcal{A}} q_{\pi_*}(s, a) \nonumber                                                                                                             \\
	       & = \max_{a \in \mathcal{A}} \mathbb{E}_{\pi_*} [G_t \mid S_t = s, A_t = a] \nonumber                                                                              \\
	       & = \max_{a \in \mathcal{A}} \mathbb{E}_{\pi_*} [R_{t+1} + \gamma G_{t+1} \mid S_t = s, A_t = a] \quad (\text{using Equation}~\ref{equ:return_recursive})\nonumber \\
	       & = \max_{a \in \mathcal{A}} \mathbb{E} [R_{t+1} + \gamma v_*(S_{t+1}) \mid S_t = s, A_t = a] \nonumber                                                            \\
	       & = \max_{a \in \mathcal{A}} \sum_{s'\in \mathcal{S}}\sum_{r\in \mathcal{R}} p(s', r \mid s, a) \left[ r + \gamma v_*(s') \right].
\end{align}

Intuitively, the Bellman optimality equation states that the value of a state, under an optimal policy $\pi_*$, must correspond to the expected return from executing the best possible action from that state. The term "max" in $\max_{a \in \mathcal{A}} f(a)$ refers to the maximum value of function $f(a)$ when evaluated over all $a \in \mathcal{A}$. The fourth line of Equation~\ref{equ:optimal-bellman} exploits the fact that the state-value function of state \(s\) under the optimal policy \(\pi_*\) represents the discounted return the agent would obtain by following \(\pi_*\) from that state.

Note that the last line of Equation~\ref{equ:optimal-bellman} does not include the summation over actions that appears in the Bellman expectation equation (Equation~\ref{equ_bellman_equation}). In the optimality equation, the action is not sampled from a policy but instead fixed when evaluating the maximization operator. Because the expectation is conditioned on this specific action $a$, its probability is effectively 1, so no averaging over the action space is required. The only remaining expectation is over the possible next states and rewards dictated by the environment dynamics.

Similarly, there is a Bellman optimality equation for the action-value function \(q_*\), which recursively assumes that the agent will always take the best possible action in all future states \cite{sutton2018reinforcement}:
\begin{align*}
	q_*(s, a) & = \mathbb{E} \left[ R_{t+1} + \gamma \max_{a'} q_*(S_{t+1}, a') \mid S_t = s, A_t = a \right]                       \\
	          & = \sum_{s'\in\mathcal{S}} \sum_{r\in\mathcal{R}}p(s', r \mid s, a) \left[ r + \gamma \max_{a'} q_*(s', a') \right].
\end{align*}

This equation can be understood as follows: the value of taking action \(a\) in state \(s\) under the optimal policy is the expected immediate reward \(R_{t+1}\) plus the discounted value of the best possible action in the next state \(S_{t+1}\). By taking the maximum over all possible next actions \(a'\), \(q_*\) captures the notion of always following the optimal choice from every future state, which is the key idea behind optimal decision-making in reinforcement learning.

The Bellman optimality equations form a system of equations, with one equation for each state. For $n$ states, the system consists of $n$ equations with $n$ unknowns. Hence, if we are provided with the dynamics function $p$, this system can theoretically be solved for $v_*$ and $q_*$. Note that the Bellman optimality equations make no reference to any specific policy, thus allowing us to build an optimal policy from them, once these have been obtained.

Given that Bellman optimality equations take into account the reward consequences of all possible behavior given a start state $s$, the optimal action at each instant can be obtained with \textit{greedy} evaluation of $v_*$, that is, with a one-step search. Hence, this one-step ahead search still encompasses the long-term consequences of the immediate choice. Concretely, the optimal action for state $s$ maximizes the expected sum of the immediate reward and the discounted value of the next state $s'$, summarized in $v_*(s')$ \cite{sutton2018reinforcement}:
\begin{equation}\label{equ:optimal-policy}
	\pi_*(s)= \arg\max_{a} \sum_{s'\in\mathcal{S}} \sum_{r\in\mathcal{R}}p(s', r \mid s, a) \left[ r + \gamma v_*(s') \right].
\end{equation}

In this case, the policy is assumed to be deterministic. That is, instead of mapping state-action pairs to probabilities, it maps each state to a single action: $\pi_*: \mathcal{S}\rightarrow\mathcal{A}$. It is natural to define this optimal policy directly in terms of the Bellman optimality equation for the action-value function. Since $q_*(s, a)$ expresses the expected return obtained by taking action
$a$ in state $s$ and then following the optimal policy thereafter, selecting the optimal action simply amounts to choosing the action that maximizes this value. Thus, the optimal policy is given by \cite{sutton2018reinforcement}:

\begin{equation*}
	\pi_*(s) = \arg\max_{a} q_*(s, a).
\end{equation*}

This definition makes explicit that $q_*$ implicitly encodes long-term return: once
$q_*$ is known, the optimal policy follows immediately by acting greedily with respect to it.

As for the \textit{Bellman equation}, the \textit{Bellman optimality equation} (Equation~\ref{equ:optimal-bellman}) can be turned into an update rule that allows iterative approximation of the \textit{optimal state-value function} $v_*$. Formally, for all $k=0,1,2,\dots$, approximation $v_{k+1}$ is obtained from the preceding approximation $v_{k}$ as follows \cite{sutton2018reinforcement}:
\begin{align*}
	v_{k+1}(s) & \doteq \max_{a \in \mathcal{A}} \mathbb{E} [R_{t+1} + \gamma v_k(S_{t+1}) \mid S_t = s, A_t = a]                                                                 \\
	           & = \max_{a \in \mathcal{A}} \sum_{s'\in \mathcal{S}}\sum_{r\in \mathcal{R}} p(s', r \mid s, a) \left[ r + \gamma v_k(s') \right] , \quad \text{for all } s \in S.
\end{align*}

If there is a $v_*$, this iterative procedure guarantees that the sequence $\{v_k\}$  converges to $v_*$ as $k\rightarrow \infty$. As before, the sequence is stopped as soon as the difference between successive approximations is below a small enough threshold. Having obtained a sufficiently good approximation of $v_*$, it can be used to determine the optimal policy $\pi_*$ by applying Equation~\ref{equ:optimal-policy}. This dynamic programming algorithm is known as \textit{value iteration} and its pseudo-code is presented in Algorithm~\ref{alg:value-iteration}

\begin{algorithm}[h]
	\caption{Value Iteration (adapted from \cite{sutton2018reinforcement})}
	\begin{algorithmic}[1]
		\STATE \textbf{Input:} the dynamics function, $p$
		\STATE \textbf{Parameters:} threshold $\psi > 0$
		\STATE \textbf{Initialize} arbitrarily $V(s)$, for all $s \in \mathcal{S}^+$, except for $V(terminal) = 0$
		\STATE \textcolor{blue}{// determine $V$, such that $V \approx v_*$}
		\STATE \textbf{Loop:}
		\begin{ALC@g}
			\STATE $\Delta \leftarrow 0$
			\STATE \textbf{For each $s \in \mathcal{S}$:}
			\begin{ALC@g}
				\STATE $v \leftarrow V(s)$
				\STATE $V(s) \leftarrow \max_{a\in\mathcal{A}} \sum_{s'\in\mathcal{S}} \sum_{r\in\mathcal{R}}  p(s', r \mid s, a) \left[ r + \gamma V(s') \right]$
				\STATE $\Delta \leftarrow \max(\Delta, |v - V(s)|)$
			\end{ALC@g}
		\end{ALC@g}
		\STATE \textbf{Until $\Delta < \psi$}
		\STATE \textcolor{blue}{// determine $\pi$, such that $\pi \approx \pi_{*}$}
		\STATE \textbf{For each $s \in \mathcal{S}$:}
		\begin{ALC@g}
			\STATE $\pi(s) \leftarrow \arg\max_{a\in\mathcal{A}} \sum_{s'\in\mathcal{S}} \sum_{r\in\mathcal{R}}  p(s', r \mid s, a) \left[ r + \gamma V(s') \right]$
		\end{ALC@g}
		\STATE \textbf{Return} $\pi$
	\end{algorithmic}
	\label{alg:value-iteration}
\end{algorithm}

\subsubsection*{Example: Gridworld (continued) (adapted from \cite{sutton2018reinforcement})}

\begin{figure}[ht]
	\centering
	\begin{subfigure}[b]{0.225\textwidth}
		\centering
		\includegraphics[width=\textwidth]{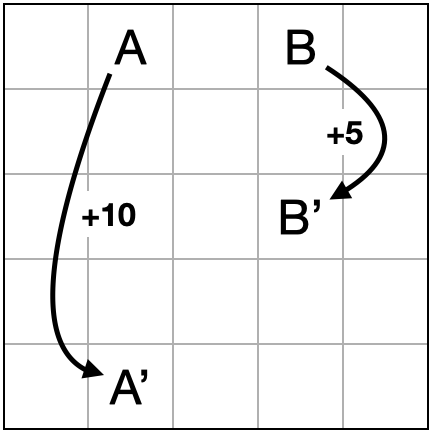}
		\caption{World rules.}
		\label{fig:sub1}
	\end{subfigure}
	\hfill
	\begin{subfigure}[b]{0.3\textwidth}
		\centering
		\includegraphics[width=\textwidth]{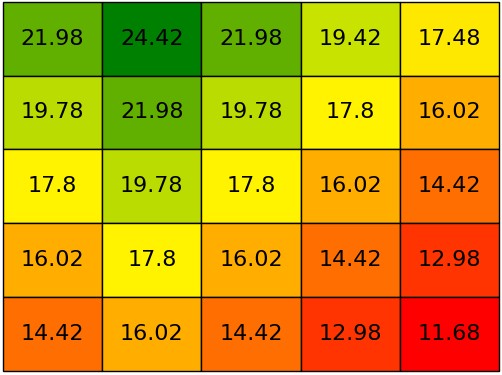}
		\caption{$v_*$, converged at $k=111$}
		\label{fig:sub1}
	\end{subfigure}
	\hfill
	\begin{subfigure}[b]{0.3\textwidth}
		\centering
		\includegraphics[width=\textwidth]{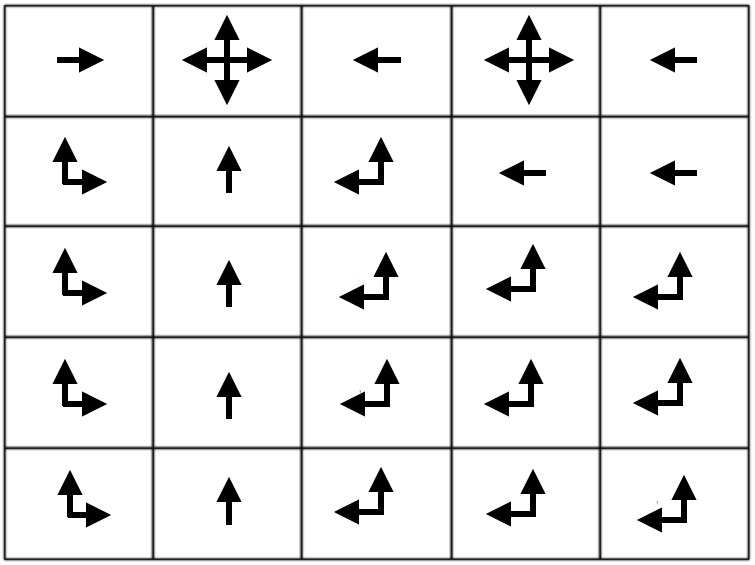}
		\caption{$\pi_*$}
		\label{fig:sub2}
	\end{subfigure}
	\caption{Value iteration for grid world with jumps and no walls. Left: the rules of the world: jumps positions $A\leftarrow A'$ and $B\leftarrow B'$. Middle: optimal value function for all states. Right: optimal policy, where the arrows indicate the optimal actions. (Adapted from Sutton and Barto \cite{sutton2018reinforcement})}
	\label{fig:value-iteration-example-A-B-no-walls}
\end{figure}

\begin{figure}[ht]
	\centering
	\begin{subfigure}[b]{0.225\textwidth}
		\centering
		\includegraphics[width=\textwidth]{figures/world_rewards.png}
		\caption{World rules.}
		\label{fig:sub1}
	\end{subfigure}
	\hfill
	\begin{subfigure}[b]{0.3\textwidth}
		\centering
		\includegraphics[width=\textwidth]{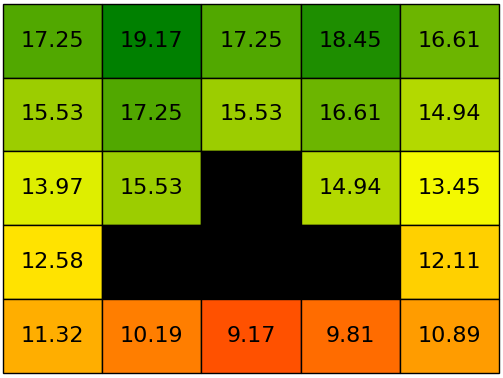}
		\caption{$v_*$, converged at $k=111$}
		\label{fig:sub1}
	\end{subfigure}
	\hfill
	\begin{subfigure}[b]{0.3\textwidth}
		\centering
		\includegraphics[width=\textwidth]{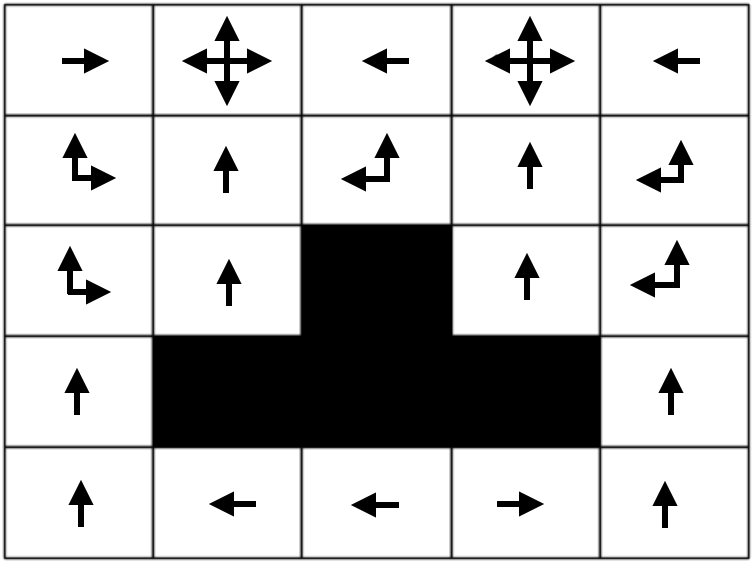}
		\caption{$\pi_*$}
		\label{fig:sub2}
	\end{subfigure}
	\caption{Value iteration for grid world with jumps and walls. Left: the rules of the world: jumps positions $A\leftarrow A'$ and $B\leftarrow B'$ and walls positions (black cells). Middle: optimal value function for all states. Right: optimal policy, where the arrows indicate the optimal actions. (Adapted and extended from Sutton and Barto \cite{sutton2018reinforcement})}
	\label{fig:value-iteration-example-A-B}
\end{figure}

Let us revisit the grid world example that contains positively rewarding jumps and negatively rewarding boundaries and walls. Figure~\ref{fig:value-iteration-example-A-B-no-walls} and Figure~\ref{fig:value-iteration-example-A-B} show the optimal value function $v_*$ and optimal policy $\pi_*$ for the grid world without and with walls, respectively. As expected, the optimal actions guide the agent to avoid walls and boundaries while attracting it towards the rewarding jumps in an endless loop. Note that in $A$ and $B$ all actions are optimal, because when reaching those states the agent deterministically jumps to $A'$ and $B'$, respectively, whatever the action it picks.

\ifincludeexercises
\clearpage
\section{Test Your Knowledge}

To consolidate your understanding, work through the exercises in Section~\ref{sec:exercises_MDP} on your own. After completing each exercise, review the step-by-step solution in Section~\ref{sec:solutions_MDP} to confirm your reasoning. Avoid consulting the detailed solutions before obtaining an answer yourself to ensure stronger problem-solving skills and deeper learning.

\subsection{Exercises}\label{sec:exercises_MDP}

\noindent \textbf{Exercise 1.} 
Consider the Markov Decision Process depicted in the followiung figure. 
Assume that the agent starts in the initial state indicated in the figure and that rewards are obtained immediately after each action execution. 

\begin{center}
	\includegraphics[width=12cm]{figures/mdp_example.png}
\end{center}

\begin{itemize}
	\item[a.] Which of the following action execution sequences yields a higher cumulative reward: \texttt{east-north-west-south} or \texttt{north-east-west-east}?
\end{itemize}

\noindent \textbf{Exercise 2.} Consider an agent moving along an ideal one-dimensional line at discrete time steps \(t = 0,1,2,\dots\). 
At each time step, the agent applies a force \(F_t \in \{-1,+1\}\) (the action), and no other forces act on the agent. 
Let the agent have mass \(m > 0\), position \(x_t\), and velocity \(v_t\). 
\begin{itemize}
	\item[a.] Can the system be modeled as a \emph{Markov Decision Process (MDP)} under the following definitions of the state:  (a) \(s_t = x_t\); (b) \(s_t = \{x_t, x_{t-1}\}\); and (c) \(s_t = (x_t, v_t)\)?
\end{itemize}

\noindent \textbf{Exercise 3.} Consider a Markov Decision Process with two states, \(\mathcal{S}=\{N,S\}\), representing \emph{North} and \emph{South}, and two actions, \(\mathcal{A}=\{\texttt{Up},\texttt{Down}\}\). High reward (food) may be found in both states, but it is more frequent in \(N\) than in \(S\). The environment dynamics joint model is reported in the following tables. 
\begin{center}
{\small\begin{tabular}{|c|c|c|c|c|} \hline \textbf{$s$} & \textbf{$a$} & \textbf{$s'$} & \textbf{$r$} & \textbf{$p(s',r \mid s,a)$} \\ \hline $N$ & Up & $N$ & $1.0$ & $0.56$ \\ \hline $N$ & Up & $N$ & $0.2$ & $0.14$ \\ \hline $N$ & Up & $S$ & $1.0$ & $0.06$ \\ \hline $N$ & Up & $S$ & $0.2$ & $0.24$ \\ \hline $N$ & Down & $N$ & $1.0$ & $0.24$ \\ \hline $N$ & Down & $N$ & $0.2$ & $0.06$ \\ \hline $N$ & Down & $S$ & $1.0$ & $0.14$ \\ \hline $N$ & Down & $S$ & $0.2$ & $0.56$ \\ \hline \end{tabular}} \hspace{0.5cm} {\small\begin{tabular}{|c|c|c|c|c|} \hline \textbf{$s$} & \textbf{$a$} & \textbf{$s'$} & \textbf{$r$} & \textbf{$p(s',r \mid s,a)$}  \\ \hline $S$ & Up & $N$ & $1.0$ & $0.64$ \\ \hline $S$ & Up & $N$ & $0.2$ & $0.16$ \\ \hline $S$ & Up & $S$ & $1.0$ & $0.04$ \\ \hline $S$ & Up & $S$ & $0.2$ & $0.16$ \\ \hline $S$ & Down & $N$ & $1.0$ & $0.16$ \\ \hline $S$ & Down & $N$ & $0.2$ & $0.04$ \\ \hline $S$ & Down & $S$ & $1.0$ & $0.16$ \\ \hline $S$ & Down & $S$ & $0.2$ & $0.64$ \\ \hline \end{tabular}}
\end{center}

\begin{itemize}
	\item[a.] What is the probability of reaching N from S is the agent selects Up?
	\item[b.] What is the probability of reaching N from S is the agent selects Down?
	\item[c.] What is the expected reward if the agent is in S and selects Up?
	\item[d.] What is the expected reward if the agent is in S and selects Down?
\end{itemize}

\noindent \textbf{Exercise 4.} Consider the grid-world example represented in the following figure?
\begin{center}
\includegraphics[width=8cm]{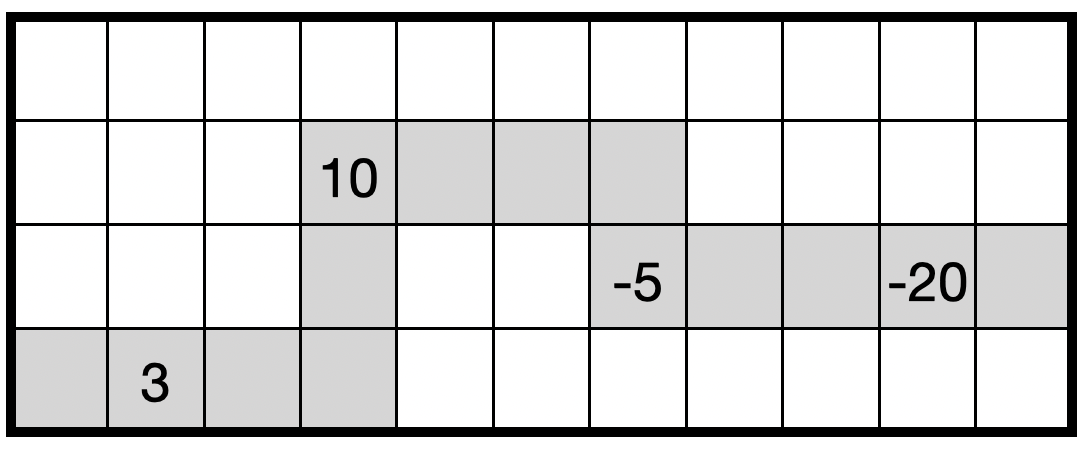}
\end{center}

\begin{itemize}
	\item[a.] What is the discounted return of the at $t=3$, assuming a discount factor of $\gamma=0.9$?
\end{itemize}

\noindent \textbf{Exercise 5.} Consider a Markov Decision Process with three states, \(\mathcal{S}=\{\texttt{A},\texttt{B},\texttt{C}\}\), representing the three cells in the following figure, and two actions, \(\mathcal{A}=\{\texttt{Left},\texttt{Right}\}\). The figure also includes the immediate reward $r$ earned by the agent in each cell. 
\begin{center}
\includegraphics[width=6cm]{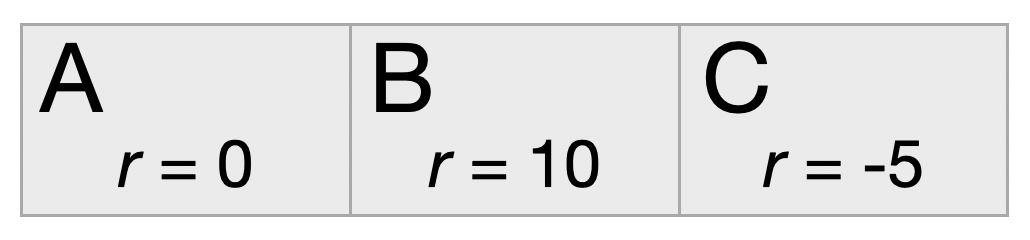}
\end{center}

\begin{itemize}
	\item[a.] What is the optimal policy $\pi$ for this MDP?
 \end{itemize}

\noindent \textbf{Exercise 6.} Consider an agent controlled by a policy $\pi_1$. Suppose we execute the agent from state $s$ under this policy. Assume that 30\%, 20\%, 10\%, and 40\% of the resulting trajectories coincide with trajectories $\tau_1$, $\tau_2$, $\tau_3$, and $\tau_4$, having cumulative rewards of 10, 5, $-5$, and 20, respectively. Now, suppose we execute the agent from the same state under a different policy $\pi_2$, resulting in a different distribution over trajectories: 10\%, 20\%, 30\%, and 40\%, respectively. 
\begin{itemize}
	\item[a.] Which policy is better?
\end{itemize}

\noindent \textbf{Exercise 7.} Consider the environment represented in the following figure, consisting of an infinite 1D grid of cells. We focus on three specific adjacent cells: $A$, $B$, and $C$. The immediate rewards associated with entering the cells are: $r(A)=0$, $r(B)=10$, and $r(C)=-5$. All other cells in the infinite sequence to the left of $A$ and to the right of $C$ have a reward of $0$. Assume the agent follows a deterministic policy $\pi$ that always moves Right, 
\( \pi(\text{Right} \mid s) = 1, \forall s \), and uses a discount factor of $\gamma = 0.9$.
\begin{center}
\includegraphics[width=10cm]{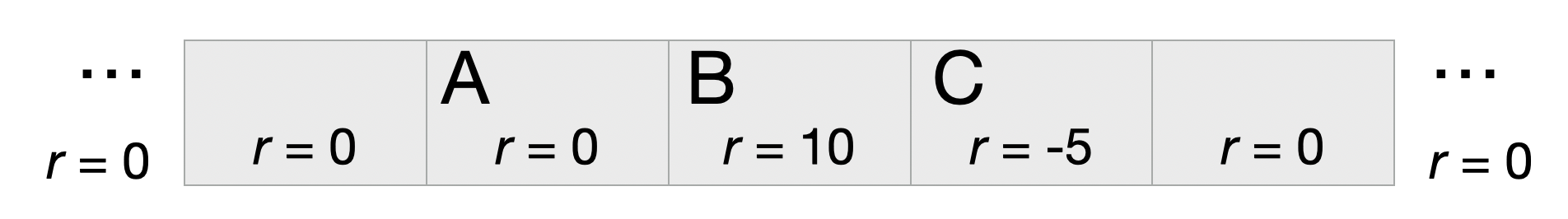}
\end{center}

\begin{itemize}
	\item[a.] What are the state-values $v_{\pi}(B)$ and $v_{\pi}(A)$?
\end{itemize}

\noindent \textbf{Exercise 8.} Consider the environment represented in the following figure, consisting of an infinite 1D grid of cells. Let us focus on three adjacent cells $A$, $B$, and $C$. The immediate rewards for entering these cells are known: $r(A)=0$, $r(B)=10$, and $r(C)=-5$. The rewards and dynamics of the cells outside this segment (to the left of A and right of C) are unknown. The agent follows a Uniform Random Policy, $\pi(\text{Left} \mid s)=\pi(\text{Right} \mid s)=0.5$, with a discount factor $\gamma = 0.9$. Assume an "Oracle" (or a separate calculation on the full map) has provided us with the true steady-state values for the neighboring cells, representing the expected discounted return from those states onwards: \( v_{\pi}(A) = 2.0 \quad \text{and} \quad v_{\pi}(C) = -3.0 \). 
\begin{center}
\includegraphics[width=10cm]{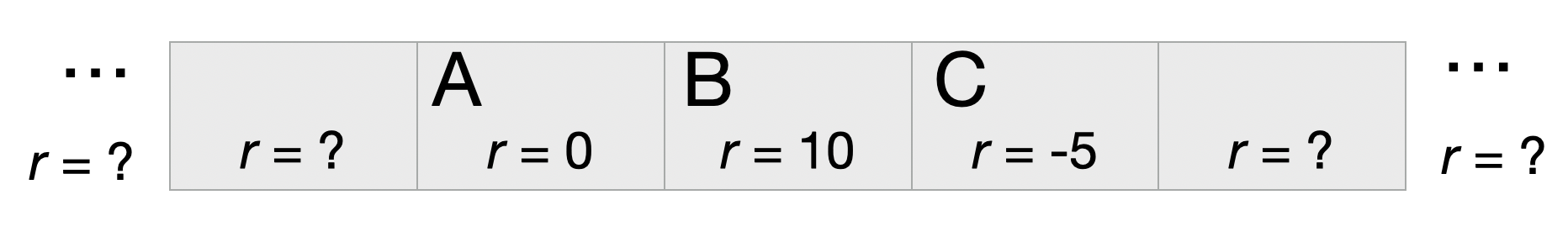}
\end{center}

\begin{itemize}
	\item[a.]What is the state-value of the center cell, $v_{\pi}(B)$?
\end{itemize}

\noindent \textbf{Exercise 9.} Consider the MDP described in Exercise 8. However, now assume that there is a strong wind blowing to the Right when the agent lands in $B$. As a result, moving Right in $B$ is aided by the wind and succeeds 100\% of the time. Moving Left fights the wind, succeeding only 60\% of the time, while 40\% of the time the agent is blown back and stays in state $B$, re-triggering the reward for $B$. 
\begin{itemize}
	\item[a.] Given the neighbor estimates $v_{\pi}(A) = 2.0$ and $v_{\pi}(C) = -3.0$, what is value for $v_{\pi}(B)$?
\end{itemize}

\subsection{Step-by-Step Solutions}\label{sec:solutions_MDP}

\noindent\textbf{Solution to Exercise 1.}  

\begin{itemize}
\item[a.]
The cumulative reward of a sequence is given by the sum of the rewards obtained at each transition.

Executing the sequence \texttt{east-north-west-south} yields the rewards
\[
0 + 10 - 100 + 0 = -90.
\]

Executing the sequence \texttt{north-east-west-east} yields the rewards
\[
-100 + 10 - 100 + 10 = -180.
\]

Since $-90 > -180$, the sequence \texttt{east-north-west-south} yields a higher cumulative reward and should therefore be preferred by the agent.
\end{itemize}

\noindent\textbf{Solution to Exercise 2.}

The dynamics of the system are:
\[
x_{t+1} = x_t + v_t, \qquad v_{t+1} = v_t + \frac{F_t}{m}.
\]

\begin{itemize}
\item[a.]  Case \(s_t = x_t\).

The next position \(x_{t+1}\) depends not only on the current position \(x_t\) but also on the current velocity \(v_t\). 
If the state includes only \(x_t\), the agent cannot unambiguously determine the next position. 
Therefore, the \emph{Markov property} is violated.

\item[b.] Case \(s_t = \{x_t, x_{t-1}\}\).

By including both the current and previous positions, the agent can estimate its current velocity as
\[
v_t \approx x_t - x_{t-1}.
\]
Using this approximation, the agent can predict the next position. 
While this approach does \emph{not} guarantee exact Markovian dynamics, it often works well in practice and can be seen as an \emph{augmented state} to partially restore the Markov property.

\item[c.] Case \(s_t = (x_t, v_t)\).

When the state explicitly includes both position and velocity, the agent can determine the next state exactly given the current state and applied force \(F_t\). 
This satisfies the \emph{Markov property exactly}, making the system a proper MDP.

\end{itemize}
\noindent\textbf{Solution to Exercise 3.}

Recall that the transition probabilities and expected rewards are obtained from the joint model as
\[
p(s' \mid s,a) = \sum_{r \in \mathcal{R}} p(s',r \mid s,a),
\]
\[
r(s,a) = \sum_{r \in \mathcal{R}} r \sum_{s' \in \mathcal{S}} p(s',r \mid s,a).
\]

\begin{itemize}
\item[a.] 	
	\begin{align*}
	p(N \mid S,\texttt{Up})
	&= \sum_{r \in \mathcal{R}} p(N,r \mid S,\texttt{Up}) \\
	&= p(N,1.0 \mid S,\texttt{Up}) + p(N,0.2 \mid S,\texttt{Up}) \\
	&= 0.64 + 0.16 \\
	&= 0.80.
	\end{align*}

\item[b.]
	\begin{align*}
	p(N \mid S,\texttt{Down})
	&= \sum_{r \in \mathcal{R}} p(N,r \mid S,\texttt{Down}) \\
	&= p(N,1.0 \mid S,\texttt{Down}) + p(N,0.2 \mid S,\texttt{Down}) \\
	&= 0.16 + 0.04 \\
	&= 0.20.
	\end{align*}

\item[c.]
\begin{align*}
	r(S,\texttt{Up})
	&= \sum_{r \in \mathcal{R}} r \sum_{s' \in \mathcal{S}} p(s',r \mid S,\texttt{Up}) \\
	&= 1.0 \cdot \bigl( p(N,1.0 \mid S,\texttt{Up}) + p(S,1.0 \mid S,\texttt{Up}) \bigr) \\
	&\quad + 0.2 \cdot \bigl( p(N,0.2 \mid S,\texttt{Up}) + p(S,0.2 \mid S,\texttt{Up}) \bigr) \\
	&= 1.0 \cdot (0.64 + 0.04) + 0.2 \cdot (0.16 + 0.16) \\
	&= 0.68 + 0.064 \\
	&= 0.744.
	\end{align*}

\item[d.]	
	\begin{align*}
	r(S,\texttt{Down})
	&= \sum_{r \in \mathcal{R}} r \sum_{s' \in \mathcal{S}} p(s',r \mid S,\texttt{Down}) \\
	&= 1.0 \cdot \bigl( p(N,1.0 \mid S,\texttt{Down}) + p(S,1.0 \mid S,\texttt{Down}) \bigr) \\
	&\quad + 0.2 \cdot \bigl( p(N,0.2 \mid S,\texttt{Down}) + p(S,0.2 \mid S,\texttt{Down}) \bigr) \\
	&= 1.0 \cdot (0.16 + 0.16) + 0.2 \cdot (0.04 + 0.64) \\
	&= 0.32 + 0.136 \\
	&= 0.456.
	\end{align*}
\end{itemize}

\noindent\textbf{Solution to Exercise 4.} 

\begin{itemize}
\item[a.] 
Recall that the expected discounted return is obtained with
\[G_t \doteq R_{t+1} + \gamma R_{t+2} + \gamma^2 R_{t+3} + \ldots\]

With $\gamma=0.9$, the return for $t=3$ is:
\[G_3 \doteq 10 \cdot 0.9^1 - 5 \cdot 0.9^5 - 20\cdot 0.9^8 = -2.56\]
\end{itemize}

\noindent\textbf{Solution to Exercise 5.} 

\begin{itemize}
\item[a.] 
\begin{align*}
\pi(A_t&=\texttt{Left} \mid S_t = \texttt{A}) = 0, \quad \pi(A_t=\texttt{Right} \mid S_t = \texttt{A}) = 1\\
\pi(A_t&=\texttt{Left} \mid S_t = \texttt{B}) = 1, \quad \pi(A_t=\texttt{Right} \mid S_t = \texttt{B}) = 0\\
\pi(A_t&=\texttt{Left} \mid S_t = \texttt{C}) = 1, \quad \pi(A_t=\texttt{Right} \mid S_t = \texttt{C}) = 0
\end{align*}
\end{itemize}

\medskip
\noindent\textbf{Solution to Exercise 6.}
\begin{itemize}
\item[a.] 
Recall that the state-value function is the expected return starting from state $s$:
\[
    v_{\pi}(s) \doteq \mathbb{E}_{\pi} [G_t \mid S_t = s].
\]
To solve this, we calculate the weighted average of the returns, where the weights are the probabilities of observing each trajectory. Based on the provided proportions:

\begin{align*}
    v_{\pi_1}(s) &= 0.3(10) + 0.2(5) + 0.1(-5) + 0.4(20) = 11.5, \\[6pt]
    v_{\pi_2}(s) &= 0.1(10) + 0.2(5) + 0.3(-5) + 0.4(20) = 8.5.
\end{align*}

Since $v_{\pi_1}(s) > v_{\pi_2}(s)$, $\pi_1$ is a better policy.
\end{itemize}

\noindent\textbf{Solution to Exercise 7.}

\begin{itemize}
\item[a.] 
Since the policy is deterministic, the randomness in the Bellman equation disappears, and $v_{\pi}(s)$ is simply the discounted sum of the sequence of rewards. The agent starts in $B$ and moves Right to $C$, where earns a reward of $r(C)=-5$. Then it moves Right from $C$ to $D$ and into infinite empty cells, where receives no reward. Hence, using the Bellman relation:

\begin{align*}
    v_{\pi}(B) &= r(C) + \gamma v_{\pi}(C) \\
               &= -5 + \gamma [r(D) + \gamma v_{\pi}(D)] \\
               &= -5 + 0.9 [0 + 0] \\
               &= -5.
\end{align*}

If the agent starts at state $A$ and moves Right to $B$ receives a reward of $r(B)=10$. The agent is now in state $B$. The value of the rest of the path is exactly $v_{\pi}(B)$, which we just calculated. Hence, using the Bellman relation:

\begin{align*}
    v_{\pi}(A) &= r(B) + \gamma v_{\pi}(B) \\
               &= 10 + 0.9(-5) \\
               &= 10 - 4.5 \\
               &= 5.5.
\end{align*}

Note how $v(A)$ is positive because it leads to the high-reward state $B$, but $v(B)$ is negative because the policy immediately forces the agent into the penalty state $C$. In conclusion, under policy $\pi$ the, it is more rewarding to visit the cell $A$ than the cell $B$.
\end{itemize}

\noindent\textbf{Solution to Exercise 8.}

\begin{itemize}
\item[a.] 
Since we are given the true steady-state values of the neighbors, we can solve for $v_{\pi}(B)$ directly using the Bellman equation. The values $v_{\pi}(A)$ and $v_{\pi}(C)$ implicitly account for all future rewards in the unknown regions of the map. Since the transition dynamics and reward function are deterministic, the Bellman equation simplifies (as only a single specific next state $s'$ and reward $r(s')$ exists for each action):
\[
v_{\pi}(B) = \sum_{a} \pi(a \mid B) \left[ r(s') + \gamma v_{\pi}(s') \right]
\]

Substituting the known transition dynamics for state $B$ and considering that the actions take the agent from $B$ to $A$ and from $B$ to $C$:
\begin{align*}
    v_{\pi}(B) &= 0.5 \left[ r(A) + \gamma v_{\pi}(A) \right] + 0.5 \left[ r(C) + \gamma v_{\pi}(C) \right] \\
    &= 0.5 \left[ 0 + 0.9(2.0) \right] + 0.5 \left[ -5 + 0.9(-3.0) \right] = -2.95.
\end{align*}
\end{itemize}

\noindent\textbf{Solution to Exercise 9.}

According to the statement, the transition dynamics is:
\begin{center}
{\small
\begin{tabular}{|c|c|c|c|c|}
\hline
$s$ & $a$ & $s'$ & $r$ & $p(s',r \mid s,a)$ \\ \hline
$B$ & Left & $A$ & $0$ & $0.6$ \\ \hline
$B$ & Left & $B$ & $10$ & $0.4$ \\ \hline
$B$ & Right & $C$ & $-5$ & $1.0$ \\ \hline
\end{tabular}
}
\end{center}

\begin{itemize}
\item[a.] 
The value of state $B$ is given by:
\[
v_{\pi}(B) = \sum_{a} \pi(a \mid B) \, q_{\pi}(B, a).
\]

The action-value is given by:
\[
q_{\pi}(B, a) = \sum_{s', r} p(s', r \mid B, a) \left[ r + \gamma v_{\pi}(s') \right].
\]

Using the table, if follows that:
\begin{align*}
    q_{\pi}(B, \text{Left}) &= 0.6 \left[ r(A) + \gamma v_{\pi}(A) \right] + 0.4 \left[ r(B) + \gamma v_{\pi}(B) \right] \\
    &= 0.6 \left[ 0 + 0.9(2.0) \right] + 0.4 \left[ 10 + 0.9 v_{\pi}(B) \right] \\
    &= 5.08 + 0.36 v_{\pi}(B)\\\\
    q_{\pi}(B, \text{Right}) &= 1.0 \left[ r(C) + \gamma v_{\pi}(C) \right] \\
    &= 1.0 \left[ -5 + 0.9(-3.0) \right] \\
    &= -7.7.
\end{align*}

Substitute these action-values in into the Bellman equation:
\begin{align*}
    v_{\pi}(B) &= 0.5 \cdot q_{\pi}(B, \text{Left}) + 0.5 \cdot q_{\pi}(B, \text{Right}) \\
&= 0.5 \left[ 5.08 + 0.36 v_{\pi}(B) \right] + 0.5 \left[ -7.7 \right] \\
&= -1.31 + 0.18 v_{\pi}(B)\\
    &\approx -1.598
\end{align*}
\end{itemize}

\fi

\chapter{Deep Reinforcement Learning}\label{cha:drl}
\label{sec:chapter_rl}

This chapter builds on the Markov Decision Process (MDP) formalism introduced earlier to present several foundational algorithms in Reinforcement Learning (RL). It begins with a discussion of some of the main types of RL algorithms, providing a conceptual overview of the field. The chapter then focuses on Policy Gradient methods, with particular emphasis on the REINFORCE algorithm, before transitioning to Deep RL techniques. Within this context, on-policy actor-critic methods are introduced and derived, followed by coverage of one of the most influential modern algorithms, Proximal Policy Optimization (PPO) \cite{schulman2017ppo}. When applicable, this chapter adopts the notation and equations of Sutton and Barto \cite{sutton2018reinforcement}, ensuring coherence with the standard reference textbook in reinforcement learning.

\section{Types of RL Algorithms}

As discussed in the previous chapter, it is theoretically possible to exactly solve the system of Bellman optimality equations to construct optimal policies. Dynamic programming methods, such as \textit{value iteration}, provide practical algorithms for computing these policies when a complete model of the environment, comprising both the \textit{dynamics function} and the \textit{reward function}, is available. These methods come with well-defined convergence guarantees, making them foundational tools in many domains.

Dynamic programming essentially reduces to planning optimal sequences of actions based on perfect knowledge of the environment. The known dynamics function specifies the consequences of each action, while the known reward function identifies which state-action pairs are desirable. Planning occurs entirely within the agent's "brain", prior to any interaction with the environment. Because the agent has full knowledge, no trial-and-error learning is necessary to discover optimal behavior.

However, exactly solving an MDP requires evaluating all state-action pairs in detail. In realistic problems, such as chess with its enormous discrete state-action space, or continuous control tasks like robotic locomotion, the number of possible state-action combinations grows exponentially. This leads to prohibitively high computational requirements, a phenomenon known as the \textit{curse of dimensionality}.

In most practical scenarios, the agent does not have access to the full environment dynamics. This is often the case in complex or hard-to-model environments, for example when physics is involved, when multiple agents interact, or when the environment is only partially observable. In these situations, the agent cannot directly solve the Bellman optimality equation using dynamic programming and therefore cannot plan its actions optimally in advance.

RL addresses these challenges by enabling the agent to gather information through interaction with the environment, rather than relying on a complete model. The agent uses trial-and-error experiences to update internal models, such as value functions or policies, which gradually improve over time. The performance of an RL algorithm is typically evaluated along two dimensions: \textit{effectiveness}, or the ability to approach the optimal policy, and \textit{efficiency}, which includes both computational cost (\textit{wall time}) and the number of interactions with the environment (\textit{sample efficiency}). These criteria often involve trade-offs. For instance, in virtual simulations, wall-time efficiency may be prioritized, whereas in real-world physical systems, minimizing environment interactions is crucial due to their cost or risk.

Although trial-and-error learning circumvents the need for complete knowledge, it introduces its own challenges. Due to the curse of dimensionality, it is usually impractical for the agent to visit every state and explore all possible actions. To make effective decisions in previously unseen states, the agent must \textit{generalize} knowledge acquired from experienced states to new ones. This is typically achieved using \textit{function approximation}, where a parameterized function estimates the value of states or the policy directly, rather than storing values in a table. Function approximators, such as neural networks, exploit their generalization capabilities to predict reasonable outputs for unobserved states, enabling RL to scale to large or continuous state spaces. Through trial-and-error, the agent refines these approximators over time, aiming to approximate the optimal solutions closely enough for practical purposes.

Approximation-based RL algorithms can be broadly categorized as \textit{model-free} or \textit{model-based} methods:

\begin{itemize}
	\item \textit{Model-free} methods do not rely on any explicit model of the environment. Instead, they learn directly from interactions with the environment, updating value functions or policies based on observed rewards and transitions. These methods are generally simpler to implement and tune, making them robust and widely applicable across a variety of problems. However, they tend to be less sample-efficient, often requiring a large number of interactions to achieve satisfactory performance.
	\item \textit{Model-based} methods, by contrast, construct and use a model of the environment to simulate interactions. By learning from these simulated experiences, they can improve sample efficiency, achieving good performance with fewer real-environment interactions. This is particularly valuable in scenarios where data collection is expensive, slow, or risky. Typically, the model is learned incrementally as the agent interacts with the environment. Despite their potential for higher sample efficiency, model-based methods are often more complex to implement and rely on accurately modeling the environment. In highly complex domains, learning an accurate model may be more challenging than directly learning an optimal mapping from states to actions.
\end{itemize}

Approximation-based RL algorithms can also be categorized as \textit{off-policy} or \textit{on-policy} methods:

\begin{itemize}
	\item \textit{Off-policy} methods use data collected from a different policy (the behavior policy) to update the policy being optimized (the target policy). Examples include Deep Q-Networks (DQN) \cite{mnih2015dqn}, Deep Deterministic Policy Gradient (DDPG) \cite{lillicrap2015ddpg}, and Soft Actor-Critic (SAC) \cite{haarnoja2018sac}. Off-policy methods are generally more sample-efficient because they can reuse past experiences multiple times to improve the policy. They also allow more flexible exploration strategies by decoupling the behavior policy from the target policy, and they can learn from diverse sources of data, such as demonstrations or experiences generated by other agents. This versatility makes off-policy methods particularly valuable in scenarios where collecting new interactions is costly or slow.
	\item \textit{On-policy} methods, in contrast, use data collected from the policy being optimized to update that same policy. In other words, the policy being improved is the one generating the data. Examples include REINFORCE \cite{williams1992reinforce} and Proximal Policy Optimization (PPO) \cite{schulman2017ppo}. On-policy methods tend to be more stable because they optimize the policy that is actively interacting with the environment. They are typically simpler to implement, as there is no need to maintain separate behavior and target policies. Additionally, on-policy methods adapt quickly to changes in the environment, and when sample collection is inexpensive, they can converge faster in terms of wall-clock time since updates are made continuously using fresh experiences.
\end{itemize}

\section{Policy Gradient Methods}\label{sec:policy_grad}

Given that \textit{model-free}, \textit{on-policy} methods are typically more stable, simpler to implement, and run in shorter wall clock times, they are ideal for pedagogical purposes. In addition, these methods have been pivotal in many of the latest breakthroughs in the field of Deep RL, making their learning essential for those who wish to contribute to the field. For these reasons, In this section, we explore \textit{model-free}, \textit{on-policy} methods for learning a \textit{parameterized policy} through trial-and-error, known as \textit{policy gradient methods}. A parameterized policy is a function governed by a set of learnable parameters. The objective of policy gradient methods is to approximate the optimal set of parameters by iteratively improving them as the agent interacts with the environment. The analysis will focus on the episodic setting; the continuing case follows a similar structure but requires a more elaborate derivation and analysis.

\subsection{Parameterized Policies}

Stochastic policies were introduced in the previous chapter as non-parametric functions of the form $\pi(a \mid s)$. They are called non-parametric because the probabilities they define are fixed, similar to entries in a table for the discrete case. In principle, these probabilities could be learned by filling in the appropriate values in the table. However, tabular representations have several limitations. First, they are suitable only for discrete action spaces. Second, large state-action spaces lead to tables that are prohibitively large. Third, reading from and writing to these tables are not differentiable operations, which complicates both the mathematical treatment and practical implementation of learning algorithms. Finally, tabular policies offer no direct mechanism for generalizing beyond the observed training data. For all these reasons, policies are typically defined as parameterized functions, where the parameters are adjusted through learning from experience to approximate the optimal policy.

Let $\boldsymbol{\theta}\in \mathbb{R}^d$ denote the vector of policy parameters, where $d$  is the dimensionality of the parameter space. In general, larger values of $d$ allow the policy to represent more complex state-action mappings. The policy is represented by a function $\pi:\mathcal{A}\times \mathcal{S}\times \mathbb{R}^d \rightarrow [0,1]$, which is defined as the probability that the agent selects action $a\in \mathcal{A}$ when in environment state $s$ under parameters $\boldsymbol{\theta}$ \cite{sutton2018reinforcement}:
\begin{equation*}
	\pi(a\mid s, \boldsymbol{\theta})\doteq P(A_t=a \mid S_t=s, \boldsymbol{\theta}_t=\boldsymbol{\theta}).
\end{equation*}

As in the non-parametric case, $\pi(\cdot \mid s, \boldsymbol{\theta})$ defines a probability distribution over $\mathcal{A}$ for each state $s \in \mathcal{S}$. In the parametric setting, however, this distribution is governed by the the parameter vector $\boldsymbol{\theta}$. The agent's action at time $t$ is determined by sampling from this distribution \cite{sutton2018reinforcement}:
\begin{equation*}
	A_t \sim \pi(\cdot \mid S_t, \boldsymbol{\theta}).
\end{equation*}

Determining an action by sampling means selecting actions stochastically rather than deterministically. That is, the probability of drawing a particular action is proportional to its probability mass or density under the distribution defined by the policy $\pi$. Hence, action selection via sampling from the policy-induced action distribution ensures that more rewarding actions are selected more frequently. Occasionally selecting less rewarding actions allows the agent to explore alternatives and gather information about the reward distribution. Balancing the exploration of new actions to gather information about their rewards and exploiting known actions to maximize cumulative reward is a cornerstone of RL. Sampling from the policy distribution effectively manages this trade-off.

The policy gradient methods, detailed in the following sections, can be applied to any policy parameterization, as long as $\pi(a \mid s, \boldsymbol{\theta})$ is differentiable with respect to its parameters $\boldsymbol{\theta}$. Among the various options, artificial neural networks form a particularly versatile family of functions. These networks are composed of simple units with no domain-specific assumptions, which makes them adaptable to a wide range of tasks. This flexibility explains why neural networks are often the preferred choice for representing parametric policies. In this chapter, policy gradient methods are first discussed in a general framework, and then, in Section~\ref{sec:non-linear-policies},  focus shifts to the use of neural networks for policy parameterization.

\subsection{Function Optimization with Gradient Ascent and Descent}

Before delving into the specifics of parametric policy functions, let us revisit the fundamentals of function optimization. In machine learning, a recurring task is to iteratively adjust a parameter vector $\boldsymbol{\theta}\in \mathbb{R}^n$ so as to (locally) maximize a differentiable objective function \(f: \mathbb{R}^n \to \mathbb{R}\), $\arg\max_{\boldsymbol{\theta}}f(\boldsymbol{\theta})$. Formally, the goal is to approximate $\arg\max_{\boldsymbol{\theta}} f(\boldsymbol{\theta})$, starting from an initial estimate $\boldsymbol{\theta}_0$. One of the simplest and most widely used approaches to this problem is \textit{gradient ascent}, which relies on evaluating the partial derivatives of $f$ at the current guess and using them to determine an improved parameter update in an iterative manner.

The gradient of \(f\) at a point \(\boldsymbol{\theta} \doteq (\theta_1, \theta_2, \ldots, \theta_n) \in \mathbb{R}^n\) is denoted by \(\nabla f(\boldsymbol{\theta})\) and is defined as a vector containing the function's partial derivatives with respect to $\boldsymbol{\theta}$ (recall that a partial derivative is just the derivative with respect to one of the variables while considering the other variables as constants):
\[
	\nabla f(\boldsymbol{\theta}) \doteq \left( \frac{\partial f}{\partial \theta_1}(\boldsymbol{\theta}), \frac{\partial f}{\partial \theta_2}(\boldsymbol{\theta}), \ldots, \frac{\partial f}{\partial \theta_n}(\boldsymbol{\theta}) \right).
\]

Intuitively, the gradient of a function is a vector that represents the direction of steepest ascent of function $f(\boldsymbol{\theta})$. That is, this vector indicates how much \(f\) changes as each component \(\theta_i\) of \(\boldsymbol{\theta}\) is varied. The direction of the gradient vector points in the direction of the greatest rate of increase of the function, whereas the magnitude of the vector represents the function's rate of change in that direction.

\begin{figure}
	\begin{center}
		\includegraphics[width=8cm]{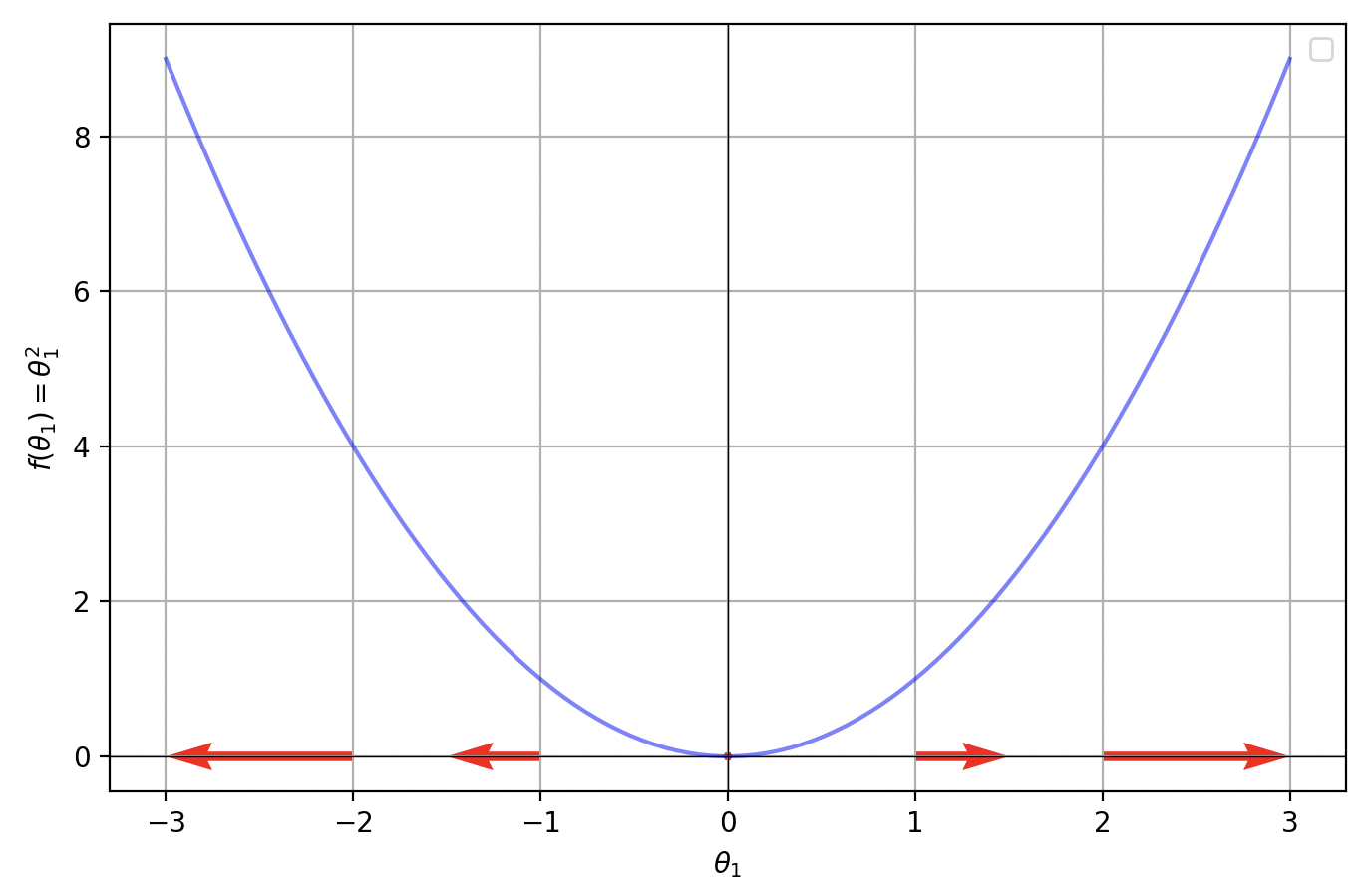}
		\caption{Gradient vectors (red arrows) of function $f(\theta_1)=\theta_1^2$ (blue curve). The origin of each vector indicates the point at which is evaluated.}
		\label{fig:grad_2d}
	\end{center}
\end{figure}

\begin{figure}
	\begin{center}
		\includegraphics[width=8cm]{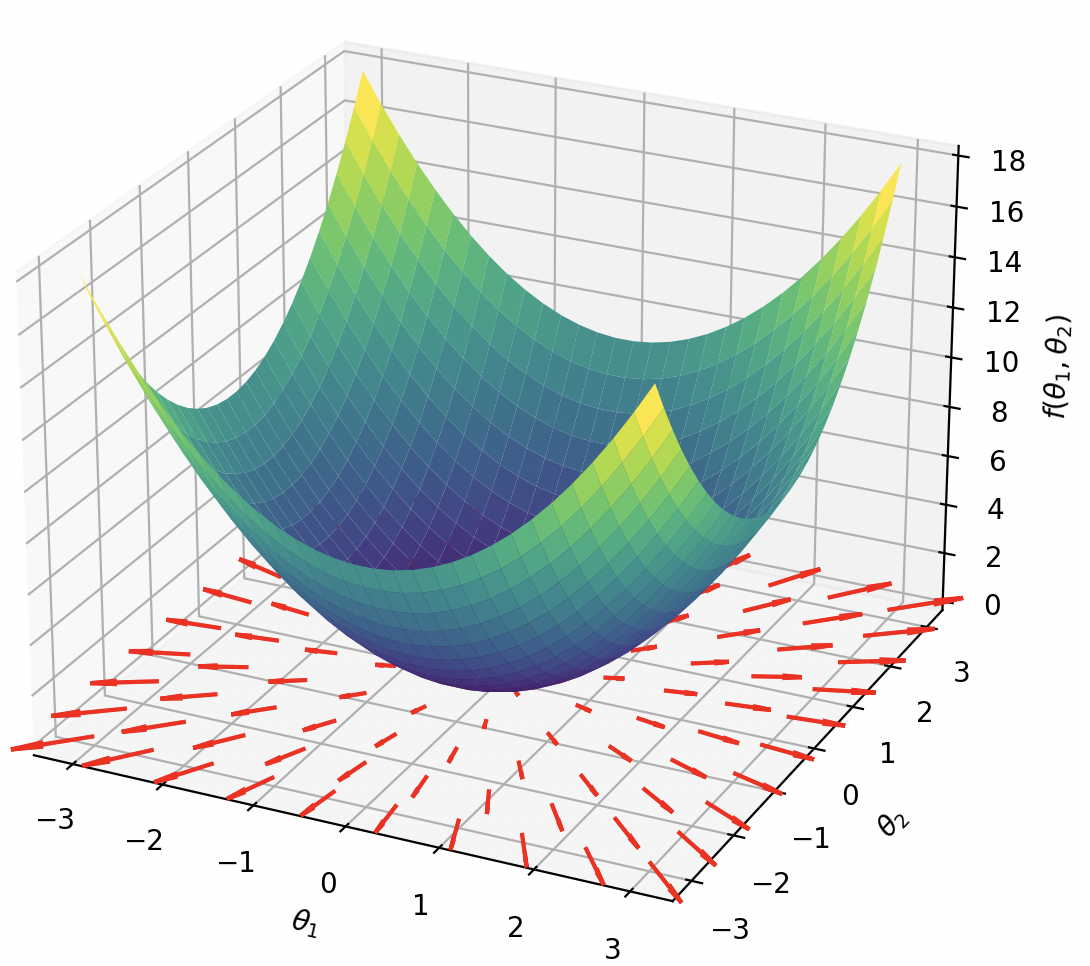}
		\caption{Gradient vectors (red arrows) of function $f(\theta_1, \theta_2)=\theta_1^2+\theta_2^2$ (greenish surface). The origin of each vector indicates the point at which is evaluated.}
		\label{fig:grad_3d}
	\end{center}
\end{figure}

To illustrate the concept of a gradient vector, Figure~\ref{fig:grad_2d} plots the gradient vectors of the function $f(\theta_1)=\theta_1^2$, which increases as the magnitude of $\theta_1$ grows. Accordingly, each gradient vector points in the direction of increasing magnitude. The length (amplitude) of the vectors also grows with $\theta_1$, since the slope of the function becomes steeper as $\theta_1$ moves away from 0, precisely captured by the derivative $f'(\theta_1)=2\theta_1$. Moving the parameter $\theta_1$ along this direction therefore increases the value of the function. The same reasoning applies in any number of dimensions. For example, Figure~\ref{fig:grad_3d} shows the gradient vectors of the two-dimensional function $f(\theta_1, \theta_2)=\theta_1^2+\theta_2^2$. Because this function depends on two parameters,  its gradient vectors are two-dimensional. As in the one-dimensional case, each vector points toward the direction in parameter space that yields the greatest increase in the function's value.

Let us now describe a process that leverages the gradient vectors of a function to iteratively move toward its maximum. Starting from an initial guess, $\boldsymbol{\theta}_0$, gradient ascent iteratively takes a small step in the direction of the gradient at each discrete time step $t = 0, 1, 2, 3, \ldots$, as it is the direction in parameter space that induces the fastest incrase in \(f\). The steps are taken with a magnitude that is proportional to the magnitude of the gradient, that is, the steeper the function, the larger the step. Formally, a gradient ascent steps is defined as follows:
\begin{equation*}
	\boldsymbol{\theta}_{t+1} \doteq \boldsymbol{\theta}_t + \alpha \nabla f(\boldsymbol{\theta}_t),
\end{equation*}

\noindent where $\alpha\in\mathbb{R}_+$ is a small enough \textit{step size}, also known as \textit{learning rate}. If $\alpha$ is too small, the convergence will be slow. A high $\alpha$ may lead to overshooting and divergence.  With a sufficiently small $\alpha$ and ensuring that certain assumptions on the function hold (e.g., continuity) the method is guaranteed to converge to a local maximum. Under stronger assumptions on the function (e.g., convexity), the method can converge to the global maximum.

The iterative process is repeated until some termination criterion is achieved, such as reaching a maximum number of iterations or the change in the function value or the parameter vector is below a certain  threshold $\epsilon$:
\begin{equation*}
	|f(\boldsymbol{\theta}_{t+1})-f(\boldsymbol{\theta}_t)|<\epsilon \quad \text{or} \quad ||\boldsymbol{\theta}_{t+1}-\boldsymbol{\theta}_{t}||<\epsilon.
\end{equation*}

\subsection*{Example}

\begin{figure}
		\begin{center}
			\includegraphics[width=10cm]{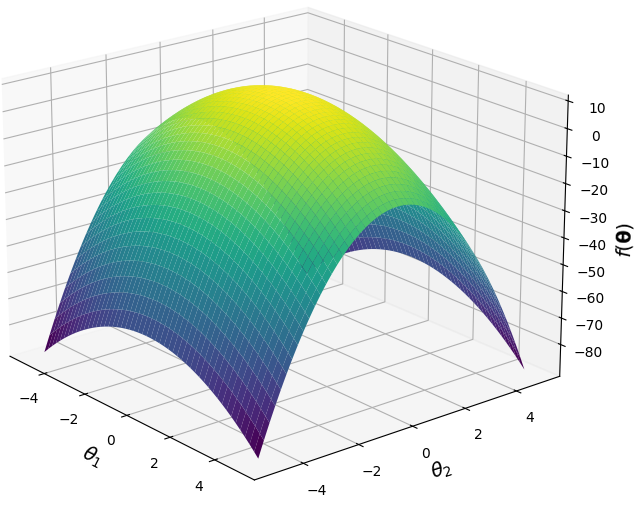}
			\caption{3D surface plot of the function $f(\boldsymbol{\theta}) = 10 - \theta_1^2 - 3\theta_2^2$.}
			\label{fig:grad-ascent-3d-example}
		\end{center}
	\end{figure}

To demonstrate an iteration of gradient ascent, let us assume a function \(f(\boldsymbol{\theta}) = 10 - \theta_1^2 - 3\theta_2^2\) with  \(\boldsymbol{\theta} = (\theta_1, \theta_2)$, plotted in Figure~\ref{fig:grad-ascent-3d-example}. The gradient of this function is
	\[
		\nabla f(\boldsymbol{\theta}) \doteq \left( \frac{\partial f}{\partial \theta_1}(\boldsymbol{\theta}), \frac{\partial f}{\partial \theta_2}(\boldsymbol{\theta}) \right)=(-2\theta_1, -6\theta_2).
	\]

\begin{figure}
		\begin{center}
			\includegraphics[width=11cm]{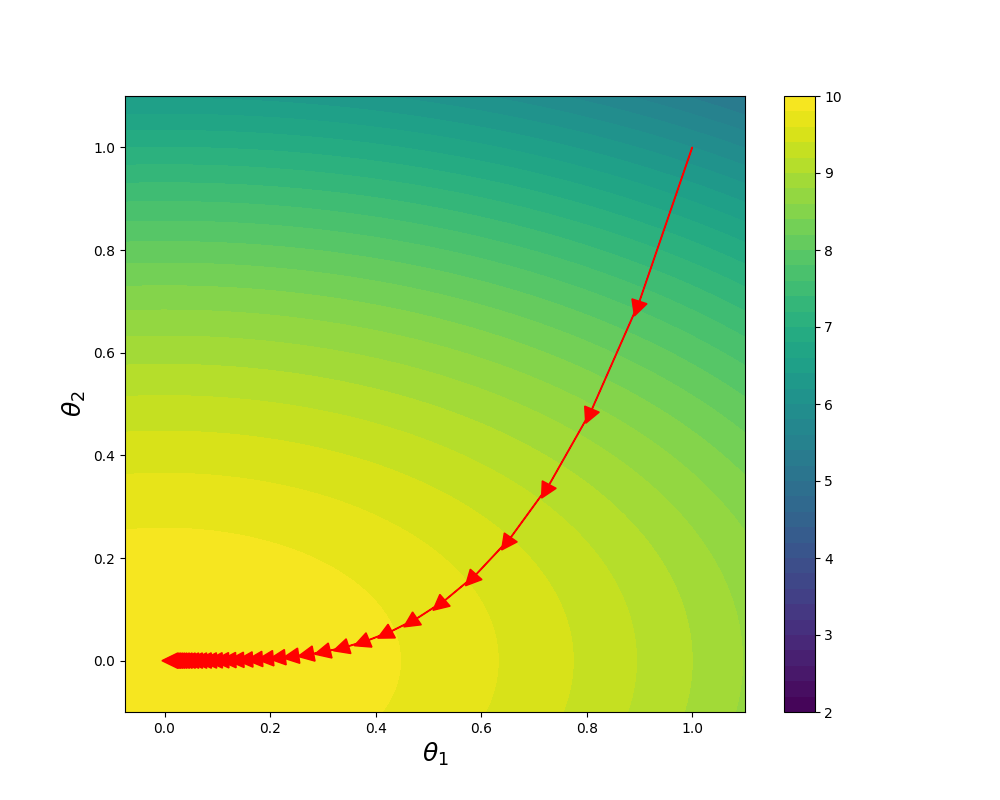}
			\caption{2D contour plot of  function $f(\boldsymbol{\theta}) = 10 - \theta_1^2 - 3\theta_2^2$, with gradient vectors as  red arrows.}
			\label{fig:grad-ascent-example}
		\end{center}
	\end{figure}

	At point \(\boldsymbol{\theta}_0=(1, 1)\), $f(\boldsymbol{\theta}_0)=10-1^2 - 3\times 1^2=6$, and $\nabla f(\boldsymbol{\theta}_0)=(-2\times 1, -6 \times 1)=(-2,-6)$. Note that the second coordinate of the gradient is three times higher than the first, indicating that a change in parameter $\theta_2$ induces a change in $f$ that is three times higher than a change in parameter $\theta_1$. Hence, if the goal is to maximize $f$, $\theta_2$ should be adjusted more strongly than $\theta_1$. Such a \textit{gradient ascent} step can be taken with $\alpha=0.05$ as follows: $\boldsymbol{\theta}_1=\boldsymbol{\theta}_0+0.05\nabla f(\mathbf{x_0})=(1,1)+0.05(-2,-6)=(0.9,0.7)$. $f$ can now be evaluated in this new point, $f(\boldsymbol{\theta}_1)=10-0.9^2 - 3\times 0.7^2=7.72$. Hence, by moving from $\boldsymbol{\theta}_0$ to $\boldsymbol{\theta}_1$, $f$ has been climbed from 6 to 7.72. Across iterations, this procedure slides both $\theta_1$ and $\theta_2$ in the direction of $0$, where the function has its maximum, 10. Figure~\ref{fig:grad-ascent-example} shows the 2D contour plot of the function with the first 35 gradient ascent vectors represented as red arrows. Note how progress in parameter space is faster along $\boldsymbol{\theta}_2$ (explained by the larger partial derivative) and how the algorithm is able to converge to the function's maximum at $\theta_1=\theta_2=0$.

	\subsubsection*{Gradient Descent}

	In some cases, instead of maximizing a given function, it may be necessary to minimize it. In the latter case, \textit{gradient descent} can be used instead of \textit{gradient ascent}. The iterative procedure is essentially the same, with the exception that the parameters are updated in the opposite direction of the function's gradient:
	\begin{equation*}
		\boldsymbol{\theta}_{t+1} \doteq \boldsymbol{\theta}_t - \alpha \nabla f(\boldsymbol{\theta}_t),
	\end{equation*}

	\noindent where \(f: \mathbb{R}^n \to \mathbb{R}\) is a differentiable function of the parameters \(\boldsymbol{\theta}\), \(\alpha\) is the learning rate, and \(\boldsymbol{\theta}_0\) is the initial guess for the parameters. The remaining procedures and convergence analysis follow those provided for the gradient ascent case.

	\subsubsection*{Learning Rate Decay}

	When optimizing a function, it is often beneficial to lower the learning rate as the optimization progresses, that is as $t$ varies. This approach allows for a more extensive exploration of the parameter space in the initial stages and gradually reduces the step sizes to refine the policy.

	There are various types of learning rate update schedules. One of the most commonly used is known as \textit{exponential decay}. In this method, the learning rate is adjusted either at every time step (for a smooth decay) or at specified intervals $\delta t$ (for a staircase decay) before the optimization step. This adjustment is based on an initial learning rate $\alpha_0$ and an exponential decay factor $\tau$ (e.g., updating the learning rate by $\tau=0.9$ every $\delta t=100$ time steps):
	\begin{equation}
		\alpha_t \doteq \alpha_0 \tau ^ {\frac{t}{\delta t}}.
	\end{equation}

	\subsection{Policy Gradient}

	Let $J(\boldsymbol{\theta}) \in \mathbb{R}$ denote a policy performance measure that evaluates how well the policy parameterized by $\boldsymbol{\theta}$, denoted $\pi_{\boldsymbol{\theta}}$, assigns higher probabilities to actions leading to high long-term returns and lower probabilities to actions yielding low returns. At this stage, we do not yet know how to update $\boldsymbol{\theta}$ so that the policy achieves high values of this performance measure. Deriving such an update rule is the objective of the following sections.

	The gradient of the performance measure, $\nabla_{\boldsymbol{\theta}} J(\boldsymbol{\theta}) \in \mathbb{R}^d$, with respect to the policy parameters $\boldsymbol{\theta}$ indicates how changes in the parameters influence the performance measure. Therefore, performance can be gradually maximized by adjusting the policy parameters in the direction of this gradient. The challenge, however, is that we do not have a differentiable expression for the performance measure to perform gradient ascent: it depends on the agent's interactions with the environment, and the dynamics function is not available. The best we can do to evaluate this function is to run the agent in the environment and collect data to estimate its value.

	Performance, and consequently its gradient, can be estimated by running the agent in the environment according to the current policy. The resulting stochastic estimate of the gradient, denoted by $\widehat{\nabla_{\boldsymbol{\theta}} J}(\boldsymbol{\theta})$, approximates the true gradient $\nabla_{\boldsymbol{\theta}} J(\boldsymbol{\theta})$ in expectation. That is, as the agent interacts more with the environment, the estimate becomes increasingly accurate. For now, we will not concern ourselves with how this estimate is actually constructed.

	Given a gradient estimate, the policy parameters can be updated by taking a \textit{stochastic gradient ascent} step on $J(\boldsymbol{\theta})$ with step size $\alpha$. Since $\widehat{\nabla_{\boldsymbol{\theta}} J}(\boldsymbol{\theta}_t)$ points in the direction of fastest increase of $J(\boldsymbol{\theta})$, this gradient ascent step adjusts $\boldsymbol{\theta}$ so that the agent becomes more likely to select actions under $\pi_{\boldsymbol{\theta}}$ that yield higher predicted performance according to $J(\boldsymbol{\theta})$. Formally, the gradient ascent update is \cite{sutton2018reinforcement}:
	\begin{equation}\label{equ:grad_asc}
		\boldsymbol{\theta}_{t+1} \doteq \boldsymbol{\theta}_t + \alpha \widehat{\nabla_{\boldsymbol{\theta}} J}(\boldsymbol{\theta}_t).
	\end{equation}

	\subsection{Policy Gradient Theorem}

	The Policy Gradient Theorem provides an analytic expression for the performance gradient with respect to the policy parameters, which can then be approximated to perform gradient ascent on $J(\boldsymbol{\theta})$ (Equation~\ref{equ:grad_asc}). In the episodic case, the theorem is derived from the following definition of the performance measure \cite{sutton2018reinforcement}:
	\begin{align}\label{equ:perform_state}
		J(\boldsymbol{\theta}) & \doteq v_{\pi_{\boldsymbol{\theta}}}(s_0),
	\end{align}

	\noindent where $s_0$ is the episode's initial state and $v_{\pi_{\boldsymbol{\theta}}}$ is the true value function for the policy determined by $\boldsymbol{\theta}$, $\pi_{\boldsymbol{\theta}}$. The agent does not know $v_{\pi_{\boldsymbol{\theta}}}$ a priori; if it did, there would be little value in learning. This definition informs the agent that maximizing the performance measure corresponds to finding the vector $\boldsymbol{\theta}$ that parameterizes the policy in such a way that trajectories starting from $s_0$ with higher expected returns become more probable. In other words, the goal is to adjust the policy so that no other policy can achieve higher long-term reward.

	Derived from Equation~\ref{equ:perform_state} by applying some calculus and by taking into account Equation~\ref{equ:state_action}, the \textit{policy gradient theorem} for the episodic case establishes that \cite{sutton2018reinforcement}:
	\begin{equation*}
		\nabla_{\boldsymbol{\theta}} J(\boldsymbol{\theta}) \propto \sum_{s\in\mathcal{S}} \mu_{\pi_{\boldsymbol{\theta}}}(s) \sum_{a\in\mathcal{A}} q_{\pi_{\boldsymbol{\theta}}}(s, a) \nabla_{\boldsymbol{\theta}} \pi(a \mid s, \boldsymbol{\theta}),
	\end{equation*}

	\noindent where \(\mu_{\pi_{\boldsymbol{\theta}}}\) is the state visitation probability distribution over \(\mathcal{S}\) under \(\pi_{\boldsymbol{\theta}}\) (i.e., \(\mu_{\pi_{\boldsymbol{\theta}}}(s)\) is the probability of visiting state \(s\)) and \(\propto\) signifies "proportional to". For the episodic scenario, the constant of proportionality is the average episode length, whereas for the continuing scenario it is 1.

	The formal proof of the Policy Gradient Theorem can be found in \cite{sutton2018reinforcement}. Nevertheless, we can provide some informal intuition underlying the theorem. According to Equation~\ref{equ:state_action}, the term $\sum_{a \in \mathcal{A}} \pi_{\boldsymbol{\theta}}(a \mid s, \boldsymbol{\theta}) q_{\pi_{\boldsymbol{\theta}}}(s, a)$ represents the value of state $s$ (i.e., the expected discounted return) under the policy $\pi_{\boldsymbol{\theta}}$, denoted $v_{\pi_{\boldsymbol{\theta}}}(s)$.

	If the gradient is disregarded for now, the performance objective $J(\boldsymbol{\theta})$ can be interpreted as a weighted average of the values of all possible states, where the weights are the probabilities of visiting those states under $\pi_{\boldsymbol{\theta}}$. This aligns with the intuition that optimizing $J(\boldsymbol{\theta})$ seeks the policy that induces the highest expected return.

	The gradient of $J(\boldsymbol{\theta})$ is simplified by assuming that only the policy $\pi_{\boldsymbol{\theta}}$ depends on the parameters $\boldsymbol{\theta}$, while all other elements of the system are treated as constants. Although state visitation and state values are influenced by the policy parameters, they are not treated as explicit functions of those parameters.

	Note that the policy gradient equation contains a sum over all states weighted by their occurrence probabilities, i.e., it computes an expected value over states according to the probabilities induced by $\pi_{\boldsymbol{\theta}}$. To rewrite the equation as an expectation the individual states $s$ are replaced by a random variable $S_t$ sampled according to the policy $\pi_{\boldsymbol{\theta}}$, respecting the state visitation probability \(\mu_{\pi_{\boldsymbol{\theta}}}\) \cite{sutton2018reinforcement}:

	\begin{align*}
		\nabla_{\boldsymbol{\theta}} J(\boldsymbol{\theta}) & \propto \sum_{s\in\mathcal{S}} \mu_{\pi_{\boldsymbol{\theta}}}(s) \sum_{a\in\mathcal{A}} q_{\pi_{\boldsymbol{\theta}}}(s, a) \nabla_{\boldsymbol{\theta}} \pi(a \mid s, \boldsymbol{\theta}) \\
		                                                    & = \mathbb{E}_{\pi_{\boldsymbol{\theta}}} \left[ \sum_{a\in\mathcal{A}} q_{\pi_{\boldsymbol{\theta}}}(S_t, a) \nabla_{\boldsymbol{\theta}} \pi(a \mid S_t, \boldsymbol{\theta}) \right] .
	\end{align*}

	To ensure that the samples $S_t$ follow the state-visitation frequency induced by the policy, they are simply the states visited by the agent when it interacts with the environment according to $\pi_{\boldsymbol{\theta}}$.

	The equation can be simplified in a few intuitive steps. First, we multiply and divide by the policy probability $\pi(a \mid S_t, \boldsymbol{\theta})$, which does not change the equality. This manipulation allows us to rewrite the sum over actions as an expectation under the policy $\pi_{\boldsymbol{\theta}}$, with each action weighted by its probability.
	Next, instead of summing over all actions, we can replace the action $a$ with a sample $A_t$ drawn from the policy, i.e., $A_t \sim \pi_{\boldsymbol{\theta}}(\cdot \mid S_t, \boldsymbol{\theta})$. This lets us explicitly express the expectation under $\pi_{\boldsymbol{\theta}}$ in a more practical, sample-based form.
	Finally, the action-value function $q_{\pi_{\boldsymbol{\theta}}}(S_t, A_t)$ can be approximated by the observed return $G_t$, since in expectation they are equal (see Equation~\ref{equ:return}). Putting these steps together, we obtain \cite{sutton2018reinforcement}:

	\begin{align}
		\nabla_{\boldsymbol{\theta}} J(\boldsymbol{\theta}) & = \mathbb{E}_{\pi_{\boldsymbol{\theta}}} \left[ \sum_{a\in\mathcal{A}} \pi(a \mid S_t, \boldsymbol{\theta}) q_{\pi_{\boldsymbol{\theta}}}(S_t, a) \frac{\nabla_{\boldsymbol{\theta}} \pi(a \mid S_t, \boldsymbol{\theta})}{\pi(a \mid S_t, \boldsymbol{\theta})} \right]\quad \text{(multiply \& divide by } \pi_{\boldsymbol{\theta}} \text{)}\nonumber \\
		                                                    & = \mathbb{E}_{\pi_{\boldsymbol{\theta}}} \left[ q_{\pi_{\boldsymbol{\theta}}}(S_t, A_t) \frac{\nabla_{\boldsymbol{\theta}} \pi(A_t \mid S_t, \boldsymbol{\theta})}{\pi(A_t \mid S_t, \boldsymbol{\theta})} \right] \quad \text{(substitute } a \text{ by  sample } A_t \sim \pi_{\boldsymbol{\theta}})\nonumber                                          \\
		                                                    & = \mathbb{E}_{\pi_{\boldsymbol{\theta}}} \left[ G_t \frac{\nabla_{\boldsymbol{\theta}} \pi(A_t \mid S_t, \boldsymbol{\theta})}{\pi(A_t \mid S_t, \boldsymbol{\theta})} \right] \quad \text{(given that } q_{\pi_{\boldsymbol{\theta}}}(S_t, A_t) = \mathbb{E}_{\pi_{\boldsymbol{\theta}}}[G_t \mid S_t, A_t])\label{equ:grad_exp}.
	\end{align}

	\subsection{The REINFORCE Algorithm}

	This section introduces a foundational model-free, on-policy algorithm that leverages the Policy Gradient Theorem to approximate an optimal policy. The algorithm, known as REINFORCE, was originally proposed by Williams \cite{williams1992reinforce}, and is presented here following the notation from Sutton and Barto \cite{sutton2018reinforcement}.

	As discussed in Section~\ref{sec:sampling-based}, expected values can be approximated by sampling from the underlying distribution and taking the mean of these samples. Hence, the expected value of the quantity inside the brackets in Equation~\ref{equ:grad_exp} can be approximated, and consequently, the gradient $\nabla J(\boldsymbol{\theta})$ can be estimated by sampling from the policy distribution ${\pi_{\boldsymbol{\theta}}}$. This estimate is then used to update the policy parameters $\boldsymbol{\theta}$ in the direction of higher expected return. Although each update is based on a noisy sample, averaging many updates over many episodes produces progressively more accurate policy improvements. The following describes this process in detail.

\begin{figure}[h]
		\centering
		\includegraphics[width=12cm]{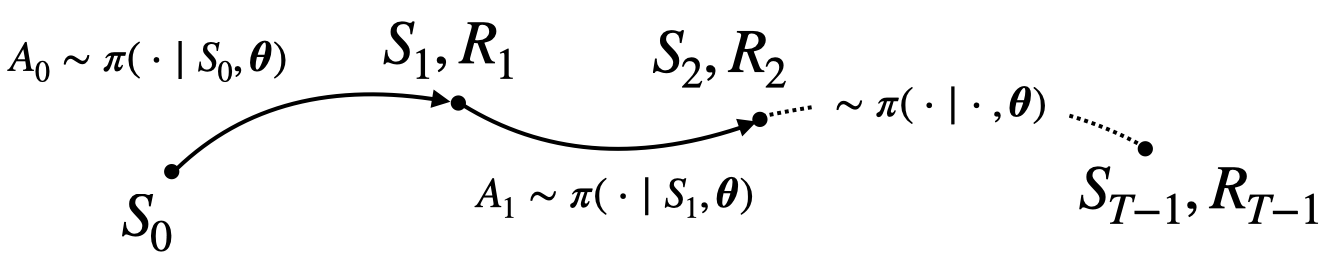}
		\caption{Illustration of a policy \textit{rollout}: the agent interacts with the environment over an episode, generating state-action-return tuples that provide unbiased samples of the policy gradient.}
		\label{fig:rollout_example}
	\end{figure}

	Sampling from the policy distribution $\pi_{\boldsymbol{\theta}}$ consists of executing the agent in the environment according to $\pi_{\boldsymbol{\theta}}$ over a complete episode of $T$ time steps. This process is commonly referred to as a policy \textit{rollout}, as illustrated in Figure~\ref{fig:rollout_example}.

	Once the rollout is complete, the generated trajectory can be analyzed to evaluate the quantity inside the brackets in Equation~\ref{equ:grad_exp} at each time step.  First, for each time step $t$ of the \textit{rollout}, the \emph{discounted return} is computed as the sum of all future rewards obtained from $t$ onward until the end of the episode \cite{sutton2018reinforcement}:
	\[G_t \doteq R_{t+1} + \gamma R_{t+2} + \gamma^2 R_{t+3} + \cdots + \gamma^{T-t-1}R_T,\]
	where $\gamma\in[0,1]$ is the \textit{discount factor}. Next, each tuple $(S_t, A_t, G_t)$ obtained from the rollout is used to compute \cite{sutton2018reinforcement}
	\begin{equation}
		G_t \, \frac{\nabla_{\boldsymbol{\theta}} \pi(A_t \mid S_t, \boldsymbol{\theta})}{\pi(A_t \mid S_t, \boldsymbol{\theta})},
		\label{equ:sample_policy_gradient}
	\end{equation}

	\noindent which, as discussed, constitutes an unbiased sample of the policy gradient $\nabla_{\boldsymbol{\theta}} J(\boldsymbol{\theta})$.

	It is now possible to perform a gradient-ascent update of the policy parameters $\boldsymbol{\theta}$ (recall Equation~\ref{equ:grad_asc}), using the sampled policy-gradient estimate in Equation~\ref{equ:sample_policy_gradient}. With learning rate $\alpha$, each sample yields the following parameter update \cite{sutton2018reinforcement}:
	\begin{align}
		\boldsymbol{\theta}_{t+1} & \doteq \boldsymbol{\theta}_t + \alpha G_t \frac{\nabla_{\boldsymbol{\theta}} \pi(A_t \mid S_t, \boldsymbol{\theta}_t)}{\pi(A_t \mid S_t, \boldsymbol{\theta}_t)}\nonumber
		\\&=\boldsymbol{\theta}_t + \alpha G_t \nabla_{\boldsymbol{\theta}}\ln \pi(A_t \mid S_t, \boldsymbol{\theta}_t) \quad (\text{because}~~ \nabla_{\boldsymbol{\theta}}\ln x= \nabla x / x).~\label{equ:update}
	\end{align}

	Let us shortly analyze the intuition behind the update rule in Equation~\ref{equ:update}. The gradient $\nabla_{\boldsymbol{\theta}}\ln \pi(A_t \mid S_t, \boldsymbol{\theta}_t)$ is a vector defined in policy parameter space that points in the direction of higher variation in the policy function $\pi(A_t \mid S_t, \boldsymbol{\theta}_t)$ for the given $S_t$ and $A_t$. That is, summing this vector to the current policy parameters $\boldsymbol{\theta}_t$ will push them in such a way that the policy function $\pi(A_t \mid S_t, \boldsymbol{\theta}_t)$ will return a higher value (probability) for the given $S_t$ and $A_t$. In other words, it will raise the probability of selecting the action $A_t$, when visiting state $S_t$.

	Given that the gradient is multiplied by the return $G_t$, when the latter is negative, the effect is the opposite, the parameters $\boldsymbol{\theta}_t$ are pushed in the opposite direction of the gradient $\nabla_{\boldsymbol{\theta}}\ln \pi(A_t \mid S_t, \boldsymbol{\theta}_t)$, in such a way that the policy function $\pi(A_t \mid S_t, \boldsymbol{\theta}_t)$ will return a lower value (probability) for the given $S_t$ and $A_t$. This way the parameters $\boldsymbol{\theta}_t$ are adjusted in order to raise the probability of selecting actions correlated with high return and to lower the probability of selecting actions correlated with low return.

\begin{figure}[h]
		\centering
		\includegraphics[width=14cm]{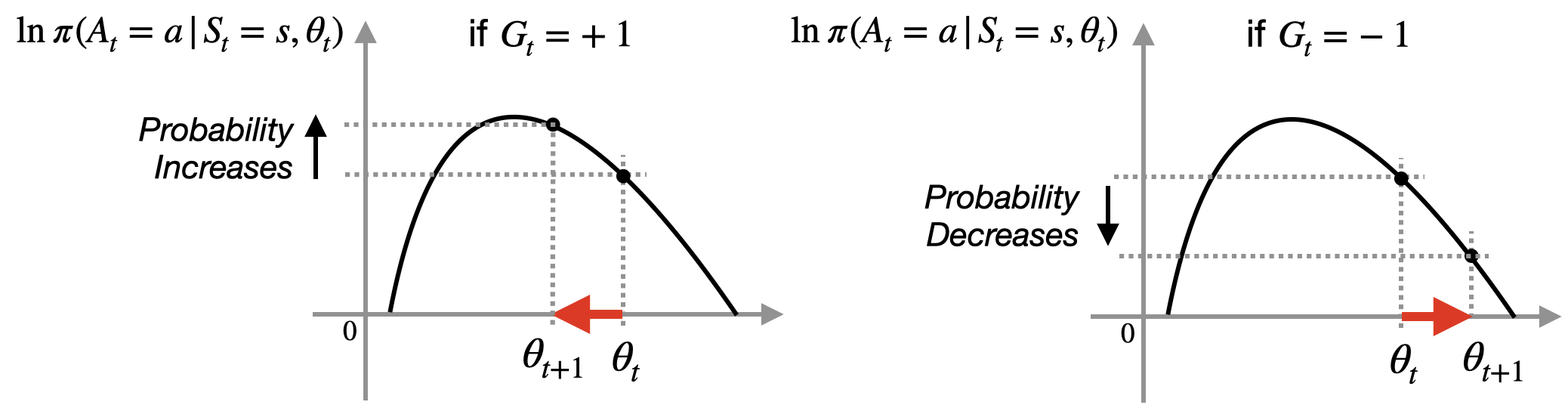}
		\caption{Visual illustration of the policy update step, where the red arrows represent the gradient vector associated with the executed action $a$ in state $s$, $\nabla_{\boldsymbol{\theta}}\ln \pi(A_t = a\mid S_t = s, \boldsymbol{\theta}_t)$.}
		\label{fig:prob_inc_dec}
	\end{figure}

	Figure~\ref{fig:prob_inc_dec} illustrates the policy-parameter update step for the simplified case of a single-parameter policy with a single maximum and a learning rate of $\alpha=1$ (a value chosen only for visualization, as it is far too large for practical use). In this example, a positive discounted return of $+1$ updates the policy parameter in the direction that \emph{increases} the probability of selecting the executed action $a$ in the state $s$ where it was taken. Conversely, if the agent instead obtained a discounted return of $-1$, the parameter update would move in the opposite direction, thereby \emph{decreasing} the probability of selecting that same action in that state.

	The description of REINFORCE \cite{williams1992reinforce} presented above is summarized in Algorithm~\ref{agorithm:reinforce}. The pseudo-code focuses on the episodic setting and incorporates an exponentially decaying learning rate. Note that, for readability, the discount factor $\gamma$ was omitted in parts of the earlier explanation but is explicitly included in the pseudo-code.

	\begin{algorithm}
		\caption{Episodic REINFORCE Algorithm with decaying learning rate (adapted from \cite{sutton2018reinforcement})}
		\begin{algorithmic}[1]
			\label{agorithm:reinforce}
			\STATE \textbf{Input:} a differentiable policy parameterization $\pi(a \mid s, \boldsymbol{\theta})$,
			\STATE \textbf{Parameters:} base learning rate $\alpha_0 > 0$,  discount factor $0\le\gamma\le 1$
			\STATE \textbf{Parameters:} learning decay rate, $0 < \tau \le 1$, decay interval, $\delta t > 0$
			\STATE \textbf{Parameters:} episode length $T$
			\STATE
			\STATE Initialize policy parameter $\boldsymbol{\theta} \in \mathbb{R}^{d}$ (e.g., randomly)
			\STATE Initialize global time step, $n\leftarrow0$
			\STATE \textbf{Loop forever}:
			\begin{ALC@g}
				\STATE \textcolor{blue}{// Generating an episode (policy rollout)}
				\STATE Observe initial state $S_{0}$
				\STATE \textbf{For each time step} $t = 0, 1, \ldots, T-1$:
				\begin{ALC@g}
					\STATE Sample action from policy, $A_t \sim \pi(\cdot \mid S_t, \boldsymbol{\theta})$
					\STATE Take action $A_t$, observe reward $R_{t+1}$ and next state $S_{t+1}$
				\end{ALC@g}
				\STATE \textcolor{blue}{// Policy update based on episode return}
				\STATE \textbf{For each step of the episode} $t = 0, 1, \ldots, T-1$:
				\begin{ALC@g}
					\STATE $n \leftarrow n + 1$
					\STATE $G \leftarrow \sum_{k=t+1}^T \gamma^{k-t-1} R_k$
					\STATE $\alpha \leftarrow \alpha_0 \tau ^ {\frac{n}{\delta t}}$
					\STATE $\boldsymbol{\theta} \leftarrow \boldsymbol{\theta} + \alpha \gamma^t G \nabla \ln \pi(A_t \mid S_t, \boldsymbol{\theta})$
				\end{ALC@g}
			\end{ALC@g}
		\end{algorithmic}
	\end{algorithm}

	\subsection{Baselines}

	As described, REINFORCE updates the policy parameters $\boldsymbol{\theta}$ in the direction of the estimated gradient $\nabla J(\boldsymbol{\theta})$, where this gradient is empirically approximated from samples generated by executing the current policy $\pi_{\boldsymbol{\theta}}$. The magnitude of each update is controlled by the learning rate $\alpha$ (or \emph{step size}). For sufficiently small $\alpha$, each update is expected to improve the performance objective $J$. Furthermore, when $\alpha$ is chosen according to standard stochastic approximation conditions (e.g., a decreasing sequence), REINFORCE is guaranteed to converge to a local optimum of $J$.

	REINFORCE is a \emph{Monte Carlo} algorithm because it estimates returns by sampling complete episodes generated under the current policy $\pi_{\boldsymbol{\theta}}$. This requires executing the agent in the environment and collecting sequences of states, actions, and rewards dictated by $\pi_{\boldsymbol{\theta}}$. By repeatedly sampling such trajectories and using their observed returns to update the policy parameters (implicitly averaging across episodes), REINFORCE approximates the expected return associated with each state--action pair, thereby guiding the policy toward higher performance.

	While REINFORCE provides an unbiased approximation of the expected return, it often suffers from high variance in its gradient estimates, which can lead to slow learning. This variance arises because the sampled return $G_t$ depends on the entire sequence of future actions and rewards collected until the end of the episode. Since such trajectories can vary substantially from one episode to another, the resulting gradient estimates fluctuate widely. Consequently, the learning signal becomes noisy, and a large number of sampled episodes may be required for the policy to approach optimality.

	A popular approach to reduce variance in the gradient estimate, and thereby make policy learning faster and more stable, is to generalize the policy gradient theorem to compare the action value against an arbitrary baseline function of the state $b(s)$ \cite{sutton2018reinforcement}:
	\begin{equation*}
		\nabla_{\boldsymbol{\theta}} J(\boldsymbol{\theta}) \propto \sum_{s\in\mathcal{S}} \mu_{\pi_{\boldsymbol{\theta}}}(s) \sum_{a\in\mathcal{A}} \left(q_{\pi_{\boldsymbol{\theta}}}(s, a)-b(s)\right) \nabla_{\boldsymbol{\theta}} \pi(a \mid s, \boldsymbol{\theta}),
	\end{equation*}

	Any function may be used as a baseline, provided it does not depend on the action $a$. Because it is independent of $a$, the baseline can be factored out of the summation over actions, which then sums to $1$ by the properties of a probability distribution. Since the gradient of a constant is zero, this term does not contribute to the gradient, and therefore the policy gradient theorem remains valid even after subtracting such a baseline \cite{sutton2018reinforcement}:
	\begin{equation*}
		\sum_{a\in\mathcal{A}} b(s) \nabla_{\boldsymbol{\theta}} \pi(a \mid s, \boldsymbol{\theta}) = b(s) \nabla_{\boldsymbol{\theta}} \sum_{a\in\mathcal{A}} \pi(a \mid s, \boldsymbol{\theta}) = b(s) \nabla_{\boldsymbol{\theta}} 1 = 0
	\end{equation*}

	Hence, using a baseline function does not alter the expected value of the gradient, provided that the baseline is independent of the action taken. In this case, the baseline term contributes nothing to the expectation of the gradient, affecting only its variance. Thus, a suitable baseline can significantly reduce variance without introducing bias. The policy gradient theorem with a baseline leads to the following update rule for the REINFORCE algorithm \cite{sutton2018reinforcement}:
	\begin{equation*}
		\boldsymbol{\theta}_{t+1} \doteq \boldsymbol{\theta}_t + \alpha \left(G_t-b(S_t)\right) \nabla_{\boldsymbol{\theta}}\ln \pi(A_t \mid S_t, \boldsymbol{\theta}_t).
	\end{equation*}

	A straightforward and effective choice for the baseline function is a differentiable approximation of the state-value function, denoted by $\hat{v}(S_t, \mathbf{w})$, where $\mathbf{w} \in \mathbb{R}^m$ is a parameter vector learned from experience (as discussed in the next section) \cite{sutton2018reinforcement}:
	\begin{equation*}
		b(S_t) \doteq \hat{v}(S_t, \mathbf{w}).
	\end{equation*}

	By using $G_t - \hat{v}(S_t, \mathbf{w})$ in the policy gradient update, actions that yield higher returns than the expected return under the current policy are reinforced. Specifically, actions for which $G_t - \hat{v}(S_t, \mathbf{w}) > 0$ increase in probability, making the agent more likely to select them in the future, thereby improving the policy. When the sampled return $G_t$ matches the expected return $\hat{v}(S_t, \mathbf{w})$, their difference is zero, resulting in no change to the action probabilities. This mechanism stabilizes learning by avoiding unnecessary updates. Directly using $G_t$ without a baseline can produce high-variance gradient estimates, leading to slow and unstable learning. Subtracting the baseline $\hat{v}(S_t, \mathbf{w})$ reduces the magnitude of the updates, lowering variance, and resulting in more stable and efficient policy improvement.

	\subsection{State-Value Function Approximation}

	Let us now discuss how the agent can incrementally learn an approximation of the true state-value function, $\hat{v}(s, \mathbf{w}) \approx v_{\pi_{\boldsymbol{\theta}}}(s)$, to be used as a baseline. This involves gradually adjusting the parameters $\mathbf{w}$ so that the predicted values $\hat{v}(s, \mathbf{w})$ closely match the true values $v_{\pi_{\boldsymbol{\theta}}}(s)$. However, the agent does not know the true state values in advance, so $v_{\pi_{\boldsymbol{\theta}}}(s)$ is not available. For the moment, let us assume it is known.

	To improve the approximation, we need a measure of the discrepancy between the predicted and true values, known as an \emph{error measure}. A common choice is the \emph{mean squared value error}, defined as follows \cite{sutton2018reinforcement}:
	\begin{align}
		\overline{VE}(\mathbf{w}) & \doteq \sum_{s \in \mathcal{S}} \mu(s) \left( v_{\pi_{\boldsymbol{\theta}}}(s) - \hat{v}(s, \mathbf{w}) \right)^2 \label{equ:squared-error}                                                                                                                  \\
		                          & = \mathbb{E}_{\pi_{\boldsymbol{\theta}}} \left[ \left( v_{\pi_{\boldsymbol{\theta}}}(S_t) - \hat{v}(S_t, \mathbf{w}) \right)^2 \right] \label{equ:squared-error-expectation} \quad \text{(replacing $s$ by the sample $S_t \sim \pi_{\boldsymbol{\theta}}$)}
	\end{align}

	\noindent where $\mu(s)$ is the steady-state probability of visiting state $s$ under policy $\pi_{\boldsymbol{\theta}}$.

	Let us gain some intuition on the mean squared value error. By weighting the squared error with the state visitation probability $\mu(s)$, this measure emphasizes prediction errors in frequently visited states more than those in rarely visited states. Consequently, even if $\hat{v}$ cannot accurately represent all states due to limited capacity (i.e., having fewer parameters than the total number of states), it prioritizes minimizing errors in the states that the agent actually encounters. The set of frequently visited states depends strongly on the policy $\pi_{\boldsymbol{\theta}}$ and is often sparse.

	The squared term in the equation ensures that all errors are positive and maintains differentiability. Additionally, it emphasizes larger errors more than smaller ones, directing the optimization process to focus on reducing significant errors in the search space.

	The weighted sum of squared errors in Equation~\ref{equ:squared-error} can be interpreted as an expected value, as explicitly written in Equation~\ref{equ:squared-error-expectation}. As discussed previously, expected values can be approximated by sampling from the underlying distribution and computing the mean of the resulting samples. In this context, this involves executing the agent in the environment according to the policy $\pi_{\boldsymbol{\theta}}$ over an episode (see Figure~\ref{fig:rollout_example}). By evaluating the quantity inside the brackets at each visited state $S_t$, for all $t \in \{0, \dots, T-1\}$, we obtain a set of samples whose average approaches the true expected value of the error.

	It is possible to define an update rule for the state-value approximation function in the same spirit as for the policy function, by using the gradient of the function being optimized. In this case, the function is a cost (or \emph{loss function}), so the objective is minimization rather than maximization. Accordingly, the update is performed using \emph{stochastic gradient descent} \cite{sutton2018reinforcement}:
	\begin{align}
		\mathbf{w}_{t+1} & \doteq \mathbf{w}_t - \frac{1}{2} \alpha \nabla_{\mathbf{w}_t} \left[ v_{{\pi_{\boldsymbol{\theta}}}}(S_t) - \hat{v}(S_t, \mathbf{w}_t) \right]^2 \nonumber                                                      \\
		                 & = \mathbf{w}_t + \alpha \left[ v_{{\pi_{\boldsymbol{\theta}}}}(S_t) - \hat{v}(S_t, \mathbf{w}_t) \right] \nabla_\mathbf{w} \hat{v}(S_t, \mathbf{w}_t) \quad \text{(applying chain rule)}\label{equ:w-update-pi}.
	\end{align}

	To gain some intuition on Equation~\ref{equ:w-update-pi}, recall that the gradient of a function is a vector in parameter space pointing in the direction of steepest increase. Thus, $\nabla_\mathbf{w} \hat{v}(S_t, \mathbf{w}t)$ is the vector that, when added to the parameters $\mathbf{w}$, most rapidly increases the predicted value $\hat{v}(S_t, \mathbf{w}t)$. This gradient is scaled by the prediction error, $v_{{\pi{\boldsymbol{\theta}}}}(S_t) - \hat{v}(S_t, \mathbf{w}_t)$.
	Hence, if the predicted value $\hat{v}(S_t, \mathbf{w}t)$ exceeds the actual value $v_{{\pi_{\boldsymbol{\theta}}}}(S_t)$, the prediction error becomes negative, effectively reversing the direction of the gradient. In such cases, the update reduces the predicted value, guiding it closer to the true value. Conversely, if the prediction underestimates the actual value, the gradient update increases it, again moving toward the correct target.

	Equation~\ref{equ:w-update-pi} assumes that $v_{\pi_{\boldsymbol{\theta}}}(S_t)$ is known, which is not the case in practice. Indeed, learning an approximation would be unnecessary if the true state values were available. Therefore, to make the update practically useful while preserving convergence guarantees, we replace $v_{\pi_{\boldsymbol{\theta}}}(S_t)$ with an unbiased, noisy estimate $U_t$, satisfying the following \cite{sutton2018reinforcement}:
	\[
		\mathbb{E}_{\pi_{\boldsymbol{\theta}}}[U_t \mid S_t = s] = v_{\pi_{\boldsymbol{\theta}}}(s).
	\]
	This substitution allows learning from sampled returns without requiring knowledge of the true state-value function. Given that the agent interacts with the environment according to $\pi_{\boldsymbol{\theta}}$, and that under these conditions the true value of a state is the expected return following it, the sampled return $G_t$ is by definition an unbiased estimate of $v_{\pi_{\boldsymbol{\theta}}}(S_t)$. Consequently, we can safely set $U_t \doteq G_t$, yielding the following update equation for the state-value approximation \cite{sutton2018reinforcement}:
	\begin{equation}\label{equ:update_w_params}
		\mathbf{w}_{t+1} \doteq \mathbf{w}_t + \alpha \left[ G_t - \hat{v}(S_t, \mathbf{w}_t) \right] \nabla_\mathbf{w} \hat{v}(S_t, \mathbf{w}_t) .
	\end{equation}

\ifincludeexercises
\subsection{Test Your Knowledge}

To consolidate your understanding, work through the following exercise on your own. After completing the exercise, review the step-by-step solution to confirm your reasoning. Avoid consulting the detailed solutions before obtaining an answer yourself to ensure stronger problem-solving skills and deeper learning.
\vspace{0.5cm}

\noindent\textbf{The Problem}

Consider a quadruped robot moving along a one-dimensional track that contains gaps in the floor (Figure~\ref{fig:quad_exercise}). Time is discrete and indexed by \(t = 0, 1, 2, \ldots\). At each time step, the robot receives a scalar observation \(x_t \in \mathbb{R}\) from a forward-facing range sensor that measures the distance to the ground ahead. Based on this continuous observation, the robot must choose one of two discrete actions: \(a_t = 0\) (Walk) or \(a_t = 1\) (Jump).
\begin{figure}[h]
\begin{center}
\includegraphics[width=8cm]{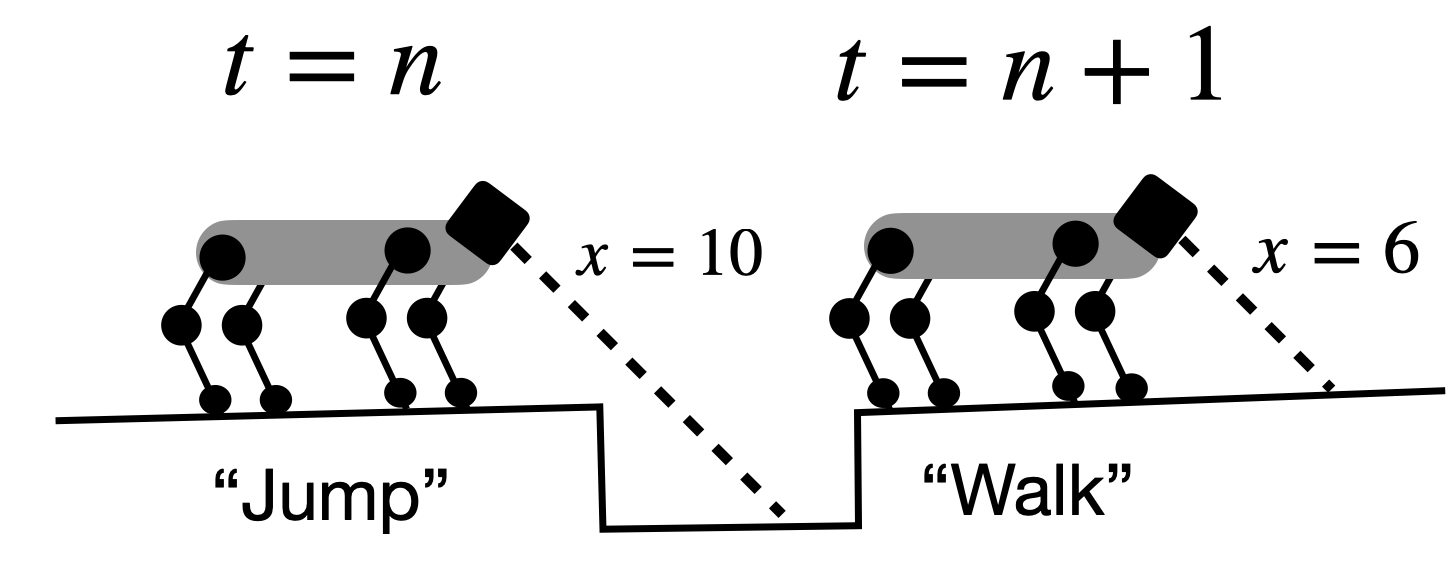}
\caption{Diagram of the quadruped robot exercise.}
\label{fig:quad_exercise}
\end{center}
\end{figure}

\noindent\textbf{The Task}

You are asked to design a stochastic, parametric policy \(\pi(a \mid x, \theta)\) that maps the sensor reading \(x\) to a probability distribution over actions. The learning objective is to adjust the policy parameters so that the probability of walking or jumping depends meaningfully on the sensor reading, for example, jumping should become more likely as the detected gap increases.

You will model the policy using a Bernoulli probability distribution. A Bernoulli random variable \(A \in \{0,1\}\) is fully characterized by a single parameter \(p \in [0,1]\), where \(p\) denotes the probability that the outcome equals \(1\). The Bernoulli distribution can be seen as the special case of the Softmax distribution for binary actions. Its probability mass function is given by
\[
\mathbb{P}(A = 1) = p, 
\qquad 
\mathbb{P}(A = 0) = 1 - p,
\quad \text{or equivalently} \quad
\mathbb{P}(A = a) = p^{a}(1-p)^{1-a}.
\]

In the context of decision making, the value \(a = 1\) can be interpreted as selecting one option (e.g., \emph{Jump}), while \(a = 0\) corresponds to selecting the alternative (e.g., \emph{Walk}). The parameter \(p\) therefore controls how likely the first option is chosen relative to the second.

To construct such a policy, you will need to map real-valued preferences to the interval \([0,1]\). This can be achieved using the sigmoid function \(\sigma : \mathbb{R} \rightarrow [0,1]\), which transforms an input \(z \in \mathbb{R}\) into a valid probability:
\[
\sigma(z) = \frac{1}{1 + e^{-z}},
\qquad
1 - \sigma(z) = \frac{e^{-z}}{1 + e^{-z}}.
\]

In practice, the input \(z\) to the sigmoid will be defined as a function of the sensor reading \(x\) and the policy parameters \(\theta\). By choosing different functional forms for \(z(x; \theta)\), the policy can assign different probabilities to the same sensor input. Figure~\ref{fig:sigmoids} illustrates the effect of different definitions of \(z\) on the resulting sigmoid curves.

\begin{figure}[h]
\begin{center}
\includegraphics[width=12cm]{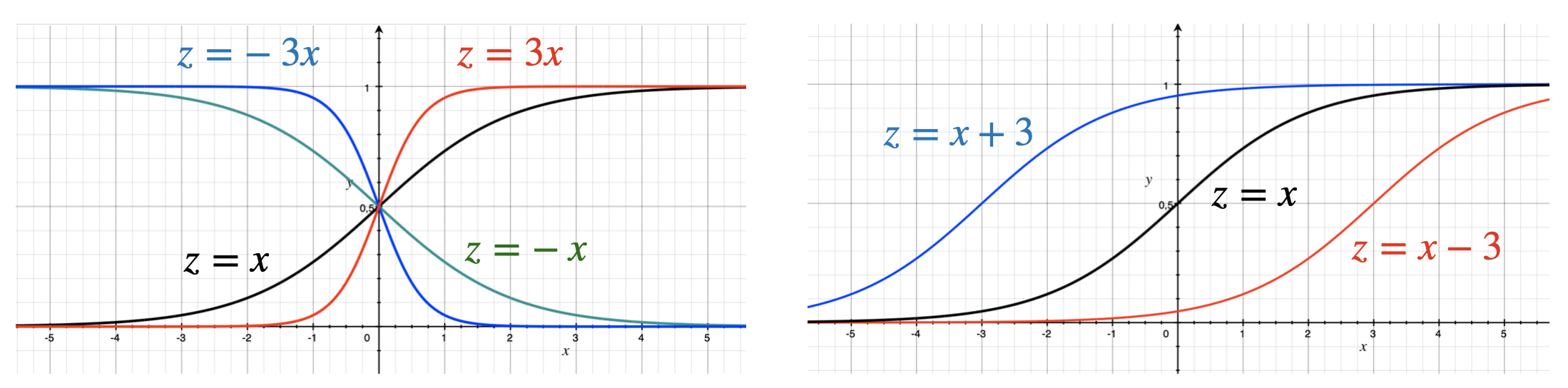}
\caption{Sigmoid function with $z$ varying according to a given $x$.}
\label{fig:sigmoids}
\end{center}
\end{figure}

\medskip
\medskip
\noindent\textbf{Questions}

\begin{itemize}
     \item[a.] Explain why a Bernoulli distribution is more appropriate for modeling the policy in this setting than a Softmax, Gaussian, or Uniform distribution. Justify your answer by referring explicitly to the properties of the state space and the action space.

    \item[b.] Assume a linear model with an explicit bias term. Explain how to parameterize the chosen distribution so that it defines a valid state-dependent stochastic policy \(\pi(a \mid x, \theta)\).

    \item[c.] We wish to train this robot using the \textsc{REINFORCE} algorithm, which requires the gradient of the log-probability of the action taken, \(\nabla_\theta \log \pi(a \mid x, \theta)\). Derive this gradient explicitly for the policy defined in (b).

    \item[d.] Which policy parameter(s) control the threshold on the sensor reading \(x\) at which jumping becomes more likely than walking? Derive this threshold as a function of the parameters. Which parameter controls how \emph{decisive} the agent is around this threshold? Discuss in which settings it is advantageous to allow the agent to learn to be less decisive, and in which settings this is not desirable.

    \item[e.] Consider the \textsc{REINFORCE} update rule,
    \[
    \theta \gets \theta + \alpha\, R \, \nabla_\theta \log \pi(a \mid x, \theta),
    \]
    with current policy parameters \(w = 1.0\), \(b = -2.0\), and learning rate \(\alpha = 0.1\). Compute the parameter updates in the following cases and briefly analyze the effect of each update on the policy:

    \begin{enumerate}
    \item 
    \(x = 3.0\), \(a = 1\) (Jump), \(R = +1\implies \)
    Increase \(p(x)\) for jumping here.

    \item 
    \(x = 1.0\), \(a = 0\) (Walk), \(R = +1 \implies\)
    Increase \(p(x)\) for walking here.

    \item 
    \(x = 3.0\), \(a = 0\) (Walk), \(R = -1 \implies\)
    Decrease \(p(x)\) for walking here.

    \item 
    \(x = 1.0\), \(a = 0\) (Jump), \(R = -1 \implies\)
    Decrease \(p(x)\) for jumping here.
\end{enumerate}

\end{itemize}

\noindent\textbf{Step-by-Step Solutions}

\begin{itemize}
\item[a.] 
\begin{itemize}
    \item Bernoulli: Naturally models binary actions and directly represents the probability of jumping versus walking.
    \item Softmax: For two actions, it reduces to a Bernoulli distribution with a logistic parameterization, making it unnecessarily general.
    \item Gaussian: Unsuitable because the action space is discrete.
    \item Uniform: Cannot depend on the state and therefore cannot represent a meaningful policy.
\end{itemize}

\item[b.] 

Let us define a stochastic policy for a binary action space using a Bernoulli distribution. To make the policy adaptive to the environment, the probability \(p\) of selecting action \(a = 1\) (Jump), and correspondingly \(1 - p\) of selecting action \(a = 0\) (Walk), must depend on the sensor observation \(x\). This dependence is governed by a set of learnable parameters. Intuitively, as the detected gap increases, the Jump action should become more likely. Consequently, the Bernoulli parameter is modeled as a parametric, state-dependent function \(p(x;\theta)\). Formally, the policy is defined as
\[
\pi(A=1 \mid x, \theta) = p(x;\theta),\quad \pi(A=0 \mid x, \theta) = 1-p(x;\theta).
\]
The learning problem is to find the parameters \(\theta\) that induce action-selection probabilities maximizing the expected cumulative reward.

Let us now describe how to parameterize \(p(x;\theta)\). As required, we begin by applying a linear transformation with an explicit bias term to the observation \(x\):
\[
z(x;\theta) = w x + b,
\]
where \(\theta = (w, b)\), with \(w \in \mathbb{R}\) denoting the slope and \(b \in \mathbb{R}\) the bias. Together, these parameters define a probabilistic decision threshold over the sensor input \(x\): the bias \(b\) determines the location of the threshold (i.e., the value of \(x\) at which jumping becomes likely), while the magnitude of \(w\) controls how sharply the policy transitions between walking and jumping.

Since \(z(x;\theta)\) is unbounded and cannot be interpreted directly as a probability, we map it to the interval \([0,1]\) using the logistic (sigmoid) function. This yields the probability of selecting the Jump action:
\[
p(x;\theta) = \sigma(z) = \sigma(wx+b) = \frac{1}{1 + e^{-(w x + b)}}.
\]

The resulting stochastic policy can therefore be written as
\[
\pi(a=1 \mid x, \theta) = p(x;\theta) = \sigma(w x + b),
\qquad
\pi(a=0 \mid x, \theta) = 1- p(x;\theta) = 1 - \sigma(w x + b).
\]

Equivalently, the policy admits the compact Bernoulli form
\[
\pi(a \mid x, \theta)
=\big[p(x;\theta)\big]^a \big[1-p(x;\theta)\big]^{1-a}=
\big[\sigma(w x + b)\big]^a
\big[1 - \sigma(w x + b)\big]^{1-a}.
\]

\item[c.]

The log-probability of the policy is given by (recall that \(\log(ab) = \log(a) + \log(b)\)):
\[
\log \pi(a \mid x,\theta)
=
a \log p(x;\theta)
+
(1-a)\log\!\big(1-p(x;\theta)\big).
\]

Let us differentiate term by term with respect to \(w\), using the linearity of differentiation:
\[
\frac{\partial}{\partial w}\log \pi(a \mid x,\theta)
=
a \frac{\partial}{\partial w}\log p(x;\theta)
+
(1-a)\frac{\partial}{\partial w}\log\!\big(1-p(x;\theta)\big).
\]

By applying the chain rule to the first we obtain (recall that \(\frac{\partial}{\partial x} \log(x) = \frac{1}{x}\)), 
\[
\frac{\partial}{\partial w}\log p(x;\theta)
=
\frac{1}{p(x;\theta)}\frac{\partial p(x;\theta)}{\partial w}.
\]
For the second term, we similarly apply the chain rule:
\[
\frac{\partial}{\partial w}\log\!\big(1-p(x;\theta)\big)
=
\frac{1}{1-p(x;\theta)}\frac{\partial}{\partial w}\big(1-p(x;\theta)\big)
=
-\frac{1}{1-p(x;\theta)}\frac{\partial p(x;\theta)}{\partial w}.
\]

Substituting back to combine the terms, we obtain:
\[
\frac{\partial}{\partial w}\log \pi(a \mid x,\theta)
=
a \frac{1}{p(x;\theta)}\frac{\partial p(x;\theta)}{\partial w}
-
(1-a)\frac{1}{1-p(x;\theta)}\frac{\partial p(x;\theta)}{\partial w}.
\]

Factoring out the common derivative \(\frac{\partial p(x;\theta)}{\partial w}\), we get:
\[
\frac{\partial}{\partial w}\log \pi(a \mid x,\theta)
=
\left(
\frac{a}{p(x;\theta)}
-
\frac{1-a}{1-p(x;\theta)}
\right)
\frac{\partial p(x;\theta)}{\partial w}.
\]

Before moving on, we need to determine \(\frac{\partial p(x;\theta)}{\partial w}\). First, let us differentiate the sigmoid function using the power rule and the chain rule, noting that the sigmoid can be rewritten as \(\sigma(z) = (1 + e^{-z})^{-1}\):
\[
\frac{d\sigma}{dz} = -1 \cdot (1 + e^{-z})^{-2} \cdot \frac{d}{dz}(1 + e^{-z})
= - (1 + e^{-z})^{-2} \cdot (- e^{-z})
= \frac{e^{-z}}{(1 + e^{-z})^2}.
\]

We can rewrite the expression by exploiting the fact that
\[
\sigma(z) = \frac{1}{1 + e^{-z}} \quad \implies \quad 1 - \sigma(z) = \frac{e^{-z}}{1 + e^{-z}},
\]

resulting in,
\[
\frac{d\sigma}{dz} = \frac{e^{-z}}{(1 + e^{-z})^2} = \frac{1}{1 + e^{-z}} \cdot \frac{e^{-z}}{1 + e^{-z}} = \sigma(z) \big(1 - \sigma(z)\big).
\]

For \(p(x;\theta) = \sigma(wx + b)\) with $z=wx+b$,
\[
\frac{\partial p(x;\theta)}{\partial w} = \frac{d\sigma}{dz} \cdot \frac{\partial z}{\partial w} = \sigma(wx+b) \big(1 - \sigma(wx+b)\big) \cdot x = p(x;\theta)\big(1-p(x;\theta)\big)x.
\]

Finally, by performing all substitutions and simplifications we obtain
\[
\frac{\partial}{\partial w}\log \pi(a \mid x,\theta)
=
\big(a-p(x;\theta)\big)x.
\]

We now need to determine the partial derivative with respect to $b$. For this purpose we proceed analogously to as before:
\[
\frac{\partial}{\partial b}\log \pi(a \mid x,\theta)
=
\left(
\frac{a}{p(x;\theta)}
-
\frac{1-a}{1-p(x;\theta)}
\right)
\frac{\partial p(x;\theta)}{\partial b},
\]

where 
\[
\frac{\partial p(x;\theta)}{\partial b}
= \frac{d\sigma}{dz} \cdot \frac{\partial z}{\partial b} =
\sigma(wx+b)\big(1-\sigma(wx+b)\big)\cdot 1
=
p(x;\theta)\big(1-p(x;\theta)\big).
\]

Finally, by performing all substitutions and simplifications we obtain

\[
\frac{\partial}{\partial b}\log \pi(a \mid x,\theta)
=
a-p(x;\theta).
\]

Collecting both partial derivatives, the gradient of the log-policy is
\[
\nabla_\theta \log \pi(a \mid x,\theta)
=
\begin{bmatrix}
\displaystyle (a-p(x;\theta))x \\[6pt]
\displaystyle a-p(x;\theta)
\end{bmatrix}.
\]

\item[d.]

\begin{itemize}
\item 
The \emph{bias} \(b\) controls the threshold on \(x\).  
    The jump probability is 
    \[
    p(x) = \sigma(wx + b) = \frac{1}{1 + e^{-(wx+b)}}.
    \] 
    The threshold \(x_\text{th}\) is defined as the value of \(x\) where the agent jumps with probability \(0.5\):
    \[
    p(x_\text{th}) = 0.5 \implies wx_\text{th} + b = 0 \implies x_\text{th} = -\frac{b}{w}.
    \]

\item The \emph{weight} \(w\) controls how sharply the jump probability changes near the threshold. Large \(|w|\) leads to a steep sigmoid, near-deterministic decision at \(x_\text{th}\). Small \(|w|\) leads to a shallow sigmoid, more stochastic behavior near \(x_\text{th}\).

\item Small \(|w|\) is beneficial in settings with \emph{sensor noise}, \emph{environmental variability}, \emph{stochastic dynamics}, or \emph{model uncertainty}, because maintaining stochasticity near the threshold allows the agent to hedge its decisions, reducing the risk of consistently choosing the wrong action and thereby improving the \emph{expected reward} over multiple runs.

\item     In a deterministic or single-run environment, where the correct action is known and the sensor is accurate, stochasticity is unnecessary. In this case, a large \(|w|\) (decisive policy) ensures the agent reliably jumps exactly at the threshold without risk, maximizing the reward for that run.
\end{itemize}

\item[e.]
Let us use the REINFORCE update
\[
\theta \gets \theta + \alpha\, R \, \nabla_\theta \log \pi(a \mid x,\theta),
\quad
\nabla_\theta \log \pi(a \mid x) = 
\begin{bmatrix} (a-p(x))x \\ a-p(x) \end{bmatrix},
\quad
p(x) = \sigma(wx+b),
\]
with current parameters \(w = 1.0\), \(b = -2.0\), and \(\alpha = 0.1\).

\medskip
\noindent
\textbf{Case 1: \(x = 3.0\), \(a = 1\) (Jump), \(R = +1\)}

\begin{align*}
\text{Before update: } & p(x) = \sigma(1\cdot 3 -2) = \sigma(1) \approx 0.731 \\
\nabla_\theta \log \pi(a\mid x) & = \begin{bmatrix} (1-0.731)\cdot 3 \\ 1-0.731 \end{bmatrix} = \begin{bmatrix} 0.807 \\ 0.269 \end{bmatrix} \\
\Delta \theta & = 0.1 \cdot 1 \cdot \begin{bmatrix}0.807 \\ 0.269\end{bmatrix} = \begin{bmatrix}0.081 \\ 0.027\end{bmatrix} \\
\text{After update: } & w = 1.081, \quad b = -1.973 \\
p(x) & = \sigma(1.081\cdot 3 -1.973) = \sigma(1.27) \approx 0.780
\end{align*}

Hence, jump probability increased from 0.731 to 0.780 at \(x=3\).

\medskip
\noindent
\textbf{Case 2: \(x = 1.0\), \(a = 0\) (Walk), \(R = +1\)}

\begin{align*}
\text{Before update: } & p(x) = \sigma(1\cdot 1 -2) = \sigma(-1) \approx 0.269 \\
\nabla_\theta \log \pi(a\mid x) & = \begin{bmatrix} (0-0.269)\cdot 1 \\ 0-0.269 \end{bmatrix} = \begin{bmatrix}-0.269 \\ -0.269 \end{bmatrix} \\
\Delta \theta & = 0.1 \cdot 1 \cdot \begin{bmatrix}-0.269 \\ -0.269\end{bmatrix} = \begin{bmatrix}-0.027 \\ -0.027\end{bmatrix} \\
\text{After update: } & w = 0.973, \quad b = -2.027 \\
p(x) & = \sigma(0.973\cdot 1 -2.027) = \sigma(-1.054) \approx 0.258
\end{align*}

Hence, walk probability increased from 0.731 (1-p) to 0.742 (1-p) at \(x=1\).

\medskip
\noindent
\textbf{Case 3: \(x = 3.0\), \(a = 0\) (Walk), \(R = -1\)}

\begin{align*}
\text{Before update: } & p(x) = \sigma(1\cdot 3 -2) = \sigma(1) \approx 0.731 \\
\nabla_\theta \log \pi(a\mid x) & = \begin{bmatrix} (0-0.731)\cdot 3 \\ 0-0.731 \end{bmatrix} = \begin{bmatrix}-2.193 \\ -0.731 \end{bmatrix} \\
\Delta \theta & = 0.1 \cdot (-1) \cdot \begin{bmatrix}-2.193 \\ -0.731\end{bmatrix} = \begin{bmatrix}0.219 \\ 0.073 \end{bmatrix} \\
\text{After update: } & w = 1.219, \quad b = -1.927 \\
p(x) & = \sigma(1.219\cdot 3 -1.927) = \sigma(1.73) \approx 0.849
\end{align*}

Hence, walk probability decreased (1-p increased) from 0.269 to 0.151 at \(x=3\).

\medskip
\noindent
\textbf{Case 4: \(x = 1.0\), \(a = 1\) (Jump), \(R = -1\)}

\begin{align*}
\text{Before update: } & p(x) = \sigma(1\cdot 1 -2) = \sigma(-1) \approx 0.269 \\
\nabla_\theta \log \pi(a\mid x) & = \begin{bmatrix} (1-0.269)\cdot 1 \\ 1-0.269 \end{bmatrix} = \begin{bmatrix}0.731 \\ 0.731 \end{bmatrix} \\
\Delta \theta & = 0.1 \cdot (-1) \cdot \begin{bmatrix}0.731 \\ 0.731\end{bmatrix} = \begin{bmatrix}-0.073 \\ -0.073\end{bmatrix} \\
\text{After update: } & w = 0.927, \quad b = -2.073 \\
p(x) & = \sigma(0.927\cdot 1 -2.073) = \sigma(-1.146) \approx 0.241
\end{align*}

Hence, jump probability decreased from 0.269 to 0.241 at \(x=1\).

\end{itemize}

\fi

	\subsection{The REINFORCE Algorithm with Baseline}

	Algorithm~\ref{agorithm:reinforce-baseline} presents the pseudo-code for the REINFORCE algorithm enhanced with an estimated state-value function used as a baseline. This version closely follows the original REINFORCE algorithm without a baseline, but now updates the parameters of both the policy and the state-value function at each iteration. Notably, the algorithm employs two separate learning rates, $\alpha^{\boldsymbol{\theta}}$ for the policy parameters and $\alpha^{\mathbf{w}}$ for the state-value approximation parameters.

	\begin{algorithm}[h]
		\caption{Episodic REINFORCE Algorithm with baseline and decaying learning rate (adapted from \cite{sutton2018reinforcement})}
		\begin{algorithmic}[1]
			\label{agorithm:reinforce-baseline}
			\STATE \textbf{Input:} a differentiable policy parameterization $\pi(a \mid s, \boldsymbol{\theta})$,
			\STATE \textbf{Input:} a differentiable state-value function parameterization $\hat{v}(s, \mathbf{w})$,
			\STATE \textbf{Parameters:} step size $\alpha^{\boldsymbol{\theta}} > 0$, $\alpha^{\mathbf{w}} > 0$, discount factor $0\le\gamma\le 1$
			\STATE \textbf{Parameters:} learning decay rates, $0 < \tau_{\boldsymbol{\theta}} \le 1$ and $0 < \tau_{\mathbf{w}} \le 1$
			\STATE \textbf{Parameters:} decay interval, $\delta t > 0$, episode length $T$
			\STATE
			\STATE Initialize policy parameter $\boldsymbol{\theta} \in \mathbb{R}^{d}$ (e.g., randomly)
			\STATE Initialize state-value weights $\mathbf{w} \in \mathbb{R}^{m}$ (e.g., randomly)
			\STATE Initialize global time step, $n\leftarrow0$
			\STATE \textbf{Loop forever}:
			\begin{ALC@g}
				\STATE \textcolor{blue}{// Generating an episode (policy rollout)}
				\STATE Observe initial state $S_{0}$
				\STATE \textbf{For each time step} $t = 0, 1, \ldots, T-1$:
				\begin{ALC@g}
					\STATE Sample action from policy, $A_t \sim \pi(\cdot \mid S_t, \boldsymbol{\theta})$
					\STATE Take action $A_t$, observe reward $R_{t+1}$ and next state $S_{t+1}$
				\end{ALC@g}
				\STATE \textcolor{blue}{// Policy update based on episode return}
				\STATE \textbf{For each step of the episode} $t = 0, 1, \ldots, T-1$:
				\begin{ALC@g}
					\STATE $n \leftarrow n + 1$
					\STATE $G \leftarrow \sum_{k=t+1}^T \gamma^{k-t-1} R_k$
					\STATE $\delta\leftarrow G-\hat{v}(S_t, \mathbf{w})$
					\STATE $\alpha^{\boldsymbol{\theta}} \leftarrow \alpha^{\boldsymbol{\theta}}_0 \tau_{\boldsymbol{\theta}}^ {\frac{n}{\delta t}}$
					\STATE $\alpha^{\mathbf{w}} \leftarrow \alpha^{\mathbf{w}}_0 \tau_{\mathbf{w}}^ {\frac{n}{\delta t}}$
					\STATE $\mathbf{w} \leftarrow \mathbf{w} + \alpha^{\mathbf{w}} \delta \nabla \hat{v}(S_t, \mathbf{w})$
					\STATE $\boldsymbol{\theta} \leftarrow \boldsymbol{\theta} + \alpha^{\boldsymbol{\theta}} \gamma^t \delta \nabla \ln \pi(A_t \mid S_t, \boldsymbol{\theta})$
				\end{ALC@g}
			\end{ALC@g}
		\end{algorithmic}
	\end{algorithm}

	\subsection{The Softmax Function}
	\label{sec:softmax}

	This section introduces the \emph{softmax} function, which is required to understand the subsequent material.

	Formally, the softmax function maps a vector of $n$ real numbers, $\mathbf{x} = (x_1, \ldots, x_n) \in \mathbb{R}^n$, to a normalized vector that represents a probability distribution. Specifically, each element of the output vector lies in the interval $[0,1]$, $\tilde{\mathbf{x}} = (\tilde{x}_1, \ldots, \tilde{x}_n) \in [0,1]^n$, and the elements sum to 1, so that $\sum_{k=1}^{n} \tilde{x}_k = 1$. Denoting the $i$-th element of a vector $\mathbf{a}$ as $\mathbf{a}[i]$, the softmax function is applied element-wise as follows:
	\begin{equation}
		\text{softmax}(\mathbf{x})[i] = \frac{e^{\mathbf{x}[i]}}{\sum_{j=1}^n e^{\mathbf{x}[j]}}, \quad \forall \, 1 \le i \le n.
	\end{equation}

	As an example, consider the vector \(\mathbf{x} = (3.0, -1.0, 0.1)\). To apply the softmax function, we first compute the exponentials of each element: \(e^{3.0} \approx 20.086\), \(e^{-1.0} \approx 0.368\), and \(e^{0.1} \approx 1.105\). The sum of these exponentials is \(20.086 + 0.368 + 1.105 \approx 21.559\). The softmax values are then calculated as:
	\begin{align*}
		\text{softmax}(\mathbf{x})[1] & = \frac{20.086}{21.559} \approx 0.932, \\ \text{softmax}(\mathbf{x})[2] &= \frac{0.368}{21.559} \approx 0.017,\\
		\text{softmax}(\mathbf{x})[3] & = \frac{1.105}{21.559} \approx 0.051.
	\end{align*}

	Thus, by applying the softmax function, the vector \(\mathbf{x} = (3.0, -1.0, 0.1)\) is transformed into the probability distribution \(\tilde{\mathbf{x}} = \text{softmax}(\mathbf{x}) = (0.932, 0.017, 0.051)\). All elements of $\tilde{\mathbf{x}}$ lie in the interval $[0,1]$, and their sum satisfies $0.932 + 0.017 + 0.051 = 1.0$, confirming that it forms a valid probability distribution.

	\subsection{Policy Parameterization for Discrete Actions}

	Until now, little attention has been paid to the exact form of the policy function $\pi_{\boldsymbol{\theta}}$. Up to this point, we have only assumed that the policy must be differentiable and define a valid probability distribution over actions. The fact that $\pi_{\boldsymbol{\theta}}$ needs to be differentiable and define a probability distribution over actions imposes some constraints on its specific implementation. However, the policy gradient update step may push and pull $\pi_{\boldsymbol{\theta}}$ such that it is no longer a probability distribution. For instance, it could happen that the policy gradient update step would discourage a bad action to the point of assigning a negative probability.

	Hence, it is necessary to ensure that the policy gradient update does not produce action probabilities that violate the fundamental properties of a probability distribution. A common solution is to decompose the policy into two components: a preference function that assigns real-valued scores to actions without being constrained to form a probability distribution, and a \emph{softmax} operation that converts these preferences into a valid probability distribution. In this setup, the policy gradient update can freely adjust the action preferences, while the softmax ensures that the resulting policy $\pi_{\boldsymbol{\theta}}$ always defines a proper probability distribution. The following section details this process.

	For discrete action spaces, we can define a parameterized numerical preference function that is not required to produce a probability distribution:
	\[
		h: \mathcal{A} \times \mathcal{S} \times \mathbb{R}^d \rightarrow \mathbb{R}.
	\]

	The purpose of the preference function is simply to assign higher numerical preferences to actions that should be selected more often. These preferences are then passed through a \emph{softmax} function to produce the policy $\pi_{\boldsymbol{\theta}}$, ensuring a valid probability distribution regardless of the raw preference values \cite{sutton2018reinforcement}:
	\begin{equation}\label{equ:soft-max-preferences}
		\pi(a \mid s, \boldsymbol{\theta}) \doteq \frac{e^{h(s,a,\boldsymbol{\theta})}}{\sum_{b\in\mathcal{A}} e^{h(s,b,\boldsymbol{\theta})}}.
	\end{equation}

	Note that the policy parameters $\boldsymbol{\theta}$ parameterize the preference function $h(\cdot, \cdot, \boldsymbol{\theta})$. Therefore, the policy gradient used to update $\boldsymbol{\theta}$, $\nabla_{\boldsymbol{\theta}} \ln \pi(\cdot \mid \cdot, \boldsymbol{\theta})$, must be propagated through $h$. To derive this gradient, we first take the logarithm of both sides of Equation~\ref{equ:soft-max-preferences}:
	\begin{align}
		\ln \pi(a|s,\theta) & = \ln \frac{e^{h(s,a,\theta)}}{\sum_{b\in\mathcal{A}} e^{h(s,b,\theta)}}\quad \text{(take the logarithm in both sides)}\nonumber    \\
		                    & =  \ln  e^{h(s,a,\theta)} - \ln \sum_{b\in\mathcal{A}} e^{h(s,b,\theta)} \quad (\text{because } \ln(a / b)=\ln(a) - \ln(b)\nonumber \\
		                    & =  h(s,a,\theta)  - \ln \sum_{b\in\mathcal{A}} e^{h(s,b,\theta)} \quad (\text{because } \ln(e^x)= x).\label{equ:ln-pii}
	\end{align}

	We can now compute the gradient of $\ln \pi(a \mid s, \boldsymbol{\theta})$ with respect to $\boldsymbol{\theta}$:
	\begin{align}
		\nabla_{\boldsymbol{\theta}} \ln & \pi(a|s,\boldsymbol{\theta}) =  \nonumber                                                                                                                                                                                                                                                               \\&=\nabla_{\boldsymbol{\theta}} h(s,a,\boldsymbol{\theta})  - \nabla_{\boldsymbol{\theta}}  \ln \sum_{b\in\mathcal{A}} e^{h(s,b,\boldsymbol{\theta})} \quad (\text{because } (f(x) + g(x))'=f'(x) + g'(x))\nonumber\\
		                                 & =  \nabla_{\boldsymbol{\theta}} h(s,a,\boldsymbol{\theta})  -  \frac{\nabla_{\boldsymbol{\theta}}  \sum_{b\in\mathcal{A}} e^{h(s,b,\boldsymbol{\theta})}}{\sum_{b\in\mathcal{A}} e^{h(s,b,\boldsymbol{\theta})}} \quad (\text{because } (\ln f(x))' =f'(x) / f(x))\nonumber                             \\
		                                 & =  \nabla_{\boldsymbol{\theta}} h(s,a,\boldsymbol{\theta})  -  \frac{  \sum_{b\in\mathcal{A}} \nabla_{\boldsymbol{\theta}} e^{h(s,b,\boldsymbol{\theta})}}{\sum_{b\in\mathcal{A}} e^{h(s,b,\boldsymbol{\theta})}}  \quad (\text{because } (f(x) + g(x))'=f'(x) + g'(x))\nonumber                        \\
		                                 & =  \nabla_{\boldsymbol{\theta}} h(s,a,\boldsymbol{\theta})  -  \frac{  \sum_{b\in\mathcal{A}} \nabla_{\boldsymbol{\theta}} h(s,b,\boldsymbol{\theta}) e^{h(s,b,\boldsymbol{\theta})}}{\sum_{b\in\mathcal{A}} e^{h(s,b,\boldsymbol{\theta})}} \quad (\text{because } [f(g(x))]' = f'(g(x))g'(x)\nonumber \\
		                                 & =  \nabla_{\boldsymbol{\theta}} h(s,a,\boldsymbol{\theta})  -  \sum_{b\in\mathcal{A}} \nabla_{\boldsymbol{\theta}} h(s,b,\boldsymbol{\theta}) \pi(b|s,\boldsymbol{\theta})  \quad (\text{substituting Equation \ref{equ:soft-max-preferences}}).\label{equ:klkll}
	\end{align}

	\subsection{Linear Policies and State-Value Functions}

	The previous section showed that the policy $\pi_{\boldsymbol{\theta}}$ is given by the softmax of a parameterized preference function $h(s, a, \boldsymbol{\theta})$. In this section, we introduce the simplest form of a parameterized preference function. Specifically, the preference for taking action $a$ in state $s$ is defined as a linear combination of features of both $a$ and $s$ \cite{sutton2018reinforcement}:
	\begin{equation}\label{equ:h-simples}
		h(s,a,\boldsymbol{\theta}) = \boldsymbol{\theta}^T \mathbf{x}(s,a),
	\end{equation}

	\noindent where $\mathbf{x}: \mathcal{S} \times \mathcal{A} \rightarrow \mathbb{R}^d$ is a $d$-dimensional feature vector representing the given state-action pair. To fully parameterize this linear preference function, at least $d$ learnable parameters are required, so that $|\boldsymbol{\theta}| = d$.

	The feature vector can be as simple as the state itself, $\mathbf{x}(s,a) \doteq s$, provided it is real-valued. More generally, a feature is a representation of the state-action pair that is expected to be useful for solving the task at hand. For example, if the agent's goal is to reach a specific target $g$ and the state encodes the agent's position in the environment, a relevant feature could be the distance from the current state to $g$. Features allow the policy to capture important non-linear aspects of the task that a purely linear policy would otherwise miss.

	Additionally, features allow us to incorporate prior domain knowledge into the reinforcement learning problem. Without features, the agent would need to discover this knowledge from scratch, requiring more complex policy functions and potentially slowing down learning. The main challenge of using features, however, is performing effective \emph{feature engineering}, which can be difficult in practice. In the subsequent sections, we will show how this problem can be circumvented by using neural networks to automatically learn useful feature representations.

	Now that the function \(h\) is defined (Equation~\ref{equ:h-simples}), its gradient with respect to the learnable parameters can be computed. In the linear case, this computation is straightforward:
	\begin{equation}\label{equ:grad-h-linear}
		\nabla_{\boldsymbol{\theta}} h(s,a,\boldsymbol{\theta}) = \nabla_{\boldsymbol{\theta}} \boldsymbol{\theta}^T \mathbf{x}(s,a) = \mathbf{x}(s,a).
	\end{equation}

	By substituting Equation~\ref{equ:grad-h-linear} into Equation~\ref{equ:klkll}, we obtain the gradient of the logarithm of the policy with respect to $\boldsymbol{\theta}$:
	\begin{equation}\label{equ:grad-log-linear-final}
		\nabla_{\boldsymbol{\theta}} \ln \pi(a|s,\boldsymbol{\theta}) =  \mathbf{x}(s,a)  -  \sum_{b\in\mathcal{A}} \mathbf{x}(s,b) \pi(b|s,\boldsymbol{\theta}).
	\end{equation}

	This gradient can then be employed in a policy gradient algorithm, such as REINFORCE, to update the policy parameters, as illustrated in Step 19 of Algorithm~\ref{agorithm:reinforce}. Let us now discuss how to evaluate this gradient for a given state-action pair $(s, a)$. For the given state $s$ and action $a$, the feature vector $\mathbf{x}(s,a)$ is first constructed (for example, it could represent the distance between $s$ and a goal $g$). Each time the policy function is evaluated, the preference function $h(s, a, \boldsymbol{\theta})$ is computed using the current policy parameters $\boldsymbol{\theta}$ (Equation~\ref{equ:h-simples}). The resulting numerical preference is then passed through the softmax function in Equation~\ref{equ:soft-max-preferences} to produce a probability. This probability is finally substituted into Equation~\ref{equ:grad-log-linear-final} to compute the gradient of the logarithm of the policy with respect to $\boldsymbol{\theta}$.

	\subsubsection*{Example: The CartPole Problem}

\begin{figure}[ht]
		\centering
		\begin{subfigure}[b]{0.45\textwidth}
			\centering
			\includegraphics[width=\textwidth]{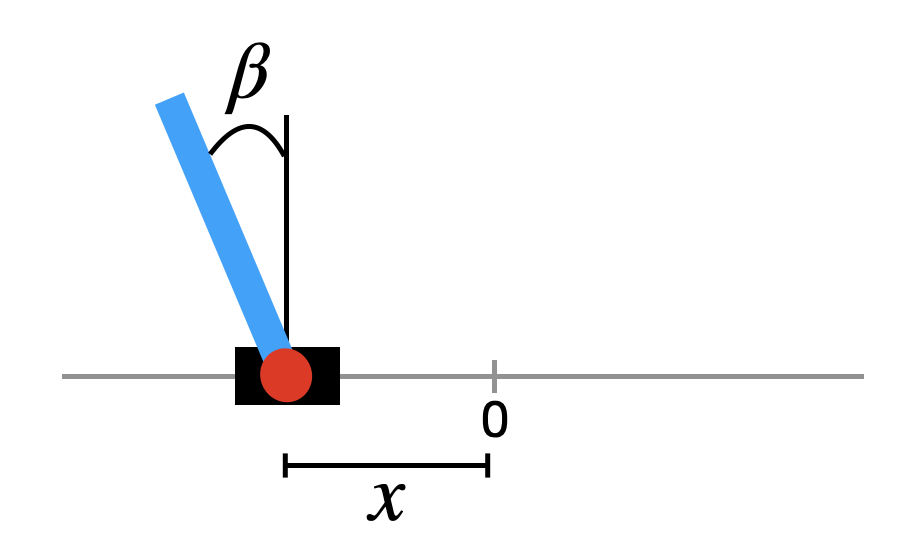}
			\caption{State}
		\end{subfigure}
		\hspace{1cm}
		\begin{subfigure}[b]{0.45\textwidth}
			\centering
			\includegraphics[width=\textwidth]{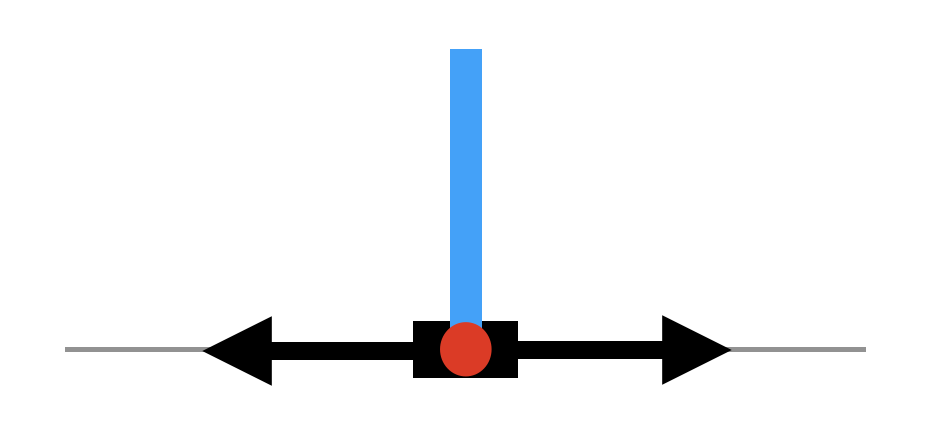}
			\vspace{0.1cm}
			\caption{Actions}
		\end{subfigure}
		\caption{State and actions of the OpenAI CartPole problem. The cart, the pole, and the passive joint are represented in black, blue, and respectively. Left: episode terminating because the position $x$ and angle $\beta$ off the limits. Right: the cart in the "desired" position and the pole in the "desired" angle, with the black arrows representing the two possible discrete actions (forces).}
		\label{fig:cart-pole-diagram}
	\end{figure}

	The CartPole problem, instantiated by OpenAI as CartPole-v1\footnote{\url{https://www.gymlibrary.dev/environments/classic_control/cart_pole/}}, is a classic problem in reinforcement learning that involves balancing a pole on a moving cart, subject to the dynamics imposed by the laws of physics. The goal is to keep a pole, attached to a cart via a passive (non-actuated) joint, balanced upright by moving the cart either to the left or to the right along a frictionless track. This is a typical toy problem often used to test and benchmark reinforcement learning algorithms. Despite its simplicity, it includes the non-linear dynamics involved in balancing the pole. Figure~\ref{fig:cart-pole-diagram} depicts the problem, including its state and action spaces.

	The state is represented by a 4-dimensional vector, $s\doteq(x, x', \beta, \beta')$ including the real-valued position of the cart on the track, $x$, the real-valued linear velocity of the cart, $x'$, the real-valued angle of the pole with the vertical, $\beta$, and the real-valued angular velocity of the pole, $\beta'$. The agent can execute one of two discrete actions, $\mathcal{A}=\{\text{left}, \text{right}\}$: applying a constant force on the cart to accelerate it to the left; applying a constant force on the cart to accelerate it to the right. The agent earns a reward of +1 for every time step until the episode terminates. Hence, the longer the episode, the larger the reward. The episode terminates when any of the following conditions occur: the pole falls past $\pm\,12$ degrees from the vertical or the cart moves beyond $\pm\,2.4$ units from the center; or a maximum of 500 time steps is reached.

	To instantiate the REINFORCE algorithm to this problem it is necessary to make some decisions. First, the feature vector is defined as simply the state, $\mathbf{x}(s,a)\doteq s$, thus independent of the action. Second, the action preferences function is defined as a linear function of the state (Equation~\ref{equ:h-simples}), using separate weights for stating the preferences for the left action and the preferences for the right action, $\boldsymbol{\theta} = \boldsymbol{\theta_1} \cup \boldsymbol{\theta_2}$:
	\begin{equation}\label{equ:h-forpole}
		h(s,a,\boldsymbol{\theta}) =
		\begin{cases}
			\boldsymbol{\theta^T_1} s & \text{if } a=\text{left}  \\
			\boldsymbol{\theta^T_2} s & \text{if } a=\text{right}
		\end{cases}.
	\end{equation}

	The policy logarithm can be derived as in Equation~\ref{equ:ln-pii} and based on Equation~\ref{equ:h-forpole},  but this time separately for the two actions (for the sake of clarity):
	\begin{align}
		\ln\pi(\text{left}~~\mid s, \boldsymbol{\theta}) & = \ln\frac{e^{h(s,\text{left},\boldsymbol{\theta})}}{e^{h(s,\text{left},\boldsymbol{\theta})}+e^{h(s,\text{right},\boldsymbol{\theta})}}=\nonumber                                                 \\
		                                                 & =\ln\frac{e^{\boldsymbol{\theta^T_1} s}}{e^{\boldsymbol{\theta^T_1} s}+e^{\boldsymbol{\theta^T_2} s}}= \boldsymbol{\theta^T_1} s-\ln(e^{\boldsymbol{\theta^T_1} s}+e^{\boldsymbol{\theta^T_2} s}),
	\end{align}
	\begin{align}
		\ln\pi(\text{right}\mid s, \boldsymbol{\theta}) & = \ln\frac{e^{h(s,\text{right},\boldsymbol{\theta})}}{e^{h(s,\text{left},\boldsymbol{\theta})}+e^{h(s,\text{right},\boldsymbol{\theta})}}=\nonumber                                               \\
		                                                & =\ln\frac{e^{\boldsymbol{\theta^T_2} s}}{e^{\boldsymbol{\theta^T_1} s}+e^{\boldsymbol{\theta^T_2} s}}=\boldsymbol{\theta^T_2} s-\ln(e^{\boldsymbol{\theta^T_1} s}+e^{\boldsymbol{\theta^T_2} s}).
	\end{align}

	Although the action preferences are defined by different sets of parameters, both influence the policy to ensure the latter defines a proper probability distribution (i.e., normalizes the action preferences to sum up to 1 and to ensure that none is negative or above 1). The next step is to take the gradients of the logarithms, as done in Equation~\ref{equ:grad-log-linear-final}, but now with respect to the different sets of parameters (recall that $\pi(\text{left}\mid s,\boldsymbol{\theta}) = 1- \pi(\text{right}\mid s,\boldsymbol{\theta})$):
	\begin{align}
		\nabla_{\boldsymbol{\theta_1}} \ln \pi(a\mid s,\boldsymbol{\theta}) & =
		\begin{cases}
			-\pi(\text{left}\mid s,\boldsymbol{\theta})s & \text{if } a=\text{right} \\
			\pi(\text{right}\mid s,\boldsymbol{\theta})s & \text{if } a=\text{left}
		\end{cases}, \\
		\nabla_{\boldsymbol{\theta_2}} \ln \pi(a\mid s,\boldsymbol{\theta}) & =
		\begin{cases}
			-\pi(\text{right}\mid s,\boldsymbol{\theta})s & \text{if } a=\text{left}  \\
			\pi(\text{left}\mid s,\boldsymbol{\theta})s   & \text{if } a=\text{right}
		\end{cases}.
	\end{align}

	\subsubsection*{Derivations (for the interested reader)}

	\noindent
	\begin{align*}
		\nabla_{\boldsymbol{\theta_1}} \ln \pi(\text{left} \mid s, \theta) & = \nabla_{\boldsymbol{\theta_1}} \left[ \boldsymbol{\theta^T_1} s - \ln (e^{\boldsymbol{\theta^T_1} s} + e^{\boldsymbol{\theta^T_2} s}) \right]                              \\
		                                                                   & = s - \nabla_{\boldsymbol{\theta^T_1}} \ln (e^{\boldsymbol{\theta^T_1} s} + e^{\boldsymbol{\theta^T_2} s})                                                                   \\
		                                                                   & = s - \frac{\nabla_{\boldsymbol{\theta^T_1}} (e^{\boldsymbol{\theta^T_1} s} + e^{\boldsymbol{\theta^T_2} s})}{e^{\boldsymbol{\theta^T_1} s} + e^{\boldsymbol{\theta^T_2} s}} \\
		                                                                   & = s - \frac{s e^{\boldsymbol{\theta^T_1} s}}{e^{\boldsymbol{\theta^T_1} s} + e^{\boldsymbol{\theta^T_2} s}}                                                                  \\
		                                                                   & = s - s \pi(\text{left} \mid s, \theta)                                                                                                                                      \\
		                                                                   & = s (1 - \pi(\text{left} \mid s, \theta))                                                                                                                                    \\
		                                                                   & = \pi(\text{right} \mid s, \theta) s
	\end{align*}

	\begin{align*}
		\nabla_{\boldsymbol{\theta_2}} \ln \pi(\text{right} \mid s, \theta) & = \nabla_{\boldsymbol{\theta_2}} \left[ \boldsymbol{\theta^T_2} s - \ln (e^{\boldsymbol{\theta^T_1} s} + e^{\boldsymbol{\theta^T_2} s}) \right]                              \\
		                                                                    & = s - \nabla_{\boldsymbol{\theta^T_2}} \ln (e^{\boldsymbol{\theta^T_1} s} + e^{\boldsymbol{\theta^T_2} s})                                                                   \\
		                                                                    & = s - \frac{\nabla_{\boldsymbol{\theta^T_2}} (e^{\boldsymbol{\theta^T_1} s} + e^{\boldsymbol{\theta^T_2} s})}{e^{\boldsymbol{\theta^T_1} s} + e^{\boldsymbol{\theta^T_2} s}} \\
		                                                                    & = s - \frac{s e^{\boldsymbol{\theta^T_2} s}}{e^{\boldsymbol{\theta^T_1} s} + e^{\boldsymbol{\theta^T_2} s}}                                                                  \\
		                                                                    & = s - s \pi(\text{right} \mid s, \theta)                                                                                                                                     \\
		                                                                    & = s (1 - \pi(\text{right} \mid s, \theta))                                                                                                                                   \\
		                                                                    & = \pi(\text{left} \mid s, \theta) s
	\end{align*}

	\begin{align*}
		\nabla_{\boldsymbol{\theta_2}} \ln \pi(\text{left} \mid s, \theta) & = \nabla_{\boldsymbol{\theta_2}} \left[ \boldsymbol{\theta^T_1} s - \ln (e^{\boldsymbol{\theta^T_1} s} + e^{\boldsymbol{\theta^T_2} s}) \right]                            \\
		                                                                   & = - \nabla_{\boldsymbol{\theta^T_2}} \ln (e^{\boldsymbol{\theta^T_1} s} + e^{\boldsymbol{\theta^T_2} s})                                                                   \\
		                                                                   & = - \frac{\nabla_{\boldsymbol{\theta^T_2}} (e^{\boldsymbol{\theta^T_1} s} + e^{\boldsymbol{\theta^T_2} s})}{e^{\boldsymbol{\theta^T_1} s} + e^{\boldsymbol{\theta^T_2} s}} \\
		                                                                   & = - \frac{s e^{\boldsymbol{\theta^T_2} s}}{e^{\boldsymbol{\theta^T_1} s} + e^{\boldsymbol{\theta^T_2} s}}                                                                  \\
		                                                                   & = - \pi(\text{right} \mid s, \theta)s
	\end{align*}

	\begin{align*}
		\nabla_{\boldsymbol{\theta_1}} \ln \pi(\text{right} \mid s, \theta) & = \nabla_{\boldsymbol{\theta_1}} \left[ \boldsymbol{\theta^T_2} s - \ln (e^{\boldsymbol{\theta^T_1} s} + e^{\boldsymbol{\theta^T_2} s}) \right]                            \\
		                                                                    & = - \nabla_{\boldsymbol{\theta^T_1}} \ln (e^{\boldsymbol{\theta^T_1} s} + e^{\boldsymbol{\theta^T_2} s})                                                                   \\
		                                                                    & = - \frac{\nabla_{\boldsymbol{\theta^T_1}} (e^{\boldsymbol{\theta^T_1} s} + e^{\boldsymbol{\theta^T_2} s})}{e^{\boldsymbol{\theta^T_1} s} + e^{\boldsymbol{\theta^T_2} s}} \\
		                                                                    & = - \frac{s e^{\boldsymbol{\theta^T_1} s}}{e^{\boldsymbol{\theta^T_1} s} + e^{\boldsymbol{\theta^T_2} s}}                                                                  \\
		                                                                    & = - \pi(\text{left} \mid s, \theta)s
	\end{align*}

\begin{figure}[t]
		\centering
		\begin{subfigure}[b]{0.8\textwidth}
			\centering
			\includegraphics[width=\textwidth]{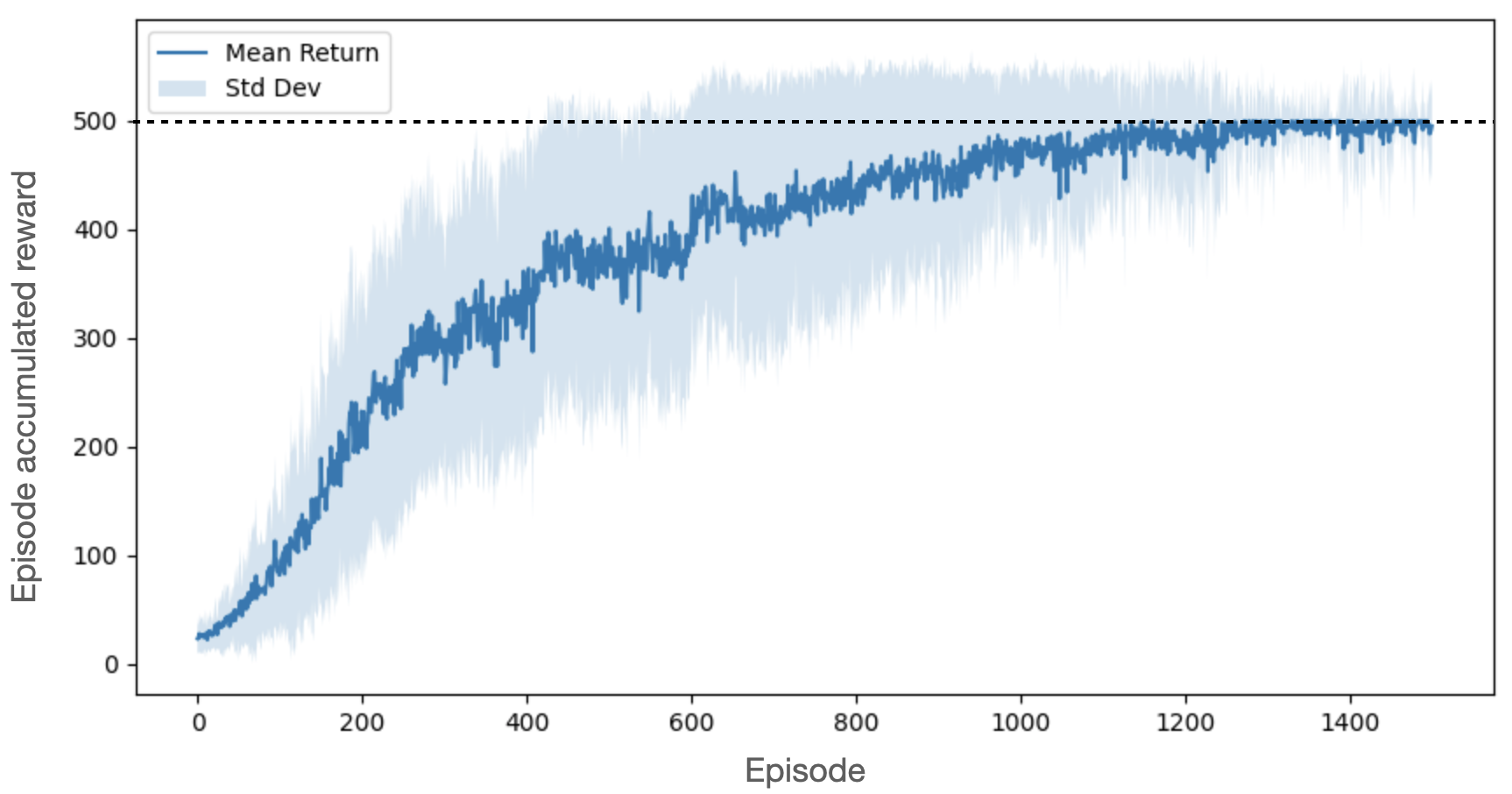}
			\caption{$\tau=0.85$ (and $\delta_t=100)$}
		\end{subfigure}
		\hspace{1cm}
		\begin{subfigure}[b]{0.8\textwidth}
			\centering
			\includegraphics[width=\textwidth]{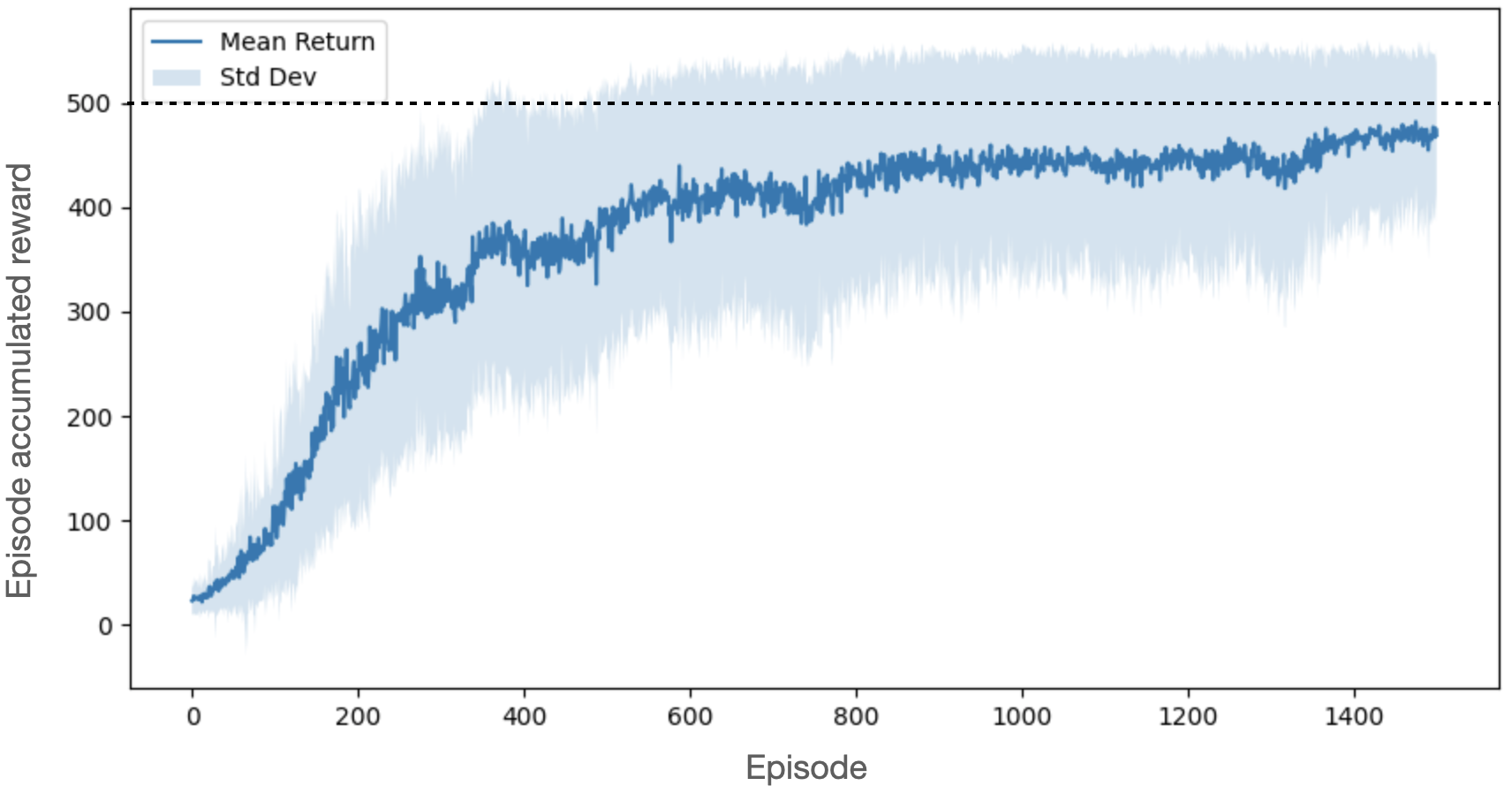}
			\caption{$\tau=1.0$}
		\end{subfigure}
		\caption{REINFORCE algorithm with decaying learning rate (but without baseline), using a base learning rate of $\alpha_0=0.001$, in the CartPole-v1 problem. The plots present the average and standard deviation of accumulated reward (undiscounted), per learning episode, over 30 runs.}
		\label{fig:learning-progress-cartpole}
	\end{figure}

	Figure~\ref{fig:learning-progress-cartpole} illustrates the learning progress of the REINFORCE algorithm with decaying learning rate (but without baseline), using a base learning rate of $\alpha_0=0.001$, for learning decay rates $\tau=0.85$ (and $\delta_t=100$) and $\tau=1.0$, in the CartPole-v1 problem. The figure shows the average and standard deviation of the accumulated (undiscounted) rewards per learning episode, averaged over 30 runs.

	Using an exponentially decaying learning rate (i.e., $\tau=0.85$) enables the accumulated reward to increase steadily, approaching the maximum possible reward of 500 more closely than without an adaptive learning rate (i.e., $\tau=1.0$). The plot also shows a large standard deviation, reflecting significant variability in results across different runs, a common characteristic of reinforcement learning. This variability partly arises from different initial conditions set by various seeds for the pseudo-random number generator, leading to varying learning trajectories and potentially causing the agent to reach local maxima instead of the global maximum.

	The observed variability is also due to the exploratory nature of the learning algorithm, which selects actions by stochastically sampling the policy at each instant. This means the agent does not always perform optimally in the sense of greedily exploiting the best action. However, running the learned policy greedily, i.e., $A_t=\arg\max_a \pi(a|S_t, \boldsymbol{\theta})$, for 30 episodes after each of the 30 runs consistently achieved the perfect accumulated reward of 500. Thus, the algorithm successfully learned the optimal policy. Nevertheless, to continue learning and adapt to changing environments, it is necessary to accept the cost of occasional suboptimal exploratory behavior.

	\subsection{Non-Linear Policies and State-Value Functions}
	\label{sec:non-linear-policies}

	Linear policies are mainly of educational interest due to their limited expressiveness, which makes them unsuitable for complex problems that require non-linear dependencies on the feature vector. This limitation also applies to linear state-value function approximations used as baselines. These too can be represented as a linear function of features (though we have not derived this representation here).

	Depending on the specific problem, learning the policy function can be easier or harder than learning the state-value function. One reason learning the policy may be easier is that it directly maps states to actions, which can simplify the learning process in environments where the optimal action for a given state is relatively clear. In contrast, the state-value function must estimate the expected return for each state under the current policy, which requires accurately accounting for long-term future rewards. This task can be more challenging, particularly in environments with high variance in rewards or delayed consequences of actions.

	When the state-value function is difficult to learn, relying on a linear model can be particularly restrictive, potentially hindering effective policy learning. This limitation arises because the sampled return is compared against an inaccurate estimate of the expected return, leading to erroneous evaluations and suboptimal policy updates. Inaccurate state-value estimates can misguide the learning process, causing the policy to deviate from the optimal trajectory. Consequently, a more expressive model, capable of capturing non-linear relationships, is often required to achieve better performance in complex problems.

	To address these limitations, this section introduces a more powerful approach, which is to utilize \textit{deep} Artificial Neural Networks (ANNs) to model the policy and state-value directly as non-linear functions of the raw state vector $s$. This method introduces non-linearity into these representations, effectively allowing the model to learn complex patterns from the state space without the need for extensive feature engineering.

	\subsubsection*{Multilayer Perceptron (MLP)}

	A commonly used ANN architecture for policy and state-value modeling is the Multilayer Perceptron (MLP). An MLP is composed of several simple information processing units, known as artificial neurons.

\begin{figure}[h]
		\centering
		\includegraphics[width=0.5\textwidth]{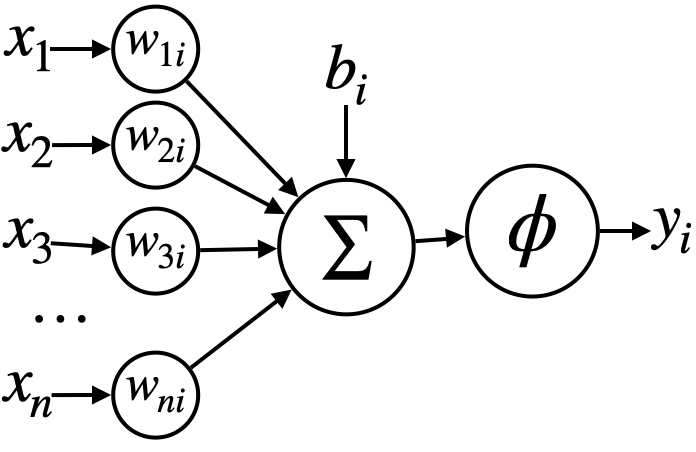}
		\caption{Diagram of an artificial neuron.}
		\label{fig:neuron}
	\end{figure}

	Each neuron generates as output a scalar resulting from a simple non-linear operation on the neuron's inputs. Formally, a neuron $i$ applies a non-linear transformation $\phi: \mathbb{R} \rightarrow \mathbb{R}$, also known as an \textit{activation function}, to the scalar bias $b_i \in \mathbb{R}$ plus the weighted sum of its inputs $\{x_j\} \in \mathbb{R}$ (see Figure~\ref{fig:neuron}):
	\begin{align*}
		y_i = \phi\left(b_i + \sum_{j}{x_j w_{ji}}\right),
	\end{align*}

	\noindent where $w_{ji}\in\mathbb{R}$ is the weight indicating how much the neuron's input $x_j$ influences its output. Learning occurs by adapting the neuron's bias and weights so the neuron's output matches the intended value.

\begin{figure}[h]
		\centering
		\includegraphics[width=0.6\textwidth]{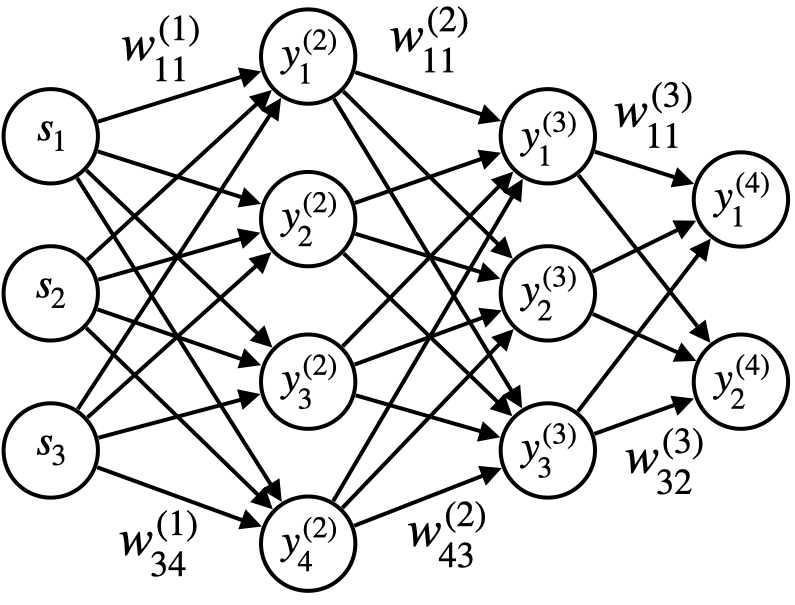}
		\caption{Diagram of an MLP. The neurons in the MLP are represented by their outputs and the layers have dimensions $n_1=3, n_2=4, n_3=3, n_4=2$.}
		\label{fig:mlp}
	\end{figure}

	In an MLP, neurons are organized in layers (see Figure~\ref{fig:mlp}). Let us denote the number of neurons in a given layer $l$ by $n_l$, where the first layer is $l=1$ and the last layer is $l=L$. The first and last layers are commonly known as \textit{input} and \textit{output} layers, respectively. All other intermediate layers are commonly known as \textit{hidden layers}. The neurons in the first layer, $l=1$, directly output the state vector $s$, that is, each neuron $i$ outputs the $i$-th element of $s$, denoted by $s_i$, where $s=(s_i)_i$. All neurons in subsequent layers, $l>1$, receive as input the outputs of all neurons in the preceding layer, $l-1$. In addition, all neurons in layer $l>1$ employ the same activation function $\phi^{(l)}$.

	The neuron equation can be redefined to explicitly represent the dependence on the layer $l$ with the superscript $(l)$, as well as on the state $s$ and the vector that encompasses all weights and biases of the MLP, $\boldsymbol{\theta}$:
	\begin{equation}\label{equ:neurons}
		y^{(l)}_i(s, \boldsymbol{\theta}) =
		\begin{cases}
			s_i                                                                                             & \text{if } l=1, \\
			\phi^{(l)}\left(b^{(l)}_i + \sum_{j}{w^{(l-1)}_{ji}} y^{(l-1)}_j(s, \boldsymbol{\theta})\right) & \text{if } l>1.
		\end{cases}
	\end{equation}

	The inclusion of the non-linear activation function is vital for the MLP's ability to model non-linear relationships in the state space. Non-linearities allow the network to capture complex dynamics and decision boundaries that are critical for effective learning and generalization. Three commonly used activation functions in neural networks are the Rectified Linear Unit (ReLU), the hyperbolic tangent (\textit{tanh}) function, and the sigmoid function (\(\sigma\)).

	In its basic formulation, the ReLU function is linear for positive values and zero otherwise, ensuring a large derivative when positive while simultaneously introducing the required non-linearity. The ReLU activation function is typically employed in all layers except the last one:
	\begin{equation*}
		\phi^{(l)}(x)=
		\begin{cases}
			\max(0, x) & \text{if } l<L, \\
			x          & \text{if } l=L.
		\end{cases}
	\end{equation*}

	The hyperbolic tangent activation function, \textit{tanh}, maps input values to the range \([-1, 1]\), introducing non-linearity while maintaining the ability to model both positive and negative relationships. It is defined as:
	\begin{equation*}
		\phi(x) = \tanh(x) = \frac{\sinh(x)}{\cosh(x)} = \frac{e^x - e^{-x}}{e^x + e^{-x}}.
	\end{equation*}

	The \textit{tanh} function is particularly important in reinforcement learning when outputs need to be bounded within a specific range, such as when actions are constrained between \([-1, 1]\). Additionally, the zero-centered output of the \textit{tanh} function can lead to faster convergence during training by improving the symmetry of gradient updates.

	The sigmoid activation function, \(\sigma\), maps input values to the range \([0, 1]\), introducing non-linearity while producing outputs that can be interpreted as probabilities. It is defined as:
	\begin{equation*}
		\sigma(x) = \frac{1}{1 + e^{-x}}.
	\end{equation*}

\begin{figure}
		\begin{center}
			\includegraphics[width=13cm]{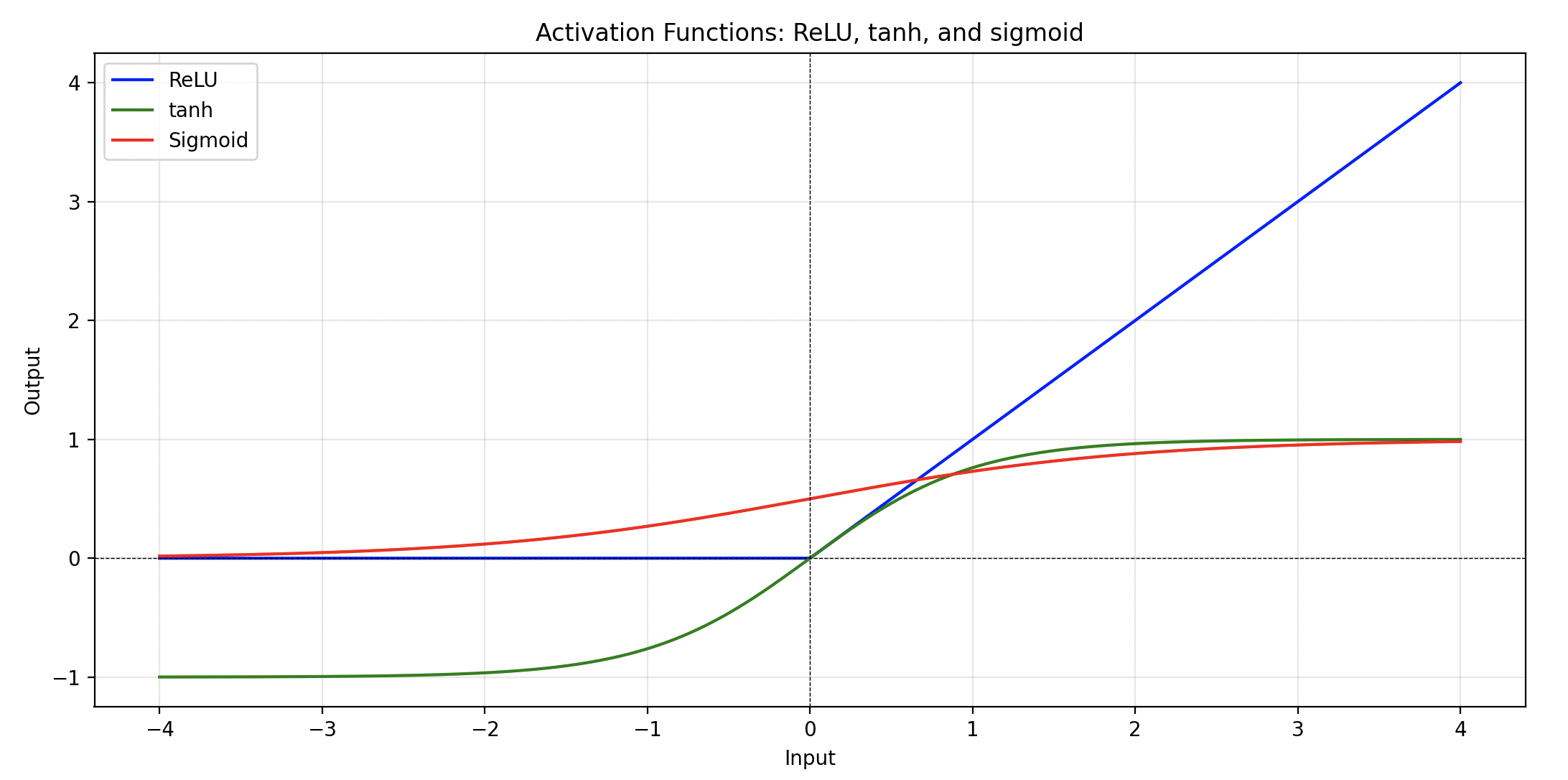}
			\caption{Tanh, ReLu, and sigmoid activation functions.}
			\label{fig:activ-functions}
		\end{center}
	\end{figure}

	Figure~\ref{fig:activ-functions} plots the \textit{tanh}, ReLU, and sigmoid activation functions. The choice among these functions often depends on the specific requirements of the problem. ReLU is computationally efficient and helps alleviate the \textit{vanishing gradient problem}, where small gradients shrink further across layers. The bounded nature of \textit{tanh} makes it well-suited for tasks where the action space or value representation is naturally constrained. The sigmoid function is particularly useful in output layers when the network needs to produce a probability. In practice, these activation functions may be combined in different parts of the network to exploit their respective advantages. The interested reader can find a thorough introduction to deep learning and artificial neural networks in \cite{goodfellow2016deep}.

	\subsubsection*{MLP for Policy Modeling}

	An MLP can be used to model a policy, such as in REINFORCE, by interpreting the outputs of the neurons in the last layer $L$ as a vector of action preferences. Each of the $n_L$ output neurons corresponds to one action, and, given a state $s$ and all weights and biases of the MLP collected in $\boldsymbol{\theta}$, these outputs define the preference scores over the action space:
	\[
		\mathbf{y}^{(L)}(s, \boldsymbol{\theta})=(y^{(L)}_1(s, \boldsymbol{\theta}), y^{(L)}_2(s, \boldsymbol{\theta}), \ldots, y^{(L)}_{n_L}(s, \boldsymbol{\theta})).
	\]

\begin{figure}[h]
		\centering
		\includegraphics[width=1\textwidth]{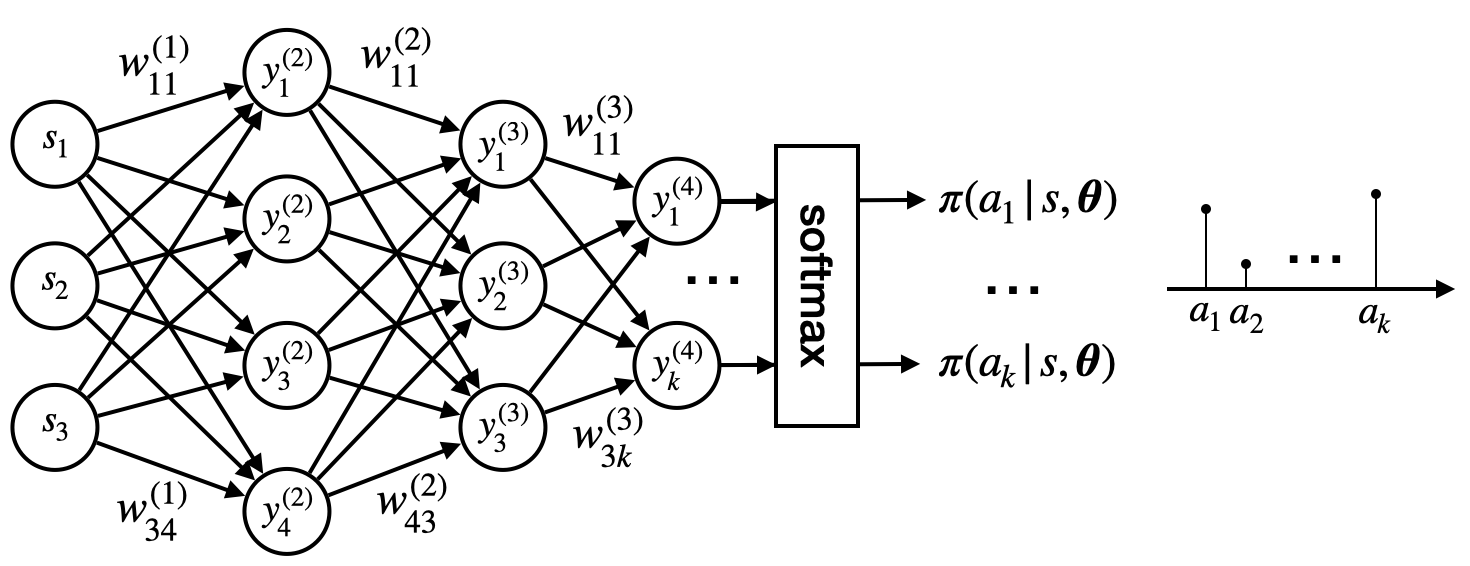}
		\caption{Architecture of a 4-Layer MLP modeling a policy over $k$ discrete actions. The plot on the right shows the resulting probability distribution (probability mass function) over the $k$ actions.}
		\label{fig:network_policy}
	\end{figure}

	The preferences vector is transformed into a probability distribution over the $n_L$ actions by applying the softmax function (as in Equation~\ref{equ:soft-max-preferences}), resulting in the following policy function:
	\begin{equation}\label{equ:mlp_with_softmax}
		\pi(a|s,\boldsymbol{\theta}) \doteq \text{softmax}(\mathbf{y}^{(L)}(s, \boldsymbol{\theta}))[a],
	\end{equation}

	\noindent where $\text{softmax}(\mathbf{y}^{(L)}(s, \boldsymbol{\theta}))[a]$ indicates the probability assigned to action $a$ (see Section~\ref{sec:softmax}). Figure~\ref{fig:network_policy} depicts an example MLP for policy function modeling.

	\subsubsection*{MLP for State-Value Function Modeling}

\begin{figure}[h]
		\centering
		\includegraphics[width=0.7\textwidth]{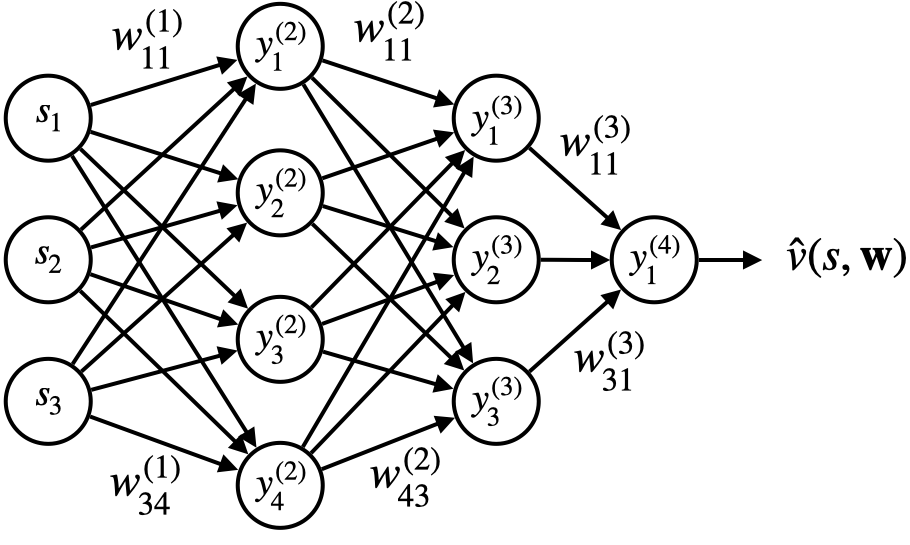}
		\caption{Diagram of an MLP for state-value function modeling.}
		\label{fig:network_value}
	\end{figure}

	An MLP can also be used to model a state-value function, serving, for instance, as a baseline in REINFORCE. Figure~\ref{fig:network_value} illustrates an example MLP architecture designed for approximating the state-value function. As the figure shows, the MLP for state-value functions is typically similar to that used for policy modeling, with one important difference in the last layer. The final layer now contains a single neuron, $n_L = 1$, since the goal is to output a scalar that estimates the value of the state $s$ rather than a probability distribution. Naturally, the MLP for modeling the state-value function requires its own set of weights and biases, represented by $\mathbf{w}$. Bearing this in mind, the MLP directly approximates $v_{\pi_{\boldsymbol{\theta}}}(s)$ as:
	\begin{equation*}
		\hat{v}(s,\mathbf{w}) \doteq y_1^{(L)}(s,\mathbf{w}).
	\end{equation*}

	\subsubsection*{Training the MLP for State-Value Function Modeling}

	As shown in Equation~\ref{equ:w-update-pi}, training a parametric state-value function amounts to updating its parameters in the direction that reduces the mean squared error between the predicted value and the true state value (Equation~\ref{equ:squared-error}). This corresponds to a gradient descent update, where the error function, commonly called the \textit{loss function}, is minimized so that the estimated value approaches the true one.
	However, since the true state value cannot be known in advance, the best we can do is rely on a sample-based estimate obtained by interacting with the environment. By executing rollouts under the current policy, we can compute the return $G_t$, which serves as an unbiased estimator of the state value. Consequently, the loss function at time $t$ is defined as:
	\begin{equation}\label{equ:loss-value}
		\mathcal{L}_t^{\hat{v}}(\mathbf{w}) = \left(G_t - \hat{v}(S_t, \mathbf{w}) \right)^2.
	\end{equation}
	The corresponding parameter update is obtained by applying gradient descent to this loss function with a learning rate $\alpha_w$:
	\begin{equation*}
		\mathbf{w}_{t+1} \doteq \mathbf{w}_t - \frac{\alpha_w}{2} \nabla_{\mathbf{w}} \mathcal{L}_t^{\hat{v}}(\mathbf{w}).
	\end{equation*}

	This expression is equivalent to the update rule in Equation~\ref{equ:update_w_params}: when taking the gradient of the squared error, the exponent contributes a factor of 2, which cancels the factor $1/2$ in the update. Although mathematically equivalent, this formulation is more conventional in the context of neural networks, where one typically defines a loss function and relies on software frameworks to automatically compute gradients and apply the corresponding gradient descent step. In the following, we examine how these gradients can be computed.

	MLP layers operate as multi-input, multi-output functions composed sequentially. From this perspective, an MLP is a composite function, where each layer \(i\) corresponds to a function \(f_i\) (as defined in Equation~\ref{equ:neurons}), parameterized by its own learnable weights and biases. This abstraction is useful for what follows. For example, a three-layer network for modeling a state-value function can be written as a composition of nested functions. Given a state vector $s$, the estimated value is computed according to the learnable parameters $\mathbf{w}$ as:
	\begin{equation*}
		\hat{v}(s,\mathbf{w}) \doteq f_3(f_2(f_1(s, \mathbf{w}^{(1)}), \mathbf{w}^{(2)}), \mathbf{w}^{(3)}), \quad \text{where } \mathbf{w} = \mathbf{w}^{(1)} \cup \mathbf{w}^{(2)} \cup \mathbf{w}^{(3)}.
	\end{equation*}

	Because $\hat{v}(S_t,\mathbf{w})$  is a composite function, the gradient of the corresponding loss function is also a composite function. Therefore, the gradient with respect to the parameters of layer $i$ can be obtained by applying the chain rule of calculus. Let  $y_i=f_i(\cdot,\cdot)$. Then:
	\begin{align*}
		\frac{d \mathcal{L}_t^{\hat{v}}}{d {\mathbf{w}}^{(1)}} & = \frac{d \mathcal{L}_t^{\hat{v}}}{d y_3} \cdot \frac{d y_3}{d y_2} \cdot \frac{d y_2}{d y_1} \cdot \frac{d y_1}{d {\mathbf{w}}^{(1)}}, \\
		\frac{d \mathcal{L}_t^{\hat{v}}}{d {\mathbf{w}}^{(2)}} & = \frac{d \mathcal{L}_t^{\hat{v}}}{d y_3} \cdot \frac{d y_3}{d y_2} \cdot \frac{d y_2}{d {\mathbf{w}}^{(2)}},                           \\
		\frac{d \mathcal{L}_t^{\hat{v}}}{d {\mathbf{w}}^{(3)}} & = \frac{d \mathcal{L}_t^{\hat{v}}}{d y_3} \cdot \frac{d y_3}{d {\mathbf{w}}^{(3)}}.
	\end{align*}

\begin{figure}
		\centering
		\includegraphics[width=10cm]{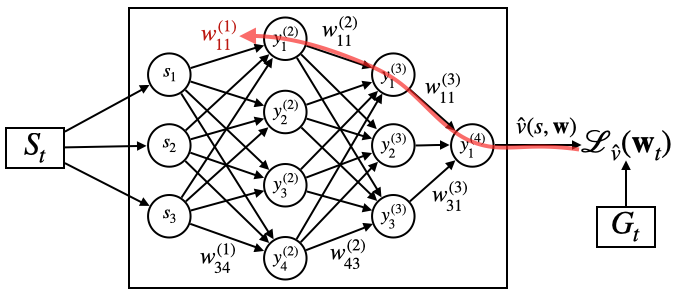}
		\caption{Schematic overview of the state-value function training process.}
		\label{fig:train_mlp}
	\end{figure}

	The gradient of the loss with respect to each parameter of the MLP can therefore be computed using the chain rule, and the parameters are updated by moving them in the direction that decreases the loss (gradient descent). Repeating this process gradually reduces the prediction error, enabling the MLP to better approximate the state-value (expected discounted return).
	Figure~\ref{fig:train_mlp} illustrates this training procedure.

	\subsubsection*{Training the MLP for Policy Function Modeling}

	Training an MLP to model a policy follows a similar procedure to that of the state-value function. In this case, the goal is to perform gradient ascent on the objective $G_t \ln \pi(A_t | S_t,\boldsymbol{\theta})$  (Equation~\ref{equ:update}), using the sampled tuples $(S_t, A_t, G_t)$ collected from policy rollouts in the environment. Intuitively, we want the policy $\pi_{\boldsymbol{\theta}}$ to assign a higher probability to the executed action $A_t$ in state $S_t$ when the observed return $G_t$ is positive, and lower probability when the return is negative.

	To use standard gradient descent machinery, we simply negate the objective, defining a loss function whose minimization corresponds to maximizing the original objective. Making explicit that the policy is the softmax of the MLP output layer (Equation~\ref{equ:mlp_with_softmax}), we obtain:
	\begin{equation}\label{equ:loss_with_softmax}
		\mathcal{L}_t^{\pi}(\boldsymbol{\theta}_t)=G_t \ln \pi(A_t | S_t,\boldsymbol{\theta}) = G_t \ln \text{softmax}(\mathbf{y}^{(L)}(S_t,\boldsymbol{\theta}))[A_t].
	\end{equation}

	As with the state-value MLP, training the policy MLP relies on applying the chain rule to the loss function: the gradient of this loss with respect to each weight of the network is computed, and the weights are then updated in the opposite direction of the gradient (gradient descent). This iterative process progressively decreases the loss, allowing the policy MLP to assign higher probabilities to actions that tend to yield higher discounted returns.

	In the case of squared-error losses, it is common to include a factor $\tfrac{1}{2}$ in the gradient descent update so that the factor of 2 from differentiating the square cancels out. However, for the policy loss, which is not a squared error, no such factor appears in the gradient, and therefore there is no need to include a $\tfrac{1}{2}$ in the update step. With a learning rate $\alpha_{\pi}$, the gradient descent update is thus written as:
	\begin{equation*}
		\boldsymbol{\theta}_{t+1} \doteq \boldsymbol{\theta}_t - \alpha_{\pi} \, \nabla_{\boldsymbol{\theta}} \mathcal{L}_t^{\pi}(\boldsymbol{\theta}).
	\end{equation*}

	\subsubsection*{Improving the training of the MLPs}

	As the network's complexity increases, finding a closed-form solution for  gradients becomes increasingly difficult. Instead, the \textit{backpropagation} algorithm \cite{rumelhart1986learning} is typically used to efficiently compute these gradients by iterating backward through the network, layer by layer. Backpropagation avoids redundant calculations of intermediate terms by storing partial derivatives, making it highly suitable for training deep networks.

	While simple gradient descent can update the parameters, more sophisticated optimization algorithms like Adam are often preferred to improve training efficiency and convergence. Adam, short for Adaptive Moment Estimation, is a popular optimization algorithm in deep learning that combines the benefits of momentum and RMSprop. Momentum accelerates convergence by smoothing the gradient updates, while RMSprop adapts the learning rate for each parameter based on recent gradient magnitudes. Adam integrates these methods by using exponentially weighted moving averages of both the gradients' mean (first moment) and variance (second moment), allowing it to adapt the learning rate dynamically for each parameter, making it more robust to noisy or sparse data.

	In training neural networks, a significant challenge is overfitting, where the model becomes overly complex and performs well on training data but fails to generalize to new, unseen data. Overfitting occurs when the model captures not just the underlying patterns but also the noise in the training set. To combat this, regularization techniques are used, with weight decay being one of the most effective and commonly applied methods. Weight decay, also known as L2 regularization, is implemented by adding a penalty term to the model's loss function that is proportional to the squared L2 norm of the model's weights.  Formally, if the original loss function is \( \mathcal{L}(\boldsymbol{\theta}) \), where \( \boldsymbol{\theta} \) represents the vector of model weights, the modified loss function with weight decay is:
	\[
		\mathcal{L}_{regularised}(\boldsymbol{\theta}) = \mathcal{L}(\boldsymbol{\theta}) + \lambda \|\boldsymbol{\theta}\|^2,
	\]

	\noindent where \( \lambda \) is a hyperparameter that controls the strength of the regularization, and \( \|\boldsymbol{\theta}\|^2 \) represents the squared L2 norm of the weight vector. This penalty term discourages the model from assigning excessively large weights to any single feature, effectively keeping the weights small.

	The benefits of weight decay go beyond just preventing overfitting. By constraining the weights, it stabilizes the training process, ensuring that the model does not develop excessively large weights, which could lead to unstable and erratic behavior during optimization. Additionally, by encouraging smaller weights, weight decay implicitly guides the model toward simpler solutions that are more likely to generalize well to new data. In essence, weight decay helps in producing models that are both robust and effective in making accurate predictions on unseen data.

	The training of an MLP can be early stopped when the rewards stabilize over a set of sequential episodes. This prevents overfitting to the specific trajectories encountered during training and ensures the model does not become overly specialized. Early stopping also saves computational resources by avoiding unnecessary training once the model has converged.

	For stable and efficient learning, it is crucial that the inputs to the MLP are zero-centered and have unit variance. This ensures that each input dimension contributes equally to the learning process, preventing any single feature from disproportionately influencing the updates. To achieve this, the state vector \( S_t = (S_{i,t})_i \) should be normalized before being fed into the MLPs. Typically, a normalized version of the state vector, \( \tilde{S}_t = (\tilde{S}_{i,t})_i \), is defined as:
	\begin{equation}\label{equ:normalisation}
		\tilde{S}_{i,t} = \frac{S_{i,t} - \mu_{{i,t}}}{\sigma_{{i,t}}},
	\end{equation}

	\noindent where \( \mu_{{i,t}} \) and \( \sigma_{{i,t}} \) represent the running mean and standard deviation, respectively, of the $i$-th component of the state vector up to time step $t$. These statistics are updated incrementally as the agent interacts with the environment, enabling the model to adjust to shifts in the data distribution over time.

	\subsubsection*{Example 1: The CartPole Problem (revisited)}

	Let us now solve the CartPole-v1 problem using REINFORCE, this time employing an MLP as the policy function and another MLP as the state-value function (for baseline), as formulated in the previous sections.

	Given that state in CartPole-v1 is represented by a 4-dimensional vector, $s\doteq(x, x', \theta, \theta')$, both MLPs have 4 neurons in the first layer, $n_1=4$. The MLP for the policy function has two hidden layers with 128 neurons each, $n_2=n_3=128$, whereas the MLP for the value function has two hidden layers with fewer neurons , $n_2=32, n_3=24$. The output layer in the MLP for the policy function has two neurons, $n_4=2$, as many as the number of possible actions, $\mathcal{A}=\{\text{left}, \text{right}\}$. For improved learning, the MLP is fed with the normalized version of the state, $\tilde{s}$ (Equation~\ref{equ:normalisation}).

	The state value changes as the policy changes. Hence, the state-value function MLP must train faster than the policy function MLP.  This is ensured by setting the learning rates for the policy and state-value functions MLPs as 1e-4 and 1e-2, respectively. The discount factor in both MLPs has been set to $\gamma = 0.99$. The hyperparameter controlling the strength of the weight decay regularization has been set to $\lambda = 0.02$. Training is early stopped as soon as a sequence of five episodes in which the maximum reward of 500 is achieved is found.

\begin{figure}[h]
		\centering
		\begin{subfigure}[b]{0.8\textwidth}
			\centering
			\includegraphics[width=\textwidth]{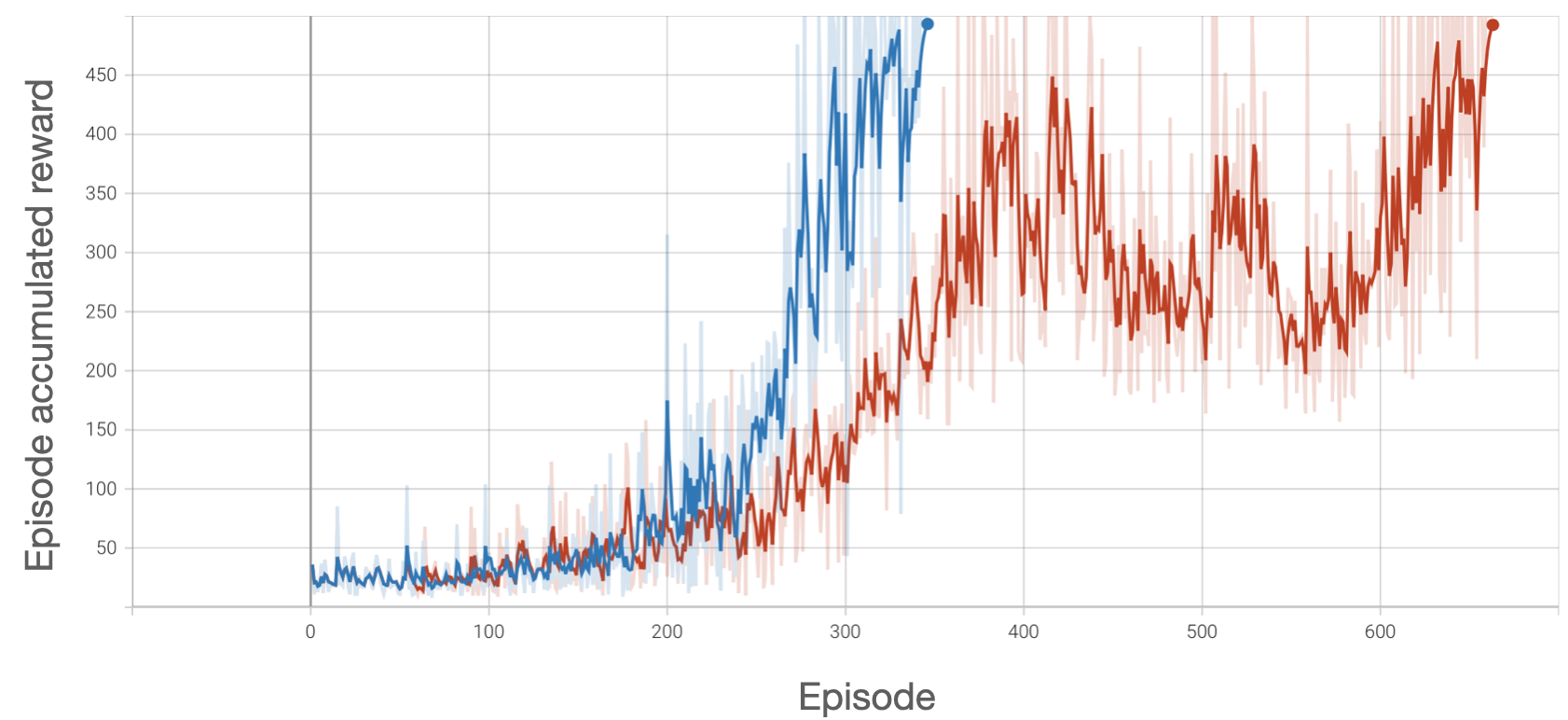}
			\caption{Seed 100.}
		\end{subfigure}
		\hspace{1cm}
		\begin{subfigure}[b]{0.8\textwidth}
			\centering
			\includegraphics[width=\textwidth]{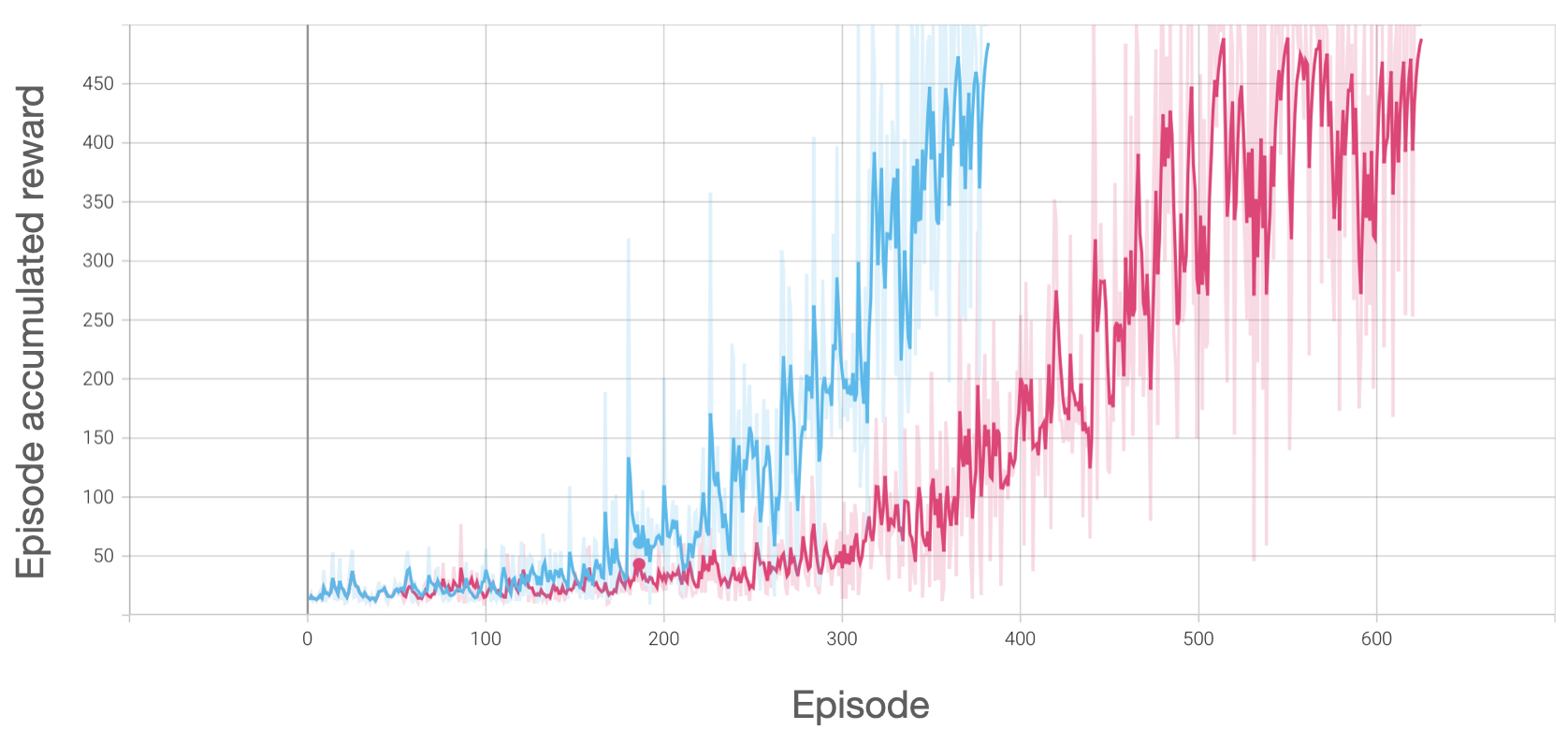}
			\caption{Seed 200.}
		\end{subfigure}
		\caption{Smoothed accumulated reward (undiscounted) per learning episode for the CartPole-v1 problem, using the REINFORCE algorithm with MLPs for policy and state-value functions. The top and bottom plots correspond to two different random seeds, with unsmoothed data shown in lower opacity. Blue plots indicate the use of the state-value function MLP as a baseline, while red plots represent results without a baseline.}
		\label{fig:cart_ann_plots}
	\end{figure}

	Figure~\ref{fig:cart_ann_plots} displays the accumulated reward (undiscounted) per learning episode for the CartPole-v1 problem, using the REINFORCE algorithm with MLPs for both the policy and state-value functions. The results demonstrate that the problem is successfully solved. Additionally, the figure highlights that incorporating the state-value function MLP as a baseline leads to faster and more stable learning compared to training without a baseline.

	\subsubsection*{Example 2: The Lunar Lander Problem}

\begin{figure}[ht]
		\centering
		\includegraphics[width=10cm]{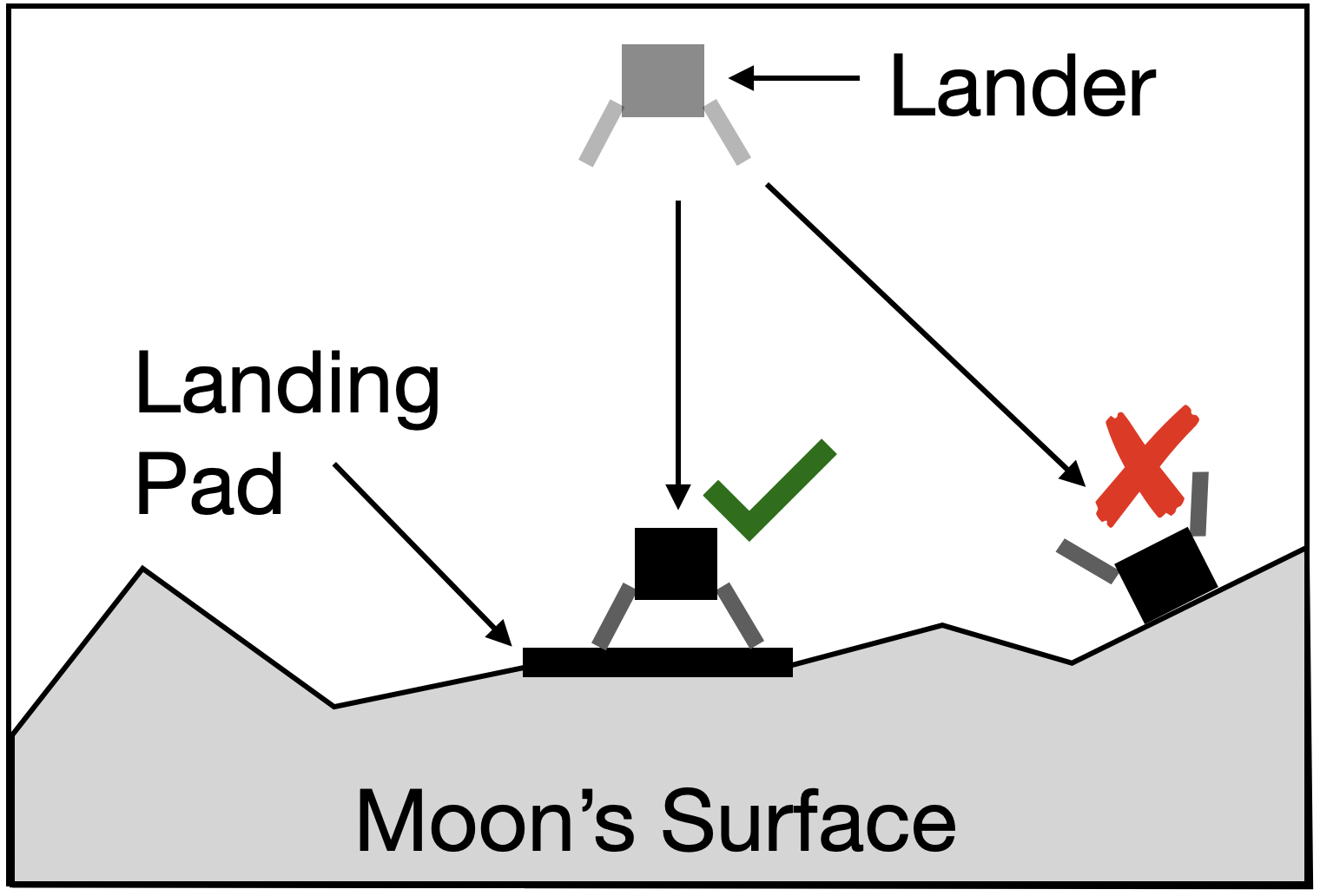}
		\caption{The OpenAI Lunar Lander problem. The objective is to safely land the lunar lander within the designated landing pad. The figure illustrates the lander's initial position (shaded at the top) and two possible outcomes: a successful landing on the pad and a crash on the lunar surface.}
		\label{fig:lunar-lander-problem}
	\end{figure}

	The Lunar Lander problem, instantiated by OpenAI as LunarLander-v2\footnote{\url{https://www.gymlibrary.dev/environments/box2d/lunar_lander/}}, is a well-known benchmark in reinforcement learning that involves safely landing a spacecraft on the surface of the moon (see Figure~\ref{fig:lunar-lander-problem}). The goal is to control the lander's descent using its main and side thrusters, ensuring a soft landing on the designated landing pad, while adhering to the dynamics imposed by the simulated environment. This problem encompasses challenges related to continuous control, delayed rewards, and the need for precision.

	The state is represented by an 8-dimensional vector, \(s \doteq (x, y, \dot{x}, \dot{y}, \theta, \dot{\theta}, l_c, r_c)\), where \(x\) and \(y\) denote the lander's horizontal and vertical positions, \(\dot{x}\) and \(\dot{y}\) are the horizontal and vertical velocities, \(\theta\) is the angle of the lander relative to the vertical, and \(\dot{\theta}\) is its angular velocity. The binary variables \(l_c\) and \(r_c\) indicate whether the left or right leg of the lander is in contact with the ground, respectively. The agent can execute one of four discrete actions, \(\mathcal{A} = \{\text{do nothing}, \text{fire left engine}, \text{fire main engine}, \text{fire right engine}\}\). Choosing to "do nothing" results in no thrust being applied. Firing the left engine generates a clockwise torque, while firing the right engine produces a counterclockwise torque. Firing the main engine applies an upward force, reducing the descent velocity and controlling the lander's vertical motion.

	The lander earns approximately 100 to 140 points for successfully descending from the top of the screen to the landing pad and coming to a complete stop. Points are deducted if the lander drifts away from the landing pad. A crash results in an additional penalty of -100 points, while safely coming to rest grants an extra +100 points. The lander also earns +10 points for each leg that makes contact with the ground. Using the main engine incurs a cost of -0.3 points per frame, and using the side engines costs -0.03 points per frame. The problem is considered solved if the agent achieves a score of 200 points.

	The lunar lander starts each episode positioned at the top center of the viewport, with a random initial force applied to its center of mass. The episode concludes under several conditions: if the lander crashes by making contact with its body in the moon's surface, if it moves outside the bounds of the viewport (specifically, if its x-coordinate exceeds the boundary), or if the lander does not move and does not collide with any other body.

\begin{figure}[h]
		\begin{center}
			\includegraphics[width=10cm]{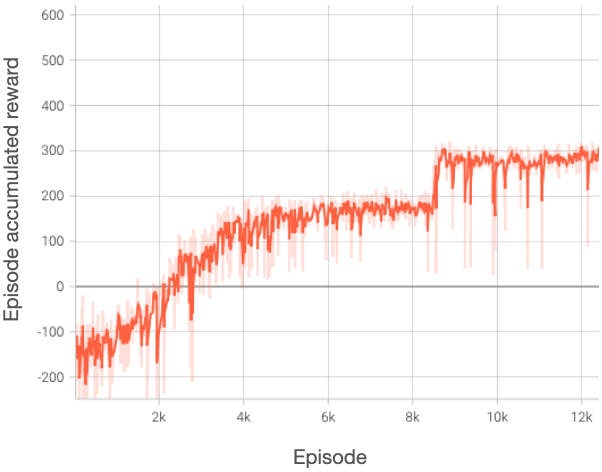}
			\caption{REINFORCE algorithm using MLPs for both the policy and state-value functions in the LunarLander-v2 problem (baseline used). The plot presents the  accumulated reward (undiscounted), per learning episode, with unsmoothed data shown in lower opacity.}
			\label{fig:results-lunar-lander}
		\end{center}
	\end{figure}

	Figure~\ref{fig:results-lunar-lander} shows the accumulated reward (undiscounted) per learning episode for the LunarLander-v2 problem, using the REINFORCE algorithm with MLPs for both the policy and state-value functions, similar to the approach used for the CartPole-v1 problem and using the baseline. The results indicate that the problem is successfully solved, with rewards consistently approaching 300, though it requires significantly more training time compared to CartPole-v1. The plot also reveals that learning stagnated during several episodes before making a final push toward higher rewards.

	\subsection{Policy Parameterization for Continuous Actions}
	\label{sec:continuous_actions}

	Up to this point, we have focused on handling discrete action spaces using stochastic policies. In this setting, sampling an action from the policy involves learning the probability distribution over the set of possible actions. This approach is feasible when the action space is discrete, as there are a finite number of possible actions, and one can assign probabilities to each. However, in continuous control problems, where the action space consists of real-valued actions (e.g., the torque to be applied to a joint in a robotic arm), the situation is more challenging. In such cases, learning the probability of each individual action would imply learning the probabilities of an infinite number of actions, as the action space consists of all real numbers within a given domain.

	An alternative to using a stochastic policy is to use a deterministic policy function. A deterministic policy maps the current state directly to a single action, which simplifies the decision-making process. Specifically, the deterministic policy function outputs a specific action given the current state, and this action is then executed. This approach has the advantage that the policy function needs to be evaluated only once per time step, as there is no need to sample from a distribution over actions.

	However, one major drawback of deterministic policies is that they do not inherently support exploration during training. Since a deterministic policy always outputs the same action for a given state, it lacks the variability needed to explore different parts of the action space. Exploration is crucial in reinforcement learning, especially during the early stages of training, to ensure that the policy does not prematurely converge to suboptimal actions. To address this, one must explicitly manage the exploration-exploitation trade-off when using deterministic policies. A common method to achieve this is the $\epsilon$-greedy strategy, in which the agent takes a random action with probability $\epsilon$ at each time step, and with probability $1-\epsilon$, it follows the deterministic policy. This way, the agent balances between exploration (random actions) and exploitation (following the policy's recommendations).

	On the other hand, stochastic policies naturally handle this trade-off. In a stochastic policy, actions are sampled from a probability distribution, which allows for exploration without needing external mechanisms. Instead of deterministically choosing an action, the policy outputs a distribution over possible actions, and actions are sampled according to their probabilities. This probabilistic sampling introduces variability in the agent's behavior, ensuring that it explores different actions even when the same state is encountered multiple times.

	In summary, stochastic policies are effective for exploration but necessitate additional mechanisms to manage continuous action spaces. The following section discusses these mechanisms.

	\subsubsection*{Gaussian Policies}

	A common approach for continuous action spaces is to learn the parameters of a probability distribution over actions, such as the Gaussian (see Section~\ref{sec:gaussian-distribution}). The idea is to model the distribution's statistics, its mean and standard deviation, as functions of the state. The mean represents the most likely action for the current state, while the standard deviation determines the spread of probabilities around the mean, effectively encoding the exploration space. This learned distribution is then sampled to produce the current action, enabling the policy to handle an infinite set of possible actions. In contrast, the discrete case directly learns the probability of each possible action.

	To employ a Gaussian distribution as a stochastic policy, both the mean and standard deviation are parametrized as functions of the state, typically modeled with MLPs. These functions are denoted as $\mu(S_{t}, \boldsymbol{\theta})$ and $\sigma(S_{t}, \boldsymbol{\theta})$ for state $S_{t}$ and learnable parameters $\boldsymbol{\theta}$. This can be implemented using either two separate MLPs, each producing a single output (one for $\mu$ and one for $\sigma$), or a single MLP with two outputs, where one output represents the mean and the other the standard deviation. The choice of architecture depends on the complexity of the problem and computational considerations. This leads to the following formulation of the stochastic policy, modeled as a Gaussian probability density function \cite{sutton2018reinforcement}:
	\begin{equation}\label{equ:policy-gaussian}
		\pi(A_{t} \mid S_{t}, \boldsymbol{\theta}) = \frac{1}{\sigma(S_{t}, \boldsymbol{\theta}) \sqrt{2 \pi}} \exp \left( -\frac{(A_{t} - \mu(S_{t}, \boldsymbol{\theta}))^2}{2 \sigma(S_{t}, \boldsymbol{\theta})^2} \right).
	\end{equation}

	The policy $\pi(A_{t} \mid S_{t}, \boldsymbol{\theta})$ is a composite function, consisting of the mean $\mu(S_{t}, \boldsymbol{\theta})$ and standard deviation $\sigma(S_{t}, \boldsymbol{\theta})$. This compositional structure does not hinder parameter optimization. The gradient of the policy function with respect to the parameters $\boldsymbol{\theta}$ propagates through both $\mu(S_{t}, \boldsymbol{\theta})$ and $\sigma(S_{t}, \boldsymbol{\theta})$ with the chain rule. Thus, during the policy gradient update, the optimization process effectively adjusts both the mean and standard deviation according to the loss function.

	Let us build some intuition about the training process. Recall that the policy loss is given by $\mathcal{L}_t^{\pi}(\boldsymbol{\theta}_t) = G_t \ln \pi(A_t \mid S_t,\boldsymbol{\theta})$ (Equation~\ref{equ:loss_with_softmax}) and that the gradient of this loss points in the direction that increases the probability of selecting action $A_t$ in state $S_t$ if the sampled return $G_t$ is positive, and to decrease it otherwise.

	When the policy is modeled as a Gaussian distribution, this means modifying the parameters $\boldsymbol{\theta}$ so that the MLP responsible for the mean $\mu(S_t, \boldsymbol{\theta})$ produces a value closer to the sampled action $A_t$ when the input is $S_t$ if $G_t>0$, and shifts it away from $A_t$ if $G_t<0$. Thus, positive returns reinforce the executed action, whereas negative returns discourage it.

	The behavior of the policy gradient with respect to the network that outputs the standard deviation $\sigma(S_t,\boldsymbol{\theta})$ differs from that of the mean. While the mean network shifts the action distribution toward or away from the sampled action $A_t$, the gradient with respect to the standard deviation governs the \emph{spread} of the policy. Positive returns $G_t>0$ push the network to reduce uncertainty around useful actions by decreasing $\sigma$, whereas negative returns $G_t<0$ encourage greater exploration by increasing $\sigma$. That is, the mean network determines \emph{what action to favor}, whereas the standard deviation network determines \emph{how confidently the policy should act} in each state.

\begin{figure}[h]
		\centering
		\includegraphics[width=14cm]{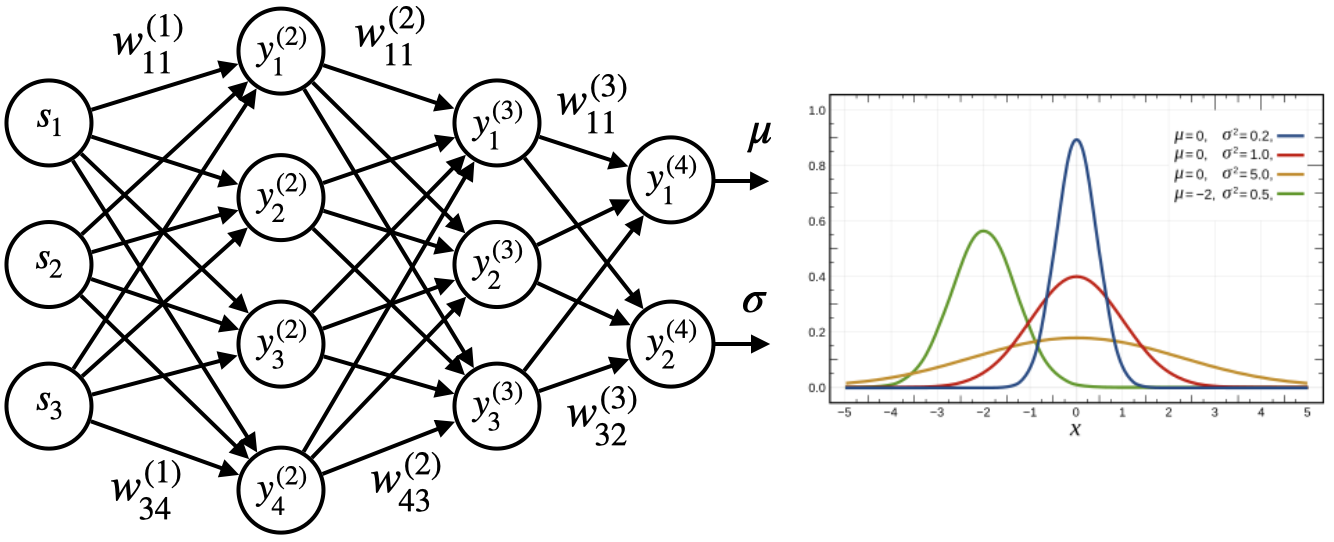}
		\caption{Architecture of a 4-Layer MLP modeling a policy over one continuous action. The plot on the right depicts possible probability density functions representing the distribution over the continuous action values.}
		\label{fig:network_policy_pdf_b1}
	\end{figure}

	Figure~\ref{fig:network_policy_pdf_b1} illustrates an example of an MLP modeling a policy over a single continuous action. In this setup, the network outputs parameters for both the mean and the standard deviation of the action distribution.

	\subsubsection*{Sampling Actions}

	With continuous action spaces, actions are sampled from a Gaussian distribution whose parameters depend on the current state. Specifically, at each time step \( t \), the action \( A_{t} \) is drawn from a normal distribution with state-dependent mean and standard deviation:
	\[
		A_{t} \sim \mathcal{N}(\mu(S_{t}, \boldsymbol{\theta}), \sigma(S_{t}, \boldsymbol{\theta})^2).
	\]

	In many environments, the action space is bounded, typically within an interval such as \( [-1, 1] \). To ensure the sampled actions respect these bounds, a squashing function, commonly the hyperbolic tangent, is applied to the output of the Gaussian sample:
	\[
		A_{t} = \tanh(\tilde{A}_{t}), \quad \tilde{A}_{t} \sim \mathcal{N}(\mu(S_{t}, \boldsymbol{\theta}), \sigma(S_{t}, \boldsymbol{\theta})^2).
	\]

	This transformation ensures that the final action remains within the allowed range, while preserving the stochasticity and flexibility of the Gaussian policy. When using such squashing functions, care must be taken when computing action probabilities, as the transformation affects the resulting probability density.

	\subsubsection*{Multidimensional Action Spaces}

	In multidimensional action spaces, such as when an agent has multiple actuators, the policy must produce actions for each dimension. A common approach is to use multiple Gaussian distributions, where each Gaussian models a single dimension of the action space. The parameters of these Gaussians, namely the means and standard deviations, can be implemented using different architectural designs depending on the problem's complexity and computational requirements.

	One approach is to use separate MLPs for each Gaussian, with each MLP independently producing the mean and standard deviation for a single dimension. This setup offers maximum flexibility for learning independent parameterizations across dimensions. Alternatively, a single MLP with multiple outputs can produce the parameters for all Gaussians. In this case, the MLP outputs the means and standard deviations for all dimensions, making it a more compact and computationally efficient solution. A third, hybrid approach, involves using an MLP with shared layers to extract common features across all dimensions of the action space, followed by specialized output layers (or heads) that produce the mean and standard deviation for each Gaussian. This design balances parameter sharing with task-specific flexibility.

	The choice of architecture depends on the degree of correlation between dimensions, the complexity of the action space, and the computational resources available. When there are dependencies or shared features across dimensions, using shared layers or a single MLP often works well. In contrast, separate MLPs may be more suitable for cases where dimensions are highly independent.

	\section{On-Policy Actor-Critic Methods}

	In the previous section, the foundational principles of policy gradient methods were explored, culminating in the algorithm REINFORCE \cite{williams1992reinforce}, which trains non-linear policies typically represented by artificial neural networks. While REINFORCE laid the groundwork, it has limitations that hinder its effectiveness in solving more complex reinforcement learning problems.

	This section gradually introduces the concepts that underly Proximal Policy Optimization (PPO) \cite{schulman2017ppo}, one of the most widely used and effective reinforcement learning algorithms. PPO is a \textit{model-free}, \textit{on-policy}, \textit{actor-critic} method, which combines policy gradients with value-based estimators. This combination addresses the shortcomings of REINFORCE, providing a more stable and scalable approach for training policies in challenging environments.

	\subsection{Advantage Functions}

	As already discussed, REINFORCE is a Monte Carlo method, meaning that it must wait until the end of the episode to obtain a single sample of the return, $G_t$. $G_t$ being a single sample, it is an unbiased high variance estimate of the true expected return. The high variance comes from the fact that $G_t$ accumulates delayed rewards over potentially long episodes. This makes REINFORCE a slow learning algorithm. It was also shown that variance can be reduced by subtracting a baseline from $G_t$, such as $\hat{v}(S_t, \mathbf{w})$. In fact, there are many alternative methods to modulate the policy gradient when updating the policy parameters. To cope with this diversity, the update equation can be reformulated in the following general form:
	\begin{equation*}
		\boldsymbol{\theta}_{t+1} \doteq \boldsymbol{\theta}_t + \alpha \Phi_t \nabla_{\boldsymbol{\theta}}\ln \pi(A_t \mid S_t, \boldsymbol{\theta}_t),
	\end{equation*}

	\noindent where $\Phi_t$ can be defined in many different ways, including:
	\begin{equation*}
		\Phi_t \doteq
		\begin{cases}
			G_t                            & \quad\text{if in REINFORCE},                                         \\
			G_t - \hat{v}(S_t, \mathbf{w}) & \quad\text{if in REINFORCE with baseline } \hat{v}(S_t, \mathbf{w}), \\
			\hat{\Delta}_t                 & \quad\text{if in Actor-Critic (e.g., PPO)},
		\end{cases}
	\end{equation*}

	\noindent where $\hat{\Delta}_t$ is an estimator of what is known as \textit{advantage function}. When using $\Phi_t \doteq \hat{\Delta}_t$, it is common to call the resulting policy gradient algorithm of an \textit{actor-critic} algorithm. This is the case for PPO. The reason why and the definition and purpose of $\hat{\Delta}_t$ are detailed in the following paragraphs.

	With $\Phi_t \doteq G_t - \hat{v}(S_t, \mathbf{w})$, the policy function parameters are updated in such a way that the executed action $A_t$ becomes more probable if the sampled return $G_t$ is above the predicted average return (i.e., $G_t-\hat{v}(S_t, \mathbf{w})>0$) and less probable if $G_t$ is below the predicted average return (i.e., $G_t-\hat{v}(S_t, \mathbf{w})<0$). Hence, $G_t - \hat{v}(S_t, \mathbf{w})$ operates as an estimator of the relative \textit{advantage} of executing action $A_t$ relative to randomly sampling an action according to policy $\pi_{\boldsymbol{\theta}}$. However, this advantage estimator still suffers from two problems resulting from depending of $G_t$: it has high variance and can only be computed by the end of the episode. High variance slows learning and waiting for the end of the episode hampers its application to non-episodic tasks.

	The idea of \textit{advantage} can be extended to define the \textit{advantage function}, which measures how much better it is, on average, to take a specific action $a$ in state $s$ compared with selecting an action according to the policy distribution $\pi_{\boldsymbol{\theta}}$. Formally, the advantage function is defined as:
	\begin{equation*}
		\Delta_\pi(s,a)\doteq q_{\pi}(s, a)-v_{\pi}(s).
	\end{equation*}

	Intuitively, a large advantage indicates that the current policy does not allocate sufficient probability mass to an action that performs better than the alternatives. Consequently, the policy should shift and concentrate its probability density around that action.

	We adopt the convention that $t$ indexes the time step within rollout (episode) $n$. The pair of indices $(n,t)$ therefore identifies a specific sample collected during training. For example, the state observed at time step $t$ in rollout $n$ is written as $S_{n,t}$.

	A well-known quantity in reinforcement learning is the \textit{Temporal-Difference error} (TD-error), denoted by $\delta_{n,t}$. It provides a one-step estimate of the advantage function and is defined as follows \cite{sutton2018reinforcement}:
	\begin{equation*}
		\delta_{n,t} \doteq R_{n,t} + \gamma \hat{v}(S_{n, t+1}, \mathbf{w}) - \hat{v}(S_{n,t}, \mathbf{w}).
	\end{equation*}

	The TD-Error also compares the expected return with the state-value function $\hat{v}(S_{n,t}, \mathbf{w})$. However, instead of estimating the expected return with $G_{n,t}$, the TD-Error estimates it with $R_{n,t} + \gamma \hat{v}(S_{n, t+1}, \mathbf{w})$. It means that the expected return is  the reward collected in $t$ plus a discounted prediction of the accumulated discounted reward from $t+1$ onwards. This alternative estimator of the expected return has lower variance than the Monte Carlo estimator $G_{n,t}$, thus improving learning.

	The Monte Carlo estimator $G_{n,t}$ exhibits higher variance than the TD-error estimator $R_{n,t} + \gamma \hat{v}(S_{n, t+1}, \mathbf{w})$ because it depends on the sum of many future rewards, whose uncertainty accumulates over multiple steps. In contrast, the approximate state-value function $\hat{v}(S_{n, t+1}, \mathbf{w})$ summarizes future outcomes, resulting in lower variance. However, being only an approximation, it introduces some bias, reflecting a trade-off between variance and bias when moving from the Monte Carlo estimator to the TD-error estimator. Using a prediction that is gradually improved, i.e., the approximate state-value function, rather than relying solely on observed returns is referred to as \textit{bootstrapping}.

	The bias in the advantage function estimator can be reduced by extending the horizon of observed rewards, instead of considering only the current one. For instance, the reward obtained at time $t$, the discounted reward obtained at $t+1$, and finally the predicted return from $t+2$ onwards according to $\hat{v}(S_{n, t+2}, \mathbf{w})$, can be summed. The further the evaluation of $\hat{v}$ is pushed into the future, the more the bias induced by $\hat{v}$ is reduced, at the cost of increased variance due to the larger number of observed reward terms.

	Bearing this in mind, the one-step TD-error-based advantage estimator can  be generalized to a $k$-step horizon, where $k>1$ observed rewards are accumulated before bootstrapping with the value estimate $\hat{v}$ \cite{sutton2018reinforcement}:
	{\small
	\begin{align*}
		\hat{\Delta}_{n,t}^{(1)} & \doteq \delta_{n,t} = - \hat{v}(S_{n,t}, \mathbf{w}) + R_{n,t} + \gamma \hat{v}(S_{n,t+1}, \mathbf{w}) ,                                                                                           \\
		\hat{\Delta}_{n,t}^{(2)} & \doteq \delta_{n,t} + \gamma \delta_{n,t+1} = - \hat{v}(S_{n,t}, \mathbf{w}) + R_{n,t} + \gamma R_{n,t+1} + \gamma^2 \hat{v}(S_{n,t+2}, \mathbf{w}),                                               \\
		\hat{\Delta}_{n,t}^{(3)} & \doteq \delta_{n,t} + \gamma \delta_{n,t+1} + \gamma^2 \delta_{n,t+2}= - \hat{v}(S_{n,t}, \mathbf{w}) + R_{n,t} + \gamma R_{n,t+1} + \gamma^2 R_{n,t+2} + \gamma^3 \hat{v}(S_{n,t+3}, \mathbf{w}), \\
		\dots                                                                                                                                                                                                                         \\
		\hat{\Delta}_{n,t}^{(k)} & \doteq \sum_{l=0}^{k-1} \gamma^l \delta_{n,t+l} = - \hat{v}(S_{n,t}, \mathbf{w}) + R_{n,t} + \gamma R_{n,t+1} + \ldots +\gamma^{k-1} R_{n,t+k-1} + \gamma^k \hat{v}(S_{n,t+k}, \mathbf{w}).
	\end{align*}
	}

	These $k$-step estimators can be combined using an exponentially-weighted average to produce the truncated \textit{Generalized Advantage Estimator} (GAE) \cite{schulman2015high}, allowing fine control over the bias-variance trade-off:
	\begin{equation}
		\hat{\Delta}_{n,t} \doteq (1-\lambda)\left(\hat{\Delta}_{n,t}^{(1)}+\lambda \hat{\Delta}_{n,t}^{(2)}+\lambda^2\hat{\Delta}_{n,t}^{(3)}+\ldots+\lambda^{k-1}\hat{\Delta}_{n,t}^{(k)}\right).
		\label{equ:unnormalised_advantage}
	\end{equation}

	The truncated GAE has two hyper-parameters to control the bias-variance trade-off. The higher the parameter $k$ the higher the control over the trade-off, at the cost of computation. The higher the parameter $\lambda$ the higher the variance and the lower the bias. This occurs because more weight is gradually given to the actual rewards observed by the agent further in time. Decreasing $\lambda$ has the opposite effect on variance and bias.

	As mentioned, the estimated advantage function \(\hat{\Delta}_{n,t}\) is used to guide policy updates by estimating the relative value of taking a particular action in a given state. However, the scale of \(\hat{\Delta}_{n,t}\) can vary during training due to several factors: (a) changes in the policy affect the distribution of rewards and value estimates, leading to variations in the computed advantages; (b) the reward structure in the environment (e.g., sparse rewards, dense rewards, or clipped rewards) influences the scale and variability of \(\hat{\Delta}_{n,t}\); and uncertainty in the estimated value function $\hat{v}$ introduce noise or inconsistencies in the advantage calculation. These factors can cause the scale of \(\hat{\Delta}_{n,t}\) to fluctuate across training iterations, potentially leading to: (a) large advantage values causing overly aggressive updates to the policy; and (b) small advantage values with diminished impact on updates, slowing down convergence.

	To ensure a consistent scale and promote more stable and efficient training, the advantage function is typically normalized. The normalized advantage function, denoted as \(\tilde{A}_{n,t}\), is computed as:
	\begin{equation}\label{equ:normalised_advantage}
		\tilde{\Delta}_{n,t} = \frac{\hat{\Delta}_{n,t} - \mu_{\pi_\theta}}{\sigma_{\pi_\theta}},
	\end{equation}

	\noindent where \(\mu_{\pi_\theta}\) and \(\sigma_{\pi_\theta}\) are the mean and standard deviation of all unnormalized advantage function values \(\hat{\Delta}_{n,t}\) collected with the current policy $\pi_\theta$ (more details below).

	\subsection{Multiple Rollouts}

	To stabilize and accelerate training, PPO is designed to leverage multiple trajectories during each policy update. It first collects a batch of \(N\) rollouts (i.e., complete episodes) using the same policy. This fixed batch is then reused for multiple policy updates, extracting more learning signal from the same interactions with the environment. By reusing data in this way, PPO produces more stable (lower-variance) gradient estimates and also improves \emph{sample efficiency}, reducing the total number of environment interactions needed to learn a good policy.

	Each rollout consists of $T$ time steps, where potential early episode terminations are ignored for simplicity. During these rollouts, all actions taken, as well as observed states and rewards, are recorded for later use. Collecting experiences from multiple rollouts helps to average out the inherent randomness and provides more reliable gradient estimates. These rollouts can be executed sequentially or in parallel. Parallel execution can greatly accelerate data collection and is commonly utilized in practice to take advantage of multi-threaded or distributed computing environments.

\begin{figure}[h]
		\centering
		\includegraphics[width=14cm]{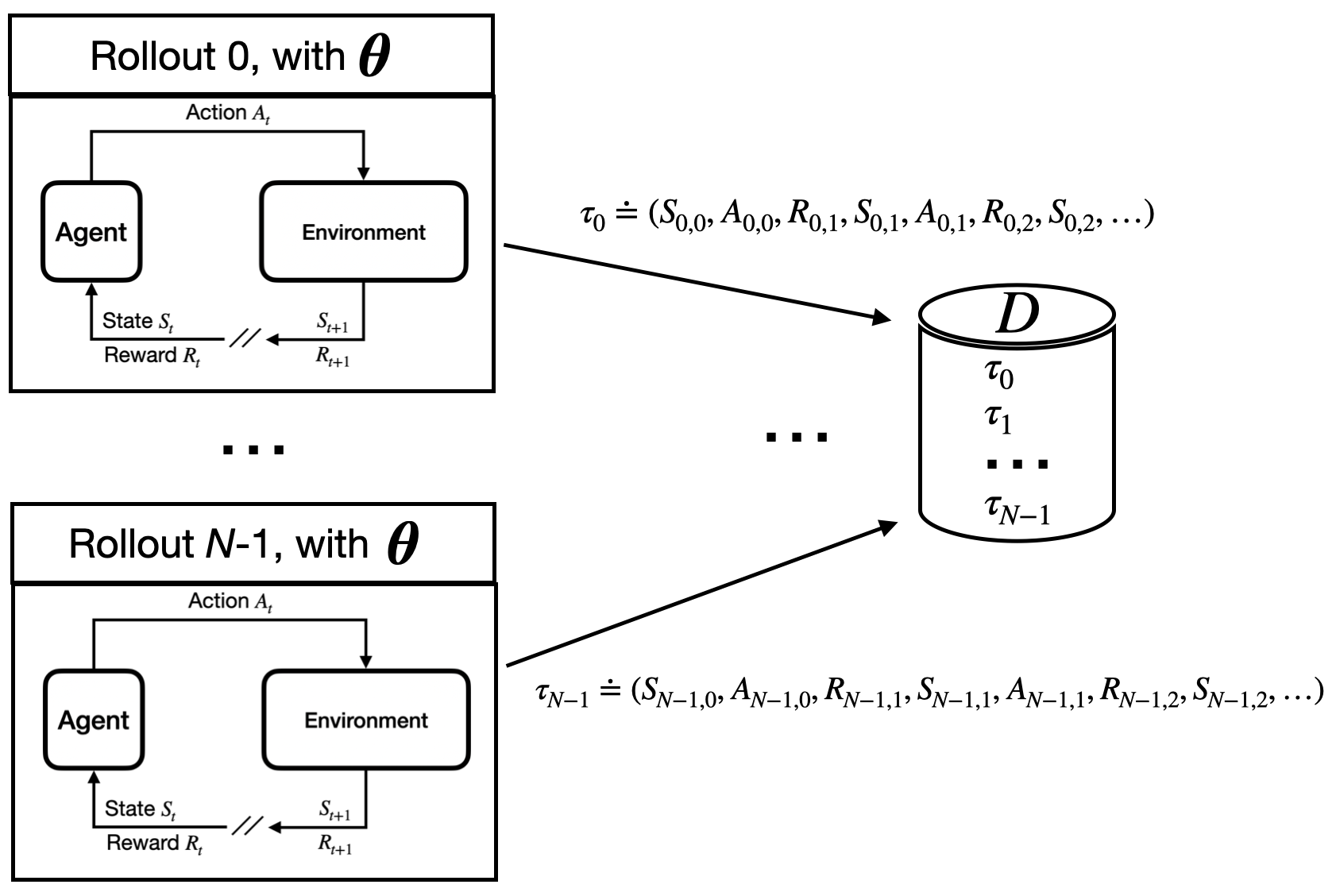}
		\caption{Collecting data by rolling out the current policy $\pi_{\theta}$ in the environment $N$ times, storing all resulting trajectories $\tau$ in the experience buffer $D$.}
		\label{fig:multiple_rollouts}
	\end{figure}

	After running the $N$ rollouts, the discounted returns for each time step are computed, as well as the corresponding unnormalized advantages. Each experience tuple, comprising the state, action, reward, discounted return, and unnormalized advantage, is stored in an \textit{experience buffer} $D$. The data is tagged with the information about the specific rollout and time step from which it was collected.  Figure~\ref{fig:multiple_rollouts} illustrates this process and Algorithm~\ref{algorithm:experiences_buffer} details how multiple rollouts can be used to build the experiences buffer.

	\begin{algorithm}[h]
		\caption{CreateExperiencesBuffer ($\pi$, $\boldsymbol{\theta}$, $\hat{v}$, $\boldsymbol{w}$, $N$, $T$, $\gamma$, $\lambda$) }
		{\footnotesize
			\begin{algorithmic}[1]
				\label{algorithm:experiences_buffer}
				\STATE \textbf{Input:}  policy function $\pi$ and its parameterization $\boldsymbol{\theta}$
				\STATE \textbf{Input:} state-value function, \( \hat{v} \), and its current parameterization, \( \boldsymbol{w} \)
				\STATE \textbf{Input:} number of rollouts $N$
				\STATE \textbf{Input:} number of time steps per rollout, $T$
				\STATE \textbf{Input:} discount factor, \( \gamma \), and weight factor for advantage, \( \lambda \)
				\STATE
				\STATE \textcolor{blue}{// Perform $N$ rollouts with the agent in the environment}
				\STATE \textbf{For each rollout} $n = 0, 1, \ldots, N-1$
				\begin{ALC@g}
					\STATE Observe initial state $S_{n,0}$
					\STATE \textbf{For each time step} $t = 0, 1, \ldots, T-1$
					\begin{ALC@g}
						\STATE Sample action from policy, $A_{n,t} \sim \pi(\cdot \mid S_{n,t}, \boldsymbol{\theta})$
						\STATE Execute action $A_{n,t}$, observe reward $R_{n,t+1}$ and next state $S_{n,t+1}$
					\end{ALC@g}
				\end{ALC@g}
				\STATE
				\STATE \textcolor{blue}{// Process the rollouts and add them to the experiences buffer}
				\STATE Reset experiences buffer, $D \leftarrow []$
				\STATE \textbf{For each rollout} $n = 0, 1, \ldots, N-1$
				\begin{ALC@g}
					\STATE \textbf{For each time step} $t = 0, 1, \ldots, T-1$
					\begin{ALC@g}
						\STATE Compute discounted return $G_{n,t} \leftarrow \sum_{k=t+1}^{T} \gamma^{k-t-1} R_{n,k}$
						\STATE Compute unnorm. advantage \( \hat{\Delta}_{n,t} \) with Equation~\ref{equ:unnormalised_advantage} (with \( \lambda \), \( \hat{v} \), \( \boldsymbol{w} \), \( \{R_{n,i}\}_{i \ge t} \), \( \{S_{n,i}\}_{i \ge t} \))

						\STATE Append data to buffer, $D\leftarrow D \cup (S_{n,t}, A_{n,t}, R_{n,t+1}, G_{n,t}, \hat{\Delta}_{n,t})$
					\end{ALC@g}
				\end{ALC@g}
				\STATE
				\STATE \textbf{Return } $D$
			\end{algorithmic}
		}
	\end{algorithm}

	\subsection{Mini-Batches}

	Experiences stored in the buffer \( D \) often exhibit temporal correlations, since they are generated sequentially by the same policy interacting with the environment. These correlations can lead to redundant information, where similar state-action transitions dominate the data and do not contribute new learning signal. Using the entire buffer for a single gradient update would be computationally expensive and wasteful, since many consecutive samples provide nearly identical information.

	To mitigate these effects, PPO estimates the gradient using mini-batches of randomly shuffled experiences. Random shuffling breaks much of the temporal correlation across rollouts and time steps, producing mini-batches that contain more diverse and less correlated samples. As a result, each gradient update becomes more informative, while remaining computationally efficient.

\begin{figure}[h]
		\centering
		\includegraphics[width=10cm]{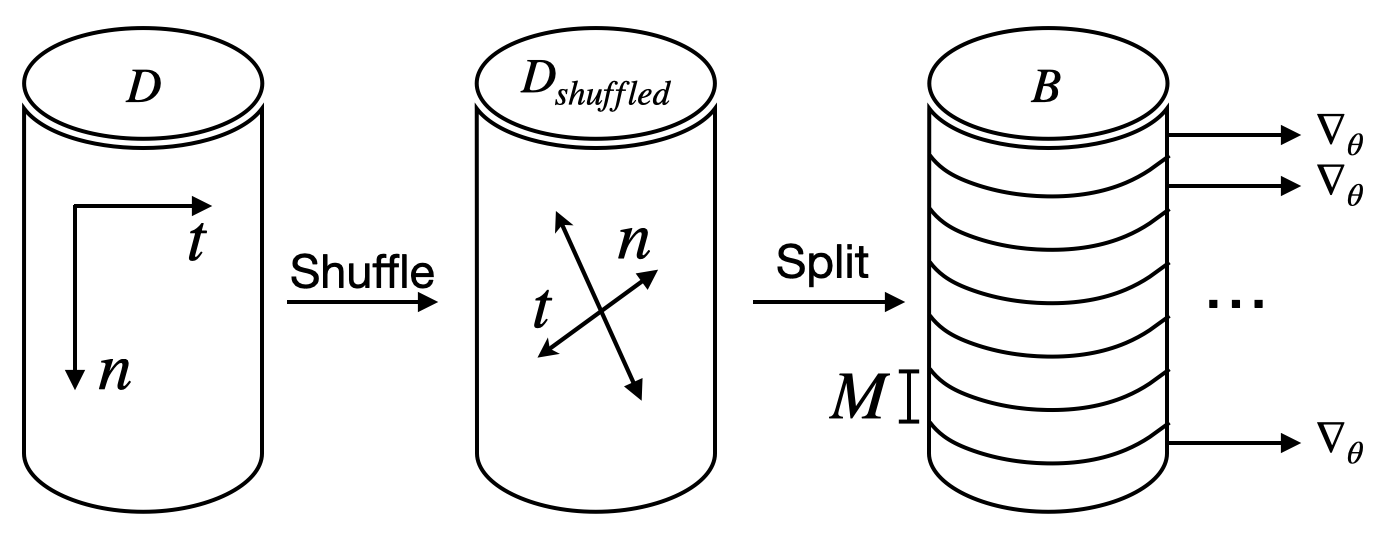}
		\caption{Creating the mini-batch list \( B \) from the experiences buffer \( D \).}
		\label{fig:mini_batches}
	\end{figure}

	To create \( M \)-length mini-batches (with \( M \) as a user-defined hyperparameter), the experiences buffer \( D \) is first shuffled to randomize the order of its elements and break existing correlations. The \textit{shuffled buffer} is denoted by \( D_{\text{shuffled}} \). The first \( M \) experiences from \( D_{\text{shuffled}} \), denoted as \( D_{\text{shuffled}}[0:M-1] \), are selected to form a \textit{mini-batch} \( D_{\text{mini}} \). Before populating \( D_{\text{mini}} \), the normalized advantage for each experience tuple in \( D_{\text{shuffled}}[0:M-1] \) is computed. Each experience tuple is then augmented with its normalized advantage and added to \( D_{\text{mini}} \). Then, the mini-batch \( D_{\text{mini}} \) is added to a list of mini-batches $B$ and the elements belonging to \( D_{\text{mini}} \) are removed from \( D_{\text{shuffled}} \). This process repeats until \( D_{\text{shuffled}} \) is left out of experiences, \( |D_{\text{shuffled}}| = 0 \). When the last remaining elements in $D_{\text{shuffled}}$ are not enough to fill a last mini-batch, the said mini-batch can be filled with randomly selected experiences from the mini-batches stored in the list $B$. Other strategies could be selected to handle these last experiences. Figure~\ref{fig:mini_batches} illustrate the creation process of the mini-batch list \( B \), and Algorithm~\ref{algorithm:minibatch_creation} outlines the corresponding procedure.

	\begin{algorithm}[h]
		\caption{CreateMiniBatches ($D$, $M$)}
		{\footnotesize
			\begin{algorithmic}[1]
				\label{algorithm:minibatch_creation}
				\STATE \textbf{Input:} experiences buffer, \( D \)
				\STATE \textbf{Input:} mini-batch size, \( M \)
				\STATE
				\STATE Initialize list of mini-batches, $B \leftarrow []$
				\STATE Shuffle the experience buffer, $D_{\text{shuffled}} \leftarrow \text{Shuffle}(D)$
				\STATE \textbf{While} \( |D_{\text{shuffled}}| > 0 \)
				\begin{ALC@g}
					\STATE \textbf{If} \( |D_{\text{shuffled}}| < M \)
					\begin{ALC@g}
						\STATE Randomly sample $M-|D_{\text{shuffled}}|$ experiences from $B$, $Z$
						\STATE Append the sampled experiences $Z$ to $D_{\text{shuffled}}$, $D_{\text{shuffled}}\leftarrow D_{\text{shuffled}} \cup Z$
					\end{ALC@g}
					\STATE Compute average of unnormalized advantages stored in $D_{\text{shuffled}}[0:M-1]$, \(\mu_{\pi_\theta}\)
					\STATE Compute standard deviation of all unnormalized advantages stored in $D_{\text{shuffled}}[0:M-1]$, \(\sigma_{\pi_\theta}\)
					\STATE Initialize a mini-batch, $D_{\text{mini}} \leftarrow []$
					\STATE \textbf{For each experience \( (S_t^{(n)}, A_t^{(n)}, R_{t+1}^{(n)}, G_t^{(n)}, \hat{\Delta}_t^{(n)}) \) in} \( D_{\text{shuffled}}[0:M-1] \)
					\begin{ALC@g}
						\STATE Compute normalized advantage $\tilde{\Delta}_t \leftarrow (\hat{\Delta}_t - \mu_{\pi_\theta})/(\sigma_{\pi_\theta})$ (Equation~\ref{equ:normalised_advantage})
						\STATE Add experience to mini-batch, $D_{\text{mini}} \leftarrow D_{\text{mini}} \cup (S_t^{(n)}, A_t^{(n)}, R_{t+1}^{(n)}, G_t^{(n)}, \tilde{\Delta}_t^{(n)})$
					\end{ALC@g}
					\STATE Add mini-batch to list of batches, $B \leftarrow B \cup D_{\text{mini}}$
					\STATE Remove the mini-batch from the shuffled buffer, $D_{\text{shuffled}} \leftarrow D_{\text{shuffled}} \setminus D_{\text{shuffled}}[0:M-1]$
				\end{ALC@g}
				\STATE
				\STATE \textbf{Return} \( B \)
			\end{algorithmic}
		}
	\end{algorithm}

	\subsection{Loss Function}

	Recall that the REINFORCE algorithm with non-linear policies optimizes the policy by maximizing the following objective function, adapted here to account for samples collected at time step $t$ of a given rollout $n$:
	\begin{equation*}
		\mathcal{J}^{\pi}_{n, t}(\boldsymbol{\theta}) \doteq (G_{n,t} - \hat{v}(S_{n,t}, \mathbf{w})) \ln \pi(A_{n,t} | S_{n,t}, \boldsymbol{\theta}),
	\end{equation*}

	\noindent where \( G_{n,t} \) is the actual return received after time \( t \) in rollout $n$,      \( \hat{v}(S_{n,t}, \mathbf{w}) \) is the estimated value function (baseline), and
	\( \ln \pi(A_{n,t} | S_{n,t}, \boldsymbol{\theta}) \) is the log-probability of taking action \( A_{n,t} \) given state \( S_{n,t} \) under the current policy parameterized by \( \boldsymbol{\theta} \).

	As discussed, a significant drawback of the REINFORCE algorithm is the high variance in the estimation of the advantage due to the dependence on individual sample returns \( G_{n,t} \). This variance can make training unstable and slow, as updates can be erratic. We discussed that a way to mitigate this is to use the normalized estimated advantage function \( \tilde{\Delta}_{n,t} \) as a more stable estimate of the advantage of performing action $A_t$, resulting in the following objective function:
	\begin{equation*}
		\mathcal{J}^{+,\pi}_{n,t}(\boldsymbol{\theta}) \doteq \tilde{\Delta}_{n,t} \ln \pi(A_{n,t} | S_{n,t}, \boldsymbol{\theta}).
	\end{equation*}

	We already know that the learning process proceeds by taking steps in parameter space along the gradient direction, with step size controlled by a learning rate $\alpha$. This learning rate must be set carefully. A step along the gradient increases the probability of selected actions that are associated with high advantage, but the magnitude of this change is determined by the local shape of the policy function. Since the policy's shape changes as learning progresses, large steps can cause substantial shifts in the action distribution, potentially leading to catastrophic forgetting. Even minor variations in parameter space can significantly affect performance, so updates must be carefully controlled to maintain stability.

	The challenge in policy optimization, then, is to allow sufficiently large steps to improve the policy efficiently while avoiding unstable updates. This requires considering not just the gradient in parameter space but also how much the policy's action distribution changes in response to those updates. PPO addresses this challenge through mechanisms that explicitly constrain policy changes. The PPO objective function is designed to prevent excessively large updates that could destabilize training. Because these mechanisms modify the original objective, the resulting function does not optimize the true objective, $\mathcal{J}^{+,\pi}_{n,t}(\boldsymbol{\theta})$, directly and is therefore commonly referred to as a \textit{surrogate} objective function. The following sections describe the derivation of this function.

	\subsubsection*{Probability Ratio}

	As discussed, to stabilize and speed up learning, and to be more sample efficient, PPO updates the policy based on mini-batches obtained from multiple rollouts stored in the experiences buffer. Hence, PPO iterations alternate between collecting rollouts and updating the policy with mini-batches.

	Let $\pi_{\boldsymbol{\theta}_\text{old}}$ denote the policy used at the beginning of an iteration to collect trajectories and fill the experience buffer. This policy remains fixed throughout the entire iteration. Let $\pi_{\boldsymbol{\theta}}$ denote the policy that PPO updates using the data in the buffer, performing multiple gradient updates over mini-batches across several learning epochs. At the end of the iteration, the updated parameters are saved as $\pi_{\boldsymbol{\theta}_\text{old}} \leftarrow \pi_{\boldsymbol{\theta}}$. In the next iteration, $\pi_{\boldsymbol{\theta}_\text{old}}$ is used to collect new rollouts, and $\pi_{\boldsymbol{\theta}}$ is updated based on those. Hence, $\pi_{\boldsymbol{\theta}}$ is the actively updated policy, whose parameters change throughout the learning process, whereas $\pi_{\boldsymbol{\theta}_\text{old}}$ is the fixed reference policy for the iteration, reflecting how $\pi_{\boldsymbol{\theta}}$ looked at the start of the iteration.

	Since the trajectories in the experience buffer were sampled under the old policy $\pi_{\boldsymbol{\theta}_\text{old}}$, the actions contained in these trajectories are not necessarily the ones that the new policy $\pi_{\boldsymbol{\theta}}$ would choose. In other words, the stored samples no longer reflect the behaviour of the policy currently being optimized.
	To account for this discrepancy, PPO applies \emph{importance sampling}: each sample's advantage $\tilde{\Delta}_{n,t}$ is weighted by a probability ratio that measures how much more (or less) probable the sampled action is under the new policy compared with the old one \cite{schulman2017ppo}:
	\begin{equation*}
		r_{n,t}(\boldsymbol{\theta}) \doteq \frac{\pi(A_{n,t} \mid S_{n,t}, \boldsymbol{\theta})}{\pi(A_{n,t} \mid S_{n,t}, \boldsymbol{\theta}_\text{old})}.
	\end{equation*}

	Weighting the advantage in this way ensures that each sample contributes to the policy update according to how probable the current policy $\pi_{\boldsymbol{\theta}}$ considers the sampled action, rather than according to the probabilities of the old policy $\pi_{\boldsymbol{\theta}_\text{old}}$ that originally collected the data. Intuitively, if the current policy $\pi_{\boldsymbol{\theta}}$ assigns a higher probability to an action, then that sample should influence learning more strongly; if it assigns a lower probability, the sample should influence learning less.

	The denominator in the probability ratio is not always intuitive and deserves additional explanation. The importance sampling ratio $r_{n,t}(\boldsymbol{\theta})$ can be interpreted as a correction for how frequently each sampled action \emph{should} appear under the current policy $\pi_{\boldsymbol{\theta}}$. If an action is more probable under the current policy $\pi_{\boldsymbol{\theta}}$ than it was under the old one $\pi_{\boldsymbol{\theta}_\text{old}}$, then $r_{n,t}(\boldsymbol{\theta}) > 1$, and its contribution is upweighted to compensate for its underrepresentation in the collected data. It is as if the experience buffer were augmented with extra copies of that sample to match the frequency expected under $\pi_{\boldsymbol{\theta}}$. (This interpretation is metaphorical: importance sampling reweights samples mathematically rather than physically adding or removing experience.) This makes sense because if during the collection phase the agent had already been governed by $\pi_{\boldsymbol{\theta}}$, it would have selected the action that induced that sample more often.

	Conversely, if an action is less probable under the current policy $\pi_{\boldsymbol{\theta}}$ than under the old one $\pi_{\boldsymbol{\theta}_\text{old}}$, then $r_{n,t}(\boldsymbol{\theta}) < 1$, and its contribution is downweighted, because it is overrepresented in the data relative to the current policy $\pi_{\boldsymbol{\theta}}$. This makes sense because if during the collection phase the agent had been governed by $\pi_{\boldsymbol{\theta}}$, it would have selected that action less often.

	Thus, the denominator $\pi(A_{n,t} \mid S_{n,t}, \boldsymbol{\theta}_\text{old})$ acts as a normalizing term that converts empirical frequencies under $\pi_{\boldsymbol{\theta}_\text{old}}$ into the frequencies that would be observed if the samples had instead been generated by $\pi_{\boldsymbol{\theta}}$. In essence, the ratio ensures that the update reflects the action preferences of the current policy $\pi_{\boldsymbol{\theta}}$, even though the data were collected by the old one $\pi_{\boldsymbol{\theta}_\text{old}}$.

\begin{figure}[h]
		\centering
		\includegraphics[width=10cm]{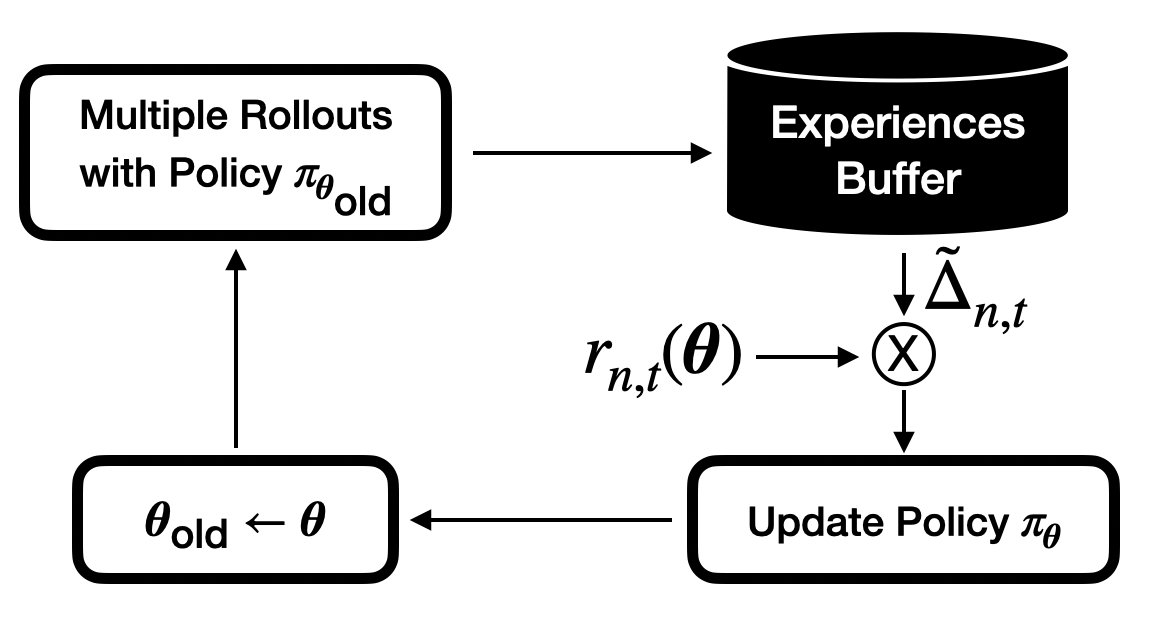}
		\caption{A PPO iteration: (1) multiple rollouts are collected using $\pi_{\boldsymbol{\theta}_\text{old}}$ and stored in the experience buffer; (2) the current policy $\pi_{\boldsymbol{\theta}}$ is updated using the buffered data, properly weighted by importance sampling; (3) the current policy parameters are saved for the next iteration, $\boldsymbol{\theta}_\text{old} \leftarrow \boldsymbol{\theta}$.}
		\label{fig:probability_ratio}
	\end{figure}

	Figure~\ref{fig:probability_ratio} illustrates the process of rollouts collection and subsequent policy update with samples weighted with importance sampling.

	\subsubsection*{Probability Ratio Clipping}

	The probability ratio $r_{n,t}(\boldsymbol{\theta})$ can become excessively large when the action probability under the old policy $\pi_{\boldsymbol{\theta}_\text{old}}$ is very small. In such cases, the ratio attempts an extreme correction: it strongly upweights samples that were rarely chosen by the old policy but appear highly preferred by the current one. Although this correction is theoretically valid from an importance sampling perspective, it can severely destabilize training, leading to sudden policy shifts and even catastrophic forgetting.

	To mitigate this effect, PPO applies a clipping operation to the ratio. The clipping function is defined as $\text{clip}(x, a, b) \doteq \min(\max(x, a), b)$, constraining its value to a bounded interval $[1 - \epsilon_{\pi}, 1 + \epsilon_{\pi}]$. The clipped ratio prevents extreme policy updates by limiting how strongly any single sample can influence learning \cite{schulman2017ppo}:
	\[
		\text{clip}\big(r_{n,t}(\boldsymbol{\theta}), 1 - \epsilon_{\pi}, 1 + \epsilon_{\pi}\big).
	\]

	In practice, this clipping limits how much the current policy $\pi_{\boldsymbol{\theta}}$ is allowed to change the probability of selecting action $A_{n,t}$ relative to the probability assigned by the old policy $\pi_{\boldsymbol{\theta}_\text{old}}$. By bounding extreme deviations according to the hyper-parameter $\epsilon_{\pi}$, PPO prevents overly large policy updates while still permitting meaningful improvement. This maintains learning stability and effectively balances exploration with controlled, safe updates.

	\subsubsection*{Clipping Objective}

	We can now present the PPO \textit{surrogate} objective function, which combines the probability ratio and the clipping function \cite{schulman2017ppo}:
	\begin{equation*}
		\mathcal{J}^{\text{CLIP},\pi}_{n,t}(\boldsymbol{\theta}; \epsilon_{\pi}) \doteq \min \left( r_{n,t}(\boldsymbol{\theta}) \tilde{\Delta}_{n,t}, \text{clip} \left( r_{n,t}(\boldsymbol{\theta}), 1 - \epsilon_{\pi}, 1 + \epsilon_{\pi} \right)\tilde{\Delta}_{n,t} \right).
	\end{equation*}

	By taking the minimum of the clipped and unclipped terms, the objective effectively forms a lower (pessimistic) bound on the unclipped objective. This ensures that updates which would lead to an excessively large improvement due to extreme changes in the probability ratio are ignored, while updates that would decrease the objective are still applied. In this way, the clipping procedure reduces the magnitude of large gradients, lowering variance in gradient estimation. These safeguards allow PPO to take relatively larger steps in parameter space, accelerating learning while maintaining stability in the action probabilities.

	\subsubsection*{KL Penalty}

	The PPO objective function can be formalized differently. Instead of employing a clipping procedure in the probability ratio, the alternative formulation imposes a penalty on those policy parameterizations $\boldsymbol{\theta}$ that explicitly induce an action probability distribution too dissimilar from the one induced by $\boldsymbol{\theta}_\text{old}$. The dissimilarity between the distributions is computed using the Kullback-Leibler (KL) divergence (see Section~\ref{sec:kl-divergence}). Intuitively, the KL divergence measures how much the current policy $\pi_{\boldsymbol{\theta}}$ diverges from the old one $\pi_{\boldsymbol{\theta}_\text{old}}$. A small KL divergence indicates similar policies, while a large value signals significant changes in action probabilities.

	Formally, given a discrete action space $\mathcal{A}$, the KL divergence between the policies induced by the current parameters $\boldsymbol{\theta}$ and the old parameters $\boldsymbol{\theta}_{\text{old}}$ quantifies the expected difference in the information content (or surprise) between the two policies. It is computed as the expected logarithmic ratio of the probabilities assigned to actions by the current policy $\pi_{\boldsymbol{\theta}}$ and the old policy $\pi_{\boldsymbol{\theta}_{\text{old}}}$ (by applying Equation~\ref{equ:KL-div-discrete}) \cite{schulman2017ppo}:
	\begin{equation}\label{equ:kl}
		D_{n,t}^{\text{KL},\pi}({\boldsymbol{\theta}}, {\boldsymbol{\theta}_{\text{old}}}) \doteq \mathbb{E}_{a\sim\pi_{\boldsymbol{\theta}}} \left[\ln \frac{\pi(a | S_{n,t}, {\boldsymbol{\theta}})}{\pi(a | S_{n,t}, \boldsymbol{\theta}_{\text{old}})}\right]= \sum_{a\in\mathcal{A}} \pi(a | S_{n,t}, {\boldsymbol{\theta}}) \ln \frac{\pi(a | S_{n,t}, {\boldsymbol{\theta}})}{\pi(a | S_{n,t}, \boldsymbol{\theta}_{\text{old}})}.
	\end{equation}

	Using this divergence, the PPO algorithm with a KL penalty defines the following surrogate objective, where the user-defined hyperparameter $\beta>0$ controls the strength of the penalty \cite{schulman2017ppo}:
	\begin{equation*}
		\mathcal{J}^{\text{KL},\pi}_{n,t}(\boldsymbol{\theta}; \beta) \doteq r_{n,t}(\boldsymbol{\theta}) \tilde{\Delta}_{n,t}-\beta D_{n,t}^{\text{KL},\pi}({\boldsymbol{\theta}}, {\boldsymbol{\theta}_{\text{old}}}).
	\end{equation*}

	\subsubsection*{Entropy Regularization}

	To further improve the stability of the training and promote exploration, an \textit{entropy} term can be added to the PPO objective. In general, entropy of a random variable quantifies the average level of uncertainty associated with the variable's possible outcomes (see Section~\ref{sec:info-entropy}).  The entropy of the policy $\pi$ parameterized by $\boldsymbol{\theta}$ and conditioned on state $S_{n,t}$, is defined as (by applying Equation~\ref{equ:entropy_discrete}) \cite{schulman2017ppo}:
	\begin{equation*}
		\mathcal{H}_{n,t}^{\pi}(\boldsymbol{\theta}) \doteq -\sum_{a\in\mathcal{A}} \pi(a | S_{n,t}, \boldsymbol{\theta}) \ln \pi(a | S_{n,t}, \boldsymbol{\theta}).
	\end{equation*}

	The entropy of the policy quantifies the randomness or uncertainty in the action selection process. It measures how spread out the probability distribution over actions is. A more uniform distribution corresponds to higher entropy, reflecting greater randomness and uncertainty about which action will be selected. Conversely, low entropy indicates that probability is concentrated around one or a few actions, resulting in a more deterministic behavior.

	To encourage exploration, the entropy term is added to the PPO objective function as a regularizer. This term penalizes low-entropy policies, as these are less likely to explore alternative actions. By maintaining higher entropy, the policy remains more exploratory, avoiding overconfidence in specific actions and reducing the risk of converging to suboptimal solutions. The entropy term promotes a balance between exploitation (selecting actions with high expected return) and exploration (trying new actions to discover better strategies).

	\subsubsection*{Aggregate Objective}

	PPO's paper \cite{schulman2017ppo} introduces two separate surrogate objectives: the clip-based and the KL-based, both of which can be combined with entropy regularization. Here, the clip-based term $\mathcal{J}^{\text{CLIP},\pi}_{n,t}(\boldsymbol{\theta}; \epsilon)$, the KL-based term $\mathcal{J}^{\text{KL},\pi}_{n,t}(\boldsymbol{\theta}; \beta)$, and the entropy regularization term $\mathcal{H}_{n,t}^{\pi}(\boldsymbol{\theta})$, are combined in a single objective:
	\begin{equation}\label{equ:objective_ppo}
		\mathcal{J}^{\text{PPO},\pi}_{n,t}(\boldsymbol{\theta}; \nu, \epsilon_{\pi}, \beta, \eta) \doteq
		\nu\mathcal{J}^{\text{CLIP},\pi}_{n,t}(\boldsymbol{\theta}; \epsilon_{\pi}) + (1-\nu)\mathcal{J}^{\text{KL},\pi}_{n,t}(\boldsymbol{\theta}; \beta) + \eta \mathcal{H}_{n,t}^{\pi}(\boldsymbol{\theta}),
	\end{equation}

	\noindent where $\nu \in\{0,1\}$ is a binary hyperparameter used to select between the clip-based and the KL-based sub-objectives and $\eta$ determines the weight of the entropy term in the objective function, controlling the level of exploration.

	\subsubsection*{From Objective to Loss}

	Although PPO is naturally formulated to perform gradient ascent on the objective function
$\mathcal{J}^{\text{PPO},\pi}_{n,t}(\boldsymbol{\theta}; \nu, \epsilon_{\pi}, \beta, \eta)$, it is common in practice to define a corresponding loss function and perform gradient descent. This convention aligns better with standard software packages for optimization. Formally, the loss function is simply the negation of the objective:
	\[
		\mathcal{L}^{\text{PPO},\pi}_{n,t}(\boldsymbol{\theta}; \nu, \epsilon_{\pi}, \beta, \eta) \doteq -\mathcal{J}^{\text{PPO},\pi}_{n,t}(\boldsymbol{\theta}; \nu, \epsilon_{\pi}, \beta, \eta).
	\]

	\subsubsection*{Value Function Objective}

	The PPO loss function $\mathcal{L}^{\text{PPO},\pi}_{n,t}(\boldsymbol{\theta}; \nu, \epsilon_{\pi}, \beta, \eta)$ depends on the normalized estimated advantage function $\tilde{\Delta}_{n,t}$, which in turn depends on the estimated state-value function $\hat{v}(S_{n,t}, \mathbf{w})$. Consequently, it is necessary to update the value function parameters $\mathbf{w}$ alongside the policy parameters.

	Similar to REINFORCE, where the parameters $\mathbf{w}$ of $\hat{v}(S_{n,t}, \mathbf{w})$ are updated by minimizing the squared error
$\left(G_t - \hat{v}(S_t, \mathbf{w}) \right)^2$ (Equation~\ref{equ:loss-value}),
	PPO also learns the value function using a squared-error loss \cite{schulman2017ppo}:
	\begin{equation}\label{equ:objective_value:ppo}
		\mathcal{L}^{\text{PPO},\hat{v}}_{n,t}(\mathbf{w}) \doteq
		\left(G_{n,t} - \hat{v}(S_{n,t}, \mathbf{w})\right)^2.
	\end{equation}
	This loss ensures that the value function provides accurate estimates of expected returns, which are then used to compute the advantage function for policy updates.

	\subsection{Complete PPO Algorithm}

	Having introduced each building block, i.e., trajectory collection, advantage estimation, importance sampling, and the clipped objective, we are now prepared to bring them together into the complete PPO algorithm.

	PPO begins with the initialization of the policy and value function networks' parameters, $\boldsymbol{\theta}$ and $\boldsymbol{w}$. The technique known as \textit{orthogonal initialization} is often used for this purpose, ensuring that the initial weights are uncorrelated and have unit variance. Compared to standard random initialization, orthogonal initialization can improve the stability and efficiency of training by providing a more balanced starting point for the gradient updates, especially in deep networks with many layers.

	The training process is repeated over $I$ (a user-defined constant) iterations, which can be early stopped as in REINFORCE. In each iteration, Algorithm~\ref{algorithm:experiences_buffer} is executed to collect an experiences buffer $D$ filled with (state, action, reward, discounted return, unnormalized advantage estimate) tuples observed over $N$ policy rollouts with horizon $T$. After filling $D$, PPO stores the current policy parameters in $\boldsymbol{\theta}_{\text{old}}$ for later use, $\boldsymbol{\theta}_{\text{old}} \leftarrow \boldsymbol{\theta}$. Thus, $\boldsymbol{\theta}_{\text{old}}$ refers to the policy parameters at the onset of the iteration. The policy $\pi_{\boldsymbol{\theta}_{\text{old}}}$ operates as a reference against which $\pi_{\boldsymbol{\theta}}$ is compared as learning proceeds during the current iteration.

	PPO then performs $E$ (a user-defined constant) learning epochs to update $\pi_{\boldsymbol{\theta}}$ before moving to the next iteration. Each learning epoch begins by shuffling $D$ and splitting it into $M$-length mini-batches, organized into a list $B$. For each mini-batch $D_{\text{mini}} \in B$, the following steps are performed: (1) The gradient of the average policy loss (Equation~\ref{equ:objective_ppo}) over the mini-batch is computed; (2) The gradient of the average state-value loss over the mini-batch is computed; (3) The gradients computed in steps 1 and 2 are used to update the parameters of the policy ($\boldsymbol{\theta}$) and state-value ($\boldsymbol{w}$) networks, respectively.

	With multiple gradient steps, PPO converges faster, but this also increases the risk of the policy drifting excessively, potentially causing instability. To assess whether the policy has changed too much during this process, the action probability distribution induced by $\pi$ under $\boldsymbol{\theta}$ can be compared to the one induced by $\boldsymbol{\theta}_{\text{old}}$. Concretely, if the average KL divergence between the action distributions induced by both $\boldsymbol{\theta}$ and $\boldsymbol{\theta}_{\text{old}}$ exceeds a user-defined threshold $\xi$, the current iteration is stopped early, prompting the agent to perform new rollouts in the environment to gather fresh data.

	Algorithm~\ref{algorithm:PPO} outlines the pseudo-code of the original PPO \cite{schulman2017ppo} extended with early stopping based on KL divergence. PPO can be extended in many directions to address the challenges of specific problems effectively. Some possible extensions include:

	\begin{itemize}
		\item Implementing a learning rate scheduler (e.g., exponential decay), as it was done in REINFORCE, can help stabilize training as learning progresses, ensuring consistent improvements without overshooting or stagnation.
		\item Similar to policy updates, applying clipping to value function updates can prevent overfitting or divergence, particularly in environments with high reward variance.
		\item Using a decaying entropy coefficient helps reduce exploration over time, similarly to a learning rate scheduler, allowing the agent to focus on exploiting learned policies as it gains confidence.
		\item Normalizing input states, as it was done in REINFORCE, especially in high-dimensional environments, can lead to faster convergence and more stable training.
		\item Distributing the processing with multiple workers to scale for solving complex problems with extensive state and action spaces.
	\end{itemize}

	Note that PPO is a complex algorithm that can be adapted in various ways to address the diversity of reinforcement learning problems. Stable Baselines 3 (SB3) \cite{sb32024} provides reliable PyTorch implementations of many reinforcement learning algorithms, including PPO. Examining its PPO source code is a practical way to complement this reading.

	\begin{algorithm}
		\caption{PPO ($\alpha_{\pi}$, $\alpha_{\mathbf{w}}$, $I$, $E$, $N$, $T$, $M$, \( \pi \), $\hat{v}$, $\xi$, $\nu$, $\epsilon_{\pi}$, $\beta$, $\eta$) (this is PPO \cite{schulman2017ppo} extended with early stopping based on KL divergence)}
		{\small
		\begin{algorithmic}[1]
			\label{algorithm:PPO}
			\STATE \textbf{Input:} policy learning rate $\alpha_{\pi} > 0$ and state-value learning rate $\alpha_{\mathbf{w}} > 0$
			\STATE \textbf{Input:} number of learning epoches $E$
			\STATE \textbf{Input:} number of iterations $I$
			\STATE \textbf{Input:} number of rollouts $N$, number of time steps per rollout $T$, and mini-batch size $M$
			\STATE \textbf{Input:} policy function, \( \pi \), and state-value function $\hat{v}$
			\STATE \textbf{Input:} KL divergence threshold $\xi > 0$
			\STATE \textbf{Input:} Selector $\nu$ between $\mathcal{L}^{\text{CLIP}}$ and $\mathcal{L}^{\text{KL}}$
			\STATE \textbf{Input:} Policy ratio clip range $\epsilon_{\pi}$
			\STATE \textbf{Input:} KL penalty weight $\beta$
			\STATE \textbf{Input:} Entropy regularization weight $\eta$
			\STATE
			\STATE \textcolor{blue}{Initialize network parameters with orthogonal initialization}
			\STATE $\boldsymbol{\theta} \leftarrow \text{OrthogonalInitialization()}$, $\mathbf{w} \leftarrow \text{OrthogonalInitialization()}$
			\STATE \textcolor{blue}{Train the networks for $I$ iterations}
			\STATE \textbf{For each iteration} $i = 1, 2, \ldots, I$:
			\begin{ALC@g}
				\STATE \textcolor{blue}{Store the current policy parameters.}
				\STATE $\boldsymbol{\theta}_{\text{old}} \leftarrow \boldsymbol{\theta}$
				\STATE \textcolor{blue}{Create a buffer of experiences using the current policy and its parameters.}
				\STATE $D \leftarrow \text{CreateExperiencesBuffer}(\pi, \boldsymbol{\theta}_\text{old}, \hat{v}, \mathbf{w}, N, T, \gamma, \lambda)$ [use Algorithm~\ref{algorithm:experiences_buffer}]
				\STATE \textcolor{blue}{Reset control flag for learning epoch early stop}
				\STATE $stop \leftarrow$ false

				\STATE \textcolor{blue}{Iterate over each learning epoch}
				\STATE \textbf{For each learning epoch} $e = 1, 2, \ldots, E$:
				\begin{ALC@g}
					\STATE \textcolor{blue}{Generate mini-batches from the experiences buffer.}
					\STATE $B \leftarrow \text{CreateMiniBatches}(D, M))$ [use Algorithm~\ref{algorithm:minibatch_creation}]
					\STATE \textcolor{blue}{Iterate over each mini-batch}
					\STATE \textbf{For each mini-batch} $D_{\text{mini}}\in B$:
					\begin{ALC@g}
						\STATE \textcolor{blue}{Calculate the gradient for the policy update using the PPO loss.}
						\STATE $g_\pi \leftarrow \nabla_\theta\left(\frac{1}{M}\sum_{\forall (S_{n,t}, A_{n,t}, R_{n,t+1}, G_{n,t}, \tilde{\Delta}_{n,t})\in D_{\text{mini}}}  \mathcal{L}^{\text{PPO},\pi}_{n,t}(\boldsymbol{\theta}; \nu, \epsilon_{\pi}, \beta, \eta)\right)$
						\STATE \textcolor{blue}{Calculate the gradient for the value function update using MSE loss.}
						\STATE $g_{\boldsymbol{w}} \leftarrow \nabla_\mathbf{w}\left(\frac{1}{M}\sum_{\forall (S_{n,t}, A_{n,t}, R_{n,t+1}, G_{n,t}, \tilde{\Delta}_t)\in D_{\text{mini}}} \mathcal{L}^{\text{PPO},\hat{v}}_{n,t}(\mathbf{w})\right)$
						\STATE \textcolor{blue}{Update the policy parameters using the policy gradient.}
						\STATE $\boldsymbol{\theta} \leftarrow \boldsymbol{\theta} - \alpha_{\pi} \cdot g_{\pi}$
						\STATE \textcolor{blue}{Update the value function parameters using the gradient of value function.}
						\STATE $\mathbf{w} \leftarrow \mathbf{w} - \alpha_{\mathbf{w}} \cdot g_{\mathbf{w}}$
						\STATE \textcolor{blue}{If average KL divergence is too excessive early stop learning epoch.}
						\STATE \textbf{If} $\frac{1}{M}\sum_{\forall (S_{n,t}, A_{n,t}, R_{n,t+1}, G_{n,t}, \tilde{\Delta}_{n,t})\in D_{\text{mini}}}D_{n,t}^{\text{KL},\pi}({\boldsymbol{\theta}}, {\boldsymbol{\theta}_{\text{old}}}) > \xi$
						\begin{ALC@g}
							\STATE $stop \leftarrow$ true
							\STATE break [leave for each mini-batch loop]
						\end{ALC@g}
					\end{ALC@g}
					\STATE \textbf{If} $stop$:
					\begin{ALC@g}
						\STATE break [leave for each learning epoch loop]
					\end{ALC@g}
				\end{ALC@g}
			\end{ALC@g}
			\STATE
			\STATE \textcolor{blue}{\textbf{Return} the final policy parameters.}
			\STATE \textbf{Return} $\boldsymbol{\theta}$
		\end{algorithmic}
		}
	\end{algorithm}

	\subsection{PPO in Continuous Action Spaces}

	As discussed in Section~\ref{sec:continuous_actions}, a common approach to policy parameterization in continuous action spaces is to learn the parameters of a probability distribution over actions, typically a Gaussian distribution. This approach can be extended to multidimensional action spaces by employing separate MLPs for each action dimension. In each dimension, one typically uses an MLP to output the mean $\mu_i(S_{n,t}, \boldsymbol{\theta})$ and another to output the standard deviation $\sigma_i(S_{n,t}, \boldsymbol{\theta})$. At each time step, these networks produce the parameters that define the Gaussian distribution estimated to best govern the agent's behavior in the current state. The goal is thus to learn the parameters of these MLPs so that their outputs induce Gaussian distributions from which the agent can sample optimal actions. The same strategy applies when using PPO. In the following, we discuss additional adjustments that enable PPO to handle continuous action spaces effectively.

	\subsubsection*{KL divergence}

	If the action space is multidimensional with \( d \) dimensions, and assuming the Gaussian distributions are factorized across dimensions (i.e., the dimensions are independent), the KL divergence between the old policy \( \boldsymbol{\theta}_{\text{old}} \) and the current policy \(\boldsymbol{\theta} \)  for the \(i\)-th dimension is given by Equation~\ref{equ:kl-continuous}:
	\[
		D^{\text{KL},\pi,i}_{n,t}( \boldsymbol{\theta},  \boldsymbol{\theta}_{\text{old}}) =
		\ln\left(\frac{\sigma_i(S_{n,t}, \boldsymbol{\theta}_{\text{old}})}{\sigma_i(S_{n,t}, \boldsymbol{\theta})}\right)
		+ \frac{\sigma_i(S_{n,t}, \boldsymbol{\theta})^2 + (\mu_i(S_{n,t}, \boldsymbol{\theta}) - \mu_i(S_{n,t}, \boldsymbol{\theta}_{\text{old}}))^2}{2\sigma_i(S_{n,t}, \boldsymbol{\theta}_{\text{old}})^2}
		- \frac{1}{2},
	\]

	\noindent where $\mu_i(S_{n,t}, \boldsymbol{\theta})$ and $\sigma_i(S_{n,t}, \boldsymbol{\theta})$ are the mean and standard deviation output by the current policy for the $i$-th action dimension and $\mu_i(S_{n,t}, \boldsymbol{\theta}_{\text{old}})$ and $\sigma_i(S_{n,t}, \boldsymbol{\theta}_{\text{old}})$ are the mean and standard deviation output by the old policy for the $i$-th action dimension.

	The total KL divergence over the $d$ dimensions, assuming independence between dimensions, is the sum of the KL divergences for each dimension:
	\[
		D^{\text{KL},\pi}_{n,t}( \boldsymbol{\theta},  \boldsymbol{\theta}_{\text{old}}) = \sum_{i=1}^d D^{\text{KL},\pi,i}_{n,t}( \boldsymbol{\theta},  \boldsymbol{\theta}_{\text{old}}).
	\]

	\subsubsection*{Information Entropy}

	If the action space is multidimensional with \( d \) dimensions, and assuming the Gaussian distributions are factorized across dimensions (i.e., the dimensions are independent), the entropy of the current policy \(\boldsymbol{\theta} \)  for the \(i\)-th dimension is given by Equation~\ref{equ:entropy_gaussian}:
	\[
		\mathcal{H}_{n,t}^{\pi, i}(\boldsymbol{\theta}) = \frac{1}{2} + \frac{1}{2} \ln(2\pi) + \ln \sigma_i(S_{n,t}, \boldsymbol{\theta}),
	\]

	\noindent where $\sigma_i(S_{n,t}, \boldsymbol{\theta})$ is the standard deviation output by the current policy for the $i$-th action dimension.

	The total information entropy over the $d$ dimensions, assuming independence between dimensions, is the sum of the information entropy for each dimension:
	\[
		\mathcal{H}_{n,t}^{\pi}(\boldsymbol{\theta}) = \sum_{i=1}^d \mathcal{H}_{n,t}^{\pi, i}(\boldsymbol{\theta}).
	\]

	\subsection{PPO in Action: Examples and Results}

\begin{figure}[h]
		\begin{center}
			\includegraphics[width=14cm]{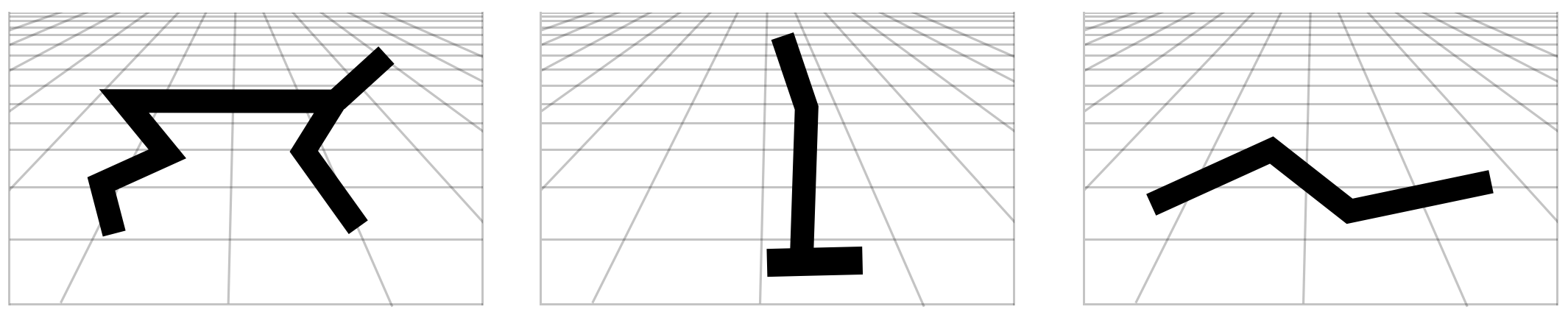}
			\caption{Schematics of three environments in MuJoCo: HalfCheetah (left), Hopper (middle), and Swimmer (right).}
			\label{fig:mujoco-envs}
		\end{center}
	\end{figure}

	The authors of PPO \cite{schulman2017ppo} compared it against REINFORCE \cite{williams1992reinforce}, also known as Vanilla Policy Gradient (VPG), across various tasks, including continuous control problems in the physics-based simulator MuJoCo\footnote{\url{https://mujoco.org}}. Figure~\ref{fig:mujoco-envs} shows snapshots of four representative MuJoCo environments: HalfCheetah, Hopper, Reacher, and Swimmer.

	The environments showcase diverse robot morphologies and objectives:
	\begin{itemize}
		\item \textbf{HalfCheetah}\footnote{\url{https://www.gymlibrary.dev/environments/mujoco/half_cheetah/}}: A 2D bipedal robot whose goal is to maximize its forward velocity.
		\item \textbf{Hopper}\footnote{\url{https://www.gymlibrary.dev/environments/mujoco/hopper/}}: A 2D monopod robot designed to hop forward as quickly as possible.
		\item \textbf{Swimmer}\footnote{\url{https://www.gymlibrary.dev/environments/mujoco/swimmer/}}: A 2D snake-like robot tasked with propelling itself forward in a simulated fluid environment.
	\end{itemize}

\begin{figure}[h]
		\begin{center}
			\includegraphics[width=13cm]{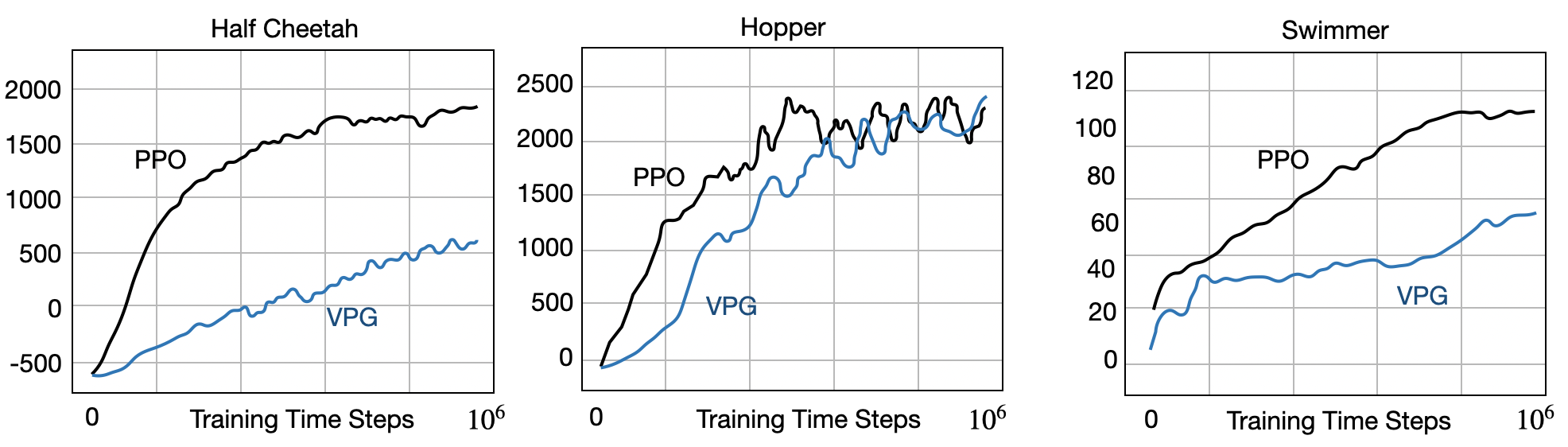}
			\caption{Smoothed average learning curves of PPO (in black) and VPG (in blue) over multiple runs across three MuJoCo environments (adapted from \cite{schulman2017ppo}). The vertical axis shows performance as reported in \cite{schulman2017ppo}.}
			\label{fig:mujoco-envs-results}
		\end{center}
	\end{figure}

	Figure~\ref{fig:mujoco-envs-results} illustrates the learning performance of PPO and VPG across these environments, as reported in \cite{schulman2017ppo}. The learning curves clearly demonstrate that PPO consistently outperforms VPG, highlighting its robustness and efficiency in optimizing policies for complex tasks. For clarity, additional comparisons to other methods have been omitted.

	\chapter{Deep Imitation Learning}\label{cha:dil}

	Imitation learning has been widely adopted in scenarios where expert demonstrations are available, designing task-specific rewards is challenging, or training through trial-and-error would be unsafe.

	In the context of virtual characters, imitation learning enables the synthesis of lifelike and reactive behaviors by learning directly from motion capture data or human demonstrations in interactive environments. This is particularly useful for animating agents in games, simulations, or training systems. In robotics, imitation learning is used to teach agents complex sensorimotor tasks, such as grasping, locomotion, and navigation. Notably, some of the earliest and most influential applications include autonomous driving from camera inputs, as demonstrated by Pomerleau's ALVINN system \cite{Pomerleau1991} and NVIDIA's end-to-end approach \cite{Bojarski2016}. These works highlight the practicality of imitation learning in real-world, safety-critical settings.

	In imitation learning, the learner is typically given a dataset consisting of expert demonstrations that show how to perform a specific task. The objective is to infer a policy that closely replicates the expert's behavior across the environment states encountered in the demonstrations, according to a predefined similarity measure, such as matching the expert's action distribution. Ideally, the learned policy not only reproduces behavior in states covered by the demonstrations but also generalizes effectively to states that were not explicitly demonstrated.

	As a concrete example, consider a learning task in which an agent must navigate a 3D environment to reach a target location while avoiding obstacles. A human operator demonstrates the task by controlling the agent through a keyboard, selecting appropriate actions, such as moving forward, turning left, or stopping, based on their perception of the environment. At each time step, the current state of the environment (e.g., the agent's position, orientation, and the relative positions of nearby obstacles) along with the action chosen by the human is recorded. These state-action pairs are collected into a dataset of expert demonstrations that will be used to train the imitation policy.

	\section{Background Concepts}\label{sec:back_concepts}

	This section offers a concise recap of the key concepts from Chapters \ref{cha:MDP} and \ref{cha:drl} that are essential for understanding Deep Imitation Learning algorithms. Readers who have already gone through these chapters may skip this section.

	\subsection{Markov Decision Processes (MDPs)}

	Markov Decision Processes (MDPs) provide a foundational framework for modeling sequential decision-making under uncertainty, of particular importance for reinforcement and imitation learning. The formalism presented in this section closely follows \cite{sutton2018reinforcement}.

	In the MDP framework, the decision-making entity is called the \emph{agent}, and everything external to it, with which it interacts, is referred to as the \emph{environment}. The environment's state space is denoted by \( \mathcal{S} \), and the agent's action space by \( \mathcal{A} \).

	Actions may range from low-level motor commands, such as the torques applied to actuators, to high-level task decisions or even internal cognitive operations, such as controlling attention. Similarly, states may include raw sensor readings (e.g., pixel values), symbolic world representations, the agent's proprioceptive information (e.g., joint angles or position), or even compressed histories when partial observability is present.

\begin{figure}[h]
		\begin{center}
			\includegraphics[width=8cm]{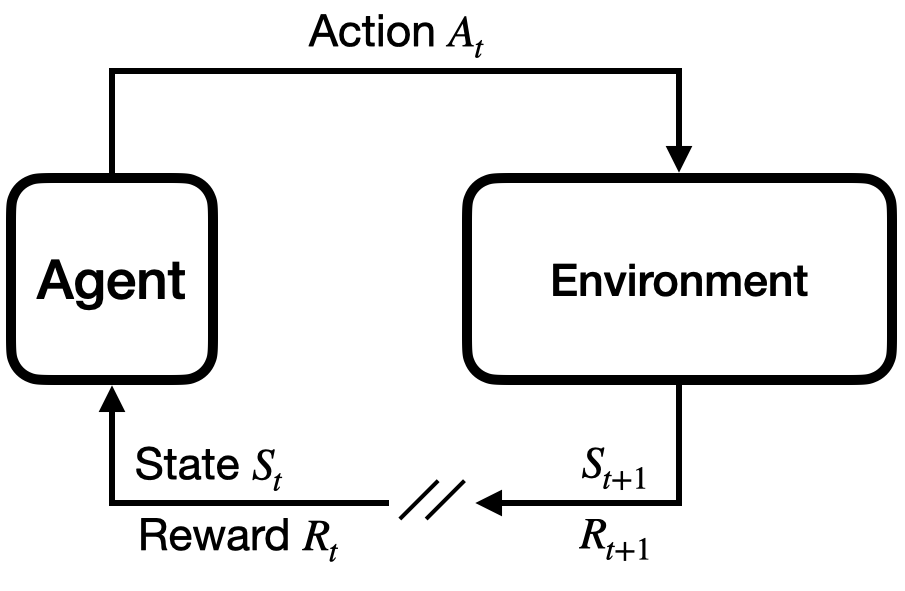}
			\caption{The agent-environment interaction cycle in an MDP (adapted from \cite{sutton2018reinforcement}).}
			\label{fig:agent-env-b}
		\end{center}
	\end{figure}

	The agent and the environment interact in a loop (Figure~\ref{fig:agent-env-b}). At each discrete time step \( t = 0, 1, 2, 3, \ldots \), the agent observes the current state \( S_t \in \mathcal{S} \) and selects an action \( A_t \in \mathcal{A} \) according to its policy (more details in the next section). The environment then returns a reward \( R_{t+1} \in \mathcal{R} \) and transitions to a new state \( S_{t+1} \in \mathcal{S} \), which depends stochastically on both the previous state and the agent's action.

	In episodic settings, the agent-environment interaction unfolds over discrete time steps \( t = 0, 1, \dots, T \), forming an \emph{episode}. The horizon \( T \) denotes the number of interaction cycles that compose the episode, which can be either fixed or variable depending on the task. At the end of each episode, the environment typically resets.

	In the discrete case (the continuous case is omitted for simplicity), the MDP formalism treats \( S_t \), \( A_t \), and \( R_t \) as discrete random variables drawn from the sets \( \mathcal{S} \), \( \mathcal{A} \), and \( \mathcal{R} \subset \mathbb{R} \), respectively. This stochastic formulation captures the uncertainty inherent in both the agent's behavior and the environment's response.

	\subsection{Policy Definition and Rollout}

	The action selection process at each time step is controlled by the agent's \emph{policy}, which defines the behavior of the agent by mapping states to probabilities over actions. Formally, a stochastic policy is a function \( \pi : \mathcal{A} \times \mathcal{S} \to [0, 1] \), where \( \pi(a \mid s) \) denotes the probability of selecting action \( a \in \mathcal{A} \) when in state \( s \in \mathcal{S} \) \cite{sutton2018reinforcement}:
	\begin{equation*}
		\pi(a \mid s) \doteq P(A_t = a \mid S_t = s), \quad \text{with} \quad \sum_{a\in\mathcal{A}} \pi(a \mid s) = 1 \quad \text{for all } s\in \mathcal{S}.
	\end{equation*}

	In practice, instead of specifying \( \pi(a \mid s) \) explicitly for each state--action pair, it is usually represented as a \emph{parameterized policy}, denoted \( \pi(a \mid s, \boldsymbol{\theta}) \), where \( \boldsymbol{\theta} \in \mathbb{R}^d \) is a vector of \( d \) learnable parameters,  which need to be learned for the agent to behave optimally. When trained appropriately (see next section), the parameterized policy approximates the true action-selection distribution \cite{sutton2018reinforcement}:
	\begin{equation*}
		\pi(a \mid s, \boldsymbol{\theta}) \approx P(A_t = a \mid S_t = s), \quad \text{with} \quad \sum_{a \in \mathcal{A}} \pi(a \mid s, \boldsymbol{\theta}) = 1.
	\end{equation*}

	Under a stochastic policy, the agent's action at time \( t \), denoted \( A_t \), is sampled from the distribution induced by the policy given the current state \( S_t \). That is, actions with higher probability under \( \pi \) are more likely to be selected \cite{sutton2018reinforcement}:
	\begin{equation*}
		A_t \sim \pi(\cdot \mid S_t, \boldsymbol{\theta}).
	\end{equation*}

	The sequence of state-action transitions experienced by the agent during an episode is called \textit{rollout} or \textit{trajectory}. Starting from an initial state \( s_0 \), the agent samples an action \(a_0\) from its policy, applies it to the environment, and receives the next state \( s_1 \). This process continues for \( T \) steps, producing a trajectory of the form:
	\[
		\tau = (s_0, a_0, s_1, a_1, \dots, s_T, a_T).
	\]

	\subsection{Policy Modeling}

	Deep artificial neural networks are commonly used to model policy functions, as they can capture non-linear relationships between states and action probabilities while remaining differentiable, which is an essential property for efficient training. Although advanced architectures such as convolutional neural networks are frequently employed, the following discussion focus on the simpler and more general case of the Multi Layer Perceptron (MLP), noting that extensions to other architectures are straightforward. An MLP consists of multiple layers of simple processing units known as artificial neurons.

	Each neuron generates as output a scalar resulting from a simple non-linear operation on the neuron's inputs. Formally, a neuron $i$ applies a non-linear transformation $\phi: \mathbb{R} \rightarrow \mathbb{R}$, also known as an \textit{activation function}, to the scalar bias $b_i \in \mathbb{R}$ plus the weighted sum of its inputs $\{x_j\} \in \mathbb{R}$:
	\begin{align*}
		y_i = \phi\left(b_i + \sum_{j}{x_j w_{ji}}\right),
	\end{align*}

	\noindent where $w_{ji}\in\mathbb{R}$ is the weight indicating how much the neuron's input $x_j$ influences its output.

	In an MLP, neurons are organized in layers. Let us denote the number of neurons in a given layer $l$ by $n_l$, where the first layer is $l=1$ and the last layer is $l=L$. The first and last layers are commonly known as \textit{input} and \textit{output} layers, respectively. All other intermediate layers are commonly known as \textit{hidden layers}. The neurons in the first layer, $l=1$, directly output the state vector $s$, that is, each neuron $i$ outputs the $i$-th element of $s$, denoted by $s_i$, where $s=(s_i)_i$. All neurons in subsequent layers, $l>1$, receive as input the outputs of all neurons in the preceding layer, $l-1$. In addition, all neurons in layer $l>1$ typically employ the same activation function $\phi^{(l)}$. The neuron equation can be redefined to explicitly represent the dependence on the layer $l$ with the superscript $(l)$, as well as on the state $s$ and the vector that encompasses all weights and biases of the MLP, $\boldsymbol{\theta}$:
	\begin{equation}\label{equ:neurons}
		y^{(l)}_i(s, \boldsymbol{\theta}) =
		\begin{cases}
			s_i                                                                                             & \text{if } l=1, \\
			\phi^{(l)}\left(b^{(l)}_i + \sum_{j}{w^{(l-1)}_{ji}} y^{(l-1)}_j(s, \boldsymbol{\theta})\right) & \text{if } l>1.
		\end{cases}
	\end{equation}

	The inclusion of the non-linear activation function is vital for the MLP's ability to model non-linear relationships in the state space. Non-linearities allow the network to capture complex dynamics and decision boundaries that are critical for effective learning and generalization. A commonly used activation function is the hyperbolic tangent activation function, $\phi^{(l)}_{\text{tanh}}(x)\doteq\text{tanh}(x)$, which maps input values to the bounded range \([-1, 1]\) (useful in bounded action spaces). Another common alternative is the sigmoid activation function, \(\sigma^{(l)}(x) = \frac{1}{1 + e^{-x}}\), which maps input values to the range \([0, 1]\), introducing non-linearity while producing outputs that can be interpreted as probabilities. Yet another common alternative is the ReLU activation function, which ensures a large derivative when positive (facilitates training):
	\begin{equation*}
		\phi^{(l)}_{\text{ReLU}}(x)=
		\begin{cases}
			\max(0, x) & \text{if } l<L, \\
			x          & \text{if } l=L.
		\end{cases}
	\end{equation*}

	\subsection{Policies for Discrete Action Spaces}
	\label{sec:discrete-actions}

	In many real-world applications, agents must choose from a finite set of distinct actions, such as moving in one of four directions or selecting from predefined commands. Handling discrete action spaces requires modeling policies that output probability distributions over a fixed set of actions.

	An MLP can be used for modeling a stochastic policy by considering the outputs of all neurons in the last layer $L$ as a vector that defines the preferences over the $n_L$ possible discrete actions, one neuron per action, given a state $s$ and the vector that encompasses all weights and biases of the MLP, $\boldsymbol{\theta}$:
	\[
		\mathbf{y}^{(L)}(s, \boldsymbol{\theta})\doteq(y^{(L)}_1(s, \boldsymbol{\theta}), y^{(L)}_2(s, \boldsymbol{\theta}), \ldots, y^{(L)}_{n_L}(s, \boldsymbol{\theta})).
	\]

	The preferences vector is transformed into a probability distribution over the $n_L$ actions by applying the \textit{softmax} function (more below), resulting in the following policy function:
	\begin{equation*}
		\pi(a|s,\boldsymbol{\theta}) \doteq \text{softmax}(\mathbf{y}^{(L)}(s, \boldsymbol{\theta}))[a].
	\end{equation*}

	The \textit{softmax} function converts a vector of \(n\) real numbers, \(\mathbf{x} = (x_1, \ldots, x_n) \in \mathbb{R}^n\), into a normalized vector that represents a probability distribution. By denoting the \(i\)-th element of any vector \(\mathbf{a}\) as \(\mathbf{a}[i]\), the softmax function is applied element-wise as follows:
	\begin{equation}
		\text{softmax}(\mathbf{x})[i] \doteq \frac{e^{\mathbf{x}[i]}}{\sum_{j=1}^n e^{\mathbf{x}[j]}}, \quad \forall \, 1 \le i \le n.
	\end{equation}

\begin{figure}[h]
		\centering
		\includegraphics[width=12cm]{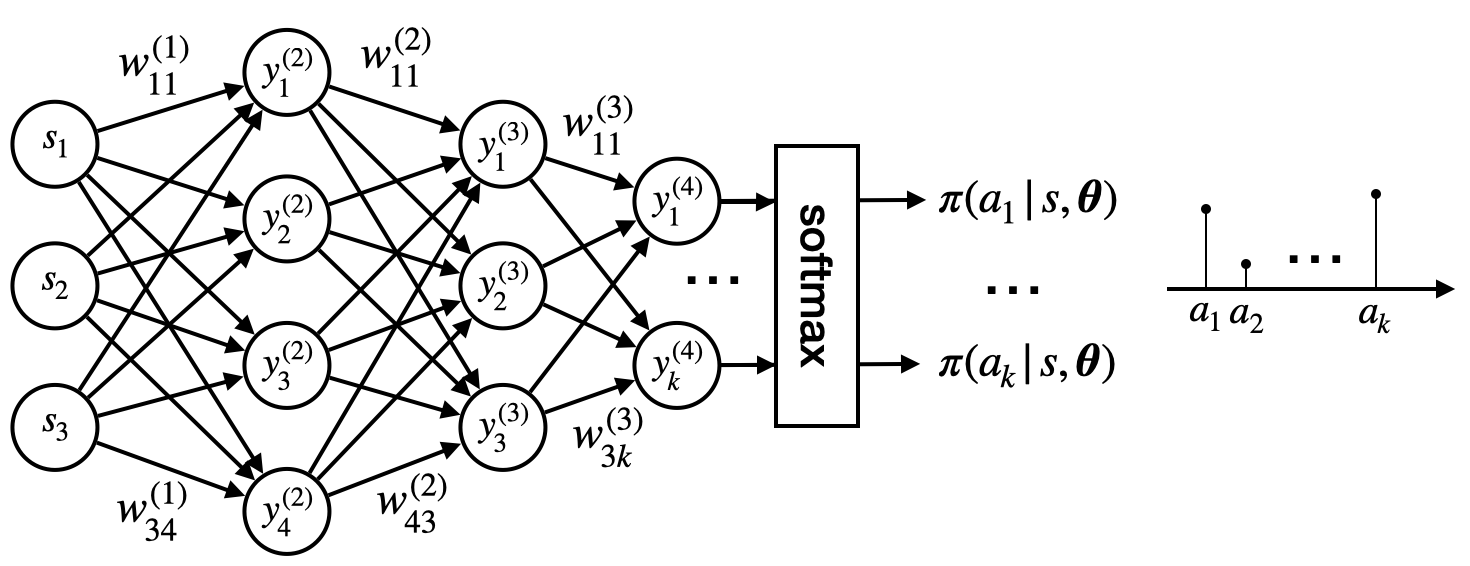}
		\caption{Architecture of a 4-Layer MLP modeling a policy over $k$ discrete actions. The plot on the right shows the resulting probability distribution (probability mass function) over the $k$ actions.}
		\label{fig:network_policy_b}
	\end{figure}

	Figure~\ref{fig:network_policy_b} depicts an example of an MLP modeling a policy over $k$ discrete actions.

	\subsection{Policies for Continuous Action Spaces}
	\label{sec:cont-action-spaces}

	In many control tasks, actions are naturally continuous, such as steering angles, joint torques, or velocity commands. Handling continuous action spaces requires policies that model probability distributions over an infinite set of possible actions, enabling smooth and precise control.

	A common approach for continuous action spaces is to learn the parameters of a probability distribution over actions, such as the mean of a Gaussian distribution. The mean represents the most likely action for the current state. This learned distribution is then sampled to produce the current action, enabling the policy to handle an infinite set of possible actions.

	To model a stochastic policy using a Gaussian distribution, the mean of the distribution is parameterized as a function of the state through an MLP. This function is denoted by \(\mu(s, \boldsymbol{\theta})\), where \(\boldsymbol{\theta}\) represents the learnable parameters of the policy network. Similarly, the standard deviation \(\sigma(s, \boldsymbol{\theta})\) can be either parameterized by the network or treated as a fixed hyperparameter. The Gaussian probability density function \(\mathcal{N}(\cdot \mid \mu, \sigma^2)\) is defined as
	\[
		\mathcal{N}(a \mid \mu, \sigma^2) \doteq \frac{1}{\sigma \sqrt{2\pi}} \exp \left(-\frac{(a - \mu)^2}{2 \sigma^2} \right).
	\]

	Using this notation, the probability of selecting action \(a\) given state \(s\), under the stochastic policy parameterized by \(\boldsymbol{\theta}\) can be expressed as \cite{sutton2018reinforcement}:
	\[
		\pi(a \mid s, \boldsymbol{\theta}) \doteq \mathcal{N}\bigl(a \mid \mu(s, \boldsymbol{\theta}), \sigma(s, \boldsymbol{\theta})^2 \bigr).
	\]

\begin{figure}[h]
		\centering
		\includegraphics[width=12cm]{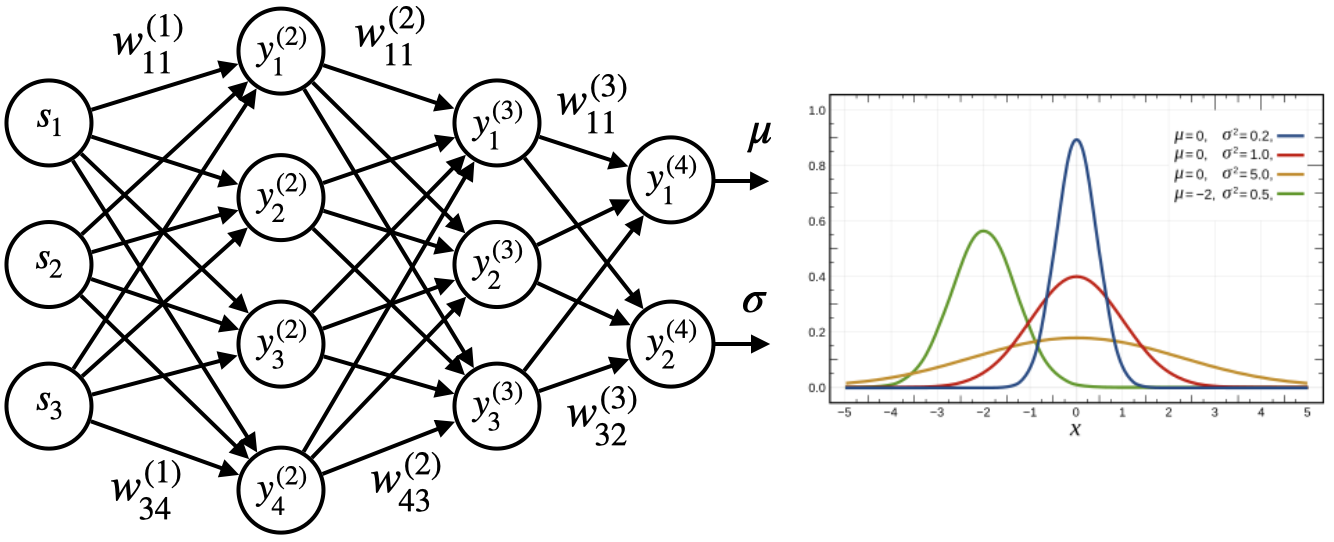}
		\caption{Architecture of a 4-Layer MLP modeling a policy over one continuous action. The plot on the right depicts possible probability density functions representing the distribution over the continuous action values.}
		\label{fig:network_policy_pdf_b}
	\end{figure}

	In summary, at each time step, the MLPs $\mu(s, \boldsymbol{\theta})$ and $\sigma(s, \boldsymbol{\theta})$ output the parameters of a Gaussian distribution that models the agent's behavior in the current state. The goal is to learn the MLP parameters so that the resulting Gaussian distributions define a policy from which the agent can sample actions appropriate for the current state. Figure~\ref{fig:network_policy_pdf_b} illustrates an example of an MLP modeling a policy over a single continuous action. In this setup, the network outputs parameters for both the mean and the standard deviation of the action distribution.

	\subsection{Handling Partial Observability}

	The standard MDP formalism assumes \emph{full observability}, meaning that the agent has direct access to the true underlying state \( s_t \in \mathcal{S} \) of the environment at each time step. This assumption implies the \emph{Markov property}, which states that the future dynamics depend only on the current state and action, not on the full history \cite{sutton2018reinforcement}:
	\[
		P(S_{t+1} = s' \mid S_t = s, A_t = a, S_{t-1}, A_{t-1}, \ldots)=P(S_{t+1} = s' \mid S_t = s, A_t = a).
	\]

	In practical scenarios where the agent receives only partial, egocentric \emph{observations} $o_t \in \mathcal{O}$ instead of the full state $s_t$, the Markov property may not hold. For example, in 3D navigation, the agent may perceive only sensory information within its field of view, while the full state includes all entities' positions and velocities. In such cases, the problem is naturally modeled as a Partially Observable Markov Decision Process (POMDP). For simplicity, this document assumes \emph{full observability}, treating the agent's input as the full state $s_t$. Nevertheless, all algorithms can be applied in partially observable settings by substituting $s_t$ with $o_t$. Performance may degrade due to missing information, which can be mitigated by augmenting observations with recent past data (\emph{observation stacking}) or with the \emph{hidden state} of a recurrent network (e.g., LSTM or GRU) that summarizes relevant history.

	\section{Expert Demonstration}

	Imitation learning starts with the collection of expert demonstrations, that is, sequences of state-action pairs that illustrate how a task should be performed. These demonstrations can be gathered in several ways, depending on the domain and the agent's embodiment. A common and intuitive approach is teleoperation, where a human expert directly controls the agent, such as a robot or a simulated character, by issuing actions in response to the current state. This setup provides state-action pairs without requiring explicit access to the expert's internal decision-making process.

	For instance, a person may control an agent via keyboard or joystick to produce expert demonstrations. Although teleoperation is perhaps the most direct form of demonstration, the expert may also be a scripted controller or a pre-trained policy acting as an oracle. In all cases, the core idea is the same: the expert's behavior is observed and recorded as a collection of trajectories, which serve as training data for the learning agent.

	Since the expert interacts with the environment through the agent (e.g., via tele-operation), in the remainder of this chapter we will distinguish explicitly between the \textbf{expert} and the \textbf{learner}, rather than referring to the learner as the agent (both directly control the same physical or simulated agent).

	Let denote the \textit{expert policy} by \( \pi_E(a \mid s) \), which specifies how an expert demonstrator selects an action \( a \in \mathcal{A} \) when in state \( s \in \mathcal{S} \). This policy represents the behavior that the learner aims to imitate, but it is not known or modeled explicitly. Instead, it is accessed interactively and should be interpreted as an \emph{oracle}, that is, an external source of supervision that can be queried for the correct action in a given state, without revealing its internal structure. In practice, \( \pi_E \) may correspond to a human operator or a scripted controller that provides action labels upon request. When the expert is deterministic, \( a = \pi_E(s) \) is used to indicate that the expert selects a single, fixed action in state \( s \). Conversely, when the expert is stochastic, \( \pi_E(a \mid s) \) represents the probability of selecting action \( a \) when in state \( s \), thus defining a distribution over possible actions.

	The goal of imitation learning is to learn a parameterized policy \(\pi(a \mid s, \boldsymbol{\theta})\) with parameters \(\boldsymbol{\theta}\) that approximates the expert policy \(\pi_E\). That is, the learning process seeks parameters \(\boldsymbol{\theta}\) such that the learned policy selects actions in a manner similar to the expert for the encountered states.

	Assuming an episodic setting, the expert policy \( \pi_E \) governs the behavior of the demonstrator and thereby induces a distribution over trajectories. Intuitively, each trajectory corresponds to a demonstration of the task, consisting of the sequence of states visited and actions taken by the expert during a single episode. Formally, a trajectory \( \tau \) is sampled according to the distribution \( p(\tau \mid \pi_E) \), which reflects both the expert's decision-making and the environment's stochastic dynamics. A \textit{trajectory} \( \tau^{(i)} \) is typically represented as a temporally ordered sequence of state-action pairs:
	\[
		\tau^{(i)} = \left( s_0^{(i)}, a_0^{(i)}, s_1^{(i)}, a_1^{(i)}, \dots, s_{T^{(i)}}^{(i)}, a_{T^{(i)}}^{(i)} \right),
	\]

	\noindent where, for a given trajectory $i$, \(s_t^{(i)}\) denotes the environment state at time step \(t\), \(a_t^{(i)}\) the corresponding action, and \(T^{(i)}\) the horizon of the episode, that is, the total number of time steps in the demonstration. Trajectories serve as the basic unit of data for imitation learning algorithms, encapsulating the expert's behavior over time in terms of what was perceived and how it was acted upon. The set of $N$ expert demonstrations collected in a \textit{demonstration dataset} is denoted as:
	\[
		\mathcal{D} \doteq \{ \tau^{(i)} \}_{i=1}^N = \left\{ (s_0^{(i)}, a_0^{(i)}, \dots, s_{T^{(i)}}^{(i)}, a_{T^{(i)}}^{(i)}) \right\}_{i=1}^N.
	\]

	Above, the dataset \(\mathcal{D}\) was defined as a collection of \(N\) expert demonstrations, where each demonstration corresponds to a full trajectory. However, in some imitation learning approaches it is common to flatten this dataset into a set of individual state-action pairs to facilitate supervised learning. Let \(M\) denote the total number of state-action pairs contained in all trajectories in \(\mathcal{D}\), $M = \sum_{i=1}^N ( T^{(i)} + 1)$ and the flattened dataset \(\mathcal{D}'\) be defined as:
	\[
		\mathcal{D}' \doteq \{ (s_j, a_j) \}_{j=1}^M, \quad \text{where } \forall (s_j, a_j) \in \mathcal{D}', \; \exists \tau^{(i)} \in \mathcal{D} \text{ such that } (s_j, a_j) \in \tau^{(i)}.
	\]

\begin{figure}[h]
		\centering
		\includegraphics[width=0.8\linewidth]{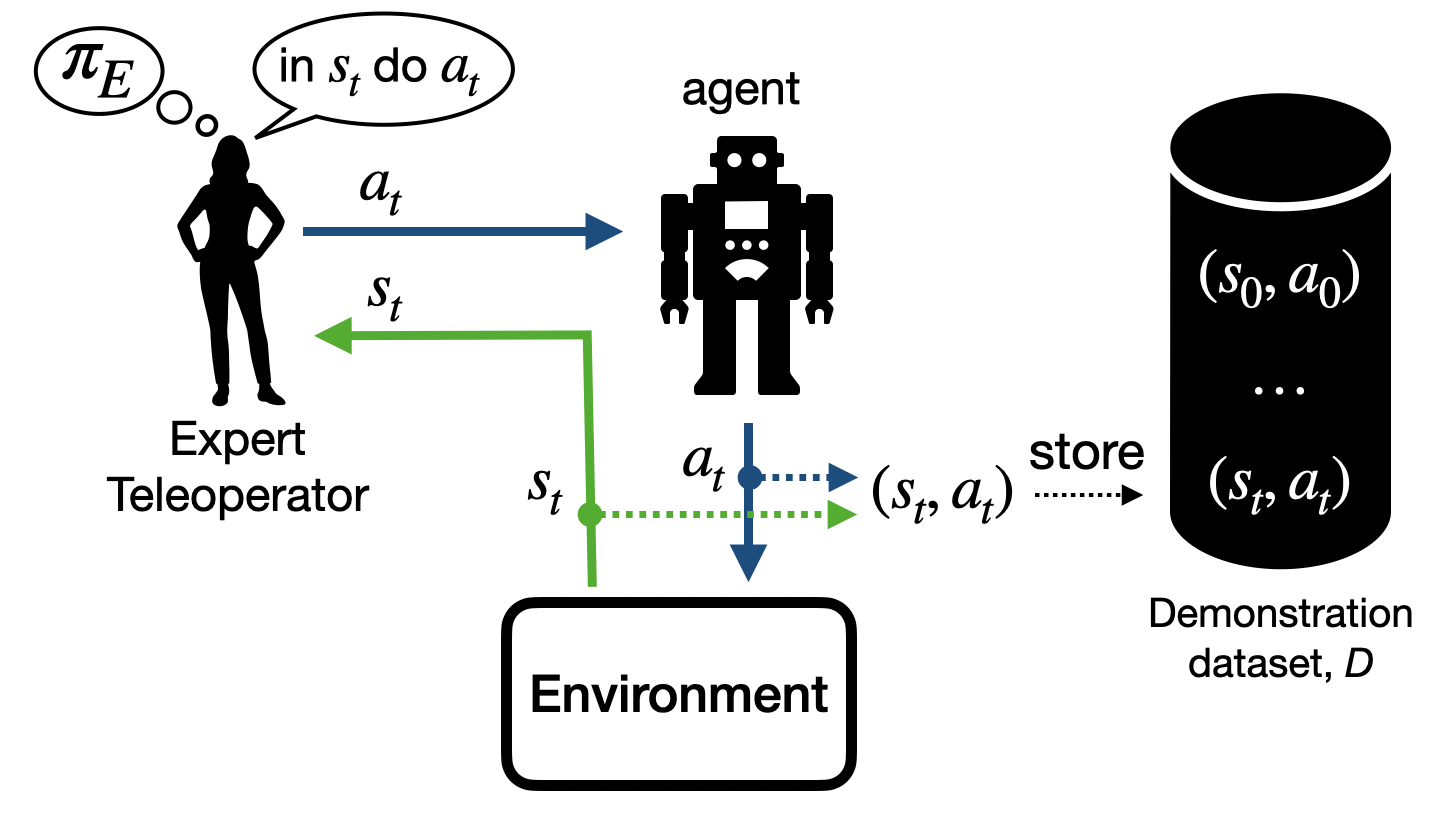}
		\caption{Construction of the demonstration dataset $\mathcal{D}$ via teleoperation.}
		\label{fig:teleoperation}
	\end{figure}

	Figure~\ref{fig:teleoperation} illustrates how the demonstration dataset $\mathcal{D}$ is constructed from expert behavior, in this case provided by a human teleoperator. At each time step $t$, the expert observes the state $s_t$, either directly or through a dedicated interface, chooses an appropriate action $a_t$, according to the expert policy $\pi_E$  (conceptually representing the expert's decision-making process), and controls the agent to execute that action. The resulting state-action pair $(s_t, a_t)$ is then recorded and stored in the demonstration dataset $\mathcal{D}$.

	\section{Behavioral Cloning}
	\label{sec:behavioral_cloning}

\begin{figure}[h]
		\centering
		\includegraphics[width=0.8\linewidth]{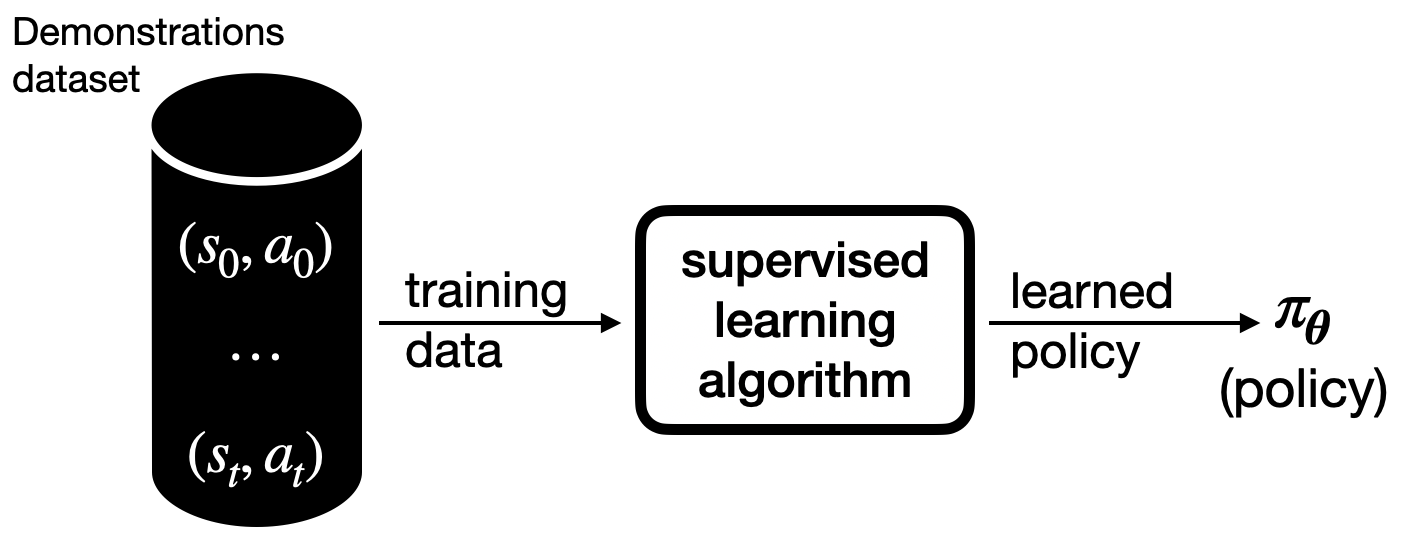}
		\caption{Behavioral Cloning treats imitation learning as a supervised learning problem. A dataset of state-action pairs $(s_t, a_t)$ is used as training data to fit a policy $\pi_{\boldsymbol{\theta}}$ via standard supervised learning.}
		\label{fig:bc_pipeline}
	\end{figure}

	A natural approach to imitation learning is to treat expert demonstrations as supervised training data and learn a policy that mimics the expert by generalizing from observed state-action pairs. This approach, known as \textit{Behavioral Cloning (BC)}, operates under the assumption that the expert acts optimally and that the correct action for any given state can be directly inferred from the demonstration data. Figure~\ref{fig:bc_pipeline} illustrates this process, showing how a supervised learning algorithm is used to fit a policy from a dataset of state-action pairs.

	Behavioral Cloning is appealing because of its conceptual simplicity, reducing the imitation learning problem to a standard supervised learning task. In this framework, the objective is to find policy parameters \(\boldsymbol{\theta}\) that minimize a \textit{loss function} in expectation, i.e., on average, over the expert demonstrations. The loss function measures the discrepancy between the actions suggested by the learned policy and those executed by the expert. Consequently, this optimization process yields a policy that closely approximates the expert's behavior as represented by the available demonstration data.

	\subsection{Definition}

	The goal of imitation learning is to find policy parameters \(\boldsymbol{\theta}^*\) that minimize the expected difference between the behavior of the learned policy and that of the expert, across observed state-action pairs. To express this formally, a general function \(f(s, \boldsymbol{\theta})\) derived from the policy \(\pi\), is introduced. This function represents the policy's output given a state \(s\), but its exact form depends on the type of action space, whether discrete or continuous, and will be explained in detail later.

	Let \(\ell(a, f(s, \boldsymbol{\theta}))\) denote a per-step loss function, which measures how well the policy's output matches the expert's chosen action \(a\). Intuitively, the loss becomes smaller when the policy better replicates the expert's behavior, for example, by assigning a higher probability to the expert's action in the discrete case, or by producing actions closer to the expert's in the continuous case.

	Putting these pieces together, the imitation learning objective is to find parameters \(\boldsymbol{\theta}^*\) that minimize the expected loss over the expert's distribution of state-action demonstrations $p_E$:
	\[
		\boldsymbol{\theta}^* = \arg\min_{\boldsymbol{\theta}} \mathbb{E}_{(s,a) \sim p_E(s,a)} \big[ \ell(a, f(s, \boldsymbol{\theta})) \big].
	\]

	The expected loss over the expert's (unknown) state-action distribution $p_E(s,a)$ cannot be computed directly, since the expert's policy, and thus the distribution of states and actions, is unknown. Nevertheless, the loss can be approximated using the empirical distribution defined by the dataset $\mathcal{D}'$. This leads to the following empirical minimization objective, where the loss is averaged over all $M$ state-action pairs in the dataset:
	\[
		\boldsymbol{\theta}^* \approx \arg\min_{\boldsymbol{\theta}} \frac{1}{M} \sum_{j=1}^M \ell(a, f(s, \boldsymbol{\theta})).
	\]

	Since exhaustively searching over all possible parameter values \(\boldsymbol{\theta}\) is computationally infeasible, and closed-form solutions are generally intractable due to the complexity of modern neural network policies, iterative optimization methods such as \textit{gradient descent} are employed.

	The intuition behind gradient descent is to treat the loss function as a surface over the parameter space and iteratively adjust the parameters \(\boldsymbol{\theta}\) in the direction that most steeply reduces the loss. This direction is given by the negative gradient of the loss with respect to the parameters.

	In its simplest form, the gradient of the empirical loss is computed over the entire dataset \(\mathcal{D}'\), and used to update the parameters as follows:
	\[
		\boldsymbol{\theta} \leftarrow \boldsymbol{\theta} - \alpha \nabla_{\boldsymbol{\theta}} \left[ \frac{1}{M} \sum_{j=1}^M \ell(a, f(s, \boldsymbol{\theta})) \right],
	\]

	\noindent where \(\alpha > 0\) is the learning rate controlling the step size of the update. By repeatedly applying this update, the policy becomes increasingly likely to produce actions that minimize the loss and, consequently, more closely resemble the expert's actions.

	When the demonstration dataset $\mathcal{D}'$ is large, computing the gradient over the entire dataset at each iteration can be computationally expensive and memory-intensive. To address this, training typically employs \emph{mini-batch} stochastic gradient descent. This involves shuffling $\mathcal{D}'$ and splitting it into segments of size $B$, each called a mini-batch. The policy update is then performed over these mini-batches. Formally, a mini-batch is a random subset $\mathcal{B} \subset \mathcal{D}'$ of size $B \ll M$. Gradient descent is applied to each mini-batch as follows:
	\[
		\boldsymbol{\theta} \leftarrow \boldsymbol{\theta} - \alpha \nabla_{\boldsymbol{\theta}} \left( \frac{1}{B} \sum_{(s,a) \in \mathcal{B}} \ell(a, f(s, \boldsymbol{\theta})) \right).
	\]

	An additional benefit of using mini-batch stochastic gradient descent is that the inherent stochasticity in the gradient estimates can improve generalization. Because each mini-batch provides only an approximate gradient of the full loss, the updates contain random fluctuations. This noise helps the optimization process avoid narrow or sharp minima that may overfit the demonstration data. In this way, mini-batch training acts as a form of implicit regularization, reducing overfitting and improving the robustness of the learned policy.

	\subsection{Loss function for continuous actions}

	For continuous action spaces, the policy outputs a deterministic action, which was denoted by \( \mu(s, \boldsymbol{\theta}) \) in Section~\ref{sec:cont-action-spaces}. That is, \( \mu(s, \boldsymbol{\theta}) \) is the action predicted by the policy neural network given the observation or state \( s \). To keep notation consistent,  the general policy output function \( f \) is defined as:
	\[
		f(s, \boldsymbol{\theta}) \doteq \mu(s, \boldsymbol{\theta}).
	\]

	A widely used and intuitive loss function in this context is the Mean Squared Error (MSE) between the expert action \( a \) and the policy's predicted action:
	\[
		\ell(a, f(s, \boldsymbol{\theta})) \doteq \| a - f(s, \boldsymbol{\theta}) \|_2^2 = \| a - \mu(s, \boldsymbol{\theta}) \|_2^2,
	\]

	\noindent where \( \| \cdot \|_2 \) denotes the Euclidean norm. Intuitively, this loss grows larger as the predicted action \( \mu(s, \boldsymbol{\theta}) \) moves further away from the expert action \( a \), thereby encouraging the policy to produce outputs that closely approximate the expert's behavior.

	\subsection{Discrete Action Spaces}

	The discrete case is a bit more intricate, requiring us to first understand the concept of Maximum Likelihood Estimation (MLE) before delving into the details.

	MLE for policy estimation builds on the fundamental question: \emph{How likely (or probable) are the expert demonstrations \((s, a)\) if the learner follows the policy parameterized by \(\boldsymbol{\theta}\)?} If this likelihood is low, it means the current policy parameters \(\boldsymbol{\theta}\) do not produce behavior similar to the expert. Conversely, if the likelihood is high, the policy closely mimics the expert's behavior. Therefore, the goal of learning is to adjust the parameters \(\boldsymbol{\theta}\) such that the policy distribution assigns high probability to the expert's observed actions in their respective states.

	Formally, this means maximizing the likelihood of the expert data under the policy. It is possible to compute the likelihood of observing the $B$ expert demonstrations $\{(s_i, a_i)\}_{i=1}^B$ from a mini-batch $\mathcal{B}$ by assuming that each state-action pair is independent, i.e., the occurrence of one pair provides no information about which pairs will occur next. While this independence assumption simplifies computation and optimization, it ignores temporal dependencies and correlations between consecutive actions within a trajectory. Nevertheless, when operating with mini-batches, these correlations are less pronounced. Under this assumption, the \emph{likelihood} is formally given by:
	\[
		L(\mathcal{B}, \boldsymbol{\theta}) = \prod_{i=1}^B \pi(a_i \mid s_i, \boldsymbol{\theta}).
	\]

	For computational convenience, rather than maximizing the likelihood, which is a product of probabilities, the \emph{log-likelihood} is optimized by applying the logarithmic identity \( \ln(xy) = \ln(x) + \ln(y) \):
	\[
		\ln L(\mathcal{B}, \boldsymbol{\theta}) = \sum_{i=1}^B \ln \pi(a_i \mid s_i, \boldsymbol{\theta}).
	\]

	By transforming the product of probabilities into a sum, the optimization becomes easier and numerically more stable. Moreover, since the logarithm is a strictly increasing (monotonic) function, maximizing the log-likelihood is equivalent to maximizing the original likelihood; that is, the logarithm does not change the location of the optimum.

	We can now formalize the MLE problem as maximizing the log-likelihood function. This involves finding the optimal parameters $\boldsymbol{\theta}^*$ such that the policy explains the expert demonstrations as well as possible, i.e., $\pi_{\boldsymbol{\theta}^*} \approx \pi_E$:
	\[
		\boldsymbol{\theta}^* = \arg\max_{\boldsymbol{\theta}} \sum_{i=1}^B \ln \pi(a_i \mid s_i, \boldsymbol{\theta}).
	\]

	Maximizing this objective encourages the policy to assign higher probabilities to the actions actually taken by the expert in the observed states. In other words, parameters \(\boldsymbol{\theta}\) are adjusted so the policy behaves more like the expert, increasing the likelihood that it reproduces expert behavior.

	Since maximizing a function is equivalent to minimizing its negative, the MLE objective can be reformulated as a minimization problem:
	\[
		\boldsymbol{\theta}^* = \arg\min_{\boldsymbol{\theta}} \left[ - \sum_{i=1}^B \ln \pi(a_i \mid s_i, \boldsymbol{\theta}) \right].
	\]

	This objective sums over a batch of expert demonstrations, with each term penalizing the model if it assigns low probability to the expert's action. As such, the quantity being minimized acts as a measure of how poorly the policy fits the expert data. This interpretation naturally motivates the definition of a \emph{loss function} for imitation learning in discrete action spaces. Because the total loss is additive over data points, it is standard to define a \emph{per-sample loss} as:
	\[
		\ell(a, s, \boldsymbol{\theta}) \doteq - \ln \pi(a \mid s, \boldsymbol{\theta}).
	\]

	To keep notation consistent, the general policy output function \( f \) is redefined as:
	\[
		f(s, \boldsymbol{\theta}) \doteq \pi(\cdot \mid s, \boldsymbol{\theta}).
	\]

	Let \( f(s, \boldsymbol{\theta})[a] \) denote the probability assigned by the policy to the expert's action \( a \) when in state \( s \), $\pi(a \mid s, \boldsymbol{\theta})$. With this notation, the per-sample loss can be written more generally as:
	\[
		\ell(a, f(s, \boldsymbol{\theta})) \doteq - \ln f(s, \boldsymbol{\theta})[a] = - \ln \pi(a \mid s, \boldsymbol{\theta}).
	\]

	This formulation corresponds to the \emph{cross-entropy loss}, a standard choice in classification tasks where the model outputs a probability distribution over discrete classes. The per-sample loss in behavioral cloning is a special case of cross-entropy loss where the expert action defines a one-hot label over actions. If expert demonstrations provide soft action probabilities instead, the full cross-entropy expression over all actions is required.

	\subsection{Behavioral Cloning Pseudocode}

	Algorithm~\ref{alg:behavioral_cloning} summarizes the Behavioral Cloning procedure described above. Starting from a dataset $\mathcal{D}$ of expert trajectories, the algorithm first flattens it into a set $\mathcal{D}'$ of individual state-action pairs. Training proceeds over a number of epochs $E$ using mini-batch stochastic gradient descent. At each iteration, a mini-batch $\mathcal{B}$ is sampled from $\mathcal{D}'$, and the policy parameters $\boldsymbol{\theta}$ are updated to minimize the loss $\ell(a, f(s, \boldsymbol{\theta}))$ between the predicted and expert actions. The loss function and the structure of $f(s, \boldsymbol{\theta})$ depend on the action space, as discussed in the previous sections. Upon convergence, the procedure yields a policy that imitates the expert by, ideally, generalizing from the provided demonstrations.

	\vspace{0.5cm}
	{\small
		\begin{algorithm}[H]
			\caption{Behavioral Cloning}
			\label{alg:behavioral_cloning}
			\SetKwInOut{Input}{Input}
			\SetKwInOut{Output}{Output}

			\Input{Demonstration dataset $\mathcal{D} = \{ \tau^{(i)} \}_{i=1}^N$, where $\tau^{(i)} = \left( s_0^{(i)}, a_0^{(i)}, \dots, s_{T^{(i)}}^{(i)}, a_{T^{(i)}}^{(i)} \right)$}
			\Input{Initial policy parameters $\boldsymbol{\theta}$}
			\Input{Loss function $\ell(\cdot, \cdot)$}
			\Input{Learning rate $\alpha$, batch size $B$, number of epochs $E$}
			\Output{Trained policy parameters $\boldsymbol{\theta}$}
			\vspace{0.2cm}
			Initialize $\boldsymbol{\theta}$ (e.g., randomly)\;
			Construct flattened dataset $\mathcal{D}' = \{(s_j, a_j)\}_{j=1}^M$ from all trajectories in $\mathcal{D}$\;

			\For{$e = 1$ \KwTo $E$}{
				Shuffle $\mathcal{D}'$\;

				Split $\mathcal{D}'$ into mini-batches of size $B$\;

				\ForEach{mini-batch $\mathcal{B} \subset \mathcal{D}'$}{
					Compute gradient: \\
					\Indp
					$\delta \leftarrow \nabla_{\boldsymbol{\theta}} \left( \frac{1}{B} \sum_{(s,a) \in \mathcal{B}} \ell(a, f(s, \boldsymbol{\theta})) \right)$\;
					\Indm
					Update parameters: \\
					\Indp
					$\boldsymbol{\theta} \leftarrow \boldsymbol{\theta} - \alpha \delta$\;
					\Indm
				}
			}
		\end{algorithm}
	}
	\subsection{Covariate Shift and Compounding Errors}

	A learner using a policy trained on expert demonstrations may deviate from the expert's behavior during execution. These deviations can arise from several sources.

	First, the policy is typically implemented as a parameterized function, such as a neural network, which may lack sufficient capacity to fully replicate the expert's decision-making. Second, the policy is trained on a finite dataset of demonstrations, which may fail to cover the full range of task variations. Third, the environment often exhibits stochastic dynamics; even in virtual settings, complex physics engines or procedurally generated layouts can cause identical actions in seemingly identical states to yield different outcomes across episodes. Additionally, learners rely on potentially noisy sensory inputs that may introduce partial observability and sensing artifacts, and control signals issued by the policy can be affected by latency, discretization, or inaccuracies in the simulated dynamics.

	Moreover, expert demonstrations are often imperfect. Human demonstrators or scripted agents may behave sub-optimally, inconsistently, or fail to respond precisely to rapidly changing environments. In navigation or locomotion tasks, for instance, demonstrators might overshoot turns, hesitate, or generate jerky movements due to limited control granularity. Consequently, even perfect imitation of such demonstrations can lead to degraded performance during autonomous execution, because the learner reproduces observed behaviors that may not fully reflect the expert's intended optimal policy $\pi_E$.

	Together, these factors create a distributional mismatch between trajectories induced by the learned policy and those induced by the expert policy $\pi_E$, which may not be fully represented by the imperfect demonstrations. This phenomenon is commonly referred to as \emph{covariate shift}. Under covariate shift, the policy can produce actions that lead the learner into states that were rarely or never visited by the expert, or that were visited only due to imperfections in the demonstrations. These states are poorly represented in the training data, leaving the policy uncertain about the appropriate action. As a result, the learner is likely to select actions that deviate from those intended by the expert. By behaving differently from the expert, the policy can drive the learner into increasingly unfamiliar states, where its accuracy further decreases, compounding deviations from the expert's intended trajectory. Consequently, small errors can accumulate over time, causing the learner's behavior to diverge from the expert's intended policy $\pi_E$ and resulting in degraded performance during autonomous execution.

\begin{figure}[h]
		\centering
		\includegraphics[width=0.8\linewidth]{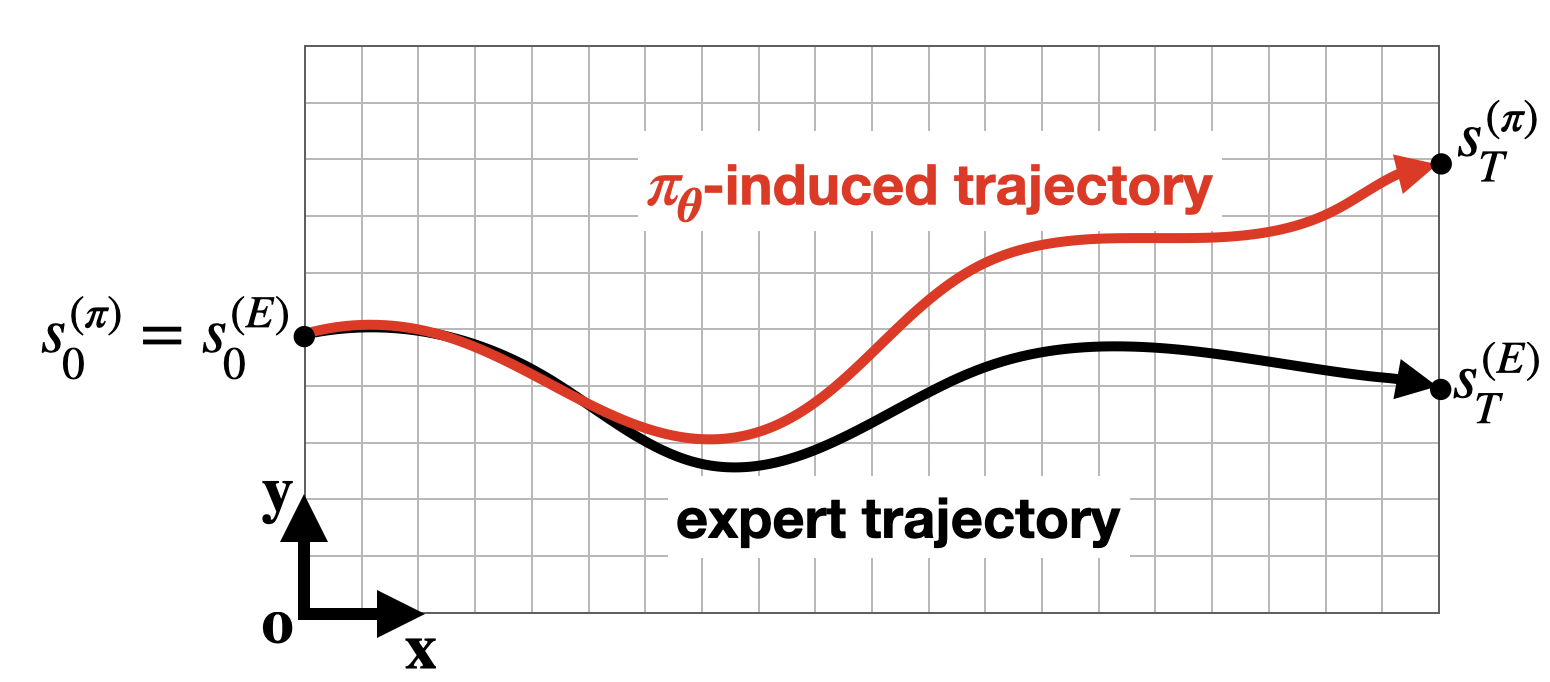}
		\caption{Illustration of covariate shift and compounding errors in Behavioral Cloning.}
		\label{fig:covariate_shift}
	\end{figure}

	The covariate shift phenomenon is illustrated in Figure~\ref{fig:covariate_shift}. In the figure, although the expert and the learner start from the same initial state, $s_0^{(E)} = s_0^{(\pi)}$, the trajectory induced by the learned policy $\pi_{\boldsymbol{\theta}}$ gradually diverges from the expert's intended trajectory due to the accumulation of small errors. After $T$ time steps, the learner reaches a state $s_T^{(\pi)}$ that differs significantly from the expert's intended state $s_T^{(E)}$, illustrating how distributional shift during execution can lead the policy into unfamiliar states where its performance deteriorates.

	The following section introduces \emph{DAgger} (Dataset Aggregation), an extension of Behavioral Cloning specifically designed to address the covariate shift problem.

	\section{DAgger: Dataset Aggregation}
	\label{sec:dagger_details}

	DAgger (Dataset Aggregation) \cite{dagger} can be viewed as \emph{Behavioral Cloning (BC) augmented with an interactive data collection loop}. Like standard BC, it relies on expert demonstrations to learn a mapping from states to actions. However, unlike BC, it does not rely on a fixed dataset of expert-generated trajectories. Instead, DAgger introduces an iterative mechanism in which the learner repeatedly interacts with the environment under its current policy, while querying the expert for corrective actions in the visited states. This allows the training distribution to gradually adapt to the one induced by the learner's own behavior during execution.

	More precisely, DAgger replaces the single supervised learning phase of behavioral cloning with a sequence of $K$ iterative refinements. In each iteration, the learner acts according to its current policy, while the expert provides corrective annotations for the states encountered, effectively answering the question: "\textit{Do I agree with the learner's action? If not, what should have been done instead?}" The policy is then retrained using the aggregated dataset, which includes  the original expert demonstrations and the new states labeled with the expert actions. This iterative data collection strategy gradually exposes the policy to states that arise due to covariate shift, states that were not part of the original demonstrations but are likely to be visited when the learner operates autonomously. As a result, the learned policy becomes robust to distributional drift and better aligned with the expert's behavior across the full range of encountered states.

	\subsection{Deterministic Policies}\label{sec:deterministic_policies}

	In DAgger, both the learner and the expert are commonly assumed to follow \emph{deterministic policies}, meaning that they always return the same action when presented with the same state. This assumption greatly simplifies the querying process: when asking the expert to correct the learner's behavior, we need a \emph{single, unambiguous reference action}, rather than a distribution over possible actions.

	Formally, a deterministic policy is a function mapping each state to a specific action,
	\[
		a_t = \pi(s_t, \boldsymbol{\theta}).
	\]

	Although policies in reinforcement learning are frequently represented as probability distributions, DAgger extracts a deterministic action from these distributions. The specific form of this action depends on the type of action space under consideration. In continuous action spaces (see Section~\ref{sec:cont-action-spaces}), policies are typically modeled as Gaussian distributions, and the deterministic action used by DAgger corresponds to the predicted mean of the distribution:
	\[
		\pi(s_t, \boldsymbol{\theta}) \doteq \mu(s_t, \boldsymbol{\theta}).
	\]

	In contrast, in discrete action spaces (see Section~\ref{sec:discrete-actions}), policies define a categorical distribution over a finite set of actions, and DAgger selects the most likely action according to this distribution:
	\[
		\pi(s_t, \boldsymbol{\theta}) \doteq \arg\max_{a \in \mathcal{A}} \pi(a \mid s_t, \boldsymbol{\theta}).
	\]

	This deterministic interpretation is essential in DAgger, as it ensures that comparisons with the expert's decisions are precise and unambiguous.

	\subsection{Algorithm Description}

	Assume that the expert behaves according to a determinist policy \( \pi_E \), which is unknown to the learner. The learner's goal is to learn this policy. Initially, the learner interacts with the environment under the full control of the expert (tele-operation), i.e., according to \( \pi_E \), and both the visited states and corresponding expert actions are recorded:
	\[
		\mathcal{D}_0' = \left\{ \left(s_t^{(i)}, a_t^{(i)} \right) \right\}_{i=1}^N, \quad \text{where } a_t^{(i)} = \pi_E(s_t^{(i)}).
	\]

	A first policy \( \pi_1 \) with parameters \( \boldsymbol{\theta}_1 \) is then trained on \( \mathcal{D}_0' \) using supervised learning, as described in Section~\ref{sec:behavioral_cloning} for Behavioral Cloning.
	This results in a policy that initially imitates the expert only within the distribution of states visited during the demonstrations carried so far. Hence, when deployed, the learner may visit unfamiliar or out-of-distribution states not present in \( \mathcal{D}_0' \). For example, a learner trained to navigate around obstacles might slightly misjudge a turn, leading it into an unvisited dead end. Since the policy has never encountered such states, it may respond poorly, leading to compounding errors. DAgger addresses this limitation by allowing the learner to act in the environment while continuing to query the expert for corrective actions in the states it actually visits, improving the policy over $K$ iterations.

	\vspace{0.5cm}
	In each iteration \( k = 1, \dots, K \), DAgger performs the following sequence of steps \cite{dagger}:

	\begin{enumerate}

		\item The learner acts in the environment over an episode under the control of the current deterministic policy \( \pi_k \)  and the visited states, $(s_0^{k},\dots,s_T^{k})$, are stored.
		\item The expert is called upon labelling the visited states $(s_0^{k},\dots,s_T^{k})$ with the correct actions $(\pi_E(s_0^{(k)}),\ldots,\pi_E(s_T^{(k)}))$
		\item The visited states and selected actions are paired into an aggregated trajectory $\tau^{(k)}=\left(\left(s_0^{(k)},\pi_E(s_0^{(k)})\right),\ldots,\left(s_T^{(k)},\pi_E(s_T^{(k)})\right)\right)$
		\item The tuples in the trajectory $\tau^{(k)}$ are stored into a new  dataset $\mathcal{D}_k^{'\text{new}}$:
		      \[
			      \mathcal{D}_k^{'\text{new}} = \left\{ \left( s_t^{(k)}, \pi_E(s_t^{(k)}) \right) \right\}_{t=0}^{T^{(k)}}.
		      \]
		\item The newly collected data is  aggregated with all previously collected data, broadening the state-action pairs coverage:
		      \[
			      \mathcal{D}_k' = \mathcal{D}_{k-1}' \cup \mathcal{D}_k^{'\text{new}}.
		      \]
		\item A new policy \( \pi_{k+1} \) is trained on the updated dataset \( \mathcal{D}_k' \), again using supervised learning, as described in Section~\ref{sec:behavioral_cloning} for Behavioral Cloning.

	\end{enumerate}

\begin{figure}[h]
		\centering
		\includegraphics[width=1\linewidth]{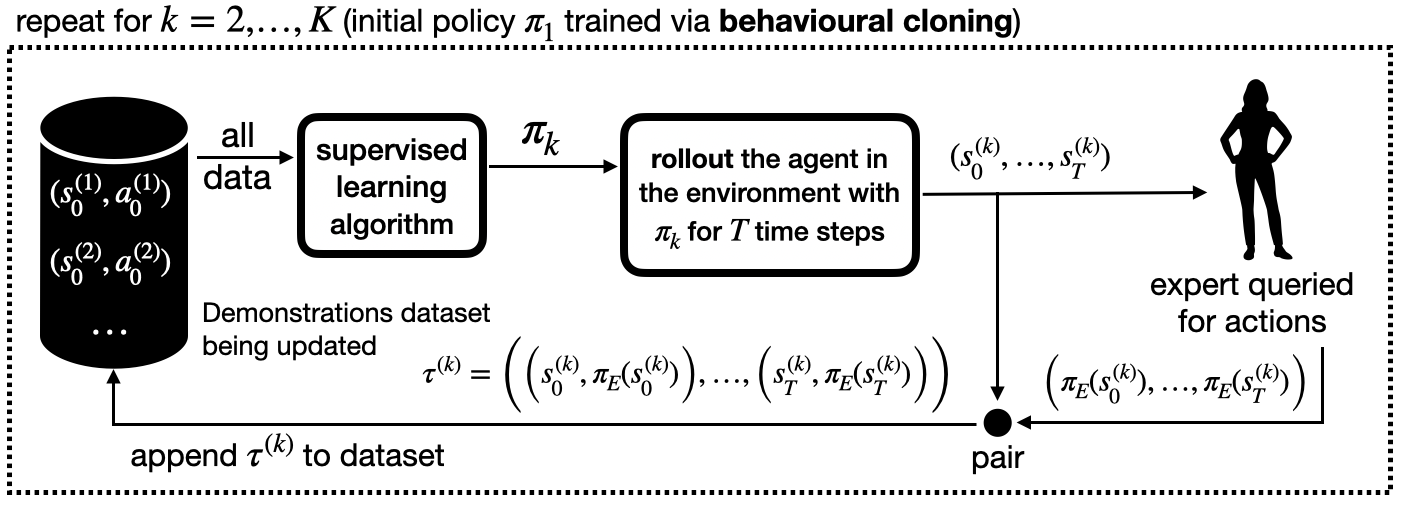}
		\caption{Diagram of the DAgger \cite{dagger} data aggregation loop for iterations \( k = 2, \dots, K \), assuming the initial policy \( \pi_1 \) was trained via Behavioral Cloning on a static expert dataset. In each iteration \( k \), the current policy \( \pi_k \) is trained on the full aggregated dataset, deployed in the environment, and queried by the expert to label the visited states. The resulting trajectory \( \tau^{(k)} \) of state-action pairs is appended to the dataset to improve robustness to distributional shift.}
		\label{fig:dagger_diagram}
	\end{figure}

	Repeating for \( K \) iterations yields a sequence of policies \( \{\pi_k\} \) that progressively improve their robustness and coverage.  After \( K \) iterations, DAgger returns the final policy \( \pi_{K+1} \), trained on the aggregated dataset containing all visited and labeled states throughout training. This iterative process ensures that the trained policy better reflects the broader state distribution encountered by the learner.  Figure~\ref{fig:dagger_diagram} illustrates the DAgger algorithm's main loop, emphasizing the iterative process by which the learner's own state distribution is aligned with the expert's behavior through repeated data aggregation.

	To improve safety and performance during early iterations, when the learned policy may still be unreliable, DAgger performs rollouts using a \emph{mixture policy} that blends the learner's decisions with those of the expert. This reduces the risk of catastrophic errors by allowing the expert to intervene whenever the learner's output is likely to be poor, while still collecting informative samples from the learner's behavior \cite{dagger}:
	\[
		\hat{\pi}_k(s_t) = \beta_k \, \pi_E(s_t^{(k)}) + (1 - \beta_k)\, \pi_k(s_t^{(k)}, \boldsymbol{\theta}_k),
	\]
	\noindent where \( \beta_k = \zeta^k \) is a time-decaying mixing coefficient defined through a user-specified parameter \( \zeta \in [0,1] \). As \( k \) increases, this decay gradually reduces the influence of the expert, allowing the learned policy to take over as it improves.

	Although the expression resembles a weighted average of policy outputs, it does \emph{not} compute any numerical combination of actions. Instead, the mixing is implemented through \emph{probabilistic switching} between the expert and the learned policy. Formally, at each decision step $t$ of iteration \( k \), the action is selected from the expert with probability \( \beta_k \), and from the learner's policy with probability \( 1 - \beta_k \), where the learner's specific action-selection rule depends on the nature of the action space (see Section~\ref{sec:deterministic_policies}):
	\[
		a_t^{(k)} \leftarrow
		\begin{cases}
			\pi_E(s_t^{(k)})                       & \text{with probability } \beta_k,     \\[6pt]
			\pi_k(s_t^{(k)},\boldsymbol{\theta}_k) & \text{with probability } 1 - \beta_k.
		\end{cases}
	\]

	\noindent This stochastic switching strategy ensures:
	\begin{itemize}
		\item \emph{Safer behavior during early iterations}, when the learned policy is unreliable, since most actions are taken from the expert.
		\item \emph{Increasing learner autonomy} as training progresses, because the contribution of the expert diminishes as \( \beta_k \) decays over iterations.
	\end{itemize}

\begin{figure}[h]
		\centering
		\includegraphics[width=0.9\linewidth]{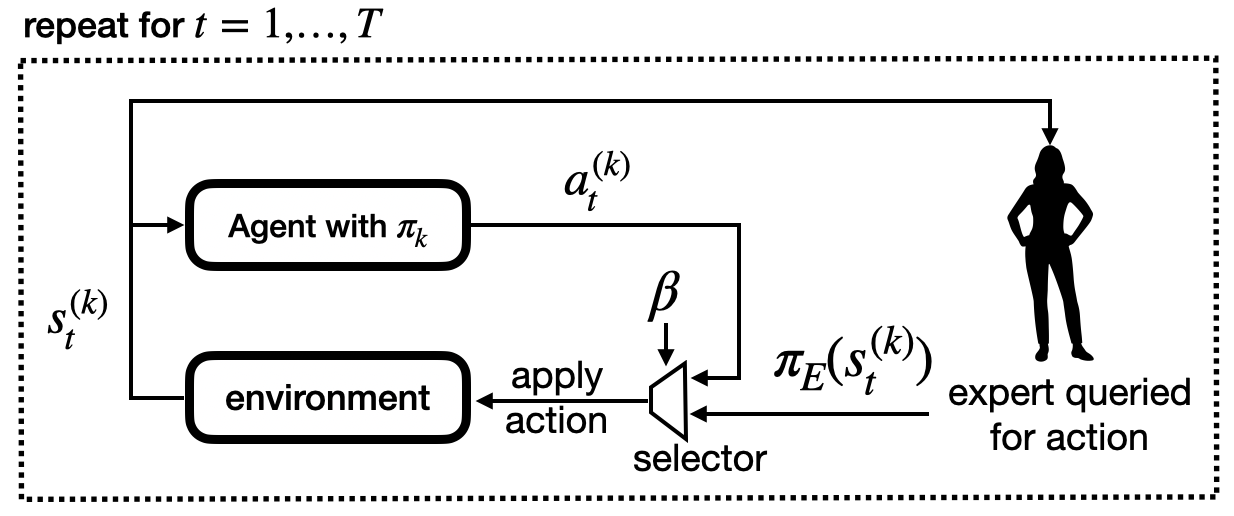}
		\caption{Zoomed-in view of the rollout loop over time steps \( t = 1, \dots, T \) in a single DAgger \cite{dagger} iteration.}
		\label{fig:dagger_diagram_inner_loop}
	\end{figure}

	The rollout module in Figure~\ref{fig:dagger_diagram} is adapted to allow the learner to interact with the environment under this mixture policy. A detailed view is shown in Figure~\ref{fig:dagger_diagram_inner_loop}, which illustrates the rollout loop over time steps \( t = 1, \dots, T \) within a single DAgger iteration. At each step \cite{dagger}:
	\begin{enumerate}
		\item The learner observes the current state \( s_t^{(k)} \) and proposes an action according to its policy \( \pi_k(s_t^{(k)}, \boldsymbol{\theta}_k) \).
		\item A probabilistic selector, controlled by \( \beta_k \), chooses between the \emph{expert action} \( \pi_E(s_t^{(k)}) \) and the \emph{learner action} \( \pi_k(s_t^{(k)}, \boldsymbol{\theta}_k) \).
		\item The selected action \( a_t^{(k)} \) is applied to the environment.
		\item The resulting next state \( s_{t+1}^{(k)} \) is fed back into the loop.
	\end{enumerate}

	Algorithm~\ref{alg:dagger} presents the complete pseudo-code of DAgger \cite{dagger}, formalizing the procedure described in this section.

	\subsection{DAgger in Action: Examples and Results}

\begin{figure}[h]
		\centering
		\includegraphics[width=1\linewidth]{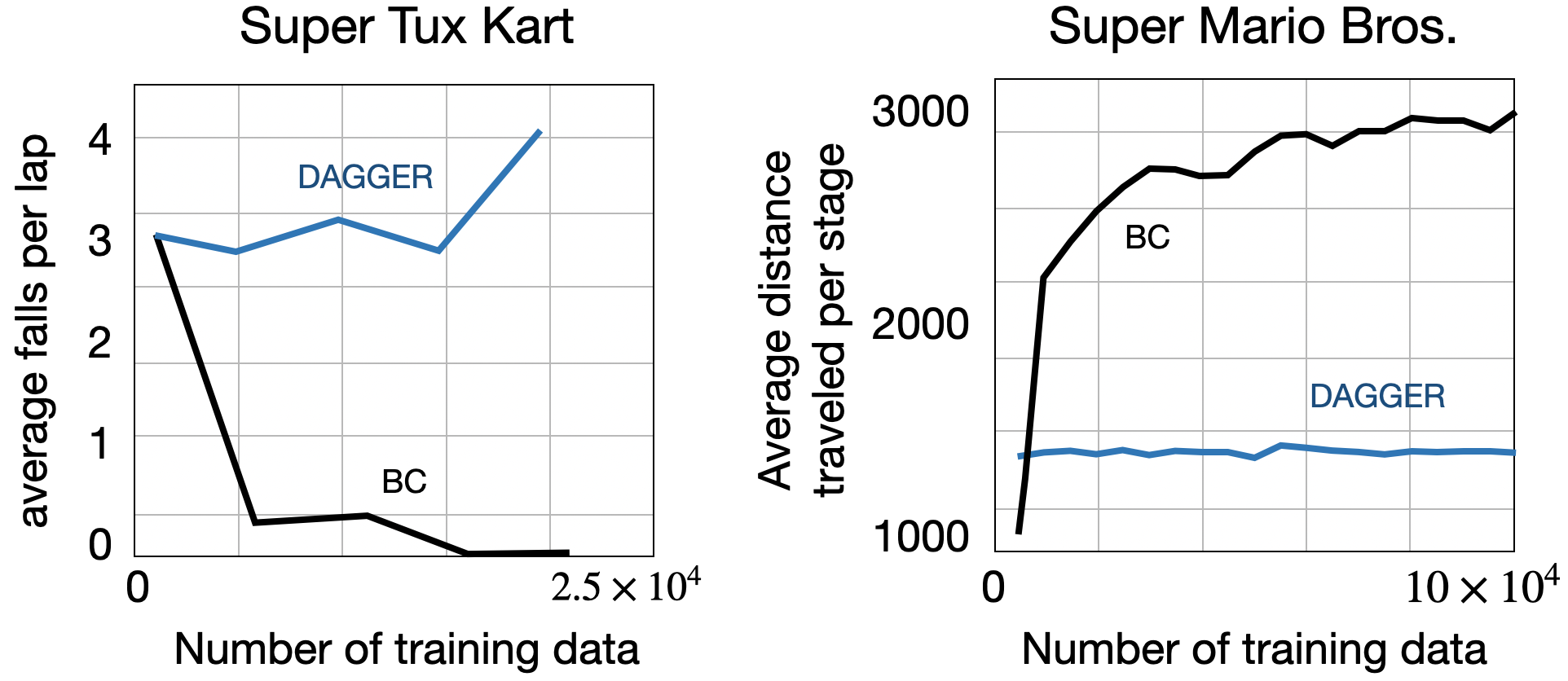}
		\caption{Comparison of average learning performance over multiple runs between DAgger (blue) and Behavioral Cloning (BC) (black) (adapted from \cite{dagger}).}
		\label{fig:dagger_results}
	\end{figure}

	The authors of DAgger evaluated their algorithm against Behavioral Cloning across a range of tasks encompassing both continuous and discrete control settings. For the continuous control problem, they used \textit{Super Tux Kart}\footnote{\url{https://supertuxkart.net}}, a 3D racing game. A human expert provided demonstrations by steering with an analog joystick while observing the corresponding game images. For the discrete control problem, they employed \textit{Super Mario Bros.}, a platform game in which the player must traverse each stage while avoiding enemies, jumping over gaps, and reaching the goal before time expires. In this case, the demonstrator was a planning algorithm that produced demonstrations by controlling a subset of actions: moving left, moving right, jumping, and accelerating. Figure~\ref{fig:dagger_results} shows that DAgger consistently outperformed Behavioral Cloning in both tasks, achieving fewer failed laps in Super Tux Kart and greater distance traveled in Super Mario Bros..

	{\small
	\begin{algorithm}[H]
		\caption{Pseudocode of DAgger \cite{dagger}}
		\label{alg:dagger}
		\KwIn{Expert policy \( \pi_E \) (e.g., human demonstrator or scripted oracle)}
		\KwIn{Number of iterations \( K \), number of episodes per iteration \( M \)}
		\KwIn{Episode horizon \( T \) (assuming fixed for simplicity)}
		\KwIn{Learning rate \( \alpha \), batch size \( B \)}
		\KwIn{Mixing decay rate \( \zeta \in (0,1) \), with \( \beta_k = \zeta^k \)}
		\KwOut{Final policy \( \pi_{K} \)}

		\vspace{0.2em}
		Initialize aggregated dataset \( \mathcal{D}_0' \leftarrow \emptyset \)

		Let the learner interact with the environment under \( \pi_E \)

		Record visited state-action pairs into \( \mathcal{D}_0' \)

		Initialize policy parameters \( \boldsymbol{\theta}_1 \) (e.g., randomly)

		Train \( \pi_1 \) on \( \mathcal{D}_0' \) using supervised learning

		\For(){$k = 1$ \KwTo $K$}{

		Initialize buffer \( \mathcal{D}_k^{'\text{new}} \leftarrow \emptyset \)

		\For(){$m = 1$ \KwTo $M$}{

		Reset environment and observe initial state \( s_0 \)

		\For(){$t = 0$ \KwTo $T$}{

		Sample action:

		$\quad a_t \leftarrow
			\begin{cases}
				\pi_E(s_t)                       & \text{with probability } \beta_k,     \\[6pt]
				\pi_k(s_t,\boldsymbol{\theta}_k) & \text{with probability } 1 - \beta_k.
			\end{cases}$

		Execute \( a_t \) in the environment; observe next state \( s_{t+1} \)

		Query expert: \( a_t^E \leftarrow \pi_E(s_t) \)

		Add \( (s_t, a_t^E) \) to \( \mathcal{D}_k^{'\text{new}} \)
		}
		}

		Aggregate data:
		\[
			\mathcal{D}_k' \leftarrow \mathcal{D}_{k-1}' \cup \mathcal{D}_k^{'\text{new}}
		\]

		Train new policy \( \pi_{k+1} \) on \( \mathcal{D}_k' \) using supervised learning
		}
	\end{algorithm}
	}

	\section{GAIL: Generative Adversarial Imitation Learning}

	Behavioral Cloning (BC) frames imitation learning as a supervised learning task, where a policy is trained to reproduce the expert's actions based on a dataset of state-action pairs. While simple and often effective, BC suffers from a well-known limitation: compounding errors due to covariate shift. Because it trains only on expert-visited states, small mistakes during execution can lead the learner into unfamiliar situations where it behaves poorly.

	DAgger \cite{dagger} addresses this issue by extending BC with an interactive component: the learner executes its current policy, and at each visited state, the expert is queried for the correct action. These newly labeled pairs are added to the dataset, and the policy is retrained from scratch. This iterative data collection helps align the training distribution with the learner's actual experience, reducing covariate shift. However, DAgger requires continual access to the expert during training, which may not be feasible in many settings.

	In this section we will study Generative Adversarial Imitation Learning (GAIL)~\cite{gail}, which is a method that does not require the expert to be queried during training, only an initial dataset of demonstrations. This is attained by integrating reinforcement learning  in the learning process. However, the trial-and-error nature of reinforcement learning requires extensive agent-environment interaction, potentially risky on physical systems and computationally demanding in simulated systems.

	Hence, each method (DAgger vs. GAIL) strikes a different balance between expert supervision and environment interaction, making them suitable under different practical constraints.

	\subsection{GAIL Overview}

	Unlike methods that directly imitate the expert's actions, GAIL learns through an \textit{adversarial interaction} between two components, inspired by Generative Adversarial Networks (GANs) \cite{gan}:

	\begin{itemize}
		\item A discriminator that learns to distinguish whether an observed trajectory comes from the expert or from the learner;
		\item A policy that aims to control the learner so effectively that the discriminator can no longer tell its behavior apart from the expert's.
	\end{itemize}

	The learner's policy is updated with reinforcement learning to maximize a reward. It is rewarded the more the discriminator believes its trajectories were generated by the expert, i.e., the more it successfully "fools" the discriminator. As the policy improves, it becomes increasingly difficult for the discriminator to detect imitation.

	The discriminator is trained to solve a classification problem that distinguishes between expert- and learner-generated trajectories. This task follows standard supervised learning, where the training data are labeled according to their origin: expert or learner. As the discriminator improves, it becomes increasingly difficult for the policy to obtain reward.

\begin{figure}[h]
		\centering
		\includegraphics[width=0.8\linewidth]{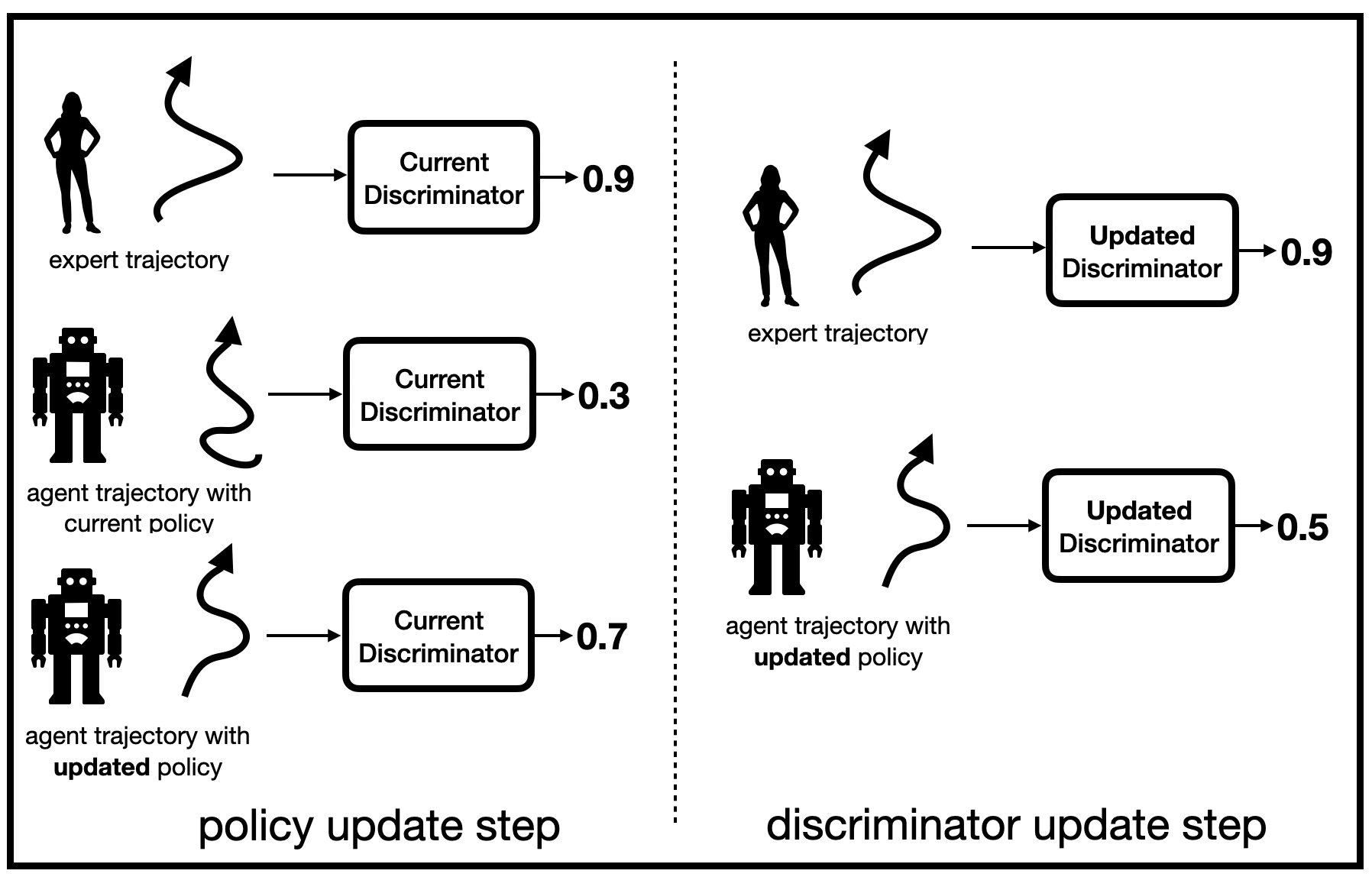}
		\caption{Conceptual illustration of a single training iteration in GAIL \cite{gail}. \textbf{Left:} In the policy update step, the current discriminator evaluates trajectories from both the expert and the learner. The learner adjusts its behavior to increase its score. \textbf{Right:} In the discriminator update step, the updated learner behavior is used to retrain the discriminator, which becomes better at identifying differences. Through this back-and-forth process, the learner learns to behave more like the expert.}
		\label{fig:gail_diagram_1}
	\end{figure}

	To build a more intuitive understanding of how GAIL works, consider what happens during a single training iteration. Figure~\ref{fig:gail_diagram_1} illustrates the two main steps in this loop. The left side shows the learner's policy update step. At this stage, the discriminator is kept fixed and used to evaluate different trajectories. The expert's trajectory receives a high score (0.9), meaning the discriminator is confident that it is authentic. A trajectory generated by the learner's current policy receives a much lower score (0.3), indicating that it is still distinguishable from the expert's behavior. Then, the learner updates its policy to achieve higher scores in trajectories generated by the learner, i.e., to behave more expert-like (to fool the discriminator). For example, a new trajectory from the updated learner's policy might now receive a significantly higher score (0.7), showing progress.

	The right side of the figure shows the discriminator update step. Here, the learner's policy is fixed, and the discriminator is retrained using both expert trajectories and those generated by the updated learner's policy. Before the update, the expert's trajectory receives a high score (0.9), meaning the discriminator is confident that it is authentic. Also, a trajectory generated by the learner's current policy receives a lower score (0.7), indicating that it is still distinguishable from the expert's behavior. Then, the learner updates its discriminator to become more precise (i.e., to be less fooled) by lowering the scores in trajectories generated by the learner ($0.7\rightarrow 0.5$).

\begin{figure}[h]
		\centering
		\includegraphics[width=0.65\linewidth]{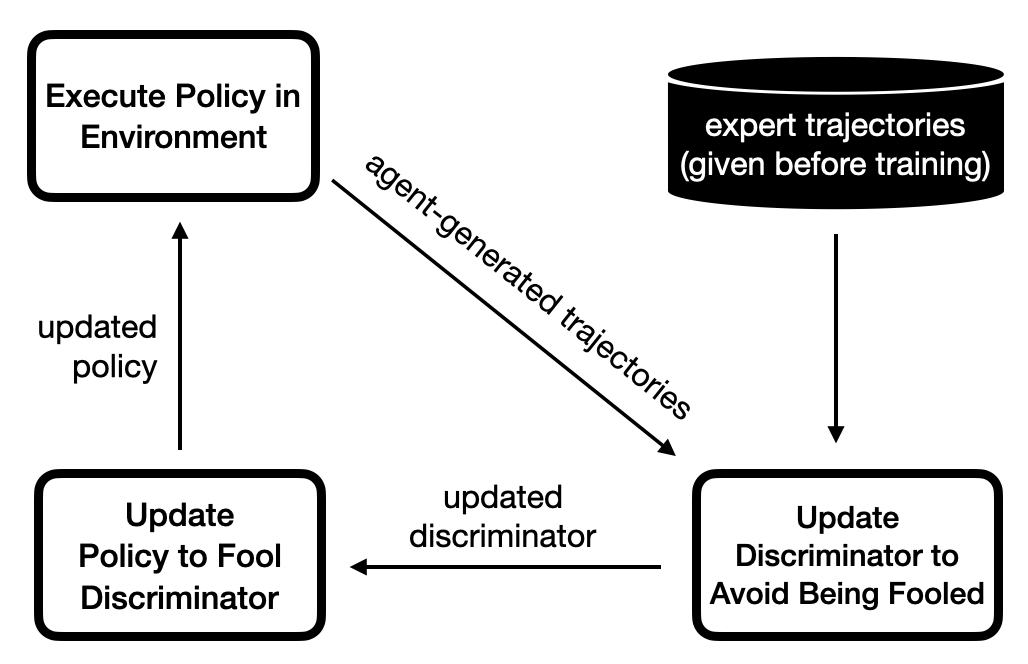}
		\caption{High-level summary of the GAIL \cite{gail} training loop. The learner's policy is executed in the environment to generate new trajectories. These are used, along with expert demonstrations collected prior to training, to update the discriminator. The updated discriminator provides a feedback signal that guides the learner's policy improvement via reinforcement learning. The process iterates, with each component improving in response to the other.}
		\label{fig:gail_2}
	\end{figure}

	The process then repeats, with each component improving in response to the other. Over time, this adversarial loop leads the learner's policy to generate behavior that is increasingly difficult to distinguish from the expert's. Ideally, training converges when the discriminator can no longer reliably tell them apart, i.e., when the learner behaves in a way that is effectively indistinguishable from the expert. The diagram in Figure~\ref{fig:gail_2} summarizes this interaction at a higher level, abstracting away individual scores and focusing on the flow of information.

	The next sections present a detailed treatment of GAIL, beginning with its mathematical formulation and followed by the algorithmic pseudocode.

	\subsection{Expert Demonstrations}

	GAIL assumes access to a collection of expert demonstrations before training, denoted \( \mathcal{D}_E \), consisting of \( N \) state-action pairs collected by observing an expert performing the task, possibly over multiple episodes. Differently from DAgger, GAIL will not query the expert for additional demonstrations. Each pair is assumed to be generated deterministically by the expert policy \( \pi_E \), as follows:
	\[
		\mathcal{D}_E \doteq \left\{ \left(s^{(i)}, a^{(i)} \right) \right\}_{i=1}^N, \quad \text{where } a^{(i)} = \pi_E(s^{(i)}).
	\]

	\subsection{Discriminator Training Step}

	In GAIL, the role of the discriminator is to determine whether a given state-action pair originates from the expert or from the learner's policy. Formally, the discriminator is a function \cite{gail}:
	\[
		D_\phi : \mathcal{S} \times \mathcal{A} \rightarrow [0,1].
	\]

	The discriminator function is parameterized by \( \phi \), which typically denotes the weights of a neural network. The final layer of $D_\phi$ applies the sigmoid activation function so that the output represents a probability.

	Given a state-action pair \( (s, a) \in \mathcal{S} \times \mathcal{A} \), the output of \( D_\phi(s, a) \) represents the predicted probability that the pair was generated by the expert. A value close to 1 indicates confidence that $(s, a)$ was generated by the expert, whereas close to 0 indicates confidence was generated by the learner.

	The original GAIL paper~\cite{gail} defines the discriminator such that it outputs high values for learner behavior and low values for expert behavior (i.e., expert = 0, learner = 1). However, most modern implementations, including the \texttt{imitation} library~\cite{imitation}, adopt the opposite convention, treating expert behavior as class 1 and learner behavior as class 0. This aligns with standard binary classification practice and simplifies the computation of the loss using common libraries.

	Training this binary classifier requires a labeled dataset composed of both expert- and learner-generated data. The expert dataset is the collection of \( N \) expert demonstrations \( \mathcal{D}_E \). The learner dataset is generated by executing the current policy \( \pi_\theta \) in the environment, collecting \( M \) state-action pairs over one or more episodes.

	Concretely, at each time step $i$ over one or several episodes:
	(a) the learner observes the current state \( s_i \);
	(b) samples an action from the current policy, \( a_i \sim \pi(\cdot \mid s_i, \boldsymbol{\theta})) \);
	(c) applies the action \( a_i \) in the environment;
	(d) records the resulting pair \( (s_i, a_i) \);
	(e) the environment transitions to the next state or resets it in a terminate state.

	This procedure produces a dataset of \( M \) state-action pairs:
	\[
		\mathcal{D}_\pi \doteq \left\{ (s_i, a_i) \right\}_{i=1}^M, \quad \text{with } a_i \sim \pi(\cdot \mid s_i, \boldsymbol{\theta}).
	\]

	Finally, the expert and learner datasets are combined into a single labeled dataset of \( M + N \) state-action pairs, where pairs generated by the expert are labeled \( 1 \) and pairs generated by the learner's policy are labeled \( 0 \)~\cite{gail}:
	\[
		\mathcal{D}_{\text{disc}} \doteq
		\left\{ ((s, a), 1) \;\middle|\; (s, a) \in \mathcal{D}_E \right\}
		\;\cup\;
		\left\{ ((s, a), 0) \;\middle|\; (s, a) \in \mathcal{D}_\pi \right\}.
	\]

	To train the discriminator it is necessary to define the \textit{loss} function that the training process will seek to minimize. For a single labeled sample \( ((s, a), y) \), the cross-entropy loss quantifies how well its predictions match the ground-truth labels \cite{gail}:
	\[
		\ell(y, D_\phi(s, a)) \doteq - y \ln D_\phi(s, a) - (1 - y) \ln(1 - D_\phi(s, a)).
	\]

	This equation is easier to understand by considering the two possible cases. If \( \mathbf{y = 1} \) (expert data), the loss reduces to:
	\[
		\ell(1, D_\phi(s, a)) \doteq - \ln D_\phi(s, a),
	\]

	\noindent which penalizes the discriminator if it predicts 'learner'. That is, the closer \( D_\phi(s, a) \) is to 0, the higher the loss: $D \rightarrow 0 \implies l \rightarrow \infty$.

	In practice, the loss remains finite because the sigmoid in $D_\phi$ ensures its output never reaches 0 or 1. Conversely, if \( \mathbf{y = 0} \) (policy data), the loss reduces to:
	\[
		\ell(0, D_\phi(s, a)) \doteq - \ln(1 - D_\phi(s, a)).
	\]

	This loss penalizes the discriminator if it predicts 'expert'. The closer \( D_\phi(s, a) \) is to 1, the higher the loss: $D \rightarrow 1 \implies l \rightarrow \infty$. As before, in practice, the loss remains finite because the sigmoid in $D_\phi$ ensures its output never reaches 0 or 1.

	In both cases, the loss encourages the training algorithm to find the discriminator parameters that allow it to assign a high score to expert pairs and a low score to learner pairs. The output \( D_\phi(s, a) \in [0, 1] \) can be interpreted as the model's confidence that the behavior is expert-like. Cross-entropy loss then pushes the model to match its confidence with the true label.

	In practice, the discriminator is trained using \textit{mini-batch} stochastic gradient descent over a fixed of $E$ learning epochs. A learning epoch proceeds as follows. First, \( \mathcal{D}_{\text{disc}} \) is shuffled and split into mini-batches of size \( B \). Then, the empirical loss over each mini-batch \(\mathcal{B} \subset \mathcal{D}_{\text{disc}}\) is computed as the average of the individual cross-entropy losses \cite{gail}:
	\begin{equation}
		\hat{\mathcal{L}}_{\text{disc}}(\phi, \mathcal{B}) \doteq \frac{1}{B} \sum_{((s,a), y) \in \mathcal{B}}^B \ell\left(y, D_\phi(s, a)\right).
		\label{eq:ldisc}
	\end{equation}

	Finally, the discriminator parameters \( \phi \) are updated in-place after processing each mini-batch, using the gradient of the mini-batch loss and a learning rate \( \alpha \) \cite{gail}:
	\[
		\phi \leftarrow \phi - \alpha \nabla_\phi \hat{\mathcal{L}}_{\text{disc}}(\phi, \mathcal{B}).
	\]

	This is the simplest possible update rule, commonly referred to as \emph{vanilla stochastic gradient descent}. In practice, however, more advanced optimizers such as \textbf{Adam} are often preferred for their adaptive learning rates and improved convergence properties. In practice, optimizers like Adam are also widely used for updating the policy in Behavioral Cloning and DAgger, not just in adversarial settings like GAIL. To reflect this flexibility and abstract away implementation details, the update step is rewritten as:
	\[
		\phi \leftarrow \text{DiscriminatorUpdate}(\phi, \nabla_\phi \hat{\mathcal{L}}_{\text{disc}}(\phi, \mathcal{B})).
	\]

	\subsection{Policy Training Step}

	While the discriminator is trained to distinguish expert behavior from learner's policy behavior, its feedback is used as a reward-like signal to improve the policy.  The goal of the policy is to produce trajectories that maximize this reward signal.

	Recall that the discriminator \( D_\phi(s, a) \) estimates the probability that a state-action pair comes from the expert. A natural idea is to treat this estimate as a measure of how good a given behavior is (i.e., how rewarding it is). In particular, the learner can receive a high reward when it behaves in a way that the discriminator classifies as expert-like: the goal is to learn a policy that fools the discriminator into thinking it is observing the expert.

	GAIL defines the policy's reward signal based on the discriminator's output \cite{gail}:
	\[
		r_\phi(s, a) \doteq -\ln \left(1 - D_\phi(s, a)\right),
	\]

	\noindent which assigns higher rewards to state-action pairs that the discriminator believes likely came from the expert: $D\rightarrow 1 \implies r\rightarrow \infty$. As before, in practice, the reward remains finite because the sigmoid in $D_\phi$ ensures its output never reaches 0 or 1.

Rather than delving into the internal mechanics of reinforcement learning algorithms, the learning process is abstracted into a general policy improvement function. This function receives two inputs: the current parameters of the policy, \( \boldsymbol{\theta} \), and a reward function \( r_\phi(s, a) \), which is derived from the discriminator's feedback. It returns a new set of parameters that define an improved policy:
\[
	\boldsymbol{\theta} \leftarrow \text{PolicyUpdate}(\boldsymbol{\theta}, r_\phi).
\]

Although compact, this abstraction encapsulates a complete learning cycle. Internally, the function first executes the policy \( \pi_{\boldsymbol{\theta}} \) in the environment, collecting a batch of state-action trajectories. At each time step of each trajectory and it evaluates the reward function \( r_\phi(s_t, a_t) \). The policy is then updated to increase the likelihood of selecting actions that lead to higher expected reward, using reinforcement learning techniques.

This process can be viewed as a black-box optimizer: it adapts the policy so that it becomes increasingly indistinguishable from the expert, as judged by the discriminator. In practice, this policy improvement function is often implemented using standard algorithms, such as PPO \cite{schulman2017ppo}, but understanding the details of those methods is not necessary at this stage. These algorithms are detailed in Chapter~\ref{sec:chapter_rl}.

\subsection{GAIL Pseudocode}

Algorithm~\ref{alg:gail} presents the pseudocode of the full training loop of GAIL. Building on the components introduced earlier, the algorithm alternates between collecting learner trajectories using the current policy, updating the discriminator to distinguish expert from learner behavior, and refining the policy based on the discriminator's feedback.

Each iteration begins by rolling out the current learning policy in the environment to generate new state-action pairs. These are combined with the fixed set of expert demonstrations to form a labeled dataset for discriminator training. The discriminator is then updated over several epochs using mini-batches of this dataset, applying standard gradient-based optimization. Once the discriminator has been updated, the policy is improved through reinforcement learning, using a reward signal derived from the discriminator's confidence in the learner's behavior. This adversarial process continues for a fixed number of iterations, ultimately yielding a policy that closely imitates the expert.

\vspace{0.5cm}
{\small
	\begin{algorithm}[h]
		\caption{Pseudocode of GAIL \cite{gail}}
		\label{alg:gail}
		\KwIn{Expert dataset \( \mathcal{D}_E \doteq \{(s^{(i)}, a^{(i)})\}_{i=1}^N \)}
		\KwIn{Batch size $B$, number of iterations \( K \), number of epochs $E$}
		\KwIn{Policy function $\pi_\theta$ and Discriminator function $D_\phi$}
		\KwIn{Reward function $r_\phi(s, a) = -\ln \left(1 - D_\phi(s, a)\right).$}
		\KwIn{Discriminator loss function $\hat{\mathcal{L}}_{\text{disc}}$ (Equation~\ref{eq:ldisc})}
		\KwOut{Final policy \( \pi_{\boldsymbol{\theta}} \)}

		Initialize policy parameters \( \boldsymbol{\theta} \) (e.g., randomly)\\
		Initialize discriminator parameters \( \phi \) (e.g., randomly)

		\For{$k = 1$ \KwTo $K$}{
		Rollout policy \( \pi_{\boldsymbol{\theta}} \) in the env. over multiple episodes to collect $M$ state-action pairs:\\
		\Indp
		$ \mathcal{D}_\pi \doteq \{(s^{(i)}, a^{(i)})\}_{i=1}^M $

		\Indm
		Construct a labeled dataset of $M+N$ state-action pairs:\\
		\Indp
		$\mathcal{D}_{\text{disc}} \doteq \left\{ ((s, a), 1) : \forall (s, a) \in \mathcal{D}_E \right\}
			\cup
			\left\{ ((s, a), 0) : \forall (s, a) \in \mathcal{D}_\pi \right\}.$

		\Indm
		\For{$e = 1$ \KwTo $E$}{
		Shuffle $\mathcal{D}_{\text{disc}}$ and split it into mini-batches of size $B$\\
		\ForEach{\text{mini-batch} $\mathcal{B} \subset \mathcal{D}_{\text{disc}}$}{
		Compute gradient:\\
		\Indp
		$\delta \leftarrow \nabla_\phi \hat{\mathcal{L}}_{\text{disc}}(\phi, \mathcal{B})$\\
		\Indm
		Update discriminator parameters: \\
		\Indp
		$\phi \leftarrow \text{DiscriminatorUpdate}(\phi, \delta)$.
		}
		}

		Update policy parameters with reinforcement learning:\\
		\Indp
		$\boldsymbol{\theta} \leftarrow \text{PolicyUpdate}(\boldsymbol{\theta}, r_{\phi}).$
		}
	\end{algorithm}
}

\subsection{GAIL in Action: Examples and Results}

\begin{figure}[h]
	\centering
	\begin{subfigure}[b]{0.4\textwidth}
		\centering
		\includegraphics[height=5cm]{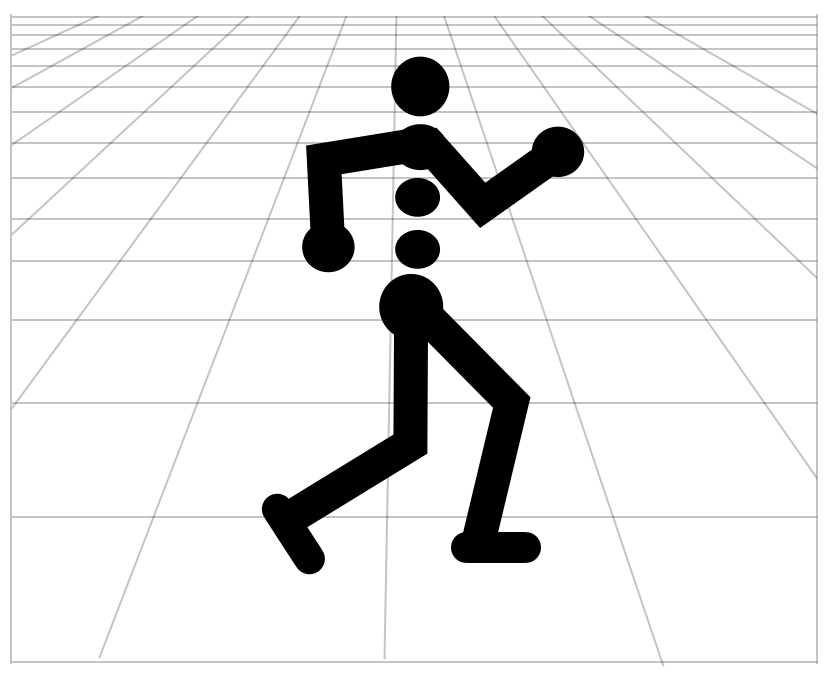}
		\label{fig:humanoid}
	\end{subfigure}\hspace{0.5cm}
	\begin{subfigure}[b]{0.4\textwidth}
		\centering
		\includegraphics[height=5cm]{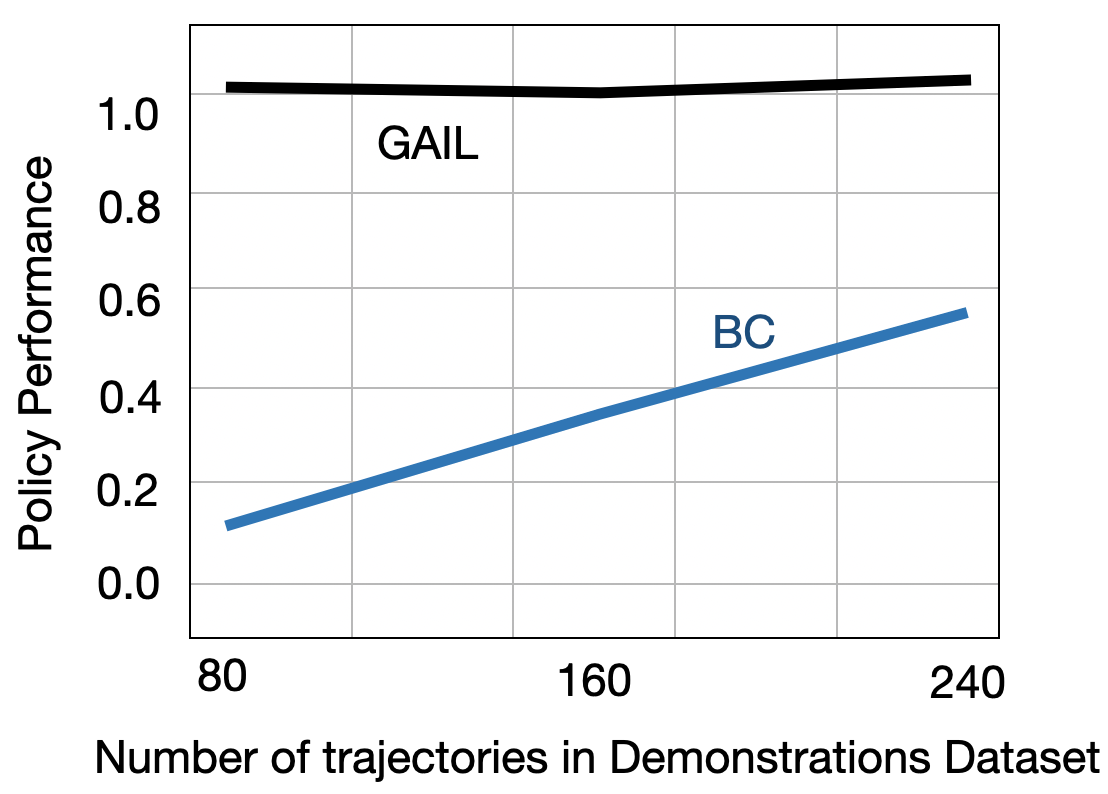}
		\label{fig:gail-results}
	\end{subfigure}
	\caption{Schematic of 3D humanoid-like simulation in MuJoCo (left), and  comparison between Behavioral Cloning (BC) and GAIL \cite{gail} on this control task (right).}
	\label{fig:mujoco}
\end{figure}

The authors of GAIL evaluated their approach against Behavioral Cloning  across several benchmark tasks \cite{gail}. The experiments include continuous-control environments simulated in MuJoCo, such as a 3D bipedal humanoid-like robot\footnote{\url{https://gymnasium.farama.org/environments/mujoco/humanoid/}} (see Figure~\ref{fig:mujoco}). In this task, the agent's objective is to move forward as fast as possible without losing balance.

Figure~\ref{fig:mujoco} shows that GAIL consistently outperforms Behavioral Cloning (BC) in this task, with the performance gap being most pronounced when only a small number of expert demonstrations are available. As the number of expert demonstrations increases, the performance of both methods improves, yet GAIL maintains a clear advantage. The original GAIL paper further demonstrates that this trend holds across multiple continuous control domains.

\clearpage
\addcontentsline{toc}{section}{References}
\bibliographystyle{plainurl}
\bibliography{references}

@article{rumelhart1986learning,
  title={Learning representations by back-propagating errors},
  author={Rumelhart, David E and Hinton, Geoffrey E and Williams, Ronald J},
  journal={Nature},
  volume={323},
  number={6088},
  pages={533--536},
  year={1986},
  publisher={Nature Publishing Group UK London}
}

@article{gan,
  title={Generative adversarial networks},
  author={Goodfellow, Ian and Pouget-Abadie, Jean and Mirza, Mehdi and Xu, Bing and Warde-Farley, David and Ozair, Sherjil and Courville, Aaron and Bengio, Yoshua},
  journal={Communications of the ACM},
  volume={63},
  number={11},
  pages={139--144},
  year={2020},
  publisher={ACM New York, NY, USA}
}

@article{schulman2015high,
  title={High-dimensional continuous control using generalized advantage estimation},
  author={Schulman, John and Moritz, Philipp and Levine, Sergey and Jordan, Michael and Abbeel, Pieter},
  journal={arXiv preprint arXiv:1506.02438},
  year={2015}
}

@article{Bojarski2016,
  title={End to end learning for self-driving cars},
  author={Bojarski, Mariusz and Del Testa, Davide and Dworakowski, Daniel and Firner, Bernhard and Flepp, Beat and Goyal, Prasoon and Jackel, Lawrence D and Monfort, Mathew and Muller, Urs and Zhang, Jiakai and others},
  journal={arXiv preprint arXiv:1604.07316},
  year={2016}
}

@article{imitation,
  title={imitation: Clean imitation learning implementations},
  author={Gleave, Adam and Taufeeque, Mohammad and Rocamonde, Juan and Jenner, Erik and Wang, Steven H and Toyer, Sam and Ernestus, Maximilian and Belrose, Nora and Emmons, Scott and Russell, Stuart},
  journal={arXiv preprint arXiv:2211.11972},
  year={2022},
  url          = {https://github.com/HumanCompatibleAI/imitation}
}

@book{goodfellow2016deep,
  author       = {I. Goodfellow and Y. Bengio and A. Courville},
  title        = {Deep Learning},
  publisher    = {MIT Press},
  year         = {2016},
  url          = {http://deeplearningbook.org/}
}

@inproceedings{haarnoja2018sac,
  title={Soft actor-critic: Off-policy maximum entropy deep reinforcement learning with a stochastic actor},
  author={Haarnoja, Tuomas and Zhou, Aurick and Abbeel, Pieter and Levine, Sergey},
  booktitle={International conference on machine learning},
  pages={1861--1870},
  year={2018},
  organization={Pmlr},
  url          = {https://arxiv.org/pdf/1801.01290}
}

@inproceedings{gail,
  title={Generative adversarial imitation learning},
  author={Ho, Jonathan and Ermon, Stefano},
  booktitle={Advances in neural information processing systems},
  volume={29},
  year={2016}
}

@article{Hussein2017,
  title={Imitation learning: A survey of learning methods},
  author={Hussein, Ahmed and Gaber, Mohamed Medhat and Elyan, Eyad and Jayne, Chrisina},
  journal={ACM Computing Surveys (CSUR)},
  volume={50},
  number={2},
  pages={1--35},
  year={2017},
  publisher={ACM New York, NY, USA}
}

@article{Juliani2018,
  title={Unity: A general platform for intelligent agents},
  author={Juliani, Arthur and Berges, Vincent-Pierre and Teng, Ervin and Cohen, Andrew and Harper, Jonathan and Elion, Chris and Goy, Chris and Gao, Yuan and Henry, Hunter and Mattar, Marwan and others},
  journal={arXiv preprint arXiv:1809.02627},
  year={2018},
  url          = {https://github.com/Unity-Technologies/ml-agents}
}

@inproceedings{Kwiatkowski2022,
  title={A survey on reinforcement learning methods in character animation},
  author={Kwiatkowski, Ariel and Alvarado, Eduardo and Kalogeiton, Vicky and Liu, C Karen and Pettr{\'e}, Julien and van de Panne, Michiel and Cani, Marie-Paule},
  booktitle={Computer graphics forum},
  volume={41},
  pages={613--639},
  year={2022},
  organization={Wiley Online Library}
}

@article{lillicrap2015ddpg,
  title={Continuous control with deep reinforcement learning},
  author={Lillicrap, Timothy P and Hunt, Jonathan J and Pritzel, Alexander and Heess, Nicolas and Erez, Tom and Tassa, Yuval and Silver, David and Wierstra, Daan},
  journal={arXiv preprint arXiv:1509.02971},
  year={2015},
  url          = {https://arxiv.org/pdf/1509.02971}
}

@article{mnih2015dqn,
  title={Human-level control through deep reinforcement learning},
  author={Mnih, Volodymyr and Kavukcuoglu, Koray and Silver, David and Rusu, Andrei A and Veness, Joel and Bellemare, Marc G and Graves, Alex and Riedmiller, Martin and Fidjeland, Andreas K and Ostrovski, Georg and others},
  journal={nature},
  volume={518},
  number={7540},
  pages={529--533},
  year={2015},
  publisher={Nature Publishing Group},
  url          = {https://www.nature.com/articles/nature14236}
}

@inproceedings{Mourot2022,
  title={A survey on deep learning for skeleton-based human animation},
  author={Mourot, Lucas and Hoyet, Ludovic and Le Clerc, Fran{\c{c}}ois and Schnitzler, Fran{\c{c}}ois and Hellier, Pierre},
  booktitle={Computer Graphics Forum},
  volume={41},
  pages={122--157},
  year={2022},
  organization={Wiley Online Library}
}

@misc{openai2018spinningup,
  author       = {OpenAI},
  title        = {OpenAI Spinning Up in Deep RL},
  year         = {2018},
  url          = {https://spinningup.openai.com/}
}

@article{Pomerleau1991,
  title={Efficient training of artificial neural networks for autonomous navigation},
  author={Pomerleau, Dean A},
  journal={Neural computation},
  volume={3},
  number={1},
  pages={88--97},
  year={1991},
  publisher={MIT Press}
}

@inproceedings{dagger,
  title={A reduction of imitation learning and structured prediction to no-regret online learning},
  author={Ross, St{\'e}phane and Gordon, Geoffrey and Bagnell, Drew},
  booktitle={Proceedings of the fourteenth international conference on artificial intelligence and statistics},
  pages={627--635},
  year={2011},
  organization={JMLR Workshop and Conference Proceedings}
}

@misc{sb32024,
  author       = {{SB3 Team}},
  title        = {Stable Baselines 3},
  year         = {2024},
  url          = {https://github.com/DLR-RM/stable-baselines3/}
}

@article{schulman2017ppo,
  title={Proximal policy optimization algorithms},
  author={Schulman, John and Wolski, Filip and Dhariwal, Prafulla and Radford, Alec and Klimov, Oleg},
  journal={arXiv preprint arXiv:1707.06347},
  year={2017},
  url          = {https://arxiv.org/pdf/1707.06347}
}

@book{sutton2018reinforcement,
  author       = {R. S. Sutton and A. G. Barto},
  title        = {Reinforcement Learning: An Introduction},
  publisher    = {MIT Press},
  year         = {2018},
  url          = {https://www.andrew.cmu.edu/course/10-703/textbook/BartoSutton.pdf}
}

@article{williams1992reinforce,
  title={Simple statistical gradient-following algorithms for connectionist reinforcement learning},
  author={Williams, Ronald J},
  journal={Machine learning},
  volume={8},
  number={3},
  pages={229--256},
  year={1992},
  publisher={Springer}
}

@article{zare2024,
  title={A survey of imitation learning: Algorithms, recent developments, and challenges},
  author={Zare, Maryam and Kebria, Parham M and Khosravi, Abbas and Nahavandi, Saeid},
  journal={IEEE Transactions on Cybernetics},
  year={2024},
  publisher={IEEE}
}

@article{Zheng2022,
  title={Imitation learning: Progress, taxonomies and challenges},
  author={Zheng, Boyuan and Verma, Sunny and Zhou, Jianlong and Tsang, Ivor W and Chen, Fang},
  journal={IEEE Transactions on Neural Networks and Learning Systems},
  volume={35},
  number={5},
  pages={6322--6337},
  year={2022},
  publisher={IEEE}
}

\end{document}